\pdfoutput=1

\documentclass[10pt,twocolumn,letterpaper]{article}
\usepackage{times}
\usepackage{xspace}
\usepackage[dvipsnames]{xcolor}
\usepackage{graphicx}
\usepackage{amsmath,amssymb,booktabs}
\usepackage[numbers,sort&compress]{natbib}
\usepackage{etoolbox}
\usepackage[hyphens]{url}
\usepackage[format=plain,labelformat=simple,labelsep=period,font=small,compatibility=false]{caption}
\usepackage[font=footnotesize,skip=3pt,subrefformat=parens]{subcaption}
\usepackage[shortlabels,inline]{enumitem}
\setlist[itemize]{noitemsep,leftmargin=*,topsep=0em}
\setlist[enumerate]{noitemsep,leftmargin=*,topsep=0em}

\renewcommand{\textfraction}{0.01}
\renewcommand{\floatpagefraction}{0.99}
\renewcommand{\topfraction}{0.99}

\renewcommand{\dblfloatpagefraction}{0.99}
\renewcommand{\dbltopfraction}{0.99}
\makeatletter
\renewcommand{\@maketitle}{%
  \newpage\null\vskip .375in
  \begin{center}
    {\Large\bfseries\@title\par}
    \vspace*{8pt}
    {\large\lineskip .5em\begin{tabular}[t]{c}\@author\end{tabular}\par}
    \vskip .5em
  \end{center}}
\renewenvironment{abstract}{%
  \centerline{\large\bfseries Abstract}\vspace*{12pt}\noindent\itshape\ignorespaces
}{\vspace*{12pt}}
\font\arxivsubheadingfont=ptmb scaled 1100
\font\arxivsubsubheadingfont=ptmb scaled 1000
\def\arxivsection{\@startsection{section}{1}{\z@}{-10pt plus -2pt minus -2pt}{7pt}{\large\bfseries}}
\def\arxivsectionstar#1{\arxivsection*{#1}}
\def\arxivsectionnumbered#1{\arxivsection{\texorpdfstring{\hskip -1em.~}{}#1}}
\renewcommand{\section}{\@ifstar\arxivsectionstar\arxivsectionnumbered}
\def\arxivsubsection{\@startsection{subsection}{2}{\z@}{-8pt plus -2pt minus -2pt}{5pt}{\arxivsubheadingfont}}
\def\arxivsubsectionstar#1{\arxivsubsection*{#1}}
\def\arxivsubsectionnumbered#1{\arxivsubsection{\texorpdfstring{\hskip -1em.~}{}#1}}
\renewcommand{\subsection}{\@ifstar\arxivsubsectionstar\arxivsubsectionnumbered}
\def\arxivsubsubsection{\@startsection{subsubsection}{3}{\z@}{-6pt plus -2pt minus -2pt}{3pt}{\arxivsubsubheadingfont}}
\def\arxivsubsubsectionstar#1{\arxivsubsubsection*{#1}}
\def\arxivsubsubsectionnumbered#1{\arxivsubsubsection{\texorpdfstring{\hskip -1em.~}{}#1}}
\renewcommand{\subsubsection}{\@ifstar\arxivsubsubsectionstar\arxivsubsubsectionnumbered}
\DeclareRobustCommand\onedot{\futurelet\@let@token\@onedot}
\def\@onedot{\ifx\@let@token.\else.\null\fi\xspace}
\def\etal{\emph{et al}\onedot}
\makeatother

\let\arxivoriginaltitle\title
\renewcommand{\title}[1]{\arxivoriginaltitle{#1}\gdef\thetitle{#1}}
\newcommand{\maketitlesupplementary}{%
  \newpage\twocolumn[\centering\Large\textbf{\thetitle}\\
  \vspace{0.5em}Supplementary Material\\\vspace{1.0em}]}

\AtEndPreamble{%
  \usepackage[capitalize]{cleveref}
  \crefname{section}{Sec.}{Secs.}\Crefname{section}{Section}{Sections}
  \crefname{table}{Tab.}{Tabs.}\Crefname{table}{Table}{Tables}}

\usepackage{amsthm}
\usepackage[final,protrusion=false]{microtype}
\usepackage{needspace}
\usepackage{placeins}
\usepackage{bm}
\usepackage{multirow}
\usepackage{array}
\usepackage{longtable}
\usepackage[ruled,vlined,linesnumbered]{algorithm2e}
\DontPrintSemicolon
\SetKwComment{Comment}{$\triangleright$\ }{}
\SetKwInOut{KwIn}{Input}\SetKwInOut{KwOut}{Output}
\SetNlSty{}{}{:}
\SetNlSkip{0.25em}
\SetArgSty{textnormal}\SetCommentSty{textnormal}
\usepackage{tikz}
\usetikzlibrary{arrows.meta,positioning,calc,fit,backgrounds,shapes.geometric,decorations.pathreplacing}
\usepackage{pgfplots}
\pgfplotsset{compat=1.18}
\usepgfplotslibrary{groupplots,fillbetween}
\usepackage{pgfplotstable}

\definecolor{oiOrange}{RGB}{230,159,0}
\definecolor{oiSky}{RGB}{86,180,233}
\definecolor{oiGreen}{RGB}{0,158,115}
\definecolor{oiYellow}{RGB}{240,228,66}
\definecolor{oiBlue}{RGB}{0,114,178}
\definecolor{oiVermilion}{RGB}{213,94,0}
\definecolor{oiPurple}{RGB}{204,121,167}
\definecolor{oiGrey}{RGB}{110,110,110}
\definecolor{hotpink}{RGB}{255,105,180}%
\colorlet{colOurs}{oiBlue}
\colorlet{colLiu}{oiVermilion}
\colorlet{colBPnP}{oiGreen}
\colorlet{colDet}{oiOrange}
\colorlet{colModel}{oiSky}

\pgfplotsset{mpPlot/.style={
  grid=none, axis lines*=left, tick pos=left, tick align=outside,
  major tick length=2pt, minor x tick num=0, minor y tick num=0,
  minor tick style={draw=none},
  axis line style={oiGrey!90, line width=0.4pt},
  tick style={oiGrey!90, line width=0.35pt},
  tick label style={font=\footnotesize}, label style={font=\footnotesize},
  title style={font=\footnotesize},
  legend style={font=\footnotesize, draw=none, fill=none},
  legend cell align=left, axis background/.style={fill=white}
}}

\newtheoremstyle{compactplain}{4pt plus 1pt minus 1pt}{4pt plus 1pt minus 1pt}{\itshape}{}{\bfseries}{.}{0.5em}{}
\newtheoremstyle{compactdefinition}{4pt plus 1pt minus 1pt}{4pt plus 1pt minus 1pt}{\normalfont}{}{\bfseries}{.}{0.5em}{}
\theoremstyle{compactplain}
\newtheorem{theorem}{Theorem}
\newtheorem{lemma}{Lemma}
\newtheorem{corollary}{Corollary}
\theoremstyle{compactdefinition}
\newtheorem{definition}{Definition}

\AtEndPreamble{\crefname{assumption}{Assumption}{Assumptions}\Crefname{assumption}{Assumption}{Assumptions}%
  \crefname{appendix}{App.}{Apps.}\Crefname{appendix}{Appendix}{Appendices}%
  \crefname{algocf}{Algorithm}{Algorithms}\Crefname{algocf}{Algorithm}{Algorithms}}
\makeatletter
\renewenvironment{proof}[1][\proofname]{\par\pushQED{\qed}\normalfont\topsep3\p@\@plus1\p@\relax
  \trivlist\item[\hskip\labelsep\itshape#1\@addpunct{.}]\ignorespaces}{\popQED\endtrivlist\@endpefalse}
\makeatother
\newcommand{\R}{\mathbb{R}}
\newcommand{\SO}{\mathrm{SO}(3)}
\newcommand{\norm}[1]{\lVert #1 \rVert}
\newcommand{\abs}[1]{\lvert #1 \rvert}
\newcommand{\mat}[1]{#1}%
\newcommand{\vect}[1]{\bm{#1}}%
\newcommand{\E}{\mathbb{E}}%
\DeclareMathOperator{\supp}{supp}%
\DeclareMathOperator{\cyc}{cyc}%
\DeclareMathOperator{\rank}{rank}
\DeclareMathOperator{\tr}{tr}

\newcommand{\dets}{\mathcal{U}}%
\newcommand{\pts}{\mathcal{Y}}%
\newcommand{\Kmat}{\mat{K}}%
\newcommand{\proj}{\operatorname{proj}}%
\newcommand{\Rot}{\mat{R}}%
\newcommand{\trans}{\vect{t}}%
\newcommand{\halfturn}{H}%
\newcommand{\opthalf}{D}%
\newcommand{\refdepth}{z_{\mathrm{ref}}}%

\newcommand{\order}{g}%
\newcommand{\mom}{\mathcal{M}}%
\newcommand{\Vset}{\mathcal{V}}%
\newcommand{\Fmap}{\mathcal{F}}%
\newcommand{\Gmod}{\Gamma_T}%

\newcommand{\noise}{s}%
\newcommand{\noisefit}{\hat s}%
\newcommand{\projrad}{\rho}%
\newcommand{\rmsrad}{\rho_{\mathrm{rms}}}%
\newcommand{\obserr}{\epsilon}%
\newcommand{\fitres}{\hat\epsilon}%
\newcommand{\fmax}{f_{\max}}%
\newcommand{\predtol}{\epsilon_{\rm pred}}%
\newcommand{\prederr}{\hat\delta}%

\newcommand{\featbound}{B}%
\newcommand{\featlip}{\mathrm{Lip}(\vect\phi)}%
\newcommand{\unmatched}{\eta}%
\newcommand{\edgelen}{\lambda}%
\newcommand{\edgemass}{\gamma_E}%
\newcommand{\edgedisp}{\delta_E}%
\newcommand{\quaderr}{\epsilon_q}%
\newcommand{\fourerr}{\Delta_\delta}%
\newcommand{\fourconst}{C_F}%
\newcommand{\posecoord}{\vect\delta}%
\newcommand{\jacsv}{\kappa_J}%
\newcommand{\nnb}{\ell}%
\newcommand{\forwardcount}{N_+}%
\newcommand{\assoc}{E}%
\newcommand{\lik}{\mathcal{L}}%

\newcommand{\degree}{^{\circ}}
\newcommand{\method}{MomentPnP\xspace}
\newcommand{\head}[1]{\par\addvspace{3pt plus 1pt minus 1pt}\noindent\phantomsection\textbf{#1}\enspace\ignorespaces}
\newcommand{\para}[1]{\head{#1}}

\ifdefined\pdfinfoomitdate\pdfinfoomitdate=1\fi
\ifdefined\pdftrailerid\pdftrailerid{}\fi

\AtBeginDocument{\urlstyle{same}}

\makeatletter
\@ifundefined{@ShipoutPicture}{}{%
  \newbox\mp@pendinglabels
  \let\mp@ShipoutPicture\@ShipoutPicture
  \renewcommand{\@ShipoutPicture}{%
    \global\setbox\mp@pendinglabels\box\@labels
    \global\let\ifmp@pendinglabel\if@inlabel \global\@inlabelfalse
    \mp@ShipoutPicture
    \global\setbox\@labels\box\mp@pendinglabels
    \global\let\if@inlabel\ifmp@pendinglabel}}
\makeatother

\makeatletter
\@ifpackageloaded{lineno}{%
  \IfHookExistsTF{build/column/before}{\RemoveFromHook{build/column/before}[lineno]}{}%
  \AtBeginDocument{%
    \def\@LN@makecol{%
      \@LN@orig@makecol
      \setbox\@outputbox\vbox{%
        \boxmaxdepth\@maxdepth
        \protected@write\@auxout{}{\string\def\string\@LN@column{\if@firstcolumn1\else2\fi}}%
        \box\@outputbox}}%
  }%
}{}
\makeatother

\definecolor{linkblue}{rgb}{0.21,0.49,0.74}
\usepackage[pagebackref,breaklinks,colorlinks,allcolors=linkblue]{hyperref}
\hypersetup{%
  pdftitle={Correspondence-Free Camera Pose Estimation by Matching Moments},
  pdfauthor={Subhransu S. Bhattacharjee, Dylan Campbell, Rahul Shome}}

\newcommand{\MNaddsOK}{99.8\%}
\newcommand{\MDdeclinedRight}{7}
\newcommand{\MNdeclinedRight}{60}
\newcommand{\EmRestarts}{210}
\newcommand{\ThresholdKappa}{0.21}
\newcommand{\TeaserScene}{Sacr\'e-C\oe ur image from the Phototourism dataset~\cite{snavely2006phototourism,jin2021imc}}
\newcommand{\SceneMaxErr}{$0.015\degree$}

\newcommand{\SceneCellRankSentence}{For the 70 complete grids used by the tuning queries in \cref{tab:scene}, the cell containing the reference camera center ranks in the top quarter in 80\% of grids. Four queries, on Electro and Playground, continue to selective subdivision.}

\newcommand{\ScaleMillionShift}{0.01}

\title{Correspondence-Free Camera Pose Estimation by Matching Moments}
\author{Subhransu S. Bhattacharjee,~Dylan Campbell,~Rahul Shome\\
School of Computing, The Australian National University, Canberra\\
Corresponding Email:~\texttt{Subhransu.Bhattacharjee@anu.edu.au}
}

\begin{document}
\raggedbottom
\twocolumn[{%
\renewcommand\twocolumn[1][]{#1}%
\maketitle
\begin{center}
  \centering
  \captionsetup{type=figure}
  \includegraphics[width=495.56255pt,height=109.60313pt]{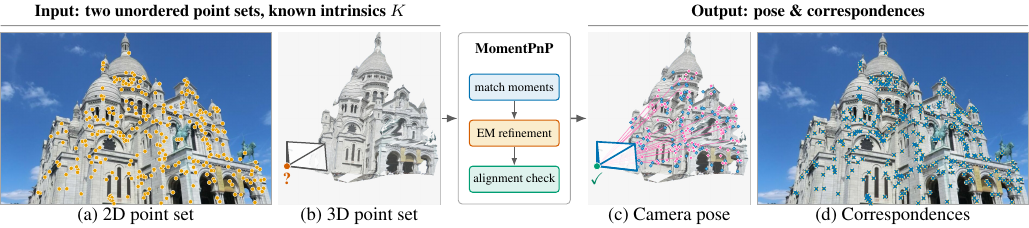}
  \captionof{figure}{\textbf{Blind perspective-$n$-point with \method.} The inputs are (a)~an unordered set of 2D SIFT points~\cite{lowe2004sift} in a \TeaserScene, tracked by its structure-from-motion reconstruction~\cite{schonberger2016sfm}, (b)~the unordered 3D point set of those tracks, and known camera intrinsics~$\Kmat$. The outputs are (c)~the camera pose, with the matched model points and their rays in \textcolor{hotpink}{pink}, and (d)~the correspondences between projected model points (\textcolor{oiBlue}{blue} crosses) and observed 2D points (\textcolor{oiOrange}{orange}).}
  \label{fig:teaser}
\end{center}%
}]
\begin{abstract}
The blind perspective-$n$-point (PnP) problem is to estimate the pose of a calibrated camera from 2D and 3D point sets with unknown correspondences. We show that, for complete, noiseless observations, fourteen permutation-invariant image moments up to order four identify a fixed camera pose for almost every 3D point set containing at least twenty points. Our learning-free method, MomentPnP, matches these moments between the observed 2D points and the projected 3D points to initialize a camera pose without correspondences and refines the pose and correspondence weights using expectation--maximization. Across the test sets of three benchmarks, our method reduces median rotation error by $5$--$170\times$ relative to the best learned baseline, both with and without outliers. We recover accurate poses even with as many outliers as inliers. On dense ETH3D scans, pose accuracy improves as the point sets grow, even as individual correspondences become harder to resolve.
\end{abstract}

\section{Introduction}
\label{sec:intro}

The blind PnP problem is to estimate a calibrated camera's pose from unordered 2D points and 3D model points, without given correspondences~\cite{morenonoguer2008blindpnp} (\cref{fig:teaser}). It enables localization from geometry alone when compatible visual descriptors are deliberately omitted~\cite{zhou2022gomatch} or unavailable, as for laser scans~\cite{campbell2020gopac}. The sets must still share scene structure: inlier 2D points are projections of model points under a common camera pose, while either set may contain unmatched points. Known pairs would give the pose by PnP~\cite{lepetit2009epnp}, but the number of possible pairings grows superpolynomially with the set sizes~\cite{grimson1987localizing}.

Local methods alternate pose estimation and correspondence updates, making their success sensitive to initialization~\cite{david2004softposit,morenonoguer2008blindpnp}. Global methods search pose space by branch and bound~\cite{brown2015globally,campbell2020gopac,campbell2019gosma}, with computational cost driven by the regions that must be examined and excluded. Learned methods predict correspondences from coordinates, using information acquired from training data~\cite{liu2020learning,campbell2020bpnpnet,zhou2022gomatch}. These limitations motivate estimating the pose from the whole point sets before resolving matches.

Because a known 3D model limits how its points can appear in an image, feature averages of the projected model constrain the pose without any pairing of points~\cite{tahri2005pose,quan2025patterns}. Mean and covariance describe position and spread, but distinct projections can share them (\cref{fig:moment_orders}). Higher-order moments, averages of products of point coordinates~\cite{hu1962moments,flusser2016moments}, capture shape beyond the mean and covariance~\cite{ho2009algebraic} and can give the pose of planar point sets. Whether a fixed number of moments determines a perspective camera pose without pairing points was, to our knowledge, open.

\head{Contributions.}
We introduce \method\ (\cref{sec:method}), a learning-free blind PnP solver that estimates the camera pose before the correspondences. It adjusts the six pose parameters until averages of bounded Fourier features over the projected model match those over the observations, after subtracting the expected effect of outliers in both sets~\cite{gribonval2021sketching}. This comparison pairs no points. On complete inputs, it matches fourteen polynomial moments, and we prove that noiseless moments determine the pose under perspective for almost every model of at least twenty points (\cref{thm:pinhole}). With outliers and noise, finitely many Fourier frequencies identify the pose and bound its error by the residual plus sampling and noise terms (\cref{thm:fourier-main}).

\head{Empirical results.}
On released blind PnP benchmarks, \method\ has a 5 to 170 times smaller median rotation error than the best learned baseline, and 82 to 97\% of frames are within $1\degree$ with as many outliers as inliers (\cref{tab:main}). On rendered ETH3D laser scans~\cite{schops2017eth3d}, its median rotation error falls to $0.00007\degree$ at one million points (\cref{fig:scaling}). In full scene reconstructions, it localizes 35 of 39 ETH3D images within $1\degree$ (Tab.~\ref{tab:scene}).

\section{Related work}
\label{sec:related}
\head{Blind PnP.}
Methods differ in how they couple pose and correspondences. Hypothesize-and-test methods such as P3P-RANSAC~\cite{fischler1981ransac} sample triples of candidate pairs, but correct triples become unlikely as the sets grow. Local methods such as SoftPOSIT~\cite{david2004softposit} and BlindPnP~\cite{morenonoguer2008blindpnp} jointly estimate pose and soft correspondences from an initial pose or pose prior, as do recent refinements~\cite{boutiyarzist2025em,song2026dynaweightpnp}. Global methods such as GOPAC~\cite{campbell2020gopac} and GOSMA~\cite{campbell2019gosma} search pose space by branch and bound~\cite{hartley2009global,brown2015globally,wang2022efficient}, at a cost growing with the search domain. Learned methods such as Deep Blind PnP~\cite{liu2020learning} and BPnPNet~\cite{campbell2020bpnpnet} predict correspondences from coordinates and then solve PnP. GoMatch~\cite{zhou2022gomatch} does so for localization against retrieved views, and MinCD-PnP~\cite{an2025mincdpnp} adds color images and surface normals. All rely on correspondences, a pose prior or pose search.

\head{Point-set registration.}
Registering point sets of equal dimension also couples transformations and correspondences. Softassign~\cite{gold1998new}, EM-ICP~\cite{granger2002emicp} and coherent point drift~\cite{clutter} alternate between soft correspondences and a transformation, which we adopt for refinement under perspective projection.
Moment-based methods avoid correspondences by comparing moments. They have been used for planar object pose~\cite{tahri2005pose}, affine registration~\cite{ho2009algebraic} and planar shapes under homographies~\cite{domokos2012nonlinear}. Sketching methods also average sines and cosines over whole datasets~\cite{gribonval2021sketching}. Because these features are bounded, estimators that match their averages remain accurate when a fraction of the data are outliers~\cite{feature}. Correspondence-free pose estimation (CFPE)~\cite{quan2025patterns} comes closest to our setting by matching averages of nonlinear features under perspective projection, but its planar experiments start near the true pose. We estimate a perspective camera pose from feature averages of whole point sets, without a pose prior, with outliers in both sets, before matching points.

\head{Kernel mean embeddings.}
Averages of feature maps represent a distribution as a kernel mean embedding, and maximum mean discrepancy measures the distance between two embeddings~\cite{gretton2012kernel}. For characteristic kernels, equal embeddings imply equal distributions~\cite{sriperumbudur2010hilbert}. Our feature averages are embeddings of this kind, but we need identification only among the projections of one known model. \cref{sec:theory} proves it from fourteen moments on complete inputs and from finitely many Fourier features with outliers.

\head{Pose verification.}
A fitted pose can be wrong in two ways. For planar targets, two poses can give nearly the same image even with known correspondences~\cite{oberkampf1996iterative,schweighofer2006robust}. Without correspondences, a wrong pose can also place projections near many observations by chance. Desolneux et al.~\cite{desolneux2000meaningful} judge a structure by how often it would arise at random, and Moisan and Stival~\cite{moisan2004probabilistic} apply this test to two-view matching. We check alignment against randomly placed projections and compare each accepted pose with its half-turn alternatives (\cref{sec:prelim}).

\section{Problem statement}
\label{sec:problem}
\head{Input \& camera.}
We are given $n$ 2D points (observations) $\dets=\{\vect u_i\}_{i=1}^n\subset\R^2$, $N$ model points $\pts=\{\vect y_j\}_{j=1}^N\subset\R^3$ and the intrinsic matrix $\Kmat$ of the camera, with zero skew and positive focal lengths. No correspondences between the two point sets are given~\cite{morenonoguer2008blindpnp,campbell2020bpnpnet}.
A pose $\xi=(\Rot,\trans)\in\SO\times\R^3$ maps a model point $\vect y$ to camera coordinates $\vect x=\Rot\vect y+\trans$. If its depth $x_3$ is positive, its image $\proj(\vect x)$ is the first two entries of $\Kmat\vect x/x_3$~\cite{hartley2004multiple}. We write $\vect p_j(\xi)=\proj(\Rot\vect y_j+\trans)$.

\head{Correspondences, outliers \& noise.}
The correspondences form an unknown map $\sigma:\{1,\dots,n\}\to\{1,\dots,N\}\cup\{\varnothing\}$. We set $\sigma(i)=j$ when observation $i$ is the image of model point $j$, and $\sigma(i)=\varnothing$ when it is the image of no model point. Distinct observations correspond to distinct model points. The $m$ corresponding pairs are inliers, and the remaining $n-m$ observations and $N-m$ model points are outliers~\cite{campbell2020bpnpnet,zhou2022gomatch}. The number $m$ is not known. Complete inputs have $n=N=m$.
At the true pose $\xi^\star$, every inlier model point has positive depth, and we model each inlier observation as
\begin{equation}
\vect u_i=\vect p_{\sigma(i)}(\xi^\star)+\vect w_i,\qquad \vect w_i\sim\mathcal N(0,\noise^2I_2),
\label{eq:model}
\end{equation}
with i.i.d. noise $\vect w_i$ of unknown standard deviation $\noise$ pixels per coordinate. This noise also approximately absorbs model errors~\cite{hartley2004multiple}.

\head{Task.}
Blind PnP is the task of estimating, from $\dets$, $\pts$ and $\Kmat$, the true pose $\xi^\star$ and the correspondences $\sigma$ of \cref{eq:model}.

\section{\method}
\label{sec:method}
In blind PnP (\cref{sec:problem}), \method\ estimates the camera pose from observations, a 3D model and the intrinsics, without correspondences. In general, both point sets contain outliers. Complete inputs, without outliers, are an exception that \method\ treats separately when this is known.

\subsection{Preliminaries}
\label{sec:prelim}
\begin{figure}[t]
\centering
\input{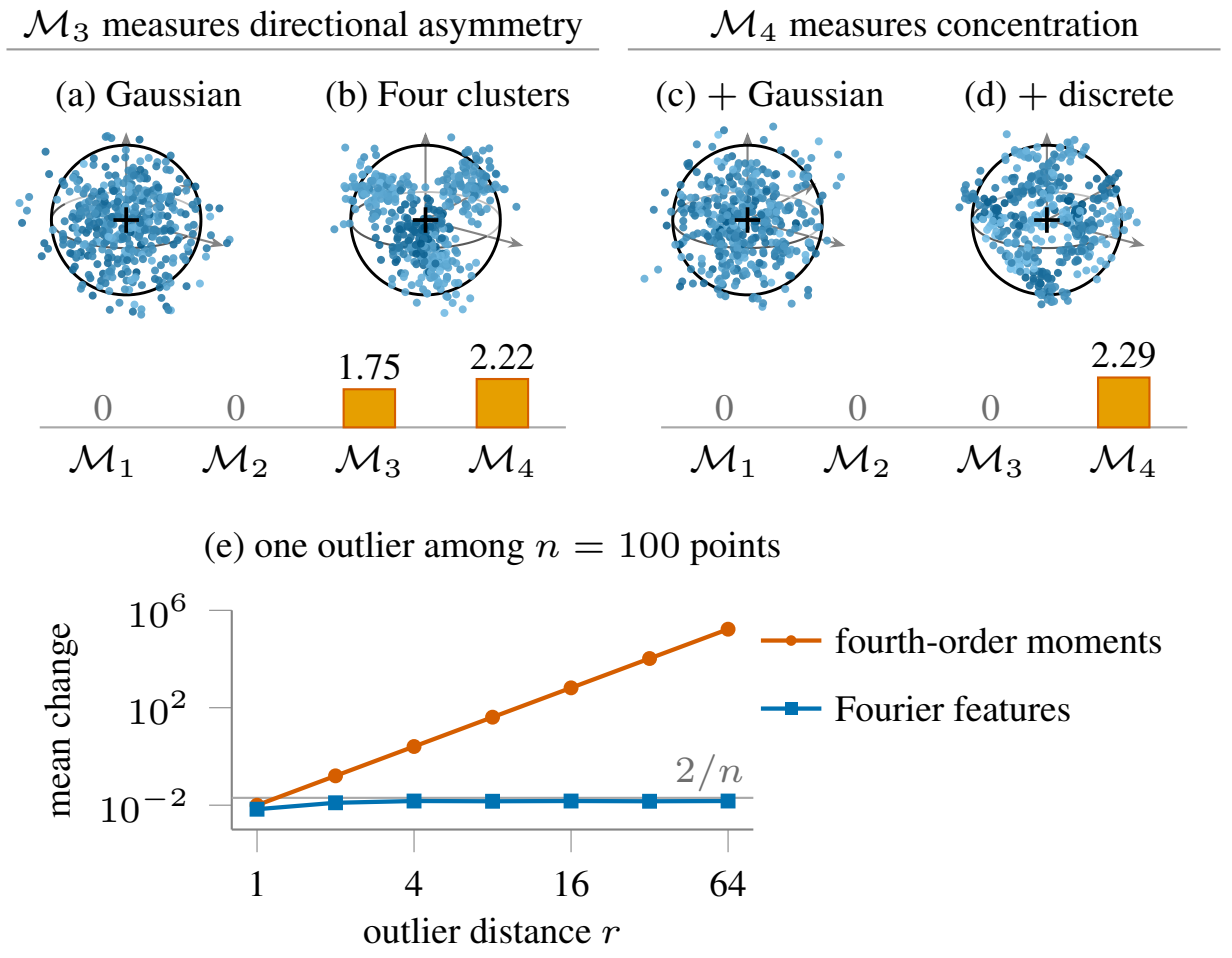}
\caption{\textbf{Higher-order moments and Fourier features.} Pairs (a, b) and (c, d) share centroid and covariance, and bars compare their moments $\mom_\order$. Third-order moments separate (a) and (b), fourth-order moments the noisy versions (c, d). (e)~An outlier at distance $r$ among $n=100$ points changes fourth-order moments in proportion to $r^4$ and Fourier feature averages by at most $2/n$.}
\label{fig:moment_orders}
\end{figure}

\head{Polynomial Moments.}
Moments are statistical averages of products of point coordinates, and their order $\order$ is the number of factors~\cite{hu1962moments,flusser2016moments}. Averages need no pairing of points, so earlier correspondence-free methods use moments~\cite{tahri2005pose,ho2009algebraic,domokos2012nonlinear}, and we use orders one to four, $\mom_\order$. Orders three and four capture asymmetry and concentration, which separate shapes with equal centroid and covariance (\cref{fig:moment_orders}a--d).

For an observation $\vect u$ in pixels, the calibrated point $\vect u_{\mathrm{cal}}$ consists of the first two coordinates of $\Kmat^{-1}(\vect u^\top,1)^\top$, where $\Kmat$ is the intrinsic matrix~\cite{hartley2004multiple}. The affine map $h(\vect u)=\mbox{$(\vect u_{\mathrm{cal}}-\bar{\vect u}_{\mathrm{cal}})/r_{\rm cal}$}$ centers calibrated points at the mean $\bar{\vect u}_{\mathrm{cal}}$ of the calibrated observations and divides by their root-mean-square radius $r_{\rm cal}$, which keeps the coordinates well conditioned~\cite{hartley1997defense}. We write $\vect z=h(\vect u)$ for the normalized point, and the same map normalizes the projection $\vect p_j(\xi)$ of each model point at a pose $\xi$ (\cref{sec:problem}).

Stacking the fourteen monomials of orders one to four of a normalized point $\vect z$ gives the feature vector $\vect\phi_4(\vect z)=(z_1,z_2,z_1^2,z_1z_2,z_2^2,\dots,z_2^4)\in\R^{14}$. Its averages $\vect\mu_{\dets}=\frac1n\sum_i\vect\phi_4(h(\vect u_i))$ over the observations and $\vect\mu_{\pts}(\xi)=\frac1N\sum_j\vect\phi_4(h(\vect p_j(\xi)))$ over the projections are the moments of both point sets, and on complete inputs they determine the pose (\cref{thm:pinhole}).

\head{Trigonometric moments.}
Monomial features fail with outliers, because one distant outlier can dominate the moments (\cref{fig:moment_orders}e). Like sketching methods~\cite{gribonval2021sketching}, \method\ therefore uses weighted cosines and sines $\vect\phi_F(\vect z)=(\cos\vect k^\top\vect z,\sin\vect k^\top\vect z)_{\vect k\in\mathrm F}$ over a finite set $\mathrm F$ of frequencies.\footnote{The set $\mathrm F$ contains the nonzero frequencies with coordinates in multiples of $\pi/2$ and length at most $\frac\pi2K_F$, one of each pair $\pm\vect k$. Here $\vect z=h_F(\vect u)$ shifts and scales the calibrated point so that the rectangle bounding the calibrated observations fits in $[-1,1]^2$.} The Fourier feature averages $\vect\mu_{\dets,F}$ and $\vect\mu_{\pts,F}(\xi)$ are defined like $\vect\mu_{\dets}$ and $\vect\mu_{\pts}(\xi)$, with $\vect\phi_F$ and $h_F$ in place of $\vect\phi_4$ and $h$. They are the trigonometric moments~\cite{mardia1999directional} of the point distribution. Because the features are bounded, one outlier has a bounded effect, and low frequencies describe the coarse layout as low orders do~\cite{huber2009robust,katznelson2004harmonic}.

\begin{figure*}[t]
  \centering
  \input{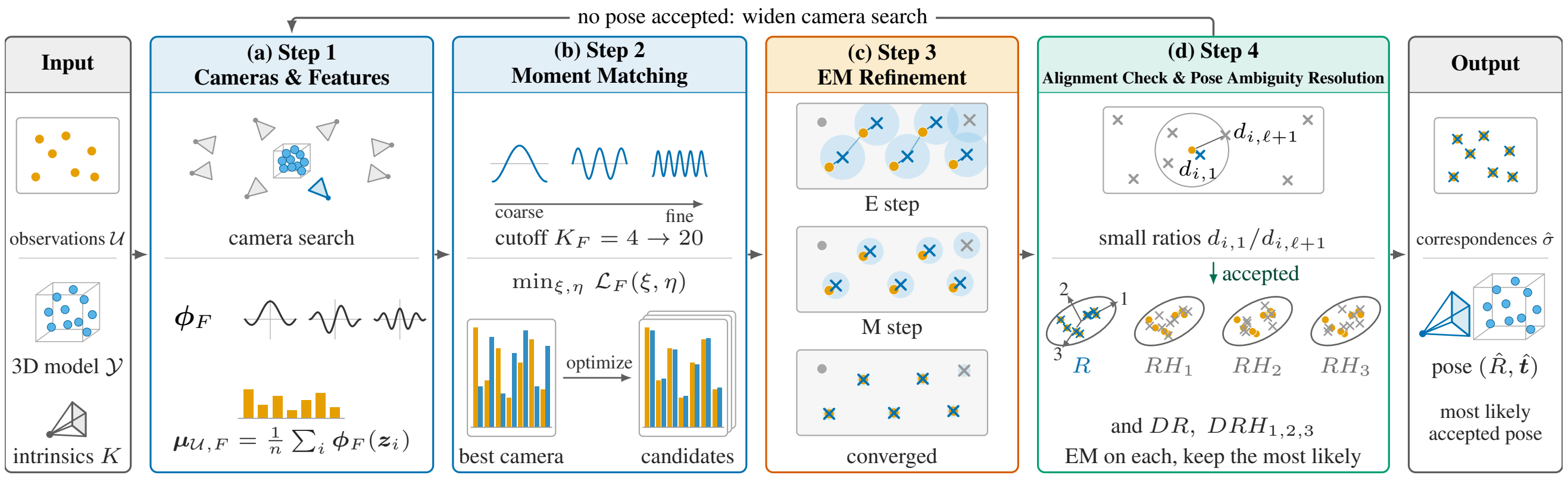}
  \caption{\textbf{Overview of \method.} (a)~Cameras are searched over centers and orientations, and observed features averaged once. (b)~Moment matching refines the best cameras, raising the cutoff from 4 to 20. (c)~EM refines fits with soft correspondences. (d)~The alignment check accepts poses whose observations lie much closer to their nearest projections than to the next, and the most likely accepted pose or alternative is returned. \textcolor{oiOrange}{Orange}: observations. \textcolor{oiBlue}{Blue}: model points and projections. \textcolor{oiGrey}{Gray}: unmatched points, unselected poses.}
  \label{fig:method}
\end{figure*}

As in the benchmark protocol~\cite{campbell2020bpnpnet}, we consider image outliers are spread uniformly over the rectangle that bounds the observations, and model outliers through the box that bounds the 3D model. On average, an image outlier adds $\vect\mu_{\dets,0}$, the average of $\vect\phi_F$ over the rectangle, and a model outlier adds $\vect\mu_{\pts,0}(\xi)$, the average of $\vect\phi_F$ over the projections of points spread through the box. With a fraction $\unmatched$ of outliers in each point set, $\vect\mu_{\dets,F}$ and $\vect\mu_{\pts,F}(\xi)$ mix the inlier averages with $\vect\mu_{\dets,0}$ and $\vect\mu_{\pts,0}(\xi)$. The inlier averages agree at the true pose, so the expected difference there is $\unmatched(\vect\mu_{\dets,0}-\vect\mu_{\pts,0}(\xi))$.

\begin{definition}[Moment matching]
\label{def:moments}
Moment matching minimizes
\begin{equation}
\label{eq:moment-matching}
\mathcal L_F(\xi,\unmatched)=\norm{\vect\mu_{\dets,F}-\vect\mu_{\pts,F}(\xi)-\unmatched\big(\vect\mu_{\dets,0}-\vect\mu_{\pts,0}(\xi)\big)}^2
\end{equation}
over the pose $\xi$ and $\unmatched$. The objective is quadratic in the unknown $\unmatched$, so the best $\unmatched$ at each pose has a closed form. On complete inputs, $\unmatched=0$ and no outlier can dominate the moments, so they replace the Fourier feature averages, and we minimize $\mathcal L_4(\xi)=\mbox{$\norm{\vect\mu_{\dets}-\vect\mu_{\pts}(\xi)}^2$}$ (\cref{thm:pinhole}).
\end{definition}

\head{Expectation--maximization (EM).}
EM~\cite{dempster1977maximum,granger2002emicp,clutter} then refines the pose with every observation, each explained as a uniform outlier over the rectangle or a noisy projection of a model point in front of the camera (\cref{fig:method}c). A refined pose is correct only if the observations lie much closer to their nearest projections than to the next, which we test against randomly placed projections~\cite{desolneux2000meaningful,moisan2004probabilistic}.

\begin{definition}[Association level]
\label{def:association}
For observation $i$, let $d_{i,1}\le d_{i,2}\le\cdots$ be the distances to the projections in increasing order. For projections placed as a homogeneous Poisson process of any density, each $\assoc_i=-\nnb\log(1-d_{i,1}^2/d_{i,\nnb+1}^2)$, which compares the nearest distance with the local spacing over $\nnb\le8$ neighbors, would be unit exponential~\cite{chiu2013stochastic}. Observation $i$ is associated when $\assoc_i\le\pi/16$, that is, when its nearest projection lies within about a quarter of the average spacing. The association level is the probability under random placement, with independent observations, of at least the observed number of associated observations.
\end{definition}

\subsection{Camera pose estimation}
\label{sec:estimation}
\method\ estimates the pose by matching moments and then verifies it (\cref{fig:method}, \cref{alg:main}). Step 2 adjusts the six pose parameters until the feature averages of the projected model agree with those of the observations, after subtracting the expected effect of outliers in both sets, which pairs no points. Because the objective has many local minima~\cite{nocedal2006numerical}, Step 1 supplies starting poses. Because averages ignore the individual observations, Step 3 refines the matched pose until it explains each observation on its own. Because a wrong pose can still align with the observations by chance, or through an ambiguity that averages cannot resolve, Step 4 checks the refined pose before it is returned.

\begin{algorithm}[!t]
\footnotesize
\SetAlCapFnt{\footnotesize}
\SetAlCapNameFnt{\footnotesize}
\LinesNumbered
\caption{\method\ (Steps 1--4; expanded in Alg.~\ref{alg:momentpnp})}
\label{alg:main}
\KwIn{Observations $\dets$, 3D model $\pts$, intrinsics $\Kmat$,
      angular grid spacing $\epsilon>0$}
\KwOut{Pose $\hat\xi$ and correspondences $\hat\sigma$}
$(\vect\mu_{\dets,F},\vect\mu_{\dets,0})\gets\mathtt{ObservedMoments}(\dets,\Kmat)$\;
$G\gets\varnothing ; \Pi\gets\varnothing$\;
\Repeat{$a$ or $G$ is at its finest level}{
    $G\gets\mathtt{WidenCameraSearch}(G,\pts,\epsilon)$\;
    $\Pi'\gets\mathtt{CameraInitialization}(G,\dets)$\;
    $\Pi'\gets\mathtt{MomentMatching}(\Pi',\vect\mu_{\dets,F},\vect\mu_{\dets,0})$\;
    $\Pi\gets\Pi\cup\mathtt{EMRefinement}(\Pi',\dets)$\;
    $(\hat\xi,a)\gets\mathtt{AlignmentCheck}(\Pi)$\tcp*{$a$: $\hat\xi$ accepted}
}
\If{$a$}{
    $\Pi'\gets\mathtt{EMRefinement}(\mathtt{Alternatives}(\hat\xi),\dets)$\;
    $(\hat\xi,a)\gets\mathtt{AlignmentCheck}(\{\hat\xi\}\cup\Pi')$\;
}
$\hat\sigma\gets\mathtt{Correspondences}(\hat\xi,\dets)$\;
\Return $(\hat\xi,\hat\sigma)$\;
\end{algorithm}

\head{Step 1: Camera initialization.}
Without a pose prior, the image may show any part of the model, so \method\ first places candidate cameras throughout it. The candidates combine camera centers on a grid over an enlarged bounding box of the model with viewing directions over the sphere and roll angles about each. We score each candidate, as in GOSMA~\cite{campbell2019gosma}, by how much more densely its projections cluster at the observations than around them, measured by the ratio of a narrow to a broad smoothing of the projections, and refine the best candidates by this score.

\head{Step 2: Moment matching.}
Step 2 fits the pose by minimizing the moment-matching objective of \cref{def:moments}. For a candidate camera $\xi$, a taper fades out projections outside the rectangle of the observations, so only the model points that project near the observations contribute to $\vect\mu_{\pts,F}(\xi)$, and the average changes smoothly as points enter or leave the view~\cite{harris1978windows}. Starting from the candidates with distinct orientations and the lowest $\mathcal L_F$, Levenberg--Marquardt~\cite{levenberg1944method,marquardt1963algorithm} minimizes $\mathcal L_F$ over the six pose parameters under perspective projection. The fit proceeds from coarse to fine. The cutoff $K_F$ rises from 4 to 20 in five stages, and after each stage only the fits of lowest $\mathcal L_F$ with distinct orientations continue. The best fits go to EM (\cref{fig:method}b).

\head{Step 3: Expectation--maximization refinement.}
For refinement we adapt EM-ICP and coherent point drift~\cite{granger2002emicp,clutter} to perspective projection. EM refines each fit with the mixture of uniform outliers and noisy projections of \cref{sec:prelim}, and estimates the noise standard deviation and the outlier probability together with the pose (\cref{fig:method}c). The refined poses are then ranked by the likelihood of the observations as a Poisson process of uniform outliers and Gaussian peaks at the projections~\cite{chiu2013stochastic}. A refined pose can still be wrong in two ways. It can place projections near many observations by chance, or it can be one of the poses that give nearly the same image as the truth. Step 4 tests for both.

\head{Step 4: Alignment check and pose ambiguity resolution.}\label{par:symmetric}
An observation is matched when its nearest projection explains it better than the outlier term does. The alignment check accepts a refined pose when enough observations are matched, their median ratio $d_{i,1}/d_{i,2}$ is small, and the association level of \cref{def:association} is very small. Acceptance does not settle the pose of a nearly symmetric or shallow model, because other poses give nearly the same image and the averages cannot separate them (\cref{sec:theory}). For a distant camera~\cite{dementhon1995model}, the poses that give the image the same mean and covariance as an accepted pose with rotation $\Rot$ are its seven alternatives $\Rot\halfturn_k$ and $\opthalf\Rot\halfturn_k$ in \cref{fig:method}d, where $\halfturn_1,\halfturn_2,\halfturn_3$ turn the model by $180\degree$ about the principal axes of its matched points, $\halfturn_0$ is the identity, and $\opthalf$ turns the camera by $180\degree$ about the optical axis. That these seven are all the poses the mean and covariance cannot separate is proved in \cref{sec:supp_proof_companions}.\footnote{When the principal variances of the model differ, these alternatives include every pose that a symmetry of the model produces, and for a planar model they include the opposite tilt~\cite{oberkampf1996iterative,schweighofer2006robust}.} Which of them EM then prefers is not proved, but on the benchmarks 63 of the 64 poses that an alternative replaced end within $1\degree$ (\cref{sec:supp_ablation}). \method\ therefore refines the most likely accepted pose and its seven alternatives by EM under perspective, checks each, and returns the most likely of them. Without an accepted pose, the search widens and then returns the most likely refined pose. At the returned pose, each matched observation corresponds to its nearest projection, and every other observation is an outlier, $\hat\sigma(i)=\varnothing$. When several observations share a nearest projection, only the closest keeps it, so distinct observations correspond to distinct model points. The refit of an accepted pose to its correspondences is given in \cref{sec:supp_config}.
\subsection{Uniqueness and accuracy of the pose}
\label{sec:theory}
\begin{figure*}[!t]
\centering
\includegraphics[width=494.77419pt,height=231.32867pt]{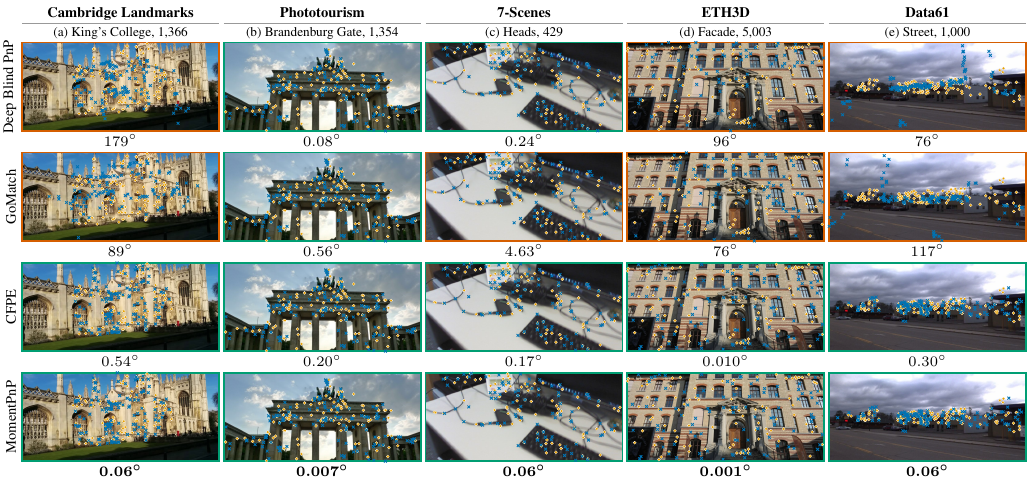}
\caption{\textbf{Qualitative results on real-world scenes.} Columns are images, titled by scene and number of observations, and rows are methods. Orange points are observations, and blue crosses are projections under the recovered pose. Below each panel is its rotation error, in bold where lowest in the column, and a green border marks an error below $1\degree$. Model points come from the released structure-from-motion reconstructions of Cambridge Landmarks~\cite{kendall2015posenet}, Phototourism~\cite{jin2021imc}, 7-Scenes~\cite{sarlin2021pixloc} and ETH3D~\cite{schops2017eth3d}, and from LiDAR for Data61~\cite{namin2015data61}.}
\label{fig:qualitative}
\end{figure*}

Moment matching recovers the pose only if two properties hold. On ideal complete inputs, its objective vanishes at the true pose alone (uniqueness). With noise and outliers, its minimizer stays within a bounded distance of the true pose (accuracy). \cref{thm:pinhole} proves uniqueness for complete inputs. \mbox{\cref{thm:fourier-main}} builds on it to prove both for $\mathcal L_F$.

\head{Uniqueness for complete inputs.}
On complete noiseless inputs, $\mathcal L_4$ vanishes at the true pose and, for a symmetric model such as a cube, at a second pose. Such models are exceptional. Equal moments at a second pose impose polynomial equations on the model points, and a dimension count shows that almost no model of $N\ge20$ points satisfies them.

\begin{theorem}[Pose uniqueness for complete inputs]
\label{thm:pinhole}
Fix the true pose $\xi^\star$ of a calibrated camera. For almost every model of $N\ge20$ points in front of the camera, every pose that keeps the points in front of the camera and gives their projections the same fourteen moments as $\xi^\star$ equals $\xi^\star$.
\end{theorem}
\begin{figure}[t]
\centering
\pgfplotsset{curveaxis/.style={mpPlot, title style={font=\footnotesize, yshift=-1.2ex},
  xlabel near ticks, ylabel near ticks, ylabel shift=-3pt, xlabel shift=-2pt,
  ymin=-4, ymax=104, ytick={0,25,50,75,100}, unbounded coords=jump, axis background/.style={fill=white}},
  cOurs/.style={colOurs, line width=0.9pt, mark=star, mark size=1.9pt}, cLiu/.style={colLiu, line width=0.65pt, mark=square*, mark size=1.1pt}, cBPnP/.style={colBPnP, line width=0.65pt, mark=triangle*, mark size=1.3pt},
  cCFPE/.style={oiOrange, line width=0.65pt, mark=*, mark size=1.3pt},
  cEM/.style={black!80, line width=0.65pt, mark=pentagon*, mark size=1.3pt},
  cGoM/.style={oiPurple, line width=0.65pt, mark=diamond*, mark size=1.3pt}}
{\footnotesize\begin{tabular}{@{}lll@{}}
\tikz[baseline=-0.6ex]{\draw[colOurs, line width=0.9pt] (0,0) -- (0.32,0); \draw[colOurs] plot[mark=star, mark size=1.9pt] coordinates {(0.16,0)};}~\method & \tikz[baseline=-0.6ex]{\draw[colLiu, line width=0.65pt] (0,0) -- (0.32,0); \draw[colLiu] plot[mark=square*, mark size=1.1pt] coordinates {(0.16,0)};}~Deep Blind PnP & \tikz[baseline=-0.6ex]{\draw[colBPnP, line width=0.65pt] (0,0) -- (0.32,0); \draw[colBPnP] plot[mark=triangle*, mark size=1.3pt] coordinates {(0.16,0)};}~BPnPNet \\
\tikz[baseline=-0.6ex]{\draw[oiOrange, line width=0.65pt] (0,0) -- (0.32,0); \draw[oiOrange] plot[mark=*, mark size=1.3pt] coordinates {(0.16,0)};}~CFPE (reproduced) & \tikz[baseline=-0.6ex]{\draw[oiPurple, line width=0.65pt] (0,0) -- (0.32,0); \draw[oiPurple] plot[mark=diamond*, mark size=1.3pt] coordinates {(0.16,0)};}~GoMatch & \tikz[baseline=-0.6ex]{\draw[black!80, line width=0.65pt] (0,0) -- (0.32,0); \draw[black!80] plot[mark=pentagon*, mark size=1.3pt] coordinates {(0.16,0)};}~EM-ICP
\end{tabular}}\\[2pt]
\includegraphics[width=\columnwidth]{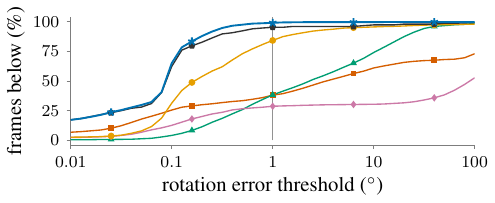}
\caption{\textbf{Rotation-error distribution without outliers.} Each curve gives the percentage of frames below each rotation error, averaged with equal weight over the MegaDepth, ModelNet40 and Data61 test sets. The vertical line marks $1\degree$.}
\label{fig:curves}
\end{figure}

Only the true pose therefore has $\mathcal L_4=0$, which justifies minimizing $\mathcal L_4$.\footnote{The twenty-point bound serves the proof and is not an operational limit. \method\ recovers poses within $1\degree$ from 8 and 16 noiseless points.}

\head{Accuracy of the Fourier fit.}
On ideal inputs, zero $\mathcal L_F$ over all frequencies determines the inlier projections and the outlier fraction, since the cosine and sine averages uniquely determine a distribution~\cite{katznelson2004harmonic}. The pose is then determined by \cref{thm:pinhole}. Observation noise and finite sampling of the outliers generally make the loss nonzero even at the true pose. \cref{thm:fourier-main} bounds the error at a finite cutoff: for every cutoff $K_F$ above a threshold, the loss and the perturbations bound the distance $\operatorname{dist}$ between any candidate $(\xi,\unmatched)$ and the truth $(\xi^\star,\unmatched^\star)$. This distance combines the rotation, translation and image-outlier fraction errors.

\begin{theorem}[Accuracy of the Fourier fit]
\label{thm:fourier-main}
Let the outliers be drawn independently and uniformly from the rectangle of the observations and the bounding box of the model, and let a compact set $\Theta$ of pairs $(\xi,\unmatched)$ contain $(\xi^\star,\unmatched^\star)$, with every pose keeping the box in front of the camera and $\unmatched\in[0,1)$. For almost every model of $m\ge20$ inlier points and every cutoff $K_F$ above a threshold $K_0$, there is a constant $\fourconst$ such that, for each $\delta\in(0,1)$, with probability at least $1-\delta$ over the outliers, every $(\xi,\unmatched)\in\Theta$ satisfies
\[
\operatorname{dist}\le\fourconst\big(\sqrt{\mathcal L_F(\xi,\unmatched)}+\fourerr\big).
\]
The error term $\fourerr$ adds the sampling and grid errors of the outlier averages, the noise error and twice the difference of the image and model outlier fractions.
\end{theorem}

At the true pair, $\sqrt{\mathcal L_F}$ is at most $\fourerr$, so the minimizing pair lies within $2\fourconst\fourerr$ of the truth, and $\mathcal L_F$ can select fits for EM. The constant $\fourconst$ is large when $\mathcal L_F$ grows slowly away from the truth on ideal inputs. The error term measures the departure from this ideal, and only its sampling part, which shrinks as the point sets grow, depends on $\delta$.

For nearly symmetric and distant planar models, the averages barely change between a pose and its alternatives in \cref{par:symmetric}, and $\fourconst$ is large. Step 4 therefore compares these alternatives by EM on the individual observations. \cref{sec:experiments} tests \method\ in practice, and the appendix tests it beyond the theorems' assumptions, with structured outliers (Fig.~\ref{fig:supp_degradation}), partial views (Tab.~\ref{tab:render}) and large depth spreads (Tab.~\ref{tab:depth}); proofs and further conditions in \cref{sec:supp_proof_pinhole,sec:supp_fourier}, further details in \cref{sec:supp_small_sets,sec:supp_ablation}.

\section{Experiments}
\label{sec:experiments}
\begin{table*}[t]
\centering\scriptsize
\setlength{\tabcolsep}{2.2pt}
\renewcommand{\arraystretch}{1.08}
\caption{\textbf{Blind PnP on released MegaDepth, ModelNet40 and Data61 test sets.}
On each test set, we sample 1{,}000 outlier-free frames and 300 frames with outliers. Searches use a 30-second limit per frame. \method\ treats inputs as complete only when they are known to be outlier-free.
Err.: median rotation error ($\degree$)\,/\,translation error (\%) over returned poses. Acc.: accuracy (\%). Time: range of the per-dataset median times without outliers (s). EM-ICP (ours): our EM refinement (Step 3) from random poses, without moment matching. Bold: best per column.}
\label{tab:main}
\begin{tabular*}{\textwidth}{@{\extracolsep{\fill}}l rrrr rrrr rrrr r@{}}
\toprule
 & \multicolumn{4}{c}{MegaDepth~\cite{li2018megadepth}} & \multicolumn{4}{c}{ModelNet40~\cite{wu2015shapenets}} & \multicolumn{4}{c}{Data61~\cite{namin2015data61}} & \\
\cmidrule(lr){2-5}\cmidrule(lr){6-9}\cmidrule(lr){10-13}
 & \multicolumn{2}{c}{no outliers} & \multicolumn{2}{c}{with outliers} & \multicolumn{2}{c}{no outliers} & \multicolumn{2}{c}{with outliers} & \multicolumn{2}{c}{no outliers} & \multicolumn{2}{c}{with outliers} & \\
\cmidrule(lr){2-3}\cmidrule(lr){4-5}\cmidrule(lr){6-7}\cmidrule(lr){8-9}\cmidrule(lr){10-11}\cmidrule(lr){12-13}
Method & Err.\,$\downarrow$ & Acc.\,$\uparrow$ & Err.\,$\downarrow$ & Acc.\,$\uparrow$ & Err.\,$\downarrow$ & Acc.\,$\uparrow$ & Err.\,$\downarrow$ & Acc.\,$\uparrow$ & Err.\,$\downarrow$ & Acc.\,$\uparrow$ & Err.\,$\downarrow$ & Acc.\,$\uparrow$ & Time (s)\,$\downarrow$ \\
\midrule
\multicolumn{14}{@{}l}{\emph{Learning-free baselines}} \\
\quad P3P-RANSAC~\cite{fischler1981ransac} & 131\,/\,8{,}482 & 0.0 & 127\,/\,7{,}091 & 0.0 & 138\,/\,161 & 0.0 & 124\,/\,123 & 0.0 & 132\,/\,5{,}536 & 0.0 & 137\,/\,4{,}734 & 0.0 & 30 \\
\quad SoftPOSIT~\cite{david2004softposit} & 0.063\,/\,0.31 & 86.0 & 9.47\,/\,36 & 34.0 & 0.27\,/\,0.26 & 80.0 & 30\,/\,14 & 0.0 & 22\,/\,67 & 41.0 & 7.63\,/\,29 & 37.0 & 1.0--30 \\
\quad GOSMA~\cite{campbell2019gosma} & 135\,/\,342 & 0.0 & 78\,/\,297 & 0.0 & 49\,/\,75 & 0.0 & 43\,/\,56 & 0.0 & 121\,/\,528 & 0.0 & 128\,/\,494 & 0.0 & 0.25--2.9 \\
\quad EM-ICP (ours)~\cite{granger2002emicp} & \textbf{0.009}\,/\,\textbf{0.011} & 99.0 & 34\,/\,130 & 31.0 & \textbf{0.16}\,/\,0.12 & 94.0 & 28\,/\,3.74 & 6.0 & \textbf{0.092}\,/\,0.20 & 93.0 & 148\,/\,415 & 0.0 & 30 \\
\quad CFPE (reproduced)~\cite{quan2025patterns} & 0.123\,/\,0.34 & 91.5 & 93\,/\,223 & 0.0 & 0.674\,/\,0.15 & 64.4 & 129\,/\,10 & 0.0 & 0.100\,/\,0.22 & 96.8 & 102\,/\,199 & 0.0 & 0.47--2.02 \\
\midrule
\multicolumn{14}{@{}l}{\emph{Learned baselines}} \\
\quad Deep Blind PnP~\cite{liu2020learning} & 0.045\,/\,0.16 & 96.8 & 0.43\,/\,2.08 & 59.0 & 3.35\,/\,0.83 & 15.6 & 27\,/\,11 & 0.0 & 130\,/\,347 & 1.8 & 138\,/\,428 & 0.0 & 0.24--0.33 \\
\quad BPnPNet~\cite{campbell2020bpnpnet} & 0.85\,/\,5.40 & 53.4 & 14\,/\,74 & 1.7 & 7.50\,/\,1.91 & 6.0 & 29\,/\,11 & 0.0 & 0.76\,/\,2.90 & 55.4 & 16\,/\,71 & 0.7 & 0.16--0.22 \\
\quad GoMatch~\cite{zhou2022gomatch} & 0.14\,/\,0.59 & 86.0 & 0.33\,/\,1.68 & 62.3 & 100\,/\,16 & 0.1 & 70\,/\,14 & 0.0 & 143\,/\,268 & 0.0 & 148\,/\,318 & 0.0 & \textbf{0.02--0.04} \\
\midrule
\method (ours) & \textbf{0.009}\,/\,0.013 & \textbf{100.0} & \textbf{0.022}\,/\,\textbf{0.051} & \textbf{91.0} & \textbf{0.16}\,/\,\textbf{0.096} & \textbf{97.3} & \textbf{0.34}\,/\,\textbf{0.20} & \textbf{82.3} & \textbf{0.092}\,/\,\textbf{0.18} & \textbf{100.0} & \textbf{0.093}\,/\,\textbf{0.22} & \textbf{97.0} & 0.03--0.06 \\
\bottomrule
\end{tabular*}
\end{table*}

\begin{figure}[!t]
\centering
\includegraphics[width=234.71445pt,height=100.95811pt]{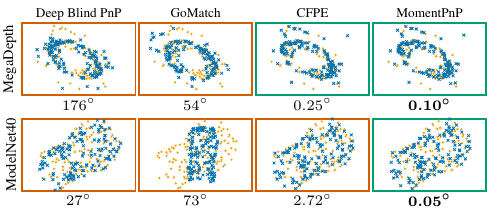}
\caption{\textbf{Qualitative results on the point-only benchmarks.} MegaDepth and ModelNet40 provide coordinates without photographs. Markers and labels follow \cref{fig:qualitative}.}
\label{fig:points_main}
\end{figure}

\begin{table}[!tp]
\centering\footnotesize
\setlength{\tabcolsep}{1.5pt}
\caption{\textbf{Blind PnP on structure-from-motion (SfM) inputs.}
Accuracy (\%) on the Cambridge Landmarks (C), Phototourism (P), 7-Scenes (S, depth-assisted SfM) and ETH3D (E) datasets (errors and times in Tab.~\ref{tab:real_full}). Bold: highest, including ties.}
\label{tab:real}
\begin{tabular*}{\columnwidth}{@{\extracolsep{\fill}}lrrrrrrrrrr@{}}
\toprule
 & \multicolumn{5}{c}{No outliers} & \multicolumn{5}{c}{With outliers} \\
\cmidrule(lr){2-6}\cmidrule(lr){7-11}
Method & C & P & S & E & All & C & P & S & E & All \\
\midrule
Deep Blind PnP & 33 & 92 & 69 & 20 & 48 & 4 & 69 & 50 & 13 & 33 \\
BPnPNet & 0 & 19 & 36 & 0 & 12 & 0 & 0 & 3 & 0 & 1 \\
GoMatch & 15 & 78 & 17 & 1 & 23 & 8 & 62 & 6 & 3 & 17 \\
CFPE (reproduced) & 81 & 92 & \textbf{100} & \textbf{100} & 95 & 0 & 0 & 0 & 0 & 0 \\
\method (ours) & \textbf{100} & \textbf{100} & \textbf{100} & \textbf{100} & \textbf{100} & \textbf{100} & \textbf{96} & \textbf{100} & \textbf{87} & \textbf{95} \\
\midrule
Images & 27 & 36 & 36 & 70 & 169 & 26 & 26 & 36 & 38 & 126 \\
\bottomrule
\end{tabular*}
\end{table}

\head{Metrics.}
We report the median rotation error, the angle in degrees between estimated and true rotations~\cite{campbell2020bpnpnet}, and the median relative translation error, the distance between the estimated and true translations divided by the true translation length~\cite{lepetit2009epnp}. Accuracy is the percentage of frames with rotation error below $1\degree$, counting missing estimates as failures.

\subsection{Blind PnP benchmarks}
\label{sec:exp_benchmarks}
\head{Benchmarks.}
We compare learned and learning-free blind PnP methods, including our reproduction of CFPE~\cite{quan2025patterns}, on the test sets of the released MegaDepth~\cite{li2018megadepth}, ModelNet40~\cite{wu2015shapenets} and Data61~\cite{namin2015data61} benchmarks. EM-ICP (ours) is our own Step 3, the EM refinement of \cref{sec:estimation}, run from random poses without Steps 1 and 2 and restarted until the time limit, so \cref{tab:main} also shows what refinement alone achieves. In the benchmark protocol~\cite{campbell2020bpnpnet}, inliers are the released 2D--3D pairs, and outliers are drawn uniformly from the bounding box of each point set. We tuned the method on the test sets, so we also test on 1{,}000 held-out frames per dataset from the training split.
\cref{tab:main} shows that no method has a lower median rotation error than \method\ on any test set. Compared with the best learned baseline, \method's median rotation error is 5 to 170 times smaller, and its median translation error is 9 to 320 times smaller. On the sampled test frames of \cref{tab:main}, its accuracy is 97.3 to 100\% without outliers and 82.3 to 97.0\% with as many outliers as inliers. Averaged over the three test sets without outliers, 99.1\% of frames are within $1\degree$, against 38.3\% for the best learned baseline. The curve of \method\ lies above those of all learned baselines at every threshold (\cref{fig:curves}, Fig.~\ref{fig:supp_curves}). On the held-out frames, its accuracy is 98.6 to 100\% without outliers and 72.4 to 91.4\% with as many outliers as inliers (Tab.~\ref{tab:supp_details}).

\head{Further datasets.}
We also evaluate on Cambridge Landmarks~\cite{kendall2015posenet}, Phototourism~\cite{snavely2006phototourism,jin2021imc}, 7-Scenes~\cite{shotton2013scene,sarlin2021pixloc} and ETH3D~\cite{schops2017eth3d}. The tracked 2D points of each image and the model points of their tracks form complete inputs (\cref{tab:real}). Learned methods use the MegaDepth benchmark weights, and \method\ uses the same settings as on the earlier benchmarks. All 169 \method\ rotation errors are below $1\degree$, compared with 95\% for reproduced CFPE and at most 48\% for learned methods. With half as many uniform outliers as inliers in each point set, \method\ reaches 95\% accuracy on the 126 images with at most 2{,}000 inliers (Tab.~\ref{tab:real_full}). The best learned baseline reaches 33\%. On all 169 images, without the inlier limit, \method\ reaches 87\% and the best learned baseline 29\% (Tab.~\ref{tab:real_uncapped}). \cref{fig:qualitative} shows real scenes (Fig.~\ref{fig:gallery}) and \cref{fig:points_main} frames of the point-only benchmarks.

\begin{figure}[!t]
\centering
\captionsetup{skip=4pt}
\includegraphics[width=237.54294pt,height=75.38083pt]{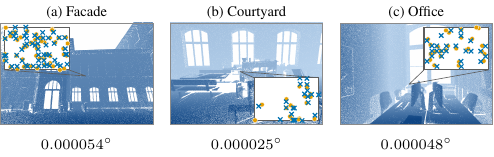}
\caption{\textbf{Recovered poses at one million points.} ETH3D laser-scan model points~\cite{schops2017eth3d} of three scans, shaded by depth and projected under the poses estimated from 1{,}000{,}000 2D points. Labels give rotation error. Each inset compares observations (orange) with projections under the recovered pose (blue) in a 30-pixel window.
}
\label{fig:scale_gallery}
\end{figure}

\begin{figure}[!t]
\centering
\includegraphics[width=237.04883pt,height=99.74976pt]{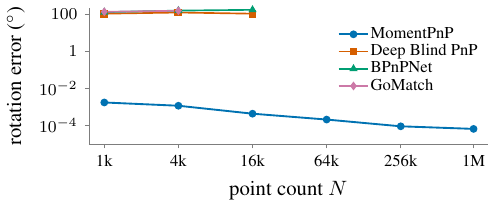}
\caption{\textbf{Rotation error against point count.} Medians over the views of three ETH3D scans~\cite{schops2017eth3d}.}
\label{fig:scaling}
\end{figure}

\subsection{Large point sets and full scenes}
\label{sec:exp_large}
\head{Scaling to large point sets.}
Beyond the benchmark sizes of up to about 12{,}000 points, more points reduce the random part of the noise in the moments but crowd the projections that each correspondence must separate. We test both effects on three ETH3D laser scans~\cite{schops2017eth3d}, each rendered into three views with one pixel of noise, ignoring occlusion. From 1{,}000 to 256{,}000 points, the median rotation error falls from $0.002\degree$ to $0.00009\degree$ (\cref{fig:scaling}), and at one million points it is $0.00007\degree$ (\cref{fig:scale_gallery}, Tab.~\ref{tab:render}), in about 20 minutes per frame with GPU memory linear in the point count (Fig.~\ref{fig:resources}). The learned baselines bring at most two of the nine poses within $1\degree$ and return no pose from 64{,}000 points (Tab.~\ref{tab:render_baselines}).

At one million points, 21 to 25\% of the 2D points lie nearer to another projection than to the one that generated them, up from none at 1{,}000 points, so a nearest-projection labeling cannot resolve them (Fig.~\ref{fig:scaling_label}). The rotation error nevertheless falls over the same range, because moment matching and EM use every point without requiring each correspondence to be correct.

\head{Full scene reconstructions.}
We also localize single images against complete ETH3D scene reconstructions, where each query image observes an unknown part of the whole scene without a pose prior. \method\ recovers all 39 tuning queries within $1\degree$ and 35 of 39 held-out test queries (Tab.~\ref{tab:scene}). On the tuning queries, the best learned baseline, Deep Blind PnP, recovers two.

\begin{table}[!tp]
\centering
\footnotesize
\setlength{\tabcolsep}{2pt}
\caption{\textbf{Global search on complete inputs.} On thirty outlier-free test frames, GOPAC, from a zero or an annealed initial bound, and GOSMA search by branch and bound over rotations and over camera centers in a cuboid that contains the true center. Finished counts the searches that ended on their own. The others were stopped after 5.2 to 6.0 hours and scored at their best pose so far. Errors cover all thirty frames, and Time is the median over finished searches. Bold: best per column. Further details in \cref{sec:supp_global_search}.}
\label{tab:global_comparison}
\begin{tabular*}{\columnwidth}{@{\extracolsep{\fill}}lrrrrr@{}}
\toprule
Method & Finished & $<1\degree$\,$\uparrow$ & Best ($\degree$)\,$\downarrow$ & Median ($\degree$)\,$\downarrow$ & Time (s)\,$\downarrow$ \\
\midrule
GOPAC zero & 12/30 & 0/30 & 5.4 & 122 & 2{,}562 \\
GOPAC annealed & 12/30 & 0/30 & 4.2 & 132 & 3{,}015 \\
GOSMA & \textbf{30/30} & 0/30 & 1.7 & 156 & 68.8 \\
\method (ours) & \textbf{30/30} & \textbf{30/30} & \textbf{0.0006} & \textbf{0.083} & \textbf{1.92} \\
\bottomrule[0.06em]
\end{tabular*}
\end{table}

\subsection{Ablation study and failure modes}
\label{sec:exp_ablation}
\head{Role of moment matching.}
Three studies isolate what moment matching contributes. (i)~From the same initial poses on 100 test frames per benchmark, moments through order three or four bring 53 to 78 of these poses within $1\degree$ and 1\% translation error before EM (Tab.~\ref{tab:supp_feature_control}). The mean and covariance bring at most two, because distinct orientations can share them. (ii)~EM-ICP (ours), our EM refinement from random poses restarted until a 30-second limit, needs those 30 seconds per frame to reach the median errors that \method\ reaches in at most 0.06 seconds without outliers (Tab.~\ref{tab:supp_timing}). With as many outliers as inliers, it reaches at most 31\% accuracy (\cref{tab:main}, Fig.~\ref{fig:supp_cumulative}). (iii)~Given a domain that contains the true camera center and about six hours per search, GOPAC and GOSMA bring none of the thirty outlier-free frames within $1\degree$. \method\ needs no such domain and brings all thirty within $1\degree$, in about 2 seconds per frame (\cref{tab:global_comparison}, Tab.~\ref{tab:supp_gopac_comparison_all}). Controls on the depth spread, the image noise and the kind of outliers are in \cref{sec:supp_ablation}. Under noise, the rotation error follows that of the pose fitted to the true correspondences, and with outliers on scene structure, every frame with up to 30\% replaced stays within $1\degree$.

\begin{figure}[!t]
\centering
\pgfplotsset{fpanel/.style={scale only axis, axis on top, enlargelimits=false, y dir=reverse, ticks=none, clip=true, axis line style={oiGrey!80, line width=0.5pt},
    title style={font=\scriptsize, at={(axis description cs:0.5,1.0)}, anchor=south, yshift=-6pt, inner sep=2pt, text height=1.5ex, text depth=0.25ex}},
  fon/.style={only marks, mark=diamond*, mark size=1.05pt, mark options={fill=oiPurple, draw=white, line width=0.22pt}},
  foff/.style={only marks, mark=*, mark size=0.85pt, mark options={fill=colDet, draw=white, line width=0.12pt}},
  fscan/.style={only marks, oiBlue, mark=*, mark size=0.6pt, mark options={fill=oiBlue, draw=oiBlue, line width=0.1pt}, opacity=0.85}}
\includegraphics[width=234.76321pt,height=146.82462pt]{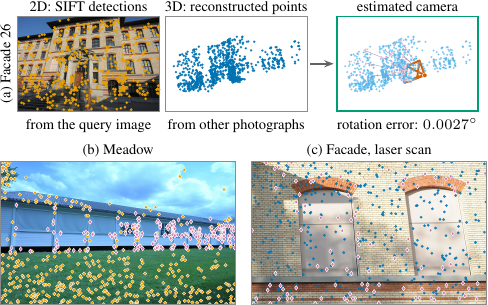}
\caption{\textbf{Independently selected inputs.} (a)~Facade 26, recovered from SIFT detections and a model built from other photographs. The camera is orange, with pink rays to matched model points. (b)~Meadow, where only the pink diamonds belong to reconstruction tracks. (c)~Facade laser scan, where pink diamonds lie within 3.4 pixels of blue projections.}
\label{fig:failure}
\end{figure}

\head{Independently selected inputs and failure modes.}
So far, every inlier 2D point is the image of a model point (\cref{fig:teaser}), but 2D points detected without reference to the 3D model may have no corresponding model point. With SIFT keypoints~\cite{lowe2004sift} in the query image and a COLMAP scene model~\cite{schonberger2016sfm} built from other photographs, the search over camera poses still recovers Facade 26 within $1\degree$ (\cref{fig:failure}(a), Tab.~\ref{tab:independent_summary}). Such inputs can fail in two ways. In Meadow, only 255 of the 12{,}069 detections belong to reconstruction tracks, and many others lie on vegetation and sky absent from the model (\cref{fig:failure}(b), Tab.~\ref{tab:supp_all_keypoints}). A dense model poses a second difficulty, because projections can lie near most 2D points even at an incorrect pose. In another Facade image, 81\% of the 2D points lie within 3.4 pixels of a projected laser-scan point at the reference pose, and 77\% still do at a pose rotated by $2\degree$ (\cref{fig:failure}(c)); further details of every study in \cref{sec:supp_experiments}. All code will be released.

\section{Conclusion}
\label{sec:conclusion}

In blind PnP, the camera pose need not wait for the correspondences. Finitely many feature averages of the two point sets determine it, and the correspondences then follow from nearest projections. The theory explains why this works on complete inputs and with uniform outliers, and the experiments show that it holds beyond these assumptions. Learned methods inherit the limits of their training data~\cite{torralba2011unbiased,sattler2019understanding}, lose accuracy outside it and run out of memory on large point sets. Using neither training nor descriptors, this inference offers an alternative to both and to branch-and-bound search.

\head{Limitations.}
\label{sec:limitations}
When 2D points have no corresponding model point, the feature averages need not match even at the true pose, so independently selected inputs remain harder than tracked ones. Adversarial outliers can prevent recovery. 
\head{Future work.} Missing structure and unequal sampling might be handled by reweighting the projections with unbalanced optimal transport~\cite{chizat2018scaling}. Moment matching may also extend to relative pose estimation and to monocular depth estimation of unknown scale~\cite{yu2025madpose,ding2025reposed}.

\FloatBarrier
\newpage
\section*{LLM Usage Statement}
We used OpenAI's Codex to assist with editing and refining prose, code and implementation, figures, and formatting. No LLMs or AI tools were used to develop the research ideas, hypotheses, scientific claims, or conclusions. The authors take full responsibility for the accuracy and integrity of the paper and its accompanying artifacts.

{\small\emergencystretch=1.5em
\bibliographystyle{ieeenat_fullname}
\bibliography{references}
}
\clearpage
\appendix
\renewcommand{\textfraction}{0.05}
\renewcommand{\topfraction}{0.95}
\renewcommand{\dbltopfraction}{0.95}
\renewcommand{\floatpagefraction}{0.9}
\renewcommand{\dblfloatpagefraction}{0.85}
\setcounter{topnumber}{3}\setcounter{dbltopnumber}{4}\setcounter{totalnumber}{5}
\crefalias{section}{appendix}\crefalias{subsection}{appendix}

\setcounter{table}{0}\renewcommand{\thetable}{A\arabic{table}}
\setcounter{figure}{0}\renewcommand{\thefigure}{A\arabic{figure}}
\setcounter{equation}{0}\renewcommand{\theequation}{A\arabic{equation}}
\setcounter{theorem}{0}\renewcommand{\thetheorem}{A\arabic{theorem}}
\setcounter{lemma}{0}\renewcommand{\thelemma}{A\arabic{lemma}}
\setcounter{corollary}{0}\renewcommand{\thecorollary}{A\arabic{corollary}}
\setcounter{definition}{0}\renewcommand{\thedefinition}{A\arabic{definition}}
\setcounter{assumption}{0}\renewcommand{\theassumption}{A\arabic{assumption}}
\setcounter{algocf}{0}\renewcommand{\thealgocf}{A\arabic{algocf}}
\makeatletter
\setlength{\@fptop}{0pt}
\setlength{\@fpsep}{8pt}
\setlength{\@fpbot}{0pt plus 1fil}
\setlength{\@dblfptop}{0pt}
\setlength{\@dblfpsep}{8pt}
\setlength{\@dblfpbot}{0pt plus 1fil}
\renewcommand{\theHtable}{A\arabic{table}}\renewcommand{\theHfigure}{A\arabic{figure}}\renewcommand{\theHequation}{A\arabic{equation}}
\@ifundefined{theHtheorem}{}{\renewcommand{\theHtheorem}{A\arabic{theorem}}}
\@ifundefined{theHlemma}{}{\renewcommand{\theHlemma}{A\arabic{lemma}}}
\@ifundefined{theHcorollary}{}{\renewcommand{\theHcorollary}{A\arabic{corollary}}}
\@ifundefined{theHdefinition}{}{\renewcommand{\theHdefinition}{A\arabic{definition}}}
\@ifundefined{theHassumption}{}{\renewcommand{\theHassumption}{A\arabic{assumption}}}
\@ifundefined{theHalgocf}{}{\renewcommand{\theHalgocf}{A\arabic{algocf}}}
\makeatother

\maketitlesupplementary

\cref{sec:supp_proofs} gives the notation, method details and proofs, \cref{sec:supp_config} the implementation, and \cref{sec:supp_experiments} the experiments. Appendix figures, tables, equations and statements carry the prefix~A, and unprefixed numbers refer to the main paper.

\section{Method details and proofs}
\label{sec:supp_proofs}
\cref{tab:supp_notation} summarizes the notation of the main paper and the appendix.

\begin{table*}[tp]
\centering\footnotesize
\setlength{\tabcolsep}{4pt}
\renewcommand{\arraystretch}{1.0}
\caption{\textbf{Notation.} The first four groups follow the main-text sections, including symbols used only in their appendix analyses. The last group covers further appendix analyses. A star marks a true value, a hat an estimate and a tilde a coordinate or moment after covariance normalization. Vector norms are Euclidean, unqualified tensor and matrix norms are Frobenius, and moment pairs use the direct-sum norm. Subscripts op, th and $\mathrm{E}$ mark the Euclidean operator norm, constants of the threshold proof and edge quantities.}
\label{tab:supp_notation}
\begin{tabular}{@{}>{\raggedright\arraybackslash}p{0.2\textwidth}>{\raggedright\arraybackslash}p{0.76\textwidth}@{}}
\toprule
Symbol & Meaning \\
\midrule
\multicolumn{2}{@{}l}{\textbf{Inputs, pose \& correspondences} (\cref{sec:problem})} \\
$n,N,m$ & Numbers of observations, of model points and of correspondences, so that $m$ is the number of inliers in each set. \\
$\dets=\{\vect u_i\},\pts=\{\vect y_j\}$ & 2D point set of observations in pixels, and 3D point set, or 3D model, of model points in model coordinates. \\
$\Kmat,\proj,\proj_{\mathrm{cal}}$ & Known intrinsics, projection to pixels, and projection $\proj_{\mathrm{cal}}(\vect x)=\vect x_{1:2}/x_3$ to calibrated image coordinates. \\
$\xi=(\Rot,\trans),\xi^\star,\hat\xi$ & Pose with rotation $\Rot\in\SO$ and translation $\trans\in\R^3$, true pose and estimated pose. \\
$\vect x_j,\vect p_j(\xi),\vect c$ & Model point $j$ in camera coordinates, its projection $\proj(\vect x_j)$, and the camera center in model coordinates. \\
$\sigma,\hat\sigma,\varnothing$ & True and estimated correspondences, and the label of an observation that images no model point. \\
$\vect w_i,\noise$ & Error of observation $i$ and its standard deviation per image coordinate in pixels, or $\noise_N$ when it varies with $N$. \\
\midrule
\multicolumn{2}{@{}l}{\textbf{Moments \& moment matching} (\cref{sec:prelim})} \\
$\order,\mom_\order,\vect m_\order$ & Moment order, moment tensor of \cref{eq:moment}, and vector of its distinct entries. \\
$\vect u_{\mathrm{cal}},\bar{\vect u}_{\mathrm{cal}},r_{\rm cal}$ & Calibrated observation, and the mean and root-mean-square radius of the calibrated observations. \\
$h$ & Image map of \cref{sec:prelim}, fixed by the observations and applied to observations and projections alike. \\
$h_F$ & Image map of the Fourier features (\cref{sec:prelim}), which shifts and scales the rectangle bounding the calibrated observations onto $[-1,1]^2$. \\
$\vect\phi_4,\vect\phi_F$ & Fourteen monomials of orders one to four, and bounded Fourier features. \\
$\vect\mu_{\dets},\vect\mu_{\pts}(\xi)$ & Moments, the averages of $\vect\phi_4\circ h$ over observations and projections. Subscript $F$ denotes the corresponding Fourier feature averages. \\
$\unmatched$ & Outlier fraction: the common fraction in both point sets in moment matching (\cref{def:moments}), the outlier probability of the uniform EM component (\cref{eq:em-mixture}), and a free parameter in \cref{sec:supp_fourier} with true values $\unmatched_U$ and $\unmatched_Y$. \\
$\vect\mu_{\dets,0},\vect\mu_{\pts,0}(\xi)$ & Expected feature values at a uniform outlier in the image and at a projected uniform outlier in the model's bounding box. \\
$\mathcal L_4,\mathcal L_F$ & Moment-matching objectives for complete inputs and inputs with outliers (\cref{def:moments}). \\
\midrule
\multicolumn{2}{@{}l}{\textbf{Refinement, association level \& pose ambiguity resolution} (\cref{sec:estimation})} \\
$\forwardcount$ & Number of model points in front of the camera with finite projections. \\
$\noisefit,\lik_{\mathrm{EM}}$ & EM estimate of $\noise$, and the Poisson log-likelihood that compares refined poses (\cref{eq:em-likelihood}). \\
$d_{i,k},\nnb,\assoc_i$ & Distance from observation $i$ to its $k$th-nearest projection, number of neighbors, and the statistic of \cref{sec:supp_association_implementation}. \\
$D_{\rm im},p_d,n_{\rm cl}$ & Rectangle around the observations that supports the uniform EM component (area $|D_{\rm im}|$), probability that a model point in front of the camera is observed, and expected number of outlier observations. \\
$r_{ij},n_{\rm m}$ & Responsibility of projection $j$ for observation $i$, and the expected number $\sum_{ij}r_{ij}$ of inlier observations. \\
$p_{\rm assoc},p^0_{\rm assoc},V_4$ & Association level as a tail probability, the same tail for a pose that EM has not refined (Gamma shape $n$ without outliers), and the first-order expected value of $\mathcal L_4$ at the true pose under noise of standard deviation $\noisefit$. \\
$\halfturn_k,\opthalf$ & Half-turns about the $k$th principal axis of the matched model points ($\halfturn_0=I_3$) and about the line through their centroid parallel to the optical axis. \\
\midrule
\multicolumn{2}{@{}l}{\textbf{Identifiability} (\cref{sec:theory})} \\
$\fourconst,\fourerr$ & Constant and error term of \cref{thm:fourier-main}. In \cref{sec:supp_fourier}, $\fourconst=1/\gamma_{K_F}$, which does not depend on $\delta$, and $\fourerr$ is the sum in \cref{eq:fourier-finite-errors} of outlier sampling, grid, image-noise and unequal-fraction errors. \\
$\operatorname{dist}$ & Distance of \cref{thm:fourier-main}, the distance $d(\theta,\theta^\star)$ of \cref{sec:supp_fourier} between rotation matrices, translations in units of the model length $\ell_{\mathcal Y}$ and outlier fractions. \\
$\mathcal P_{\pts}(\xi)$ & Average of $\vect\phi_4$ over the calibrated projections at pose $\xi$, without the map $h$ (\cref{sec:supp_proof_pinhole}). \\
$\bar{\vect y},\Sigma_{\pts},\Sigma_{\dets_{\mathrm{cal}}}$ & Mean and covariance of the model points, and covariance of the calibrated observations. \\
$\tilde\dets,\tilde\pts$ & Calibrated observations and model points, each centered and normalized by its own covariance. \\
$A,\vect b$ & Rank-two affine camera matrix and its image translation. \\
$Q,\vect\gamma$ & Relative rotation $\Rot(\Rot^\star)^\top$ and offset $\trans-Q\trans^\star$ of a competing pose. \\
$\mathcal H_\order,\mathcal S_\order,R_\perp,Q_{\rm s}$ & Binary and ternary forms of degree $\order$, quarter-turn of the image plane, model symmetry. \\
$O,\Vset$ & Orientation after normalization, and the set of $2\times3$ matrices with orthonormal rows. \\
$T,T_{\dets},\Fmap_T$ & Normalized cubic and quartic moments of the model and of the observations, and projected moments as a function of $O$. \\
$\Gmod$ & Separation constant, the lower Lipschitz constant of $\Fmap_T$ on $\Vset$ (\cref{eq:statement-4_method-2}). \\
$\obserr,\fitres$ & Observation error and fit residual of the normalized moments (\cref{sec:supp_proof_identifiability}). \\
\midrule
\multicolumn{2}{@{}l}{\textbf{Appendix analyses}} \\
$\vect v_i$ & Measurement error of a model point, in model coordinates. \\
$\refdepth$ & Reference depth of the affine initial poses and of the depth study (\cref{tab:depth}). \\
$\projrad,\rmsrad$ & Bound on the norm of the projections, and root-mean-square radius of the observations about their centroid. \\
$\fmax,\predtol,\prederr$ & Bound on the density of the projections, tolerance on the prediction error, and maximum prediction error. \\
$\featbound,\featlip$ & Norm bound and Lipschitz constant of a feature map. \\
$\edgelen_j,\vect\mu_E,\vect\mu_C$ & Projected length of model edge $j$, and feature averages over projected model edges and detected image edges. \\
$\edgemass,\edgedisp,\quaderr$ & Fraction of the coupling's mass on pairs within image distance $\edgedisp$, that distance, and the combined quadrature error. \\
$\posecoord,J^\star,\jacsv$ & Local pose coordinates in $\R^6$, Jacobian of the edge feature average at the true pose, and its least singular value. \\
\bottomrule
\end{tabular}
\end{table*}

\subsection{Moments \& measurement error}
\label{sec:supp_proof_moments}

This section bounds how much image noise changes the moments of the observations and describes the effect of errors in the model points. In the bound, the terms that are linear in the noise have zero mean, and their contribution decreases as $1/\sqrt N$ when the point sets grow (\cref{sec:theory,sec:experiments}). The second part concerns the assumption that the model points are exact (\cref{sec:problem}). We first illustrate what moments distinguish and fix the coordinates in which we measure them.

\para{A scalar example.}
The sets of real numbers $\{-2,-1,3\}$ and $\{2,1,-3\}$ are reflections of each other about zero. Both have mean zero and second moment $14/3$, so their first two moments do not distinguish them. Their third moments are $6$ and $-6$, because $(-2)^3+(-1)^3+3^3=18$ and the reflection changes the sign of every cube. An average does not depend on the order of the points, so the third moment distinguishes the two sets without any correspondence between their points. Moment matching uses the same property in the image, where the moments of orders three and four distinguish orientations whose projections have equal mean and covariance (\cref{fig:moment_orders}). The moment of order $\order$ of a 2D point set has several distinct entries, one for each monomial of degree $\order$, and we now fix how we index them.

\para{Moment coordinates.}
\label{sec:supp_moment_coordinates}%
For a nonempty finite collection $\mathcal A$ of points in a Euclidean space and a positive integer $\order$, the tensor power $\vect a^{\otimes\order}$ of a point $\vect a\in\mathcal A$ contains all products of $\order$ coordinate components, including repeated entries. The moment tensor is their average,
\begin{equation}
\label{eq:moment}
\mom_\order(\mathcal A)=\frac{1}{|\mathcal A|}\sum_{\vect a\in\mathcal A}\vect a^{\otimes\order}.
\end{equation}
We index the points of $\mathcal A\subset\R^{d_{\mathcal A}}$ as $\vect a_1,\ldots,\vect a_{|\mathcal A|}$. A multi-index $\alpha=(\alpha_1,\ldots,\alpha_{d_{\mathcal A}})$ of nonnegative integers with $|\alpha|=\alpha_1+\cdots+\alpha_{d_{\mathcal A}}=\order$ names one distinct entry of the symmetric tensor $\mom_\order(\mathcal A)$ of \cref{eq:moment}, namely the average
\[
m_\alpha(\mathcal A)=\frac1{|\mathcal A|}\sum_{i=1}^{|\mathcal A|}\prod_{r=1}^{d_{\mathcal A}} a_{i,r}^{\alpha_r}.
\]
We collect these entries, in a fixed order, in the vector $\vect m_\order(\mathcal A)=(m_\alpha(\mathcal A))_{|\alpha|=\order}$, which has $\binom{d_{\mathcal A}+\order-1}{\order}$ entries. The tensor repeats the entry $m_\alpha$ once for each ordering of its indices, that is, $\binom{\order}{\alpha}=\order!/(\alpha_1!\cdots\alpha_{d_{\mathcal A}}!)$ times, so its Frobenius norm is
\[
\norm{\mom_\order(\mathcal A)}^2
=\sum_{|\alpha|=\order}\binom{\order}{\alpha}\,m_\alpha(\mathcal A)^2.
\]
In 2D, the multiplicities are $(1,3,3,1)$ for cubic and $(1,4,6,4,1)$ for quartic moments. The fourteen averages of $\vect\phi_4\circ h$ in moment matching (\cref{sec:prelim}) are the distinct entries of $\mom_1$ to $\mom_4$ of the mapped points, without these multiplicities. \cref{sec:supp_proof_identifiability} uses the form $\vect\zeta\mapsto\langle\mom_\order(\mathcal A),\vect\zeta^{\otimes\order}\rangle$ in place of the tensor. Its coefficients $\binom{\order}{\alpha}m_\alpha$ also differ from $\vect m_\order$. We state the noise bound below in the tensor norm, which \cref{thm:identifiability} also uses.

\para{Image noise.}
The bound below concerns complete inputs in pixel coordinates, before the map $h$ is applied, under the measurement model of \cref{eq:model}. Its first term shrinks as $1/\sqrt N$, and its second term, of order $\noise^2$, does not depend on $N$.

\needspace{7\baselineskip}
\begin{lemma}[Moments under image noise]
\label{lem:moments}
Let the inputs be complete, with $N\ge1$ exact model points at positive depth under the true pose $\xi^\star$. For each observation $i$, let $\vect u_i^\star=\vect p_{\sigma(i)}(\xi^\star)$ be its noiseless position, the projection of the model point that it images, with $\norm{\vect u_i^\star}\le\projrad$. Let $\vect u_i=\vect u_i^\star+\vect w_i$ with independent noise vectors $\vect w_i\sim\mathcal N(0,\noise^2I_2)$ and $0\le\noise\le1$ pixels, and let $\dets^\star=\{\vect u_i^\star\}$. For every order $\order\ge1$, the moments $\mom_\order(\dets)$ and $\mom_\order(\dets^\star)$ agree almost surely when $\noise=0$ and satisfy
\begin{equation}
\label{eq:moment-noise}
\sqrt{\E\norm{\mom_\order(\dets)-\mom_\order(\dets^\star)}^2}\le \order\projrad^{\order-1}\noise\sqrt{2/N}+C_{\order,\projrad}\noise^2.
\end{equation}
Here $C_{\order,\projrad}\ge0$ depends only on $\order$ and $\projrad$, and the expectation is over the image noise, in pixel coordinates before normalization.
\end{lemma}

\begin{proof}[Proof of \cref{lem:moments}]
We fix an integer $\order\ge1$. The proof first shows that $\mom_\order(\dets^\star)$ is a quantity that moment matching can evaluate, and then separates the noise terms that averaging reduces from those that it does not.

\emph{Step 1. The projected moments do not depend on the correspondences.} Since $\sigma$ is a bijection of $\{1,\ldots,N\}$, substituting $j=\sigma(i)$ reorders the sum,
\begin{equation}
 \mom_\order(\dets^\star)=\frac1N\sum_{i=1}^N\vect p_{\sigma(i)}(\xi^\star)^{\otimes \order}
 =\frac1N\sum_{j=1}^N\vect p_j(\xi^\star)^{\otimes \order}.
\end{equation}
The right-hand side depends on the pose and the model but not on $\sigma$. The lemma therefore compares the observed moments with the moments of the projections at the true pose, which can be computed without correspondences.

\emph{Step 2. Expansion in the noise.} For $\noise=0$, the noise vanishes almost surely, so $\dets=\dets^\star$ and the two moments agree almost surely. We assume $\noise>0$ from now on. We expand $(\vect u_i^\star+\vect w_i)^{\otimes \order}$ by choosing the noiseless position or the noise vector in each of its $\order$ factors. Subtracting $(\vect u_i^\star)^{\otimes \order}$ leaves the $2^\order-1$ products that contain at least one noise factor. Products with one noise factor have zero mean, so averaging over the $N$ independent observations reduces their root mean square. Products with several noise factors can have a nonzero mean that averaging does not remove, but they are of order $\noise^2$. We bound the two groups separately and add their root-mean-square bounds by Minkowski's inequality.

\emph{Step 3. One noise factor.} For each of the $\order$ positions of the noise factor, the products are linear in $\vect w_i$, independent across $i$ and of zero mean, with second moment $\norm{\vect u_i^\star}^{2(\order-1)}\E\norm{\vect w_i}^2\le2\noise^2\projrad^{2(\order-1)}$. Independence and zero means make the cross terms vanish when we square their average over $i$. Each average therefore has root mean square at most $\projrad^{\order-1}\noise\sqrt{2/N}$, and the $\order$ positions contribute the first term of \cref{eq:moment-noise}.

\emph{Step 4. Several noise factors.} A product with $a\ge2$ noise factors has norm $\norm{\vect u_i^\star}^{\order-a}\norm{\vect w_i}^a\le\projrad^{\order-a}\norm{\vect w_i}^a$. Because $\norm{\vect w_i}^2/\noise^2$ is exponentially distributed with mean two, $\E\norm{\vect w_i}^{2a}=2^aa!\,\noise^{2a}$, so the product has root mean square at most $\projrad^{\order-a}\sqrt{2^aa!}\,\noise^a$. Minkowski's inequality gives the same bound for its average over $i$, and there are $\binom{\order}{a}$ choices of positions for the $a$ noise factors. Summing both groups gives
\begin{equation}\label{eq:supp-moment-noise}
\begin{aligned}
 &\Bigl(\E\norm{\mom_\order(\dets)-\mom_\order(\dets^\star)}^2\Bigr)^{1/2}\\
 &\quad\le \order\projrad^{\order-1}\noise\sqrt{2/N}
 +\sum_{a=2}^\order\binom{\order}{a}\projrad^{\order-a}\sqrt{2^aa!}\,\noise^a.
\end{aligned}
\end{equation}
Since $\noise^a\le \noise^2$ for $a\ge2$ and $0\le \noise\le1$, we may take
\begin{equation}
\label{eq:moment-remainder-constant}
C_{\order,\projrad}=\sum_{a=2}^\order\binom{\order}{a}\projrad^{\order-a}\sqrt{2^aa!}.
\end{equation}
This constant does not depend on $\noise$ or $N$, and $C_{1,\projrad}=0$ because the sum is empty. When $\projrad=0$, only the term $a=\order$ remains for $\order\ge2$, with $\projrad^0=1$.
\end{proof}

\para{Errors in the model points.}
The lemma treats the model points as exact, as \cref{eq:model} does. Suppose now that model point $\vect y_{\sigma(i)}$ is measured with an error $\vect v_i$ of covariance $\Sigma_{\mathrm{3D},i}$ and that the image error $\vect w_i$ has covariance $\Sigma_{\mathrm{2D},i}$, with $\vect v_i$ and $\vect w_i$ independent. To first order at the true pose, the difference between observation $i$ and the projection of the measured model point is $\vect w_i-J_{\proj,i}\Rot^\star\vect v_i$. Here $J_{\proj,i}$ is the Jacobian of the pixel projection $\proj$ at the true model point $\vect x_{\sigma(i)}$ in camera coordinates. Independence makes the cross-covariances vanish, so this difference has covariance
\begin{equation}
\Sigma_{\mathrm{lin},i}=\Sigma_{\mathrm{2D},i}+J_{\proj,i}\Rot^\star\Sigma_{\mathrm{3D},i}(\Rot^\star)^\top J_{\proj,i}^\top.
\end{equation}
The 3D term depends on the viewing direction through $J_{\proj,i}$ and is in general anisotropic. At a fixed viewing direction and a fixed covariance in model units, it scales with the inverse square of the depth. The model assumes independent errors and exact intrinsics, so it omits the correlations that shared calibration errors introduce. Because the expression is linearized, the exact projected error need not be Gaussian.

\subsection{Proof of \texorpdfstring{\cref{thm:pinhole}}{the pinhole identifiability theorem}}
\label{sec:supp_proof_pinhole}

We use calibrated image coordinates, in which a camera point $\vect x$ with $x_3>0$ projects to $\proj_{\mathrm{cal}}(\vect x)=\vect x_{1:2}/x_3$, and we identify a model $\pts=\{\vect y_j\}_{j=1}^N$ with the vector $(\vect y_1,\ldots,\vect y_N)\in\R^{3N}$. For a pose $\xi=(\Rot,\trans)$ under which every model point has positive depth, the raw projected moments are
\[
\mathcal P_{\pts}(\xi)=\frac1N\sum_{j=1}^N\vect\phi_4\bigl(\proj_{\mathrm{cal}}(\Rot\vect y_j+\trans)\bigr)\in\R^{14},
\]
the fourteen moments of orders one to four of the calibrated projections, without centering or normalization.

\begin{trivlist}\item[\hskip\labelsep\textbf{Theorem~\ref{thm:pinhole} (restated).}]\itshape
Fix a pose $\xi^\star$ and an integer $N\ge20$. There is a set $\mathcal Z\subset\R^{3N}$ of Lebesgue measure zero with the following property. Let $\pts\notin\mathcal Z$ be a model whose $N$ points all have positive depth under $\xi^\star$, and let $\xi$ be a pose under which all its points have positive depth. If $\mathcal P_{\pts}(\xi)=\mathcal P_{\pts}(\xi^\star)$, then $\xi=\xi^\star$.
\end{trivlist}

An invertible affine map of the image preserves the equality of moments through order four in both directions, because composition with the map or its inverse takes every polynomial of degree at most four to another such polynomial. The restated theorem therefore also covers the moments in pixels and the moments after the map $h$ of \cref{sec:prelim}, which the main text compares. On complete noiseless inputs, it shows that the true pose is generically the only zero of $\mathcal L_4$ among the poses that keep all model points in front of the camera, so the minimum that Step 2 searches for is unique. The proof shows that every kind of alternative pose imposes on each model point an equation that a generic point does not satisfy. It then counts dimensions to show that the models admitting such a pose form a null set. After the proof, \cref{cor:pinhole-stability} turns uniqueness into a stability statement for the perspective moment fit.

Uniqueness cannot hold for every model. A quarter-turn of a cube about the axis through the centers of two opposite faces maps its eight corners onto themselves, so a camera pose and its composition with this rigid motion give the same image. The theorem shows that such symmetric models are exceptional.

\para{Coordinates.}
We fix the true pose $\xi^{\star}=(\Rot^{\star},\trans^{\star})$ before choosing the model and express the model points in its camera coordinates as $\vect x_j=\Rot^{\star}\vect y_j+\trans^{\star}$. This invertible affine change of coordinates of $\R^{3N}$ preserves sets of Lebesgue measure zero, so it suffices to prove the theorem for the configurations $\mathcal X=\{\vect x_j\}_{j=1}^N$ with all $x_{j3}>0$. The true projection is then $\proj_{\mathrm{cal}}(\vect x_j)$, and a competing pose $\xi=(\Rot,\trans)$ gives
\[
\proj_{Q,\vect\gamma}(\vect x_j)=\proj_{\mathrm{cal}}(Q\vect x_j+\vect\gamma),
\]
where $Q=\Rot(\Rot^\star)^\top\in\SO$ and $\vect\gamma=\trans-Q\trans^{\star}$. The map $\xi\mapsto(Q,\vect\gamma)$ is a bijection from the poses onto $\SO\times\R^3$ that sends $\xi^\star$ to $(I_3,0)$. In the true camera coordinates, the competing camera center is $-Q^\top\vect\gamma$, so the two cameras have the same center exactly when $\vect\gamma=0$.

\para{Moment differences as polynomials.}
Let $\mathbb P_4^0$ denote the fourteen-dimensional space of bivariate polynomials of degree at most four with zero constant coefficient. Each $P\in\mathbb P_4^0$ is the inner product $\vect\kappa\cdot\vect\phi_4$ of its coefficient vector $\vect\kappa\in\R^{14}$ with $\vect\phi_4$. The linear combination of the fourteen moment differences with weights $\vect\kappa$ is therefore the average, over the points, of $P(\proj_{Q,\vect\gamma}(\vect x_j))-P(\proj_{\mathrm{cal}}(\vect x_j))$. The proof distinguishes alternative poses by whether the camera center moves and whether the optical axis tilts. For each kind, the next lemma bounds the dimension of the space of polynomials whose values cannot detect the change of pose.

\needspace{7\baselineskip}
\begin{lemma}[Polynomial features under a pinhole camera]
\label{supp:lem-pinhole-pullback}
Let $Q\in\SO$ and $\vect\gamma\in\R^3$, let $\vect u=(u_1,u_2)\in\R^2$ be calibrated image coordinates, and let $x_3>0$ be the depth along the true camera ray $(x_3\vect u,x_3)$.
\textup{(a)} If $\vect\gamma\ne0$, the subspace
\begin{equation}
\label{eq:supp-pinhole-identifiability}
\begin{aligned}
 \mathcal K_{Q,\vect\gamma}=\{P\in\mathbb P_4^0:\;&
       P(\proj_{Q,\vect\gamma}(x_3u_1,x_3u_2,x_3))\\
       &\text{ is independent of }x_3\}
\end{aligned}
\end{equation}
has dimension at most four and equals $\{0\}$ if $\gamma_3\ne0$.
Here the composition is a rational function of $(\vect u,x_3)$, and independence means that its partial derivative with respect to $x_3$ vanishes identically.
\textup{(b)} If $\vect\gamma=0$ and $(Q_{31},Q_{32})\ne(0,0)$, then
$P(\proj_{Q,\vect\gamma}(\vect u,1))-P(\vect u)$ is a nonconstant function of $\vect u$ for every nonzero
$P\in\mathbb P_4^0$.
\textup{(c)} If $\vect\gamma=0$ and $(Q_{31},Q_{32})=(0,0)$, then
$\proj_{Q,\vect\gamma}(\vect u,1)=G\vect u$ for some $G\in\mathrm O(2)$. If $Q\ne I_3$, then
$G\ne I_2$, and the invariant subspace
$\mathcal K_G=\{P\in\mathbb P_4^0:P\circ G=P\}$ has dimension at most
eight. For $P\notin\mathcal K_G$, the polynomial
$\vect u\mapsto P(G\vect u)-P(\vect u)$ is nonconstant.
\end{lemma}
\begin{proof}
\emph{Part (a).} We write $\vect a=Q(\vect u,1)$, so that $Q(x_3\vect u,x_3)+\vect\gamma=x_3\vect a+\vect\gamma$, and $\vect z(\vect u,x_3)=\proj_{Q,\vect\gamma}(x_3\vect u,x_3)$. With $\vect e_3=(0,0,1)$, we have $x_3\vect a+\vect\gamma=(x_3Q+\vect\gamma\vect e_3^\top)(\vect u,1)$. Because $\det(x_3Q+\vect\gamma\vect e_3^\top)=x_3^2\bigl(x_3+(Q^\top\vect\gamma)_3\bigr)$, the map $\vect u\mapsto\vect z(\vect u,x_3)$ is an invertible projective transformation of the plane for all but at most one value of $x_3>0$. The points $\vect z(\vect u,x_3)$ therefore fill an open subset of $\R^2$.
We first suppose that $\gamma_3\ne0$ and put $\vect u_{\rm c}=\vect\gamma_{1:2}/\gamma_3$, the image of the true camera center in the competing camera. Differentiation gives
\[
\partial_{x_3}\vect z=\frac{\gamma_3}{x_3(x_3a_3+\gamma_3)}\bigl(\vect z-\vect u_{\rm c}\bigr),
\]
so as $x_3$ varies, the point $\vect z$ moves along the line through $\vect u_{\rm c}$. If $P\in\mathcal K_{Q,\vect\gamma}$, the chain rule gives $(\vect z-\vect u_{\rm c})\cdot\nabla P(\vect z)=0$ at every point $\vect z(\vect u,x_3)$. This polynomial in $\vect z$ vanishes on an open set
and hence identically. We expand $P$ in homogeneous terms about
$\vect u_{\rm c}$. By Euler's identity, the operator $(\vect z-\vect u_{\rm c})\cdot\nabla$ multiplies the term of degree $k$ by $k$, so every term of positive degree is zero.
Thus $P$ is constant, and its zero constant coefficient gives $P=0$.
If $\gamma_3=0$, then $\vect\gamma_{1:2}\ne0$ and
\[
 \proj_{Q,\vect\gamma}(x_3\vect u,x_3)
 =\frac{(Q(\vect u,1))_{1:2}}{(Q(\vect u,1))_3}
  +\frac{\vect\gamma_{1:2}}{x_3(Q(\vect u,1))_3},
\]
so the point moves along the fixed direction $\vect\gamma_{1:2}$ as $x_3$ varies. For $P\in\mathcal K_{Q,\vect\gamma}$, the chain rule gives
$\vect\gamma_{1:2}\cdot\nabla P(\vect z)=0$ on the same open set, hence as a polynomial
identity. Such a $P$ is a polynomial of degree at most four in the linear coordinate
perpendicular to $\vect\gamma_{1:2}$, with zero constant coefficient, and these polynomials form a space of dimension four.

\emph{Part (b).} We put $z_Q(\vect u)=(Q(\vect u,1))_3=Q_{31}u_1+Q_{32}u_2+Q_{33}$, an affine polynomial that is nonconstant because $(Q_{31},Q_{32})\ne(0,0)$. The competing camera sends the line $z_Q=0$ to infinity. A nonzero $P\in\mathbb P_4^0$ has a degree $d_P\ge1$ and a nonzero leading homogeneous
part $P_{d_P}$, and its composition with $\proj_{Q,\vect\gamma}$ is a quotient with denominator $z_Q^{d_P}$.
On the line $z_Q=0$, the numerator equals
$P_{d_P}((Q(\vect u,1))_{1:2})$.
Along this line, the two coordinates $(Q(\vect u,1))_{1:2}$ trace an affine line not
through the origin, because otherwise the invertible matrix $Q$ would map a
vector $(\vect u,1)$ to zero. A nonzero homogeneous polynomial cannot
vanish on this entire line, because scalar multiples of the line contain
an open subset of $\R^2$. The numerator is therefore not divisible by the irreducible polynomial
$z_Q$, so the composition has a pole on the line $z_Q=0$. Subtracting the polynomial
$P(\vect u)$ cannot remove this pole, so the difference is not a polynomial and in particular is nonconstant.

\emph{Part (c).} The third row of $Q$ is $(0,0,Q_{33})$, so orthogonality gives $Q_{33}=\pm1$ and $Q_{13}=Q_{23}=0$. Hence
$G=Q_{1:2,1:2}/Q_{33}\in\mathrm O(2)$. If $G=I_2$, then $Q=Q_{33}I_3$, and
$\det Q=1$ gives $Q=I_3$.
If $G$ is a reflection, we choose orthonormal coordinates along its axis and perpendicular to it. The reflection then fixes the eight of the fourteen monomials whose exponent in the
perpendicular coordinate is even and negates the other six. If $G$ is a rotation by an angle $\theta\notin2\pi\mathbb Z$, it multiplies the complex monomial
$(u_1+\mathrm i u_2)^a(u_1-\mathrm i u_2)^b$ by $e^{\mathrm i(a-b)\theta}$. A nonidentity rotation therefore changes both
linear polynomials and the two cubic polynomials with $a-b=\pm1$. A half-turn also changes the other two cubic polynomials, and
every other rotation changes the two quadratic polynomials with $a-b=\pm2$. Thus at most eight independent polynomials are unchanged by a nonidentity $G$.
If $P\notin\mathcal K_G$, then $P(G\vect u)-P(\vect u)$ is nonzero, and
its zero constant coefficient makes it nonconstant.
\end{proof}

\begin{proof}[Proof of \cref{thm:pinhole}]
For $\mathcal X=\{\vect x_j\}_{j=1}^N$, the difference between
the competing and true moment vectors is
\begin{equation}\label{eq:supp-pinhole-difference}
\begin{split}
 \vect\Phi(\mathcal X,Q,\vect\gamma)
 &=\frac1N\sum_{j=1}^N
   \bigl[\vect\phi_4(\proj_{Q,\vect\gamma}(\vect x_j))\\
 &\qquad-\vect\phi_4(\proj_{\mathrm{cal}}(\vect x_j))\bigr].
\end{split}
\end{equation}
We consider only configurations and competing poses that make all true and
competing depths positive, so $\vect\Phi$ is
smooth on its domain. Its zeros are those of
$\mathcal P_{\pts}(\xi)-\mathcal P_{\pts}(\xi^{\star})$, expressed in
calibrated image and transformed model coordinates. A zero with
$(Q,\vect\gamma)\ne(I_3,0)$ is an alternative pose. We write $\partial_{\mathcal X}\vect\Phi$ for the $14\times3N$ derivative of $\vect\Phi$ with respect to the point coordinates. A coefficient vector $\vect\kappa\in\R^{14}$ with $\vect\kappa^\top\partial_{\mathcal X}\vect\Phi=0$ defines the polynomial $P=\vect\kappa\cdot\vect\phi_4$. The average of $P\circ\proj_{Q,\vect\gamma}-P\circ\proj_{\mathrm{cal}}$ over the points then has zero derivative with respect to every point. We show that the configurations $\mathcal X$ that admit an alternative pose form a set of dimension less than $3N$.

\emph{Dimension facts.} We use standard facts about semialgebraic sets~\cite{bochnak1998real}.
Sets described by finitely many polynomial equalities and inequalities,
and projections of such sets, are semialgebraic. Projection does not
increase dimension. A semialgebraic set has dimension at most $d$ if each of its points has a neighborhood in which the set lies in a smooth submanifold of dimension $d$. A family has dimension at most its number of parameters plus the largest dimension of its solution set at fixed parameters.
A nonzero polynomial in $k$ variables has a zero set of dimension at
most $k-1$. A semialgebraic subset of $\R^{3N}$ of dimension less than
$3N$ has Lebesgue measure zero. In each of the three cases below, a high rank of $\partial_{\mathcal X}\vect\Phi$ bounds the dimension directly through the implicit-function theorem. If the rank is low, \cref{supp:lem-pinhole-pullback} gives a coefficient vector $\vect\kappa$ whose polynomial detects the change of pose, and that polynomial yields one equation for every point.

\emph{Case 1. Distinct camera centers.}
We suppose $\vect\gamma\ne0$, so the competing pose has six parameters.
At a solution with $\rank \partial_{\mathcal X}\vect\Phi\ge10$, ten of the equations have independent derivatives, so near this solution the solution set lies in a submanifold of dimension $3N+6-10<3N$.
If the rank is less than ten, the coefficient vectors $\vect\kappa$ with $\vect\kappa^\top\partial_{\mathcal X}\vect\Phi=0$ form a space of dimension greater than four.
By \cref{supp:lem-pinhole-pullback}(a), we can therefore choose such a $\vect\kappa$ whose polynomial
$P$ lies outside $\mathcal K_{Q,\vect\gamma}$, and we normalize $\vect\kappa$ to the unit sphere $S^{13}$.
We write $\vect x_j=(x_{j3}\vect u^\star_j,x_{j3})$ with true projection $\vect u^\star_j$ and depth $x_{j3}>0$.
Moving a point along its true camera ray leaves $\proj_{\mathrm{cal}}(\vect x_j)$ unchanged, so the zero derivative of the average along this direction gives
\[
 \left.\frac{d}{dx_3}P(\proj_{Q,\vect\gamma}(x_3\vect u^\star_j,x_3))\right|_{x_3=x_{j3}}=0,
 \quad j=1,\ldots,N.
\]
For fixed $(Q,\vect\gamma,\vect\kappa)$, the left-hand side is a rational function of $\vect x_j$ whose denominator is a product of powers of the true and competing depths. Because $P\notin\mathcal K_{Q,\vect\gamma}$, it is not identically zero, so its numerator is a nonzero polynomial in the three coordinates of $\vect x_j$. Each point therefore lies in a
set of dimension at most two. The equations involve separate points, so their joint solution set has dimension at most $2N$.
The six pose parameters and thirteen parameters of $\vect\kappa$ give total
dimension at most $2N+19$, which is less than $3N$ exactly when $N\ge20$. This case sets the bound on $N$ in \cref{thm:pinhole}.

\emph{Case 2. Same center, tilted optical axis.}
We suppose $\vect\gamma=0$ and $(Q_{31},Q_{32})\ne(0,0)$. These rotations form an open subset of $\SO$, of
dimension three. When the derivative with respect to the point coordinates has full rank fourteen, the local dimension of the solution set is at most
$3N+3-14<3N$.
Otherwise, we choose a unit vector $\vect\kappa$ with $\vect\kappa^\top\partial_{\mathcal X}\vect\Phi=0$. The ray coordinates
$(\vect u^\star_j,x_{j3})$ are smooth coordinates for positive
depth, and $\proj_{Q,\vect\gamma}(x_{j3}\vect u^\star_j,x_{j3})=\proj_{Q,\vect\gamma}(\vect u^\star_j,1)$ because the two cameras share their center.
The derivatives with respect to the two
image coordinates therefore give
\[
 \nabla_{\vect u^\star_j}
   \bigl[P(\proj_{Q,\vect\gamma}(\vect u^\star_j,1))-P(\vect u^\star_j)\bigr]=0,
 \quad j=1,\ldots,N.
\]
By \cref{supp:lem-pinhole-pullback}(b), the function in brackets is a
nonconstant rational function of $\vect u^\star_j$, so at least one of its partial derivatives is a nonzero rational
function. The numerator of that partial derivative is a nonzero polynomial in two variables, so the critical set has dimension at most one in $\vect u^\star_j$,
and the free depth adds one dimension, which gives at most $2N$ dimensions for the point set.
Adding three rotation parameters and thirteen parameters of $\vect\kappa$ bounds the
dimension by $2N+16<3N$.

\emph{Case 3. Same center, aligned optical axis.}
We suppose $\vect\gamma=0$ and $(Q_{31},Q_{32})=(0,0)$, with $Q\ne I_3$.
By the proof of \cref{supp:lem-pinhole-pullback}(c), $Q_{33}=\pm1$. A rotation with $Q_{33}=-1$ gives every point the competing depth $-x_{j3}<0$ and is excluded. The rotations with $Q_{33}=1$ form a family of dimension one. A rank of at least six bounds the local dimension
of the solution set by $3N+1-6<3N$.
Otherwise, the coefficient vectors $\vect\kappa$ with $\vect\kappa^\top\partial_{\mathcal X}\vect\Phi=0$ form a space of dimension greater than eight,
so \cref{supp:lem-pinhole-pullback}(c) gives a unit $\vect\kappa$ whose polynomial satisfies
$P\notin\mathcal K_G$.
The argument of Case~2 requires each $\vect u^\star_j$ to be a
critical point of the nonconstant polynomial
$\vect u\mapsto P(G\vect u)-P(\vect u)$.
Its critical set has dimension at most one, which again gives at most $2N$
dimensions for the point set.
The one rotation parameter and thirteen parameters of $\vect\kappa$ give total
dimension at most $2N+14<3N$.

\emph{Conclusion.} All sets in these arguments are semialgebraic, because the rotation constraints,
positive depths, coefficient normalization and rank conditions are
polynomial equalities or inequalities after denominators are cleared.
The conditions $P\notin\mathcal K_{Q,\vect\gamma}$ and
$P\notin\mathcal K_G$ also have this form, because each requires at least one
coefficient of the corresponding rational or polynomial identity to be
nonzero. The local dimension bounds hold throughout the sets of high rank. The parameter counts of the three cases cover the remaining ranks. The finitely many unions and projections of the solution sets
therefore have dimension less than $3N$.
For the fixed true pose, we obtain a null set $\mathcal Z$ of configurations
outside which no other pose with positive depths has the same moments through order four.
This exceptional set is semialgebraic of dimension less than $3N$, so it also lies in
the zero set of a nonzero polynomial in the model coordinates.
Every joint density on $\R^{3N}$, including one for dependent points, assigns probability zero to this set.
\end{proof}

\para{Perspective stability.}
\cref{thm:pinhole} concerns exact moments, whereas Step 2 fits moments that contain error. Uniqueness together with compactness bounds the pose error of the fit by the moment errors, through a modulus that the corollary defines but does not compute. This bound is the perspective counterpart of \cref{eq:orientation-stability}, and it concerns the moment fit before EM refinement. We hold the image map $h$ fixed throughout the comparison, even if it was computed from the observations. We write $\vect\mu_{\pts}(\xi;h)$ and $\vect\mu_{\dets}(h)$ for the averages of $\vect\phi_4\circ h$ over the projections at pose $\xi$ and over the observations. Because $h$ does not change which poses give equal moments, \cref{thm:pinhole} provides the uniqueness that the corollary assumes, for a generic model of $N\ge20$ points and any compact set of poses with positive depths that contains $\xi^\star$.

\needspace{7\baselineskip}
\begin{corollary}[Stability of the perspective pose]
\label{cor:pinhole-stability}
We fix a 3D model and an invertible affine image map $h$, and write $M(\xi)=\vect\mu_{\pts}(\xi;h)$. Let $\mathcal E_\xi$ be a compact set of poses that contains $\xi^\star$ and places every model point at positive depth, and suppose that $\xi^\star$ is the unique solution of $M(\xi)=M(\xi^\star)$ in $\mathcal E_\xi$. We fix a model length $\ell_{\mathcal Y}>0$ and use the pose distance
\[
d_{\mathcal Y}(\xi,\xi')^2
=\norm{\Rot-\Rot'}^2+\norm{\trans-\trans'}^2/\ell_{\mathcal Y}^2.
\]
For $a_\xi>0$ with $\{\xi\in\mathcal E_\xi:d_{\mathcal Y}(\xi,\xi^\star)\ge a_\xi\}$ nonempty, we define $\Gamma_\xi(a_\xi)$ as the minimum of $\norm{M(\xi)-M(\xi^\star)}$ over that set. For an observed vector $\vect\mu_{\dets}(h)$ and a fitted pose $\hat\xi\in\mathcal E_\xi$, the observation error is $\epsilon_{\mathrm{obs}}=\norm{\vect\mu_{\dets}(h)-M(\xi^\star)}$ and the fit residual is $\epsilon_{\mathrm{res}}=\norm{M(\hat\xi)-\vect\mu_{\dets}(h)}$. Then $\Gamma_\xi(a_\xi)>0$, and
\begin{equation}
\label{eq:pinhole-stability-gap}
\epsilon_{\mathrm{obs}}+\epsilon_{\mathrm{res}}<\Gamma_\xi(a_\xi)
\quad\text{implies}\quad d_{\mathcal Y}(\hat\xi,\xi^\star)<a_\xi.
\end{equation}
In particular, if the residual of the fitted pose exceeds the residual $\epsilon_{\mathrm{obs}}$ of the true pose by at most $\epsilon_{\mathrm{fit}}\ge0$, so that $\epsilon_{\mathrm{res}}\le\epsilon_{\mathrm{obs}}+\epsilon_{\mathrm{fit}}$, then $2\epsilon_{\mathrm{obs}}+\epsilon_{\mathrm{fit}}<\Gamma_\xi(a_\xi)$ suffices.
\end{corollary}
\begin{proof}
Positive depths make $M$ continuous on $\mathcal E_\xi$. The poses of $\mathcal E_\xi$ at distance at least $a_\xi$ from $\xi^\star$ form a compact set, and uniqueness makes their moment difference strictly positive, so the minimum $\Gamma_\xi(a_\xi)$ is positive. The triangle inequality gives $\norm{M(\hat\xi)-M(\xi^\star)}\le\epsilon_{\mathrm{res}}+\epsilon_{\mathrm{obs}}$. If $d_{\mathcal Y}(\hat\xi,\xi^\star)\ge a_\xi$, the definition of $\Gamma_\xi(a_\xi)$ gives $\norm{M(\hat\xi)-M(\xi^\star)}\ge\Gamma_\xi(a_\xi)$, which contradicts $\epsilon_{\mathrm{res}}+\epsilon_{\mathrm{obs}}<\Gamma_\xi(a_\xi)$. Substituting $\epsilon_{\mathrm{res}}\le\epsilon_{\mathrm{obs}}+\epsilon_{\mathrm{fit}}$ gives the final assertion.
\end{proof}

On complete inputs, image noise enters through $\epsilon_{\mathrm{obs}}$, which \cref{lem:moments} bounds before the map $h$ is applied. If the observations are images of only some of the model points, the difference between their average and the average over all projections also enters $\epsilon_{\mathrm{obs}}$. With outliers, moment matching replaces the monomials by bounded features, which limit the influence of the outliers on the averages but need not remove their bias (\cref{supp:lem-contamination,lem:colocated}). When $\Gamma_\xi(a_\xi)$ is small, a small moment discrepancy can coexist with a pose error of at least $a_\xi$. A large fit residual can indicate a failed search.

\subsection{Orientation from normalized moments}
\label{sec:supp_proof_identifiability}

On complete inputs, Step 1 also places initial cameras at orientations fitted under an affine camera of rank two, such as scaled orthography. This section shows that these orientations have errors at most proportional to the moment errors. We center both point sets and scale each to identity covariance, giving $\tilde\pts$ and $\tilde\dets$. Their means and covariances then agree, and only the orientation remains unknown. Complete noiseless affine observations satisfy $\tilde\dets=O\tilde\pts$ for an orientation $O\in\Vset=\{O\in\R^{2\times3}:OO^\top=I_2\}$, the first two rows of a rotation~\cite{hyvarinen2000ica}. We therefore determine the orientation from the cubic and quartic moments. Let the moment pair $T\in\R^{25}$ collect the 10 cubic and 15 quartic moments of $\tilde\pts$, and let $\Fmap_T(O)$ and $T_{\dets}$ collect those of the projected model $O\tilde\pts$ and of $\tilde\dets$. For an estimate $\hat O$ of the true orientation $O^\star$, the fit residual $\fitres=\norm{\Fmap_T(\hat O)-T_{\dets}}$ measures how well the estimate fits the observed moments, and the observation error $\obserr=\norm{T_{\dets}-\Fmap_T(O^\star)}$ measures their noise.

\begin{theorem}[Orientation from normalized moments]
\label{thm:identifiability}
Let both point sets have positive definite covariance. For every moment pair $T$ outside a set of zero volume in $\R^{25}$, distinct orientations give distinct moments $\Fmap_T(O)$, and there is a constant $\Gmod>0$ such that for all $O^\star,\hat O\in\Vset$
\begin{equation*}
\norm{\hat O-O^\star}\le(\fitres+\obserr)/\Gmod.
\end{equation*}
A model of $N\ge34$ points drawn from a joint density has such a moment pair with probability one.
\end{theorem}

Here $\Gmod$ is the greatest lower bound of the change of the projected moments per unit change of orientation. It is zero for symmetric models such as the cube (\cref{supp:lem-gamma}) and small for nearly symmetric ones (\cref{cor:near-symmetry}). The theorem bounds the orientation error linearly by the moment errors, with the factor $1/\Gmod$, before any correspondence is formed. Under perspective projection, the normalized observations are in general not the image of the normalized model under any orientation, because the depths of the model points vary. The observation error $\obserr$ then contains this departure from an affine camera together with the image noise. Step 1 therefore uses the affine orientations as starting values, and the perspective fit of Step 2 refines them with the individual depths. The uniqueness of the perspective moment fit bounds the pose error by the moment errors, through a modulus without an explicit form (\cref{cor:pinhole-stability}). In the depth study of \cref{tab:depth}, the affine initial poses have median rotation errors of $9\degree$ to $161\degree$ before perspective fitting and EM, and the full solver is within $1\degree$ on 88 of 90 poses.

\para{Normalized point sets.}
We center the model points at their mean $\bar{\vect y}$ and normalize them by their empirical covariance $\Sigma_{\pts}=N^{-1}\sum_j(\vect y_j-\bar{\vect y})(\vect y_j-\bar{\vect y})^\top$. This gives the normalized points $\tilde{\vect y}_j=\Sigma_{\pts}^{-1/2}(\vect y_j-\bar{\vect y})$, which form the normalized 3D point set $\tilde\pts$. We normalize the calibrated observations in the same way by their own mean and covariance, which gives $\tilde\dets$. Both covariances are positive definite by assumption, so both normalizations are defined. The next lemma proves the reduction stated before \cref{thm:identifiability}, namely that, for complete noiseless affine observations, the two normalized point sets differ only by an orientation.

\begin{lemma}[Affine normalization]
\label{supp:lem-normalization}
Suppose that the 3D point set $\pts=\{\vect y_j\}_{j=1}^N$ has positive definite covariance $\Sigma_{\pts}$, that $A\in\R^{2\times3}$ has rank two, and that $\vect b\in\R^2$ is an image translation. Let $\sigma$ be a permutation, and let $\vect p_i=A\vect y_{\sigma(i)}+\vect b$ be complete affine projections without measurement error, with mean $\bar{\vect p}$. Then their covariance $\Sigma_p=A\Sigma_{\pts}A^\top$ is positive definite, and the normalized projections $\tilde{\vect p}_i=\Sigma_p^{-1/2}(\vect p_i-\bar{\vect p})$ satisfy
\begin{equation}
\tilde{\vect p}_i=O\tilde{\vect y}_{\sigma(i)},\qquad
O=\Sigma_p^{-1/2}A\Sigma_{\pts}^{1/2},\qquad OO^\top=I_2.
\end{equation}
\end{lemma}
\begin{proof}
The permutation $\sigma$ reorders the points, so the projections have mean $A\bar{\vect y}+\vect b$ and covariance $A\Sigma_{\pts}A^\top$, and centering removes $\vect b$. For a nonzero vector $\vect r\in\R^2$, the rank of $A$ gives $A^\top\vect r\ne0$, and hence $\vect r^\top\Sigma_p\vect r=(A^\top\vect r)^\top\Sigma_{\pts}(A^\top\vect r)>0$. Substituting this covariance into $OO^\top$ gives $OO^\top=\Sigma_p^{-1/2}\Sigma_p\Sigma_p^{-1/2}=I_2$. Finally, $\tilde{\vect p}_i=\Sigma_p^{-1/2}A(\vect y_{\sigma(i)}-\bar{\vect y})=\Sigma_p^{-1/2}A\Sigma_{\pts}^{1/2}\tilde{\vect y}_{\sigma(i)}=O\tilde{\vect y}_{\sigma(i)}$.
\end{proof}

Suppose that the observations are complete and noiseless and that the calibrated observations are $\vect u_{\mathrm{cal},i}=A\vect y_{\sigma(i)}+\vect b$ for a matrix $A\in\R^{2\times3}$ of rank two and a vector $\vect b\in\R^2$. \cref{supp:lem-normalization} with $\vect p_i=\vect u_{\mathrm{cal},i}$ then gives the normalized observations $\tilde{\vect u}_i=O^\star\tilde{\vect y}_{\sigma(i)}$ for a matrix $O^\star$ with orthonormal rows, that is, for an orientation $O^\star\in\Vset$. Moments do not depend on the order of the points, so the correspondences $\sigma$ drop out, and the normalized observed moments are $T_{\dets}=\Fmap_T(O^\star)$.

\para{Pixel \& calibrated coordinates.}
\cref{thm:identifiability} normalizes the calibrated observations. Normalizing the pixel coordinates gives the same orientation up to an orthogonal map of the image. Removing the intrinsics is an invertible affine map $\vect u\mapsto C_K\vect u+\vect b_K$, with an invertible matrix $C_K\in\R^{2\times2}$, that sends a pixel $\vect u$ to its calibrated point $\vect u_{\mathrm{cal}}$. If the pixel covariance $\Sigma_{\dets}$ is positive definite, the normalized pixel coordinates are $\vect e_{\mathrm{pix}}=\Sigma_{\dets}^{-1/2}(\vect u-\bar{\vect u})$. The calibrated covariance is $C_K\Sigma_{\dets}C_K^\top$, so normalization after calibration gives $\tilde{\vect u}=Q_K\vect e_{\mathrm{pix}}$ with $Q_K=(C_K\Sigma_{\dets}C_K^\top)^{-1/2}C_K\Sigma_{\dets}^{1/2}$, which is orthogonal because $Q_KQ_K^\top=I_2$. If $\tilde{\vect u}_i=O^\star\tilde{\vect y}_{\sigma(i)}$, the normalized pixel coordinates of observation~$i$ are therefore $Q_K^\top O^\star\tilde{\vect y}_{\sigma(i)}$, and $Q_K^\top O^\star$ is again an orientation in $\Vset$. Perspective fitting does not need this comparison, because $\mathcal L_4$ applies one fixed map $h$ to the observations and to the projections.

\para{Separation constant.}
The constant of \cref{thm:identifiability} is the greatest lower bound of the ratio between a change of the projected moments and the change of orientation that causes it, in the Frobenius norm,
\begin{equation}
\label{eq:statement-4_method-2}
\Gmod=\inf_{\substack{O,O'\in\Vset\\O\ne O'}}
\frac{\norm{\Fmap_T(O)-\Fmap_T(O')}}{\norm{O-O'}}.
\end{equation}
Outside the exceptional set defined in the proof, the derivative of $\Fmap_T$ in orientation also has rank three at every $O\in\Vset$. Step~4 of the proof uses this to show $\Gmod>0$.

\para{Outline of the proof.}
The proof has four parts. First, we write each moment tensor as a homogeneous polynomial, so that $\Fmap_T(O)$ becomes the restriction of two polynomials to the plane spanned by the rows of $O$. Second, \cref{supp:lem-ranks} counts the independent linear conditions on $T$ under which two distinct orientations have equal moments, and those under which the derivative of $\Fmap_T$ vanishes in a tangent direction. Third, the proof of \cref{thm:identifiability} combines these counts with the dimensions of the families of orientation pairs and tangent directions. Steps~1 to~3 of that proof show that the exceptional moment pairs $T$ form a set of dimension at most $24$ in the $25$-dimensional space of moment pairs. Outside this set, compactness gives $\Gmod>0$ (Step~4), and the definition of $\Gmod$ gives the error bound (Step~5). Fourth, a separate proof shows that a model of $N\ge34$ points drawn from a joint density avoids the exceptional set with probability one. After the proof, \cref{supp:lem-gamma} shows that $\Gmod$ depends continuously on $T$ and vanishes for symmetric models. Together, these two properties imply that nearly symmetric models have small $\Gmod$ (\cref{cor:near-symmetry}).

\para{Moments as forms.}
A symmetric tensor of order $\order$ on $\R^3$ is equivalent to a form of degree $\order$, which is a polynomial in three variables whose monomials all have total degree $\order$. We use forms because restricting a polynomial to a plane is simpler to analyze than contracting a tensor. We give each form the Frobenius norm of its symmetric tensor, and a pair of forms the direct-sum norm of the two tensors, as in \cref{thm:identifiability}. The value of $\Gmod$ depends on this choice, because the Frobenius norm weights each distinct entry $m_\alpha$ by its multiplicity (\cref{sec:supp_moment_coordinates}). Unweighted norms of the distinct polynomial coefficients need not give the same value. In this representation, the normalized model moment $\mom_\order(\tilde\pts)$ is the form
\begin{equation}\label{eq:supp-form}
 F_\order(\vect\zeta)=\bigl\langle\mom_\order(\tilde\pts),\vect\zeta^{\otimes \order}\bigr\rangle=\frac1N\sum_{j=1}^N(\vect\zeta\cdot\tilde{\vect y}_j)^\order,\quad\vect\zeta\in\R^3 .
\end{equation}
We write $\mathcal S_\order$ for the space of forms of degree $\order$ on $\R^3$, of dimension $\binom{\order+2}2$, and $\mathcal H_\order$ for the space of forms of degree $\order$ on $\R^2$, of dimension $\order+1$. The pair $T=(F_3,F_4)$ is then a point of $\mathcal T=\mathcal S_3\oplus\mathcal S_4$. This space has dimension $10+15=25$ and is the space of moment pairs in \cref{thm:identifiability}.

\para{Projected moments as restrictions.}
The two orthonormal rows of an orientation $O\in\Vset$ span a plane in $\R^3$. We write $\vect\nu$ for a unit normal of this plane, so that $O\vect\nu=0$ and $O^\top O=I_3-\vect\nu\vect\nu^\top$ is the orthogonal projection onto the plane $\vect\nu^\perp$. For image coordinates $\vect u\in\R^2$, we have $\vect u\cdot O\tilde{\vect y}_j=O^\top\vect u\cdot\tilde{\vect y}_j$. The projected moment of order $\order$ is therefore the restriction of $F_\order$ to the plane, written in image coordinates,
\begin{equation}\label{eq:supp-restrict}
 (\mathcal R_OF_\order)(\vect u)=F_\order(O^\top\vect u),\quad\vect u\in\R^2.
\end{equation}
The cubic and quartic projected moments form the pair
\begin{equation}\label{eq:supp-MT}
 \Fmap_T(O)=\bigl(\mathcal R_OF_3,\mathcal R_OF_4\bigr)\in\mathcal H_3\oplus\mathcal H_4,
\end{equation}
in a space of dimension $4+5=9$. The map $\Fmap_T$ is linear in $T$ and polynomial in $O$, and it is defined for every $T\in\mathcal T$, whether or not $T$ comes from a point set. Each restriction $\mathcal R_O$ maps $\mathcal S_\order$ onto $\mathcal H_\order$, because a form $f\in\mathcal H_\order$ has the extension $\vect\zeta\mapsto f(O\vect\zeta)$, whose restriction is $f$. Two orientations $O$ and $O'$ have equal projected moments exactly when $T$ lies in the kernel of the linear operator $\Delta_{O,O'}:\mathcal T\to\mathcal H_3\oplus\mathcal H_4$ given by
\begin{equation}\label{eq:supp-delta}
 \Delta_{O,O'}T=\Fmap_T(O)-\Fmap_T(O').
\end{equation}
The rank of $\Delta_{O,O'}$ is the number of independent linear conditions on $T$ under which the two orientations share their moments. The following lemma computes this rank. It also computes the rank of the derivative of $\Fmap_T$ in a tangent direction, which plays the same role for two orientations that are infinitely close.

\begin{lemma}[Rank of the moment difference of two orientations]\label{supp:lem-ranks}
Let $O$ and $O'$ be orientations with unit normals $\vect\nu$ and $\vect\nu'$, distinct in parts~(a) and~(b).

\textup{(a)} If $\vect\nu'\ne\pm\vect\nu$, let $\vect d_{\rm L}$ be a unit vector orthogonal to both normals, and put $\vect o=O\vect d_{\rm L}$ and $\vect o'=O'\vect d_{\rm L}$. The rank of $\Delta_{O,O'}$ is $9$ if $\vect o'\ne\pm\vect o$. It is $7$ if $\vect o'=\vect o$, and $8$ if $\vect o'=-\vect o$.

\textup{(b)} If $\vect\nu'=\pm\vect\nu$, then $O'=GO$ for an orthogonal matrix $G\ne I_2$. The rank of $\Delta_{O,O'}$ is $4$ if $G$ is a reflection or the rotation by $\pi$. It is $6$ if $G$ is a rotation by $\pm\pi/2$ or $\pm2\pi/3$, and $8$ for every other rotation $G\ne I_2$.

\textup{(c)} Let $R_\perp$ be the rotation of the plane by $\pi/2$. The tangent space of $\Vset$ at $O$ consists of the matrices $\dot O=\omega R_\perp O+\vect q\vect\nu^\top$ with $\omega\in\R$ and $\vect q\in\R^2$. The derivative of $\Fmap_T$ at $O$ in the direction $\dot O$, as a linear map of $T$, has rank $9$ if $\omega\ne0\ne\vect q$. Its rank is $7$ if $\omega=0\ne\vect q$, and $8$ if $\vect q=0\ne\omega$.
\end{lemma}
\begin{proof}
Each operator acts on the cubic and the quartic component of $T$ separately. In all three parts, we therefore compute the rank for each degree $\order\in\{3,4\}$ and add the two ranks.

\emph{Part (a).} The two planes are distinct and meet in the line spanned by $\vect d_{\rm L}$. We first determine which pairs $(\mathcal R_OF,\mathcal R_{O'}F)$ arise as $F$ ranges over $\mathcal S_\order$, and then which differences between the restrictions arise. A form $F\in\mathcal S_\order$ satisfies $\mathcal R_OF=0$ exactly when it vanishes on the plane $\vect\nu^\perp$, that is, when it is divisible by $\vect\nu\cdot\vect\zeta$. A form that vanishes on both planes is therefore divisible by $(\vect\nu\cdot\vect\zeta)(\vect\nu'\cdot\vect\zeta)$. Dividing out these factors identifies the kernel of $F\mapsto(\mathcal R_OF,\mathcal R_{O'}F)$ with $\mathcal S_{\order-2}$, so this map has rank $\binom{\order+2}2-\binom\order2=2\order+1$. The restrictions $\mathcal R_OF$ at $\vect o$ and $\mathcal R_{O'}F$ at $\vect o'$ both take the value $F(\vect d_{\rm L})$, because $O^\top\vect o=\vect d_{\rm L}=O'^\top\vect o'$. The image of the map is therefore contained in
\begin{equation}
 \mathcal W_\order=\bigl\{(f,f')\in\mathcal H_\order\times\mathcal H_\order:f(\vect o)=f'(\vect o')\bigr\}.
\end{equation}
Since $\mathcal W_\order$ also has dimension $2\order+1$, the image equals $\mathcal W_\order$.

The degree-$\order$ component of $\Delta_{O,O'}$ subtracts the two restrictions, so its image is $\{f-f':(f,f')\in\mathcal W_\order\}$. Suppose first that some form of $\mathcal H_\order$ takes different values at $\vect o$ and $\vect o'$. The image is then all of $\mathcal H_\order$. Indeed, given $f_0\in\mathcal H_\order$, we choose $f'\in\mathcal H_\order$ with $f'(\vect o')-f'(\vect o)=f_0(\vect o)$ and put $f=f_0+f'$, so that $(f,f')\in\mathcal W_\order$ and $f-f'=f_0$. Suppose now that all forms of degree $\order$ agree at $\vect o$ and $\vect o'$. Every difference then vanishes at $\vect o$, and every $f$ with $f(\vect o)=0$ is the difference between the components of $(f,0)\in\mathcal W_\order$. The image is then defined by one linear condition, so its dimension drops by one. Because $\vect o$ and $\vect o'$ are unit vectors, all forms of degree $\order$ agree at them exactly when $\vect o'=\vect o$, or when $\vect o'=-\vect o$ and $\order$ is even. The ranks are therefore $4+5$, $3+4$ and $4+4$ in the three cases.

\emph{Part (b).} The two orientations share their plane and differ only in the image coordinates on it. The rows of $O'$ lie in the row space $\vect\nu^\perp$ of $O$, so $O'=O'O^\top O=GO$ with $G=O'O^\top$. The matrix $G$ is orthogonal because $GG^\top=O'(I_3-\vect\nu\vect\nu^\top)O'^\top=I_2$, and $G\ne I_2$ because $O'\ne O$. By \cref{eq:supp-restrict}, $\mathcal R_{O'}F=(\mathcal R_OF)\circ G^\top$. Since $\mathcal R_O$ is onto, the degree-$\order$ rank of $\Delta_{O,O'}$ is $\order+1$ minus the dimension of the space of forms $f\in\mathcal H_\order$ with $f\circ G^\top=f$. It remains to count these invariant forms.

For a rotation $G$ by an angle $\theta$, complex coordinates make the action diagonal. In the coordinate $z=u_1+\mathrm iu_2$, the monomials $z^j\bar z^{\order-j}$ with $0\le j\le \order$ span the complex forms of degree $\order$, and $f\mapsto f\circ G^\top$ multiplies them by $e^{-\mathrm i(2j-\order)\theta}$. This operator is real, so its fixed subspaces over $\R$ and over $\mathbb C$ have the same dimension, namely the number of indices $j$ with $(2j-\order)\theta\in2\pi\mathbb Z$. Since $G\ne I_2$, the angle $\theta$ is not a multiple of $2\pi$. For $\order=3$, the integers $2j-\order$ are $\pm1$ and $\pm3$, so this dimension is $2$ if $3\theta\in2\pi\mathbb Z$ and $0$ otherwise. For $\order=4$, the integers are $0$, $\pm2$ and $\pm4$, so the dimension is $1$, plus $2$ if $2\theta\in2\pi\mathbb Z$, plus $2$ if $4\theta\in2\pi\mathbb Z$. For a reflection $G$, choosing coordinate axes along and perpendicular to its axis gives $(u_1,u_2)\mapsto(u_1,-u_2)$. The invariant forms are then those even in $u_2$, with dimensions $2$ for $\order=3$ and $3$ for $\order=4$. Subtracting these dimensions from $4+5$ gives the ranks of part~(b).

\emph{Part (c).} A tangent vector is a first-order change of an orientation that keeps its rows orthonormal. Differentiating $OO^\top=I_2$ shows that the tangent vectors are the matrices $\dot O$ with $\dot O O^\top+O\dot O^\top=0$, a space of dimension three. The matrices $\omega R_\perp O+\vect q\vect\nu^\top$ satisfy this equation because $O\vect\nu=0$ and $R_\perp^\top=-R_\perp$. They also span three dimensions, so they form the tangent space. The term $\omega R_\perp O$ rotates the image coordinates within the plane, and the term $\vect q\vect\nu^\top$ tilts the plane in $\R^3$.

To differentiate the restriction, we expand $F\in\mathcal S_\order$ in powers of the coordinate normal to the plane,
\begin{equation}
 F(\vect\zeta)=\sum_{j=0}^\order(\vect\nu\cdot\vect\zeta)^jF^{(j)}(O\vect\zeta),\quad F^{(j)}\in\mathcal H_{\order-j}.
\end{equation}
As $F$ ranges over $\mathcal S_\order$, the pair $(F^{(0)},F^{(1)})$ ranges over $\mathcal H_\order\times\mathcal H_{\order-1}$. Since $(O+\tau\dot O)^\top\vect u=O^\top(\vect u+\tau\omega R_\perp^\top\vect u)+\tau(\vect q\cdot\vect u)\vect\nu$, the derivative of the restriction is
\begin{equation}
\begin{gathered}
 \frac{d}{d\tau}\mathcal R_{O+\tau\dot O}F\Big|_{\tau=0}=\omega\partial_{R_\perp}F^{(0)}+(\vect q\cdot\vect u)F^{(1)},\\
 (\partial_{R_\perp} f)(\vect u)=\nabla f(\vect u)\cdot R_\perp^\top\vect u.
\end{gathered}
\end{equation}
The first term comes from the rotation within the plane and the second from the tilt of the plane. Because $F^{(0)}$ and $F^{(1)}$ vary independently, the image of the derivative is the sum of the images of the two terms.

Differentiating the rotation factors of part~(b) at $\theta=0$ shows that $\partial_{R_\perp}$ has the eigenvalues $-\mathrm i(2j-\order)$ for $0\le j\le \order$. The operator is therefore invertible on $\mathcal H_3$, and its image in $\mathcal H_4$ consists of the forms with zero mean on the unit circle, a space of dimension four. Multiplication by $\vect q\cdot\vect u$ with $\vect q\ne0$ is injective from $\mathcal H_{\order-1}$ to $\mathcal H_\order$, so the tilt term has rank $\order$. If $\omega\ne0\ne\vect q$, the cubic component is onto because $\partial_{R_\perp}$ is invertible on $\mathcal H_3$. The quartic form $(\vect q\cdot\vect u)^4$ lies in the image of the tilt term and has positive mean on the unit circle. It therefore supplies the direction missing from the image of $\partial_{R_\perp}$, so the quartic component is also onto and the rank is $4+5$. If $\omega=0\ne\vect q$, only the tilt term remains, with rank $3+4$. If $\vect q=0\ne\omega$, only the rotation term remains, with rank $4+4$.
\end{proof}

\begin{table}[tp]
\centering\footnotesize
\setlength{\tabcolsep}{4pt}
\caption{\textbf{Families of distinct orientation pairs} in the proof of \cref{thm:identifiability}. The columns give the dimension of each family $\mathcal D$, the rank of $\Delta_{O,O'}$ on the family (\cref{supp:lem-ranks}), and the bound of \cref{eq:supp-count} on the dimension of the set of moment pairs $T$ for which some pair of orientations in the family has equal moments.}
\label{tab:supp_strata}
\begin{tabular*}{\columnwidth}{@{\extracolsep{\fill}}lccc@{}}
\toprule
Orientation pairs $(O,O')$ & $\dim\mathcal D$ & Rank & Bound \\
\midrule
$\vect\nu'\ne\pm\vect\nu$, $\vect o'\ne\pm\vect o$ & $6$ & $9$ & $22$\\
$\vect\nu'\ne\pm\vect\nu$, $\vect o'=\vect o$ & $5$ & $7$ & $23$\\
$\vect\nu'\ne\pm\vect\nu$, $\vect o'=-\vect o$ & $5$ & $8$ & $22$\\
$\vect\nu'=\pm\vect\nu$, rotation by $\pm\pi/2$ or $\pm2\pi/3$ & $3$ & $6$ & $22$\\
$\vect\nu'=\pm\vect\nu$, rotation by $\pi$ & $3$ & $4$ & $24$\\
$\vect\nu'=\pm\vect\nu$, any other rotation & $4$ & $8$ & $21$\\
$\vect\nu'=\pm\vect\nu$, reflection $\operatorname{diag}(1,-1)$ & $3$ & $4$ & $24$\\
\bottomrule
\end{tabular*}
\end{table}

\begin{proof}[Proof of \cref{thm:identifiability}]
Steps~1 to~3 show that the moment pairs $T$ for which $\Fmap_T$ fails to be one-to-one, or has a derivative of rank less than three, form a set of dimension at most $24$. The sets involved are semialgebraic, that is, built by finite unions and intersections from polynomial equations and inequalities. As in the proof of \cref{thm:pinhole}, we use three facts about their dimension~\cite{bochnak1998real}. First, a semialgebraic map does not increase dimension. Second, the dimension of its domain is at most the dimension of its image plus the largest dimension of a fiber, the set of points mapped to one point. Third, a semialgebraic set of smaller dimension than its ambient Euclidean space has Lebesgue measure zero and lies in the zero set of a nonzero polynomial.

\emph{Step 1. Two orientations with the same moments.} Let $\mathcal T_{\mathrm{inj}}$ be the set of $T\in\mathcal T$ with $\Delta_{O,O'}T=0$ for some $O\ne O'$, so that $\Fmap_T$ is one-to-one exactly when $T\notin\mathcal T_{\mathrm{inj}}$. We divide the pairs of distinct orientations into the families of \cref{tab:supp_strata}. On each family, the rank of $\Delta_{O,O'}$ is constant by \cref{supp:lem-ranks}. We consider the triples $(O,O',T)$ with $(O,O')$ in a family $\mathcal D$ and $\Delta_{O,O'}T=0$. The fiber of these triples over a fixed pair $(O,O')$ is the kernel of $\Delta_{O,O'}$, of dimension $25$ minus the rank, and the pairs contribute at most $\dim\mathcal D$ further parameters. By the first two facts, the set of $T$ that occur in these triples has dimension at most
\begin{equation}\label{eq:supp-count}
 \dim\mathcal D+25-\rank\Delta_{O,O'}.
\end{equation}

The families are semialgebraic. We choose the normals as the cross products of the rows of $O$ and of $O'$, which are polynomial in the entries. Then $\vect\nu'=\pm\vect\nu$ exactly when $O'(I_3-O^\top O)=0$. On the families with distinct planes, the cross product $\vect d_{\times}=\vect\nu\times\vect\nu'$ is a nonzero multiple of $\vect d_{\rm L}$, so $\vect o'=\pm\vect o$ exactly when $O'\vect d_{\times}=\pm O\vect d_{\times}$. The dimensions of the families follow from $\dim\Vset=3$. All pairs form a set of dimension $6$. A pair with $\vect o'=\pm\vect o$ is determined, up to finitely many choices, by $O$, the unit vector $\vect o$ and the angle between the two planes about $\vect d_{\rm L}=O^\top\vect o$, which gives dimension $3+1+1=5$. A pair with $\vect\nu'=\pm\vect\nu$ has the form $(O,GO)$, which gives dimension $3$ when $G$ is one of finitely many rotations and $4$ when $G$ ranges over all rotations.

The reflection pairs $(O,GO)$ also have dimension $4$, and \cref{eq:supp-count} would give $25$ for them, so we remove one redundant parameter. Rotating the image coordinates of both orientations by the same planar rotation $G_1$ preserves equal moments. Indeed, \cref{eq:supp-restrict} gives $\Delta_{G_1O,G_1GO}T=(\Delta_{O,GO}T)\circ G_1^\top$, and composition with $G_1^\top$ is invertible, so the two kernels agree. As the rotation $G_1$ varies, the conjugate $G_1GG_1^{-1}$ runs over all reflections. It therefore suffices to consider the pairs $(O,\operatorname{diag}(1,-1)\,O)$, a family of dimension $3$ with kernels of dimension $25-4$. The largest bound in \cref{tab:supp_strata} is $24$, so $\dim\mathcal T_{\mathrm{inj}}\le24$.

\emph{Step 2. An orientation with a degenerate derivative.} Let $\mathcal T_{\mathrm{imm}}$ be the set of $T$ for which, at some orientation, the derivative of $\Fmap_T$ vanishes in some nonzero tangent direction $\dot O$. For $T\notin\mathcal T_{\mathrm{imm}}$, the derivative is injective on the three-dimensional tangent space at every orientation, that is, it has rank three. Step~4 needs this property, because two orientations that approach each other produce such a direction in the limit. The condition does not change when $\dot O$ is multiplied by a nonzero scalar, so only the direction of $\dot O$ contributes parameters. The families of orientations and tangent directions have dimension $3+2$ when $\omega\ne0\ne\vect q$, dimension $3+1$ when $\omega=0$, and dimension $3$ when $\vect q=0$. By \cref{supp:lem-ranks}(c), the corresponding derivatives have ranks $9$, $7$ and $8$. The count of \cref{eq:supp-count} gives $21$, $22$ and $20$, so $\dim\mathcal T_{\mathrm{imm}}\le22$.

\emph{Step 3. A set of measure zero.} The exceptional set $\mathcal T_{\mathrm{exc}}=\mathcal T_{\mathrm{inj}}\cup\mathcal T_{\mathrm{imm}}$ is semialgebraic of dimension at most $24$ in the $25$-dimensional space $\mathcal T$. By the third fact, it has Lebesgue measure zero and lies in a proper algebraic subset of $\mathcal T$. For $T\notin\mathcal T_{\mathrm{exc}}$, the map $\Fmap_T$ is one-to-one on $\Vset$ and its derivative has rank three at every orientation. The first property is the first claim of \cref{thm:identifiability}, and Step~4 uses the second.

\emph{Step 4. The lower bound is positive.} Injectivity alone does not give $\Gmod>0$, because the ratio in \cref{eq:statement-4_method-2} could tend to zero along pairs of distinct orientations that approach each other. Compactness of $\Vset$ and Step~2 exclude this. We fix $T\notin\mathcal T_{\mathrm{exc}}$ and suppose that $\Gmod=0$. There are then distinct orientations $O_i,O_i'\in\Vset$ for $i=1,2,\ldots$ such that
\begin{equation}\label{eq:supp-modulus-sequence}
 \frac{\norm{\Fmap_T(O_i)-\Fmap_T(O_i')}}{\norm{O_i-O_i'}}\xrightarrow[i\to\infty]{}0.
\end{equation}
By compactness, we may pass to a subsequence with $O_i\to O$ and $O_i'\to O'$. If $O\ne O'$, the numerators tend to zero because $\Vset$ is bounded, and continuity gives $\Fmap_T(O)=\Fmap_T(O')$, which contradicts injectivity. If $O=O'$, we pass to a further subsequence along which the normalized differences $(O_i'-O_i)/\norm{O_i'-O_i}$ converge to a matrix $\dot O$ of unit norm. Subtracting the equations $O_iO_i^\top=I_2$ and $O_i'O_i'^\top=I_2$ and dividing by $\norm{O_i'-O_i}$ gives $\dot O O^\top+O\dot O^\top=0$ in the limit, so $\dot O$ is tangent at $O$. The map $\Fmap_T$ extends to a polynomial, hence continuously differentiable, map on $\R^{2\times3}$. The moment difference therefore equals the derivative at $O$ applied to $O_i'-O_i$, up to a remainder of order $o(\norm{O_i'-O_i})$. By \cref{eq:supp-modulus-sequence}, the derivative at $O$ vanishes in the direction $\dot O$, which contradicts Step~2. Hence $\Gmod>0$.

\emph{Step 5. The error bound.} Let $T\notin\mathcal T_{\mathrm{exc}}$, and let $O^\star$ and $\hat O$ lie in $\Vset$. The definition of $\Gmod$ gives $\Gmod\norm{\hat O-O^\star}\le\norm{\Fmap_T(\hat O)-\Fmap_T(O^\star)}$. The triangle inequality through $T_{\dets}$ bounds the right-hand side by $\fitres+\obserr$, and dividing by $\Gmod>0$ proves the bound
\begin{equation}
\label{eq:orientation-stability}
\norm{\hat O-O^\star}\le\frac{\fitres+\obserr}{\Gmod}
\end{equation}
of \cref{thm:identifiability}. For complete noiseless affine observations, $T_{\dets}=\Fmap_T(O^\star)$ by the normalization above, so $\obserr=0$. Any orientation $\hat O$ that fits the observed moments exactly has $\fitres=0$ and therefore equals $O^\star$, so these observations determine the normalized orientation.
\end{proof}

\begin{proof}[Proof of \cref{thm:identifiability} for sampled models]
Step~3 of the proof above shows that $\mathcal T_{\mathrm{exc}}$ has measure zero among moment pairs. A model, however, is a point of $\R^{3N}$, and a set of measure zero among moment pairs need not have a preimage of measure zero among models. We therefore show that the map from point coordinates to $T$ is a submersion at almost every point set, meaning that its derivative maps onto the space of moment pairs. The pair $T$ depends on the point set through its $34$ power sums, one sum over $j$ for each monomial of degree one to four in the three coordinates, and we treat the map to the power sums and the map from them to $T$ in turn.

\emph{Step 1. From the point set to its power sums.} The Jacobian of the power-sum map has one row per monomial, and the row of a monomial lists its gradients at $\vect y_1,\dots,\vect y_N$. These rows are linearly dependent exactly when a nonzero linear combination of the $34$ monomials has zero gradient at every $\vect y_j$. Such a combination is a nonconstant polynomial, so the set where its gradient vanishes is a proper algebraic subset of $\R^3$, of dimension at most two. Up to scale, these combinations form a family of dimension $33$. Allowing each of the $N$ points to lie in the corresponding set gives a semialgebraic set of dimension at most $33+2N$, which contains every point set at which the Jacobian loses rank. This set has measure zero in $\R^{3N}$ because $33+2N<3N$.

\emph{Step 2. From the power sums to $T$.} The derivative of the map from power sums to moment pairs has full row rank wherever the covariance is positive definite. With the sums of degrees one and two held fixed, normalization is a fixed invertible affine map of $\R^3$, and $(F_3,F_4)$ is an invertible affine function of the sums of degrees three and four.

\emph{Step 3. Conclusion.} For $N\ge34$, the covariance is positive definite with probability one. By Steps~1 and~2, the map to $T$ is a submersion on an open set whose complement has measure zero. After smooth changes of coordinates, a submersion acts locally as a coordinate projection, so the preimage of the null set $\mathcal T_{\mathrm{exc}}$ is null~\cite{lee2012smooth}. The distribution of the point set has a density on $\R^{3N}$, so it assigns probability zero to this preimage.
\end{proof}

\cref{cor:near-symmetry} uses two further properties of $\Gmod$, which the next lemma proves.

\begin{lemma}[Continuity of the lower bound and symmetric models]
\label{supp:lem-gamma}
There is a constant $C$ such that $\abs{\Gmod-\Gamma_{T'}}\le C\norm{T-T'}$ for all moment pairs $T$ and $T'$. If $T=(F_3,F_4)$ satisfies $F_3\circ Q_{\rm s}=F_3$ and $F_4\circ Q_{\rm s}=F_4$ for an orthogonal matrix $Q_{\rm s}\ne I_3$, then $\Gmod=0$ and $T\in\mathcal T_{\mathrm{exc}}$.
\end{lemma}
\begin{proof}
\emph{Continuity.} The map $\Fmap_T(O)$ is linear in $T$ and polynomial in $O$, and $\Vset$ is bounded. There is therefore a constant $C$ such that $\norm{\Fmap_T(O)-\Fmap_T(O')}\le C\norm T\norm{O-O'}$ for all orientations and all $T$. Applying this bound to $T-T'$, and the triangle inequality to the definition of $\Gmod$, gives $\abs{\Gmod-\Gamma_{T'}}\le C\norm{T-T'}$. In particular, the set $\{T:\Gmod>0\}$ is open.

\emph{Symmetry.} We choose an orientation $O$ with $OQ_{\rm s}\ne O$, which exists because $Q_{\rm s}\ne I_3$. The matrix $OQ_{\rm s}$ is also an orientation. The forms are invariant under $Q_{\rm s}^\top=Q_{\rm s}^{-1}$ as well, so $\mathcal R_{OQ_{\rm s}}F_\order(\vect u)=F_\order(Q_{\rm s}^\top O^\top\vect u)=F_\order(O^\top\vect u)$. Hence $\Fmap_T(OQ_{\rm s})=\Fmap_T(O)$, so $\Fmap_T$ is not one-to-one and $\Gmod=0$. The invariance holds when $Q_{\rm s}$ maps the normalized model to itself, and a symmetry of the model gives such a matrix. Suppose that an invertible matrix $L\ne I_3$ and a permutation $\pi$ satisfy $L(\vect y_j-\bar{\vect y})=\vect y_{\pi(j)}-\bar{\vect y}$ for every $j$. A mirror plane or a nontrivial rotational symmetry of the model has this form, because an isometry that maps a finite set to itself fixes its centroid. Reordering the points does not change their covariance, so $L\Sigma_{\pts}L^\top=\Sigma_{\pts}$. The conjugate $Q_{\rm s}=\Sigma_{\pts}^{-1/2}L\Sigma_{\pts}^{1/2}$ is therefore orthogonal, because $Q_{\rm s}Q_{\rm s}^\top=\Sigma_{\pts}^{-1/2}L\Sigma_{\pts}L^\top\Sigma_{\pts}^{-1/2}=I_3$. It differs from $I_3$ and maps $\tilde{\vect y}_j$ to $\tilde{\vect y}_{\pi(j)}$. Its inverse $Q_{\rm s}^\top$ also permutes the normalized points, so \cref{eq:supp-form} gives $F_\order\circ Q_{\rm s}=F_\order$, and the moment pair of the model lies in $\mathcal T_{\mathrm{exc}}$.
\end{proof}

\para{Depth variation \& affine initial poses.}
The affine initial poses of Step 1 replace the depth of every model point by a common reference depth $\refdepth>0$, which adds a feature error before perspective fitting. To bound it, we write the camera coordinates of model point $j$ as $(\vect x_j^{\rm lat},\refdepth(1+\vartheta_j))$, with lateral coordinates $\vect x_j^{\rm lat}\in\R^2$ and relative depth deviation $\vartheta_j$. The affine projection of the point in calibrated coordinates is $\vect p^{\rm aff}_j=\vect x_j^{\rm lat}/\refdepth$. We assume that $|\vartheta_j|\le \vartheta_{\max}<1$ and $\norm{\vect p^{\rm aff}_j}\le\rho_{\rm cal}$ for every $j$, so that $\rho_{\rm cal}$ bounds the radius of the calibrated affine projections. The perspective projection of the point is $\vect p^{\rm aff}_j/(1+\vartheta_j)$, so the two are at most $\rho_{\rm cal}\vartheta_{\max}/(1-\vartheta_{\max})$ apart. For a fixed feature map $\vect\phi$ with Lipschitz constant $\featlip$ on the line segments joining these pairs, Lipschitz continuity and averaging give
\begin{equation}
\label{eq:affine-perspective-feature-error}
\left\lVert\frac1N\sum_j\left[
\vect\phi\!\left(\frac{\vect p^{\rm aff}_j}{1+\vartheta_j}\right)
-\vect\phi(\vect p^{\rm aff}_j)\right]\right\rVert
\le \frac{\featlip\,\rho_{\rm cal}\vartheta_{\max}}{1-\vartheta_{\max}}.
\end{equation}
Monomial features composed with a fixed affine image map, and the Fourier features of \cref{sec:prelim}, have finite Lipschitz constants on every bounded region of the image, so the bound applies to both. The bound concerns the feature averages at a fixed pose and grows with the relative depth deviation $\vartheta_{\max}$. Perspective fitting uses the individual depths, so on complete noiseless inputs it has no error from the affine approximation at the true pose. \cref{tab:depth} compares the affine initial poses with the full solver as the depth variation changes.

\subsection{Near-symmetry \& bounded features}
\label{sec:supp_auxiliary}

This section proves two facts that the main text uses. \cref{cor:near-symmetry} shows that $\Gmod$ is small when the moment pair of a model is close to that of a symmetric model, so the bound of \cref{eq:orientation-stability} is weak for nearly symmetric models. \cref{supp:lem-contamination} proves that, at the true pose, the expected feature averages over the observations and over the projections differ by $\unmatched(\vect\mu_{\dets,0}-\vect\mu_{\pts,0})$, the term that moment matching with outliers subtracts in \cref{def:moments}.

\para{Near-symmetry.}
A model with a symmetry has two distinct orientations with the same projected moments, so its moment pair lies in the exceptional set of \cref{thm:identifiability} (\cref{supp:lem-gamma}). The next corollary shows that moment pairs near such a pair have a small constant $\Gmod$, and hence weak orientation bounds.

\begin{corollary}[Near-symmetric models]
\label{cor:near-symmetry}
Let $T_0$ be a moment pair for which two distinct orientations give identical projected moments. Then $\Gamma_{T_0}=0$, and every moment pair $T$ satisfies $\Gmod\le C\norm{T-T_0}$ with the constant $C$ of \cref{supp:lem-gamma}.
\end{corollary}
\begin{proof}
At $T_0$, the two orientations give a zero numerator in the definition of $\Gamma_{T_0}$ in \cref{eq:statement-4_method-2}, so $\Gamma_{T_0}=0$. \cref{supp:lem-gamma} shows that $\abs{\Gmod-\Gamma_{T'}}\le C\norm{T-T'}$ for all moment pairs $T$ and $T'$, and the choice $T'=T_0$ gives the bound.
\end{proof}

\para{Bounded features with outliers.}
Moment matching with outliers (\cref{def:moments}) uses the fact that, at the true pose, the inliers contribute equally to the average over the observations and to the average over the projections, so the two averages differ through the outliers alone. The next lemma states this identity for the feature means of two distributions. It also bounds the change of a feature average when a fraction of the points is replaced. Without a bound on the features, one replaced point could change the average arbitrarily, so moment matching uses bounded features when outliers may be present.

\begin{lemma}[Bounded features with outliers]
\label{supp:lem-contamination}
Let $\vect\phi:\R^2\to\R^{d_\phi}$ be a measurable feature map with $\sup_{\vect u\in\R^2}\norm{\vect\phi(\vect u)}\le1$. Replacing a fraction $0\le\unmatched\le1$ of the points of a nonempty finite collection of 2D points by other 2D points changes the average of $\vect\phi$ over the collection by at most $2\unmatched$. Consider two distributions on $\R^2$, each a mixture of an inlier component with weight $1-\unmatched$ and an outlier component with weight $\unmatched$. Let the two inlier components have the same feature mean $\vect\phi_{\mathrm{in}}$, and let the outlier components have feature means $\vect\mu_{\dets,0}$ and $\vect\mu_{\pts,0}$. Then the feature means $\vect\mu_1$ and $\vect\mu_2$ of the two distributions satisfy
\begin{equation}
\begin{aligned}
\vect\mu_1&=(1-\unmatched)\vect\phi_{\mathrm{in}}+\unmatched\vect\mu_{\dets,0},\\
\vect\mu_2&=(1-\unmatched)\vect\phi_{\mathrm{in}}+\unmatched\vect\mu_{\pts,0},\\
\vect\mu_1-\vect\mu_2&=\unmatched(\vect\mu_{\dets,0}-\vect\mu_{\pts,0}).
\end{aligned}
\end{equation}
\end{lemma}
\begin{proof}
Each replaced point changes the feature sum by a vector of norm at most two, so division by the size of the collection gives $2\unmatched$. Each mixture mean is the weighted average of its inlier and outlier means, and subtracting the two cancels the common inlier term.
\end{proof}

If both point sets undergo replacements, the difference of their feature averages changes by at most twice the sum of the replaced fractions. These bounds hold for a fixed feature map and a fixed image normalization, and they do not shrink as the number of points grows. \cref{lem:colocated} gives a bound that does shrink when the outlier model points have the projected distribution of the inliers.

\subsection{Poisson reference for the association level}
\label{sec:supp_association}

\cref{def:association} compares the distance from each observation to its nearest projection with the local spacing of the projections. This section proves that, if the projections around an observation form a homogeneous Poisson process independent of it, the statistic $\assoc_i$ has a unit exponential distribution at every intensity. The association level compares the observed statistics with this reference, as \cref{sec:supp_association_implementation} specifies.

\begin{lemma}[Exponential nearest-neighbor statistic]
\label{supp:lem-association}
Let a 2D point be chosen independently of a homogeneous Poisson process of positive intensity on $\R^2$. For an integer $\nnb\ge1$, let $d_1\le\cdots\le d_{\nnb+1}$ be the distances from this point to its $\nnb+1$ nearest points of the process. Then
\begin{equation}
\assoc=-\nnb\log(1-d_1^2/d_{\nnb+1}^2)\ \sim\ \operatorname{Exp}(1).
\end{equation}
\end{lemma}
\begin{proof}
Conditional on $d_{\nnb+1}$, the $\nnb$ nearer points of a homogeneous Poisson process are independent and uniform in the disk of radius $d_{\nnb+1}$ about the fixed point. The fraction of the area of that disk within distance $d$ of its center is $d^2/d_{\nnb+1}^2$, so the ratios $d_j^2/d_{\nnb+1}^2$ for $1\le j\le\nnb$ are the order statistics of $\nnb$ independent uniform random variables on $[0,1]$, whatever the intensity. For $x\ge0$, their minimum exceeds $1-e^{-x/\nnb}$ with probability $(e^{-x/\nnb})^\nnb=e^{-x}$, and this event is $\{\assoc>x\}$.
\end{proof}

The lemma concerns one point chosen independently of the Poisson process. In \method, we estimate the pose from the observations, and the neighborhoods of different observations can share projections. Sums of these statistics therefore need not follow a Gamma distribution, and \cref{sec:supp_association_implementation} compares the association level with fixed thresholds.

\subsection{Finite Fourier matching with uniform outliers}
\label{sec:supp_fourier}

\cref{supp:lem-contamination} controls the contribution of outliers at the true pose. We now show that, under its mixture model with uniform outliers, finitely many of the method's Fourier features determine the pose and the outlier fraction, and we bound the error caused by finitely many sampled outliers. The results concern the objective of \cref{def:moments}, before EM refinement. We first fix the setting and restate \cref{thm:fourier-main} in its notation.

\para{Coordinates and features.}
We work in the box coordinates $\vect z$ of \cref{sec:supp_outlier_features}, defined by the image rectangle $D_{\rm F}$ of positive area. As defined there, the taper is $W_F(\vect z)=W_{F,1}(z_1)W_{F,2}(z_2)$. Each factor $W_{F,r}$ vanishes outside the interval $[-h_{F,r}-\delta_F,\,h_{F,r}+\delta_F]$ given by the half-width $h_{F,r}$ of $D_{\rm F}$ and the padding $\delta_F$, and it rises linearly over a distance $\tau_F$ from each end. The weights are $a_{\vect k}=\exp(-\zeta_F^2\norm{\vect k}^2/2)$ with spectral width $\zeta_F$, and $\mathrm F(K_F)$ contains one vector from each pair $\pm\vect k$ of nonzero vectors $\vect k\in\frac\pi2\mathbb Z^2$ with $\norm{\vect k}\le\frac\pi2K_F$. Throughout, $D_{\rm F}$, the box normalization, the taper $W_F$ and the weights $a_{\vect k}$ are fixed data. The method takes $D_{\rm F}$ to be the bounding rectangle of the calibrated observations. Holding it fixed matches the method when $D_{\rm F}$ is the bounding rectangle of the inlier observations, because image outliers drawn on it leave it unchanged. The same holds for the bounding box of the model points. For a pose $\xi$, the map $p_\xi:\R^3\to\R^2$ sends a model point at positive depth to its projection in box coordinates, that is, the pixel projection of \cref{tab:supp_notation} followed by calibration and the box normalization. In this section only, $\vect p_j(\xi)=p_\xi(\vect y_j)$ denotes this box projection. For a cutoff $K_F$, let $\mathrm F(K_F)$ be the frequency set of \cref{sec:supp_outlier_features}, let $\featbound_{K_F}=(\sum_{\vect k\in\mathrm F(K_F)}a_{\vect k}^2)^{1/2}$, and let $\vect\phi_F$ be the feature map, read as a function of box coordinates, with $n_\phi=2|\mathrm F(K_F)|$ real coordinates. The sets $\mathrm F(K_F)$ grow with $K_F$, every weight $a_{\vect k}$ is positive, and every nonzero $\vect k\in\frac\pi2\mathbb Z^2$ or its negative lies in $\mathrm F(K_F)$ once $K_F$ is large enough. The spectral width $\zeta_F$, and hence every weight, stays fixed when $K_F$ varies. The quantities below depend on $K_F$, and a subscript $K_F$ marks this dependence where two cutoffs are compared.

\para{Population model.}
Let $\vect y_1,\ldots,\vect y_m$ be the inlier model points, and let $\mathcal B$ be their bounding box, which has positive volume and is held fixed. Let $U$ be the uniform distribution on $D_{\rm F}$, expressed in box coordinates. At a pose $\xi$, the inlier projections have the empirical distribution $P^{\rm in}_\xi$, which puts mass $1/m$ at each $\vect p_j(\xi)$, and $P^{\mathcal B}_\xi$ is the distribution of $p_\xi(\vect y)$ for $\vect y$ uniform in $\mathcal B$. Let $\mathcal E$ be a compact set of poses that contains $\xi^\star$ and under which every point of $\mathcal B$ has positive depth, and let $\mathcal I\subset[0,1)$ be a compact interval. A parameter is a pair $\theta=(\xi,\eta)\in\mathcal E\times\mathcal I$ of a pose and an outlier fraction. For the true parameter $\theta^\star=(\xi^\star,\eta^\star)$ with $\eta^\star\in\mathcal I$, we use the pose distance $d_{\mathcal Y}$ of \cref{cor:pinhole-stability} and put
\[
d(\theta,\theta^\star)^2
=d_{\mathcal Y}(\xi,\xi^\star)^2+|\eta-\eta^\star|^2.
\]
If both point sets contain the fraction $\eta^\star$ of uniform outliers, the observations have the population distribution $P^{\rm obs}=(1-\eta^\star)P^{\rm in}_{\xi^\star}+\eta^\star U$, and the model points projected at $\xi$ have the distribution $(1-\eta^\star)P^{\rm in}_\xi+\eta^\star P^{\mathcal B}_\xi$. The population residual corresponding to \cref{eq:moment-matching} is therefore
\begin{equation}
\label{eq:fourier-population-residual}
\begin{aligned}
\vect r_{K_F}(\xi,\eta)
={}&(1-\eta^\star)\bigl(\vect\mu^{\rm in}(\xi^\star)-\vect\mu^{\rm in}(\xi)\bigr)\\
&+(\eta^\star-\eta)\bigl(\vect\mu^U-\vect\mu^{\mathcal B}(\xi)\bigr),
\end{aligned}
\end{equation}
where the feature means are $\vect\mu^{\rm in}(\xi)=\int\vect\phi_F\,dP^{\rm in}_\xi$ for the inliers, $\vect\mu^U=\int\vect\phi_F\,dU$ for the image outliers and $\vect\mu^{\mathcal B}(\xi)=\int\vect\phi_F\,dP^{\mathcal B}_\xi$ for the model outliers. The residual vanishes at $\theta^\star$.

\para{Geometric assumptions.}
We fix $\xi^\star$ before choosing the inlier model of $m\ge20$ points. The model lies outside the exceptional set of \cref{thm:pinhole} for $N=m$ and outside the set on which the $2m\times6$ matrix of the derivatives of the projections $\vect p_j$ at $\xi^\star$ has rank less than six. Both sets have measure zero, the second by the paragraph on the rank condition in the proof of \cref{thm:fourier-identification}. The support of $W_F$ lies strictly inside $(-2,2)^2$, which holds for the taper of \cref{sec:supp_outlier_features} when $h_{F,r}+\delta_F<2$ for $r=1,2$. The true projections $\vect z_j^\star=\vect p_j(\xi^\star)$ are distinct, the taper is positive at each of them, and none lies on a line where a factor $W_{F,r}$ changes slope, so $W_F$ is differentiable at each of them. Finally, the two outlier distributions differ after weighting by $W_F$:
\begin{equation}
\label{eq:fourier-background-separation}
W_F(U-P^{\mathcal B}_{\xi^\star})\ne0.
\end{equation}
Here $W_F\mu$ denotes a measure $\mu$ weighted by the function $W_F$. This condition makes the outlier fraction identifiable. Without it, the residual at the true pose stays zero when the fraction changes.

\para{Finitely many outliers.}
The observations are $n$ points with box coordinates $\vect z_1,\ldots,\vect z_n$, and the model has $N$ points. For the analysis, we number the points so that observation $i\le m$ images the inlier model point $\vect y_i$, possibly with noise. The other $n_{\rm out}=n-m$ observations are image outliers drawn uniformly on $D_{\rm F}$, and the other $N_{\rm out}=N-m$ model points are model outliers drawn uniformly in $\mathcal B$. All outliers are independent of each other and of the inlier observations, which we hold fixed. Either number of outliers may be zero. The outlier fractions are $\eta_U=n_{\rm out}/n$ and $\eta_Y=N_{\rm out}/N$, and we set $\eta^\star=\eta_U$. With the midpoint quadratures $\vect\mu_{\dets,0}$ of $\vect\mu^U$ and $\vect\mu_{\pts,0}(\xi)$ of $\vect\mu^{\mathcal B}(\xi)$ from \cref{sec:supp_outlier_features}, the empirical residual is the vector of \cref{eq:moment-matching},
\[
\widehat{\vect r}_{K_F}(\theta)=\vect\mu_{\dets,F}-\vect\mu_{\pts,F}(\xi)-\eta\bigl(\vect\mu_{\dets,0}-\vect\mu_{\pts,0}(\xi)\bigr),
\]
so that $\norm{\widehat{\vect r}_{K_F}(\theta)}^2=\mathcal L_F(\xi,\eta)$.

\para{Error terms.}
The projection $p_\xi(\vect y)$ is smooth in $(\xi,\vect y)$ near the compact set $\mathcal E\times\mathcal B$, and $\vect\phi_F$ is Lipschitz, so there is a finite $L_\xi\ge0$ such that every $\vect y\in\mathcal B$ and all poses $\xi,\xi'\in\mathcal E$ satisfy
\[
\|\vect\phi_F(p_\xi(\vect y))-\vect\phi_F(p_{\xi'}(\vect y))\|
\le L_\xi\, d_{\mathcal Y}(\xi,\xi').
\]
We fix $\epsilon_{\rm net}>0$ and $N_{\rm net}$ poses of $\mathcal E$ such that every pose of $\mathcal E$ lies within $d_{\mathcal Y}$-distance $\epsilon_{\rm net}$ of one of them. For $0<\delta<1$, we put
\begin{equation}
\label{eq:fourier-finite-errors}
\begin{aligned}
\Delta_U&=2\sqrt{\frac{\log(4n_\phi/\delta)}{n_{\rm out}}},\\
\Delta_Y&=2\sqrt{\frac{\log(4n_\phi N_{\rm net}/\delta)}{N_{\rm out}}}+2L_\xi\epsilon_{\rm net},\\
\fourerr&=\eta_U\Delta_U+\eta_Y\Delta_Y+2|\eta_U-\eta_Y|\\
&\quad+\eta_{\max}(q_U+q_Y)+\epsilon_{\rm im},
\end{aligned}
\end{equation}
where $\eta_{\max}=\max\mathcal I$. When $n_{\rm out}=0$, we set $\eta_U\Delta_U=0$, and likewise $\eta_Y\Delta_Y=0$ when $N_{\rm out}=0$. The quadrature errors satisfy $\norm{\vect\mu_{\dets,0}-\vect\mu^U}\le q_U$ and $\norm{\vect\mu_{\pts,0}(\xi)-\vect\mu^{\mathcal B}(\xi)}\le q_Y$ for every $\xi\in\mathcal E$. The image noise of the inliers enters through $\epsilon_{\rm im}\ge\norm{n^{-1}\sum_{i=1}^m(\vect\phi_F(\vect z_i)-\vect\phi_F(\vect z_i^\star))}$.

\begin{trivlist}\item[\hskip\labelsep\textbf{Theorem~\ref{thm:fourier-main} (restated).}]\itshape
Fix $\xi^\star$ and $m\ge20$, and let the inlier model lie outside the null set of the geometric assumptions. Assume the remaining geometric assumptions and the sampling model above, with $\eta_U\in\mathcal I$. There is a cutoff $K_0$ such that every cutoff $K_F\ge K_0$ admits a constant $\gamma_{K_F}>0$ with the following property. The constant depends on the inlier model, $\xi^\star$, $\eta_U$, $\mathcal E$, $\mathcal I$, $D_{\rm F}$, $\mathcal B$, the taper, the weights and $K_F$, and not on the outlier positions, the noise or $\delta$. For every $\delta\in(0,1)$ and every net fixed before the draws, with probability at least $1-\delta$ over the outlier positions, every $\theta=(\xi,\eta)\in\mathcal E\times\mathcal I$ satisfies
\[
d(\theta,\theta^\star)\le\frac{\sqrt{\mathcal L_F(\xi,\eta)}+\fourerr}{\gamma_{K_F}},
\qquad \theta^\star=(\xi^\star,\eta_U).
\]
Here $\mathcal L_F$ uses the cutoff $K_F$, and $\eta_U$ is the fraction of image outliers.
\end{trivlist}

\cref{thm:fourier-identification} provides $K_0$ and $\gamma_{K_F}$ for the population residual, and \cref{thm:fourier-finite} gives the bound with probability at least $1-\delta$. The constant $\fourconst$ of the main text is $1/\gamma_{K_F}$, its distance $\operatorname{dist}$ is $d(\theta,\theta^\star)$, with translations in units of $\ell_{\mathcal Y}$, its error term is $\fourerr$ of \cref{eq:fourier-finite-errors}, and its compact set $\Theta$ lies in such a product $\mathcal E\times\mathcal I$ of compact sets of poses and fractions. Only $\Delta_U$ and $\Delta_Y$ depend on $\delta$. The minimizer bound of the main text is the case $\epsilon_{\rm fit}=0$ of \cref{thm:fourier-finite}.

\begin{theorem}[Finite Fourier identification]
\label{thm:fourier-identification}
Under the geometric assumptions, let $\eta^\star\in\mathcal I$. There is a finite cutoff $K_0$ such that every cutoff $K_F\ge K_0$ admits a constant $\gamma_{K_F}>0$ satisfying
\begin{equation}
\label{eq:fourier-global-lower}
\|\vect r_{K_F}(\theta)\|\ge\gamma_{K_F}\, d(\theta,\theta^\star),
\qquad \theta\in\mathcal E\times\mathcal I.
\end{equation}
Consequently, $\theta^\star$ is the unique zero of $\vect r_{K_F}$ on $\mathcal E\times\mathcal I$. Both $K_0$ and $\gamma_{K_F}$ may depend on all fixed data, including $\eta^\star$.
\end{theorem}

\begin{proof}
The proof first identifies the parameter from all Fourier coefficients and then shows that finitely many suffice, both near the truth and away from it.

\emph{Fourier coefficients.} The square $(-2,2)^2$ is a fundamental cell of the torus $\mathbb T=\R^2/4\mathbb Z^2$, so we view measures supported inside it as measures on $\mathbb T$. The characters of $\mathbb T$ are $\chi_{\vect k}(\vect z)=\exp(\mathrm i\vect k^\top\vect z)$ for $\vect k\in\frac\pi2\mathbb Z^2$, and a finite signed measure $\nu$ has the coefficients $\hat\nu(\vect k)=\int\chi_{\vect k}\,d\nu$. For $\theta\in\mathcal E\times\mathcal I$, we define the signed measure
\[
\nu_\theta=W_F\big[(1-\eta^\star)P^{\rm in}_\xi
 +(\eta^\star-\eta)P^{\mathcal B}_\xi+\eta U\big],
\]
which is supported in the support of $W_F$, and $\nu_{\theta^\star}=W_FP^{\rm obs}$. By \cref{eq:fourier-population-residual}, $\vect r_{K_F}(\theta)$ is the integral of $\vect\phi_F$ against $P^{\rm obs}$ minus its integral against $(1-\eta^\star)P^{\rm in}_\xi+(\eta^\star-\eta)P^{\mathcal B}_\xi+\eta U$. For every measure $\mu$, the integral $\int\vect\phi_F\,d\mu$ lists the real and imaginary parts of $\widehat{W_F\mu}(\vect k)$ for $\vect k\in\mathrm F(K_F)$, each multiplied by $a_{\vect k}/\featbound_{K_F}$. Hence $\vect r_{K_F}(\theta)$ lists the scaled real and imaginary parts of $\widehat{W_FP^{\rm obs}}(\vect k)-\hat\nu_\theta(\vect k)$. Real measures satisfy $\hat\nu(-\vect k)=\overline{\hat\nu(\vect k)}$, so $\vect r_{K_F}(\theta)=0$ for every cutoff exactly when $\hat\nu_\theta(\vect k)=\widehat{W_FP^{\rm obs}}(\vect k)$ for every nonzero $\vect k$.

\emph{Identification from all coefficients.} Suppose that these coefficients agree. A finite signed measure on $\mathbb T$ whose coefficients vanish at every nonzero frequency is a constant multiple of the uniform measure~\cite{katznelson2004harmonic}. The difference $\nu_\theta-W_FP^{\rm obs}$ vanishes on the complement of the support of $W_F$, a nonempty open subset of $\mathbb T$, so the multiple is zero and $\nu_\theta=W_FP^{\rm obs}$. The measures $U$ and $P^{\mathcal B}_\xi$ have densities. For $P^{\mathcal B}_\xi$, the change of camera coordinates $(u_1,u_2,x_3)\mapsto(x_3u_1,x_3u_2,x_3)$ has volume element $x_3^2\,du_1\,du_2\,dx_3$, integration over positive depth gives a density of the calibrated projection, and the invertible box normalization preserves it. The atomic parts of the two sides therefore coincide. After division by $(1-\eta^\star)/m>0$, the atomic part on the left puts mass $W_F(\vect p_j(\xi))$ at each $\vect p_j(\xi)$, with the masses of coinciding points added. The atomic part on the right puts mass $W_F(\vect z_j^\star)>0$ at each of the $m$ distinct points $\vect z_j^\star$. The left side has at most $m$ points of positive mass, so the $\vect p_j(\xi)$ are distinct, each lies where $W_F$ is positive, and they are the points $\vect z_j^\star$. Thus $P^{\rm in}_\xi=P^{\rm in}_{\xi^\star}$, so the inlier projections agree as multisets in calibrated coordinates as well. \cref{thm:pinhole}, applied to the $m$ inlier points, which have positive depth under $\xi$ and $\xi^\star$, gives $\xi=\xi^\star$. Then $\nu_\theta-W_FP^{\rm obs}=(\eta^\star-\eta)W_F(P^{\mathcal B}_{\xi^\star}-U)$, and \cref{eq:fourier-background-separation} gives $\eta=\eta^\star$. Every point with positive mass lies in the same fundamental cell, so no point is confused with a translate by a period.

\emph{Derivative at the truth.} We use local pose coordinates $(\vect\omega,\vect\tau)\in\R^6$ for the poses obtained from $\xi^\star$ by the camera motion $\vect x\mapsto\exp([\vect\omega]_\times)\vect x+\vect\tau$, and write a direction as $\dot\theta=(\dot\xi,\dot\eta)\in\R^6\times\R$. Let $\partial\vect p_j\in\R^{2\times6}$ be the derivative of $\vect p_j$ at $\xi^\star$ in these coordinates. For a smooth function $f$ on $\mathbb T$, let $F_f(\theta)=\int f\,d\nu_\theta$. Its atomic term is differentiable at $\theta^\star$ because $W_F$ is differentiable at each $\vect z_j^\star$, and its uniform term is linear in $\eta$. Its box term $(\eta^\star-\eta)\int W_Ff\,dP^{\mathcal B}_\xi$ is the product of a factor that vanishes at $\theta^\star$ and a function of $\xi$ that is continuous by dominated convergence, because $W_Ff$ is bounded and continuous and depths stay positive on $\mathcal E$. This term is therefore differentiable at $\theta^\star$ with derivative $-\dot\eta\int W_Ff\,dP^{\mathcal B}_{\xi^\star}$, without differentiating the box integral. With $\nu^{\rm out}=W_F(U-P^{\mathcal B}_{\xi^\star})$, the derivative of $F_f$ at $\theta^\star$ in the direction $\dot\theta$ is
\begin{equation}
\label{eq:fourier-measure-derivative}
\begin{aligned}
\Upsilon_{\dot\theta}(f)
={}&\frac{1-\eta^\star}{m}\sum_{j=1}^m
 \nabla(W_Ff)(\vect z_j^\star)^\top\partial\vect p_j\,\dot\xi\\
&+\dot\eta\int f\,d\nu^{\rm out}.
\end{aligned}
\end{equation}
The same argument shows that $\vect r_{K_F}$ is differentiable at $\theta^\star$. Its derivative in the direction $\dot\theta$ lists, with the opposite sign and the factors $a_{\vect k}/\featbound_{K_F}$, the real and imaginary parts of $\Upsilon_{\dot\theta}(\chi_{\vect k})$ for $\vect k\in\mathrm F(K_F)$.

\emph{Injectivity.} Suppose that $\Upsilon_{\dot\theta}(\chi_{\vect k})=0$ for every nonzero $\vect k$. Linearity then gives $\Upsilon_{\dot\theta}(f)=\Upsilon_{\dot\theta}(1)\langle f\rangle$ for every trigonometric polynomial $f$, where $\langle f\rangle=\frac1{16}\int_{(-2,2)^2}f(\vect z)\,d\vect z$ is the mean over a period. To remove the atomic terms, we use a trigonometric polynomial that vanishes together with its gradient at every true projection. On $\mathbb T$, let
\[
\begin{aligned}
b_j(\vect z)&=2-\cos\!\left(\tfrac\pi2(z_1-z_{j1}^\star)\right)
-\cos\!\left(\tfrac\pi2(z_2-z_{j2}^\star)\right),\\
f_0(\vect z)&=\prod_{j=1}^m b_j(\vect z)^2.
\end{aligned}
\]
Each $b_j$ is nonnegative and vanishes on $\mathbb T$ exactly at $\vect z_j^\star$, so $f_0$ is nonnegative and vanishes, together with its gradient, exactly at the true projections. For every trigonometric polynomial $f$, the product $f_0f$ also vanishes with its gradient there, so the atomic terms of $\Upsilon_{\dot\theta}(f_0f)$ vanish. Hence $\dot\eta\int f_0f\,d\nu^{\rm out}=\Upsilon_{\dot\theta}(1)\langle f_0f\rangle$ for every trigonometric polynomial $f$. The finite signed measures $\dot\eta f_0\nu^{\rm out}$ and $\frac1{16}\Upsilon_{\dot\theta}(1)f_0\,d\vect z$ therefore have the same coefficients and are equal~\cite{katznelson2004harmonic}. Outside the support of $W_F$, the measure $\nu^{\rm out}$ vanishes and $f_0$ is positive, so $\Upsilon_{\dot\theta}(1)=0$. Hence $\dot\eta f_0\nu^{\rm out}=0$. The measure $\nu^{\rm out}$ has a density, which is nonzero on a set of positive measure by \cref{eq:fourier-background-separation}, and $f_0$ vanishes only at finitely many points. The density of $f_0\nu^{\rm out}$ is therefore nonzero on a set of positive measure, and $\dot\eta=0$.

It remains to show that $\dot\xi=0$. We now have $\Upsilon_{\dot\theta}(f)=0$ for every trigonometric polynomial $f$, and only the atomic terms remain. For a point index $i$ and a coordinate $a\in\{1,2\}$, let
\[
f_{i,a}(\vect z)=\sin\!\left(\tfrac\pi2(z_a-z_{ia}^\star)\right)
\prod_{j\ne i}b_j(\vect z)^2.
\]
This function vanishes at every true projection, its gradient vanishes at $\vect z_j^\star$ for $j\ne i$, and its gradient at $\vect z_i^\star$ is $\frac\pi2\prod_{j\ne i}b_j(\vect z_i^\star)^2$ times the $a$th coordinate vector. This factor is positive because the true projections are distinct points of the cell. Hence $\Upsilon_{\dot\theta}(f_{i,a})$ equals $(1-\eta^\star)W_F(\vect z_i^\star)/m$ times this factor times $(\partial\vect p_i\,\dot\xi)_a$, and all factors except the last are positive. Thus $(\partial\vect p_i\,\dot\xi)_a=0$ for every $i$ and $a$, and the stacked derivatives have rank six, which gives $\dot\xi=0$.

The complex linear functionals $\dot\theta\mapsto\Upsilon_{\dot\theta}(\chi_{\vect k})$ on $\R^7$, for nonzero $\vect k$, therefore have no common nonzero kernel vector. Their kernels are subspaces of $\R^7$ with zero intersection, so finitely many of them already intersect in zero. Let $K_1$ be a cutoff for which $\mathrm F(K_1)$ contains each of these frequencies or its negative. The derivative of $\vect r_{K_1}$ at $\theta^\star$ is then injective, with least singular value $\sigma_1>0$ in the coordinates $(\vect\omega,\vect\tau,\eta)$.

\emph{Rank condition.} The rank condition on the projections is generic. At the identity camera, with $\vect u=\vect x_{1:2}/x_3$ for a camera point $\vect x$, the derivatives with respect to $(\vect\omega,\vect\tau)$ are
\[
\begin{aligned}
\partial u_1&=(-u_1u_2,1+u_1^2,-u_2,x_3^{-1},0,-u_1/x_3),\\
\partial u_2&=(-(1+u_2^2),u_1u_2,u_1,0,x_3^{-1},-u_2/x_3).
\end{aligned}
\]
For the camera points $(0,0,1)$, $(2,0,2)$ and $(0,3,3)$, the six rows form a matrix with determinant $29/36$. The $6\times6$ minor formed by the rows of three points is a rational function of their coordinates whose denominator is a power of their positive depths. Its numerator is a nonzero polynomial, so it vanishes only on a set of measure zero in $\R^9$. Rank less than six forces this minor to vanish, so the configurations of $m\ge3$ points with rank less than six form a null set. Expressing the points in the camera coordinates of $\xi^\star$ is an invertible affine change of the model coordinates, and the box normalization multiplies each $\partial\vect p_j$ by the same invertible matrix. The exceptional set of inlier models is therefore null as well.

\emph{Local bound.} Differentiability at $\theta^\star$ gives $\|\vect r_{K_1}(\theta)\|\ge\frac12\sigma_1|\dot\theta|$ for every $\theta$ in a neighborhood of $\theta^\star$ with coordinates $\dot\theta$. The chart is smooth, so $d(\theta,\theta^\star)\le c_1|\dot\theta|$ near $\theta^\star$ for a constant $c_1$. There is therefore an open neighborhood $\mathcal O_1$ of $\theta^\star$ in $\mathcal E\times\mathcal I$ on which $\|\vect r_{K_1}(\theta)\|\ge\gamma_1d(\theta,\theta^\star)$ with $\gamma_1=\sigma_1/(2c_1)$.

\emph{Away from the truth.} The set $\mathcal O_1^{\rm c}=(\mathcal E\times\mathcal I)\setminus\mathcal O_1$ is compact. For each $\theta\in\mathcal O_1^{\rm c}$, the identification step gives a nonzero frequency $\vect k_\theta$ with $\hat\nu_\theta(\vect k_\theta)\ne\widehat{W_FP^{\rm obs}}(\vect k_\theta)$. The coefficient $\hat\nu_\theta(\vect k)$ is continuous in $\theta$, the atomic part because $W_F\chi_{\vect k}$ is continuous and the box part by dominated convergence. The inequality therefore holds on an open neighborhood of $\theta$, and finitely many of these neighborhoods cover $\mathcal O_1^{\rm c}$. Let $K_0\ge K_1$ be a cutoff for which $\mathrm F(K_0)$ contains each of the selected frequencies or its negative.

\emph{Every cutoff $K_F\ge K_0$.} The sets satisfy $\mathrm F(K_1)\subset\mathrm F(K_F)$, and only the normalizing constant changes, so the coordinates of $\vect r_{K_F}$ include those of $\vect r_{K_1}$ multiplied by $\featbound_{K_1}/\featbound_{K_F}>0$. Hence $\|\vect r_{K_F}(\theta)\|\ge(\featbound_{K_1}/\featbound_{K_F})\gamma_1d(\theta,\theta^\star)$ on $\mathcal O_1$. If $\mathcal O_1^{\rm c}$ is empty, this bound proves \cref{eq:fourier-global-lower}. Otherwise, $\vect r_{K_F}$ is continuous on $\mathcal O_1^{\rm c}$, and it is nonzero there because $\mathrm F(K_F)$ contains each selected frequency or its negative, so its norm has a positive minimum $\underline r_{K_F}$ there. With $d_{\max}>0$ the largest value of $d(\theta,\theta^\star)$ on $\mathcal E\times\mathcal I$, every $\theta\in\mathcal O_1^{\rm c}$ satisfies $\|\vect r_{K_F}(\theta)\|\ge(\underline r_{K_F}/d_{\max})\,d(\theta,\theta^\star)$. Taking $\gamma_{K_F}$ as the smaller of the two constants proves \cref{eq:fourier-global-lower}. Because $d$ is a metric, $\vect r_{K_F}(\theta)=0$ forces $\theta=\theta^\star$.
\end{proof}

\begin{theorem}[Finite-sample Fourier pose bound]
\label{thm:fourier-finite}
Assume the setting above with $\eta^\star=\eta_U\in\mathcal I$, and let $K_F\ge K_0$ and $\gamma_{K_F}$ be as in \cref{thm:fourier-identification}. Fix $\epsilon_{\rm net}>0$, a net of $N_{\rm net}$ poses as above, and $\delta\in(0,1)$. With probability at least $1-\delta$ over the outlier draws, the empirical residual satisfies, simultaneously for every $\theta\in\mathcal E\times\mathcal I$,
\begin{equation}
\label{eq:fourier-finite-pose}
d(\theta,\theta^\star)
\le\frac{\|\widehat{\vect r}_{K_F}(\theta)\|+\fourerr}{\gamma_{K_F}}.
\end{equation}
On the same event, every parameter $\hat\theta$ whose residual norm exceeds the infimum over $\mathcal E\times\mathcal I$ by at most $\epsilon_{\rm fit}\ge0$ satisfies $d(\hat\theta,\theta^\star)\le(2\fourerr+\epsilon_{\rm fit})/\gamma_{K_F}$.
\end{theorem}

\begin{proof}
Each real coordinate of $\vect\phi_F$ belongs to one frequency $\vect k$ and lies in $[-a_{\vect k}/\featbound_{K_F},a_{\vect k}/\featbound_{K_F}]$, because the taper is at most one. Each frequency contributes a cosine and a sine with the same amplitude, so the squared amplitudes of all coordinates sum to $2$. Let $X_1,\ldots,X_{n_{\rm s}}$ be the values of one coordinate, with amplitude $a_{\vect k}/\featbound_{K_F}$, at $n_{\rm s}$ independent outliers, and let $\bar X$ be their mean. For $\alpha>0$, Hoeffding's inequality~\cite{hoeffding1963probability} gives
\[
\Pr\bigl\{|\bar X-\E X_1|>\alpha\,a_{\vect k}/\featbound_{K_F}\bigr\}
\le2e^{-n_{\rm s}\alpha^2/2}.
\]
When every coordinate satisfies the opposite inequality, the squared norm of the vector error is at most $2\alpha^2$. Let $\widehat{\vect\mu}^U$ be the average of $\vect\phi_F$ over the image outliers and $\widehat{\vect\mu}^{\mathcal B}(\xi)$ the average of $\vect\phi_F\circ p_\xi$ over the model outliers. A union bound over the $n_\phi$ coordinates, with $2n_\phi e^{-n_{\rm out}\alpha^2/2}=\delta/2$, gives $\norm{\widehat{\vect\mu}^U-\vect\mu^U}\le\sqrt2\,\alpha=\Delta_U$ with probability at least $1-\delta/2$. A union bound over the $n_\phi$ coordinates and the $N_{\rm net}$ net poses gives $\norm{\widehat{\vect\mu}^{\mathcal B}(\xi')-\vect\mu^{\mathcal B}(\xi')}\le\Delta_Y-2L_\xi\epsilon_{\rm net}$ at every net pose $\xi'$ with probability at least $1-\delta/2$. The same model-outlier sample is used at every pose, and the union bound needs no independence between coordinates or between net poses. Both $\widehat{\vect\mu}^{\mathcal B}$ and $\vect\mu^{\mathcal B}$ are $L_\xi$-Lipschitz in the pose, so moving from the nearest net pose to any $\xi\in\mathcal E$ adds at most $2L_\xi\epsilon_{\rm net}$. Both events hold with probability at least $1-\delta$. An event is not needed when the corresponding number of outliers is zero.

On these events, we compare $\widehat{\vect r}_{K_F}$ with $\vect r_{K_F}$ at $\eta^\star=\eta_U$. We split the feature average over the observations into the contributions of the inliers and the image outliers, and the feature average over the model points into those of the inliers and the model outliers. Because $1-\eta_U=m/n$ and $1-\eta_Y=m/N$, this gives
\[
\begin{aligned}
\widehat{\vect r}_{K_F}&(\theta)-\vect r_{K_F}(\theta)
=\frac1n\sum_{i=1}^m\bigl(\vect\phi_F(\vect z_i)-\vect\phi_F(\vect z_i^\star)\bigr)\\
&+\eta_U\bigl(\widehat{\vect\mu}^U-\vect\mu^U\bigr)
-\eta_Y\bigl(\widehat{\vect\mu}^{\mathcal B}(\xi)-\vect\mu^{\mathcal B}(\xi)\bigr)\\
&+(\eta_Y-\eta_U)\bigl(\vect\mu^{\rm in}(\xi)-\vect\mu^{\mathcal B}(\xi)\bigr)\\
&-\eta\bigl[(\vect\mu_{\dets,0}-\vect\mu^U)-(\vect\mu_{\pts,0}(\xi)-\vect\mu^{\mathcal B}(\xi))\bigr].
\end{aligned}
\]
The five terms have norms at most $\epsilon_{\rm im}$, $\eta_U\Delta_U$, $\eta_Y\Delta_Y$, $2|\eta_Y-\eta_U|$ and $\eta_{\max}(q_U+q_Y)$, where the fourth bound uses $\norm{\vect\phi_F}\le1$. The triangle inequality therefore gives
\[
\sup_{\theta\in\mathcal E\times\mathcal I}
\|\widehat{\vect r}_{K_F}(\theta)-\vect r_{K_F}(\theta)\|\le\fourerr.
\]
\cref{eq:fourier-global-lower} and this inequality give $\gamma_{K_F}d(\theta,\theta^\star)\le\|\vect r_{K_F}(\theta)\|\le\|\widehat{\vect r}_{K_F}(\theta)\|+\fourerr$, which is \cref{eq:fourier-finite-pose}. Because $\vect r_{K_F}(\theta^\star)=0$, the empirical residual at $\theta^\star$ has norm at most $\fourerr$. A parameter $\hat\theta$ as in the theorem therefore has residual norm at most $\fourerr+\epsilon_{\rm fit}$, and \cref{eq:fourier-finite-pose} at $\hat\theta$ gives the final bound.
\end{proof}

\para{Finite quadrature bounds.}
We bound $q_U$, $q_Y$ and $\epsilon_{\rm im}$ for the midpoint rules used by the method. With $\beta_{\vect k}=a_{\vect k}/\featbound_{K_F}$, the feature map $\vect\phi_F$ is Lipschitz in box coordinates with constant
\[
L_\phi\le \frac{2\sqrt2}{\tau_F}
 +\Bigl(\sum_{\vect k}\beta_{\vect k}^2\|\vect k\|^2\Bigr)^{1/2}.
\]
Each factor $W_{F,r}$ is a product of two functions with values in $[0,1]$ and Lipschitz constant $1/\tau_F$, so the gradient of $W_F$ has norm at most $2\sqrt2/\tau_F$. The untapered feature vector has norm one, and the inequality $|e^{\mathrm i\vartheta}-e^{\mathrm i\vartheta'}|\le|\vartheta-\vartheta'|$ for real $\vartheta$ and $\vartheta'$ bounds its Lipschitz constant by the second term. Let $L_{\rm proj}$ bound the norm of the derivative of $\vect y\mapsto p_\xi(\vect y)$ on $\mathcal E\times\mathcal B$, which is finite by compactness and positive depths. For an image rectangle with side lengths $w_{U,r}$ in box coordinates and $J_{U,r}$ midpoint cells along axis $r$, and a model box with side lengths $w_{Y,r}$ and $J_{Y,r}$ cells, the bounds are
\[
\begin{aligned}
q_U&\le\frac{L_\phi}{2}
 \Bigl(\sum_{r=1}^2(w_{U,r}/J_{U,r})^2\Bigr)^{1/2},\\
q_Y&\le\frac{L_\phi L_{\rm proj}}{2}
 \Bigl(\sum_{r=1}^3(w_{Y,r}/J_{Y,r})^2\Bigr)^{1/2}.
\end{aligned}
\]
Every point of a cell lies within half the cell diagonal of its midpoint, and the box is convex, so the integrand differs from its midpoint value by at most $L_\phi$, or $L_\phi L_{\rm proj}$, times that distance. Averaging over each cell and summing proves these bounds. Likewise, $\epsilon_{\rm im}\le L_\phi n^{-1}\sum_{i=1}^m\|\vect z_i-\vect z_i^\star\|$.

\para{Relation to the fitted fraction.}
The method sets the fraction at each pose to the least-squares value $\hat\unmatched(\xi)$ of Step 2 in \cref{sec:supp_config}, clipped to $[0,0.75]$. The theorems cover this choice with $\mathcal I=[0,0.75]$ and $\eta_{\max}=0.75$ whenever $\eta_U\le0.75$, because their bounds hold for every $\theta\in\mathcal E\times\mathcal I$. At the population level, \cref{eq:fourier-population-residual} reads $\vect r_{K_F}(\xi,\eta)=\vect r_{K_F}(\xi,\eta^\star)+(\eta^\star-\eta)\vect\mu^{\rm out}(\xi)$. Here $\vect\mu^{\rm out}(\xi)=\vect\mu^U-\vect\mu^{\mathcal B}(\xi)$ is the population value of the difference $\vect\mu_{\dets,0}-\vect\mu_{\pts,0}(\xi)$ used by the method. For $K_F\ge K_0$, the vector $\vect\mu^{\rm out}(\xi^\star)$ is nonzero, because $-\vect\mu^{\rm out}(\xi^\star)$ is the derivative of $\vect r_{K_F}$ with respect to $\eta$ at $\theta^\star$, a column of an injective derivative. Where $\vect\mu^{\rm out}(\xi)\ne0$, the population version of $\hat\unmatched(\xi)$ is $\eta^{\rm pr}(\xi)=\eta^\star+\langle\vect r_{K_F}(\xi,\eta^\star),\vect\mu^{\rm out}(\xi)\rangle/\norm{\vect\mu^{\rm out}(\xi)}^2$. The profiled residual is $\vect r^{\rm pr}(\xi)=\vect r_{K_F}(\xi,\eta^{\rm pr}(\xi))=\Pi^\perp_{\vect\mu^{\rm out}(\xi)}\vect r_{K_F}(\xi,\eta^\star)$, where $\Pi^\perp$ with a vector as subscript denotes the orthogonal projection onto the orthogonal complement of that vector. If $0<\eta^\star<0.75$, the continuity of $\eta^{\rm pr}$ at $\xi^\star$, where it equals $\eta^\star$, makes the clipping inactive near $\xi^\star$. Because $\vect r_{K_F}(\cdot,\eta^\star)$ vanishes and is differentiable at $\xi^\star$, and $\vect\mu^{\rm out}$ is continuous there, the profiled residual has the derivative
\[
-(1-\eta^\star)\Pi^\perp_{\vect\mu^{\rm out}(\xi^\star)}\,\partial\vect\mu^{\rm in}(\xi^\star)
\]
at $\xi^\star$. This derivative has rank six. Indeed, a direction $\dot\xi$ in its kernel makes $\partial\vect\mu^{\rm in}(\xi^\star)\dot\xi$ a multiple of $\vect\mu^{\rm out}(\xi^\star)$, and the injective derivative of $\vect r_{K_F}$ then gives $\dot\xi=0$. Pose changes therefore remain distinguishable after the fraction is fitted, and the squared norm of the profiled residual grows at least quadratically in the pose error near $\xi^\star$. If, in addition, $\vect r^{\rm pr}$ is continuously differentiable with a Lipschitz derivative near $\xi^\star$, the local convergence theory of Gauss--Newton for zero-residual problems with an injective derivative applies~\cite{nocedal2006numerical}. \cref{thm:fourier-finite} bounds the error of every returned parameter by its residual, so it does not require the search to attain the global minimum.

\para{Scope.}
The theorems hold for each fixed set of weights, that is, for each fixed spectral width $\zeta_F$. The schedule of the method pairs each cutoff with its own spectral width (\cref{tab:supp_search_settings}), so the theorems apply to each stage of the schedule separately. The error terms of \cref{eq:fourier-finite-errors} account for finite samples and the grid approximation.

\subsection{Ambiguous poses}
\label{sec:supp_proof_companions}

Step 4 refines seven alternatives to the accepted pose, obtained by half-turns about the principal axes of the model points and about the optical axis (\cref{par:symmetric}). This section shows why these are the poses to test. Under scaled orthography, the accepted pose and its seven alternatives give images with the same mean and covariance for every model (\cref{supp:lem-companions}(a)). An alternative gives the same scaled-orthographic image as the accepted pose when the model has the corresponding symmetry (part~(b)). For a planar model, the alternative that mirrors the depths about the centroid does so (part~(c)). A final bound shows that, under perspective projection, mirroring the depths moves each image point by at most an amount that grows with the depth variation of the model, so the two images stay close for shallow models.

\para{Setting.}
We use calibrated image coordinates, with $\Kmat=I_3$, and state the results for a 3D model $\pts$. Step 4 applies them to all model points on complete inputs and to the model points matched at the accepted pose when outliers may be present. Let $\bar{\vect y}$ and $\Sigma_{\pts}$ be the centroid and covariance of the model points, and let $\vect e_1,\vect e_2,\vect e_3$ be an orthonormal basis of eigenvectors of $\Sigma_{\pts}$, the principal axes. The half-turn about the $k$th principal axis is $\halfturn_k=2\vect e_k\vect e_k^\top-I_3$ for $k=1,2,3$, and $\halfturn_0=I_3$. The half-turn about the optical axis is $\opthalf=\operatorname{diag}(-1,-1,1)$, and $-\opthalf=\operatorname{diag}(1,1,-1)$ reverses depth. A pose $(\Rot,\trans)$ maps $\vect y_j$ to $\vect x_j=\Rot\vect y_j+\trans$ in camera coordinates and the centroid to $\bar{\vect x}=\Rot\bar{\vect y}+\trans$, with positive depth $\bar x_3$. The scaled-orthographic image of $\vect x_j$ is $(x_{j1},x_{j2})/\bar x_3$, and its pinhole image is $\proj_{\mathrm{cal}}(\vect x_j)=(x_{j1},x_{j2})/x_{j3}$ when $x_{j3}>0$. The eight poses of Step 4 are the accepted pose and its seven alternatives. For $k=0,\dots,3$, they are
\begin{equation}\label{eq:supp-eight}
\begin{gathered}
 \Rot_k^{+}=\Rot \halfturn_k,\quad\Rot_k^{-}=\opthalf\Rot \halfturn_k,\\
 \trans_k^{\pm}=\bar{\vect x}-\Rot_k^{\pm}\bar{\vect y},
\end{gathered}
\end{equation}
where each translation keeps the centroid at $\bar{\vect x}$. They combine the four choices of $\halfturn_k$ with the presence or absence of $\opthalf$.

\begin{lemma}[Second-order ambiguous poses and model symmetries]\label{supp:lem-companions}
\textup{(a)} Each $\Rot_k^{\pm}$ is a rotation, each of the eight poses maps the centroid to $\bar{\vect x}$, and the scaled-orthographic images of the eight poses have the same mean and the same covariance.

\textup{(b)} Suppose that the eigenvalues of $\Sigma_{\pts}$ are distinct. Then every orthogonal matrix $Q_{\rm s}$ that maps the centered model to itself, as a multiset, is one of $\pm \halfturn_k$. If $\halfturn_k$ is such a symmetry, the pose $(\Rot_k^{+},\trans_k^{+})$ gives the same multiset of points in camera coordinates as $(\Rot,\trans)$, and hence the same image. If $-\halfturn_k$ is such a symmetry, the pose $(\Rot_k^{-},\trans_k^{-})$ gives the multiset $\{\bar{\vect x}-\opthalf(\vect x_j-\bar{\vect x})\}$. These are the points produced by $(\Rot,\trans)$ with their depths mirrored about $\bar x_3$, so the scaled-orthographic image is the same.

\textup{(c)} If the model lies in a plane, we number the principal axes so that $\vect e_3$ is its unit normal. The reflection $I_3-2\vect e_3\vect e_3^\top=-\halfturn_3$ in the plane then fixes every centered model point. The pose $(\Rot_3^{-},\trans_3^{-})$ maps each $\vect y_j$ to $\bar{\vect x}-\opthalf(\vect x_j-\bar{\vect x})$ and hence gives the same scaled-orthographic image as $(\Rot,\trans)$, whether or not the eigenvalues are distinct.
\end{lemma}
\begin{proof}
The translations keep the centroid fixed, so every part reduces to how $\halfturn_k$ and $\opthalf$ act on the centered points.

\emph{Part (a).} Each $\halfturn_k$ is either the identity or a symmetric orthogonal matrix with eigenvalues $1,-1,-1$, so it is a rotation. The matrix $\opthalf$ is also a rotation, and so is every product $\Rot_k^{\pm}$. By the choice of $\trans_k^{\pm}$, the pose $(\Rot_k^{\pm},\trans_k^{\pm})$ maps the model points to
\begin{equation}\label{eq:supp-campoints}
 \bar{\vect x}+\Rot_k^{\pm}(\vect y_j-\bar{\vect y}),
\end{equation}
with centroid $\bar{\vect x}$ and covariance $\Rot_k^{\pm}\Sigma_{\pts}(\Rot_k^{\pm})^\top$. A half-turn about a principal axis negates two principal coordinates and preserves their variances, because $\halfturn_k\vect e_i=\pm\vect e_i$ and $\halfturn_k\Sigma_{\pts}\halfturn_k=\Sigma_{\pts}$. For $\Rot_k^{+}$, the covariance is therefore $\Rot\Sigma_{\pts}\Rot^\top$, and for $\Rot_k^{-}$ it is $\opthalf\Rot\Sigma_{\pts}\Rot^\top\opthalf$. The scaled-orthographic image keeps the first two coordinates, divided by the common depth $\bar x_3$. The matrix $\opthalf$ negates both, which leaves their covariance unchanged. The image covariance is therefore the upper-left $2\times2$ block of $\Rot\Sigma_{\pts}\Rot^\top/\bar x_3^2$ in all eight cases, and the image mean is $(\bar x_1,\bar x_2)/\bar x_3$.

\emph{Part (b).} The multiset $\{Q_{\rm s}(\vect y_j-\bar{\vect y})\}$ has covariance $Q_{\rm s}\Sigma_{\pts}Q_{\rm s}^\top$, so a symmetry satisfies $Q_{\rm s}\Sigma_{\pts}Q_{\rm s}^\top=\Sigma_{\pts}$ and commutes with $\Sigma_{\pts}$. A matrix that commutes with $\Sigma_{\pts}$ maps each eigenspace of $\Sigma_{\pts}$ to itself. Distinct eigenvalues make every eigenspace a line, so the orthogonal matrix $Q_{\rm s}$ satisfies $Q_{\rm s}\vect e_i=\pm\vect e_i$. Thus $Q_{\rm s}$ is diagonal in the principal-axis basis with entries $\pm1$, which gives the eight matrices $\pm \halfturn_k$. If $\halfturn_k$ is a symmetry, then by \cref{eq:supp-campoints} the pose $(\Rot_k^{+},\trans_k^{+})$ gives the multiset
\begin{equation}
 \{\bar{\vect x}+\Rot \halfturn_k(\vect y_j-\bar{\vect y})\}=\{\bar{\vect x}+\Rot(\vect y_j-\bar{\vect y})\}=\{\vect x_j\}.
\end{equation}
If $-\halfturn_k$ is a symmetry, the pose $(\Rot_k^{-},\trans_k^{-})$ gives the multiset
\begin{equation}
\begin{aligned}
 &\{\bar{\vect x}+\opthalf\Rot \halfturn_k(\vect y_j-\bar{\vect y})\}\\
 &\quad=\{\bar{\vect x}-\opthalf\Rot(-\halfturn_k)(\vect y_j-\bar{\vect y})\}\\
 &\quad=\{\bar{\vect x}-\opthalf\Rot(\vect y_j-\bar{\vect y})\}=\{\bar{\vect x}-\opthalf(\vect x_j-\bar{\vect x})\}.
\end{aligned}
\end{equation}
The matrix $-\opthalf$ negates the third coordinate and keeps the first two, so the scaled-orthographic image is unchanged.

\emph{Part (c).} The unit normal of the model plane is an eigenvector of $\Sigma_{\pts}$ with eigenvalue zero, so we can choose it as $\vect e_3$. For a planar model, $\vect e_3\cdot(\vect y_j-\bar{\vect y})=0$ for every $j$. Hence $(I_3-2\vect e_3\vect e_3^\top)(\vect y_j-\bar{\vect y})=\vect y_j-\bar{\vect y}$, and $\halfturn_3(\vect y_j-\bar{\vect y})=-(\vect y_j-\bar{\vect y})$. By \cref{eq:supp-campoints}, the pose $(\Rot_3^{-},\trans_3^{-})$ maps $\vect y_j$ to $\bar{\vect x}+\opthalf\Rot \halfturn_3(\vect y_j-\bar{\vect y})=\bar{\vect x}-\opthalf(\vect x_j-\bar{\vect x})$. As in part~(b), the matrix $-\opthalf$ leaves the scaled-orthographic image unchanged.
\end{proof}

Part~(b) requires distinct eigenvalues of $\Sigma_{\pts}$. A model with repeated eigenvalues can have further symmetries, and parts~(a) and~(c) still hold.

\para{Perspective projection.}
A pinhole camera divides by the depth of each point, so mirroring the depths can change the image. We bound this change when the depths of the model points vary little compared with the depth of their centroid. By parts~(b) and~(c), the alternative $(\Rot_k^{-},\trans_k^{-})$ maps the model points onto the multiset of the points $\vect x'_j=\bar{\vect x}-\opthalf(\vect x_j-\bar{\vect x})$ when $-\halfturn_k$ is a symmetry of the model, or when the model is planar and $k=3$. The point $\vect x'_j$ keeps the first two coordinates of $\vect x_j$ and has depth $2\bar x_3-x_{j3}$. Let $\zeta=\max_j\abs{x_{j3}-\bar x_3}$ be the largest deviation of a depth from that of the centroid, and assume $\zeta<\bar x_3$. Every $\vect x'_j$ then has positive depth, and the pinhole image of the alternative is the multiset of the points $\proj_{\mathrm{cal}}(\vect x'_j)$, which satisfy
\begin{equation}\label{eq:supp-relief}
\begin{aligned}
 \norm{\proj_{\mathrm{cal}}(\vect x'_j)-\proj_{\mathrm{cal}}(\vect x_j)}&=\norm{\proj_{\mathrm{cal}}(\vect x_j)}\frac{2\abs{x_{j3}-\bar x_3}}{2\bar x_3-x_{j3}}\\
 &\le\norm{\proj_{\mathrm{cal}}(\vect x_j)}\frac{2\zeta}{\bar x_3-\zeta}.
\end{aligned}
\end{equation}
If the accepted pose fits the observations and this displacement is small compared with the spacing of the observations in calibrated image units, the mirrored pose also places projections near the observations. A test of the accepted pose alone then cannot exclude it, so Step 4 refines both poses by EM and compares their likelihoods.

\subsection{Matching under an accurate pose}
\label{sec:supp_proof_plugin}

At the selected pose, \method\ labels each observation with its nearest projection. We bound the probability that any of these labels is wrong when every projection is predicted within a tolerance $\predtol$. The bound grows with the density of the projections and with the number of points, which makes precise the remark of \cref{sec:experiments} that each correspondence depends on the spacing between nearby projections. It tends to zero as $N\noise$ and $N\predtol$ tend to zero, and \cref{sec:supp_proof_threshold} compares this with the noise that a joint search over the camera and the correspondences tolerates.

\para{Setting.}
We consider complete inputs with $n=N\ge2$ in which every model point lies at positive depth at the true pose $\xi^\star$. We number the observations so that the true correspondence is the identity. Then $\vect u_i=\vect p_i+\vect w_i$ with $\vect p_i=\vect p_i(\xi^\star)$. We assume that the projections $\vect p_i\in\R^2$ are independent, each with a density bounded by $\fmax$, that they are jointly independent of the errors $\vect w_1,\dots,\vect w_N$, and that $\E\norm{\vect w_i}^2=2\noise^2$. The bound uses no other property of the projection, so it holds for any camera model. A prediction $\hat{\vect p}_j$ of $\vect p_j$ is any measurable function of the data. For an estimated pose $(\hat\Rot,\hat\trans)$ that places every model point at positive depth, the prediction is $\hat{\vect p}_j=\proj(\hat\Rot\vect y_j+\hat\trans)$. The predictions have maximum error $\prederr=\max_j\norm{\hat{\vect p}_j-\vect p_j}$. Nearest-projection matching assigns to observation $i$ a label $\hat\sigma(i)$ that minimizes $\norm{\vect u_i-\hat{\vect p}_j}$ over $j$, with ties resolved by a fixed measurable rule.

\needspace{7\baselineskip}
\begin{lemma}[Matching error under an accurate pose]
\label{prop:plugin}
Under the assumptions above, every fixed tolerance $\predtol\ge0$ on the prediction error satisfies
\begin{equation}\label{eq:supp-plugin}
\begin{aligned}
 &\Pr\bigl[\hat\sigma\ne\mathrm{id}\ \text{and}\ \prederr\le\predtol\bigr]\\
 &\quad\le8\pi\fmax N(N-1)(2\noise^2+\predtol^2).
\end{aligned}
\end{equation}
\end{lemma}

\begin{proof}
We show that an observation can receive a wrong label only if another projection lies close to its own. We then bound the probability of such a close pair by the density bound.

\emph{Step 1. A well-separated observation is labeled correctly.} We put $d_i=\min_{j\ne i}\norm{\vect p_i-\vect p_j}$, and let $\Omega_i=\{d_i>2(\norm{\vect w_i}+\predtol)\}$ be the event that observation $i$ is well separated. On $\Omega_i\cap\{\prederr\le\predtol\}$, every $j\ne i$ satisfies
\begin{equation}
\begin{aligned}
 \norm{\vect u_i-\hat{\vect p}_j}&\ge\norm{\vect p_i-\vect p_j}-\norm{\vect w_i}-\norm{\hat{\vect p}_j-\vect p_j}\\
 &\ge d_i-\norm{\vect w_i}-\predtol\\
 &>\norm{\vect w_i}+\predtol\ge\norm{\vect u_i-\hat{\vect p}_i}.
\end{aligned}
\end{equation}
The first line is the triangle inequality, and the last uses $\vect u_i-\hat{\vect p}_i=\vect w_i+\vect p_i-\hat{\vect p}_i$. Observation $i$ is therefore strictly closest to prediction $i$, and $\hat\sigma(i)=i$ whatever the tie rule. Hence a wrong label with $\prederr\le\predtol$ requires at least one $\Omega_i$ to fail.

\emph{Step 2. Probability of a close pair.} We condition on $\vect p_i$ and $\vect w_i$, since the other projections are independent of them. A disk of radius $d$ has area $\pi d^2$, so by the density bound each other projection falls in the disk of radius $d$ about $\vect p_i$ with probability at most $\pi\fmax d^2$. Taking $d=2(\norm{\vect w_i}+\predtol)$ and using $(\norm{\vect w_i}+\predtol)^2\le2\norm{\vect w_i}^2+2\predtol^2$, we obtain
\begin{equation}
\begin{aligned}
 \Pr[\Omega_i^c]&\le(N-1)\pi\fmax\E\bigl[4(\norm{\vect w_i}+\predtol)^2\bigr]\\
 &\le8\pi\fmax(N-1)(2\noise^2+\predtol^2).
\end{aligned}
\end{equation}
A union bound over the $N$ observations gives \cref{eq:supp-plugin}. At the true pose, the predictions equal the projections, so $\prederr=0$ and Step~2 applies with $\predtol=0$ and $\E[4\norm{\vect w_i}^2]=8\noise^2$. Nearest-projection matching then errs with probability at most $8\pi\fmax N(N-1)\noise^2$.
\end{proof}

\subsection{Exact affine matching threshold}
\label{sec:supp_proof_threshold}

Shuffled regression estimates a linear map without known input-output pairings~\cite{unnikrishnan2018unlabeled}, and moment-based methods recover this map by comparing averages~\cite{tsakiris2020algebraic,liang2026field}. Exact recovery of every pairing in shuffled regression~\cite{lufkin2024sharp,zhang2022diversity} and geometric planted matching~\cite{kunisky2022planted,schwengber2026geometric} requires the noise to stay below a threshold. We examine when every correspondence can be recovered. We compare estimating the correspondences after estimating the camera with a joint search over both.

\method\ estimates the pose first and then labels each observation with its nearest projection (\cref{sec:estimation}). \cref{thm:threshold} concerns a joint least-squares search over an affine camera and all permutations of the model points, which differs from the perspective fitting and EM of \method. For random model points, it shows that this search recovers every correspondence with probability tending to one when $N\noise\to0$, and that no estimator does so when $N\noise$ does not tend to zero, even with the camera known. For fixed noise $\noise>0$, exact recovery of the correspondences therefore fails as the point sets grow. \cref{prop:plugin} gives the corresponding statement for nearest-projection matching. Under its assumptions, which do not depend on the camera model, nearest-projection matching recovers every label with probability tending to one if $N\noise\to0$ and if the largest error $\prederr$ of the predicted projections satisfies $\Pr[\prederr\le\predtol]\to1$ for tolerances $\predtol$ with $N\predtol\to0$. Given an accurate pose, labeling after pose estimation therefore tolerates noise of the same order as the joint search. Under the random model of \cref{thm:threshold}, no estimator tolerates noise of a larger order.

\begin{theorem}[Exact affine matching threshold]
\label{thm:threshold}
Fix a distribution on $\R^3$ that either has a bounded density on a bounded set or is a nondegenerate Gaussian, and fix an affine camera $\vect y\mapsto A\vect y+\vect b$ with $A\in\R^{2\times3}$ of rank two and $\vect b\in\R^2$. Draw $N$ model points $\vect y_1,\dots,\vect y_N$ independently from this distribution, and observe $\vect u_i=A\vect y_{\sigma(i)}+\vect b+\vect w_i$ for $i=1,\dots,N$ and an unknown permutation $\sigma$. The noise vectors $\vect w_i\sim\mathcal N(0,\noise^2I_2)$ are independent of each other and of the model. The noise level $\noise=\noise_N\ge0$ may depend on $N$ as $N\to\infty$. If $N\noise\to0$, the correspondence that minimizes the least-squares reprojection error jointly with the unknown camera is $\sigma$ with probability tending to one. In the bounded case, once $N\noise$ is below a constant, this probability is at least $1-C_{\rm th}(N^2\noise^2+e^{-c_{\rm th}N})$, where the constants $C_{\rm th},c_{\rm th}>0$ depend on the distribution and the camera. If $\limsup N\noise>0$, no estimator recovers a uniformly random $\sigma$ with probability tending to one, even when the camera is known.
\end{theorem}

\para{Scope.}
If the projections have unit variance per coordinate, the signal-to-noise ratio is $\mathrm{SNR}=1/\noise^2$. The threshold in \cref{thm:threshold} then lies at $\mathrm{SNR}$ of order $N^2$, and exact recovery requires $\mathrm{SNR}/N^2\to\infty$. The theorem assumes a fixed model dimension, two image coordinates, Gaussian noise and outlier-free point sets under an affine camera. It concerns the event that every label is correct, whose probability \cref{fig:threshold_all_sizes} measures. One wrong label makes this event fail. \cref{fig:scaling_label} measures the percentage of wrong labels, which grows with $N$ because each label depends on the density of the projections near it (\cref{prop:plugin}). Over the same range of point counts, pose accuracy keeps improving up to one million points (\cref{fig:scaling}). This behavior is consistent with moments averaging over all points, and it motivates estimating the pose before the correspondences.

\para{Outline of the proof.}
Least squares selects the true correspondence when every incorrect correspondence fits worse after the noise is added. We therefore first show that, without noise, every incorrect correspondence leaves a loss that no affine camera removes, and then that noise rarely cancels this loss. The argument depends on how many points the correspondence moves. When it moves few points, its loss is at least half its total squared displacement, because each point has little influence on the fitted camera (\cref{supp:lem-leverage,supp:lem-design}). When it moves many points, its loss exceeds a fixed constant, because a fixed affine map rarely fits it. A count of permutations and a finite covering of the affine maps extend this bound to all maps at once (\cref{supp:lem-path,supp:lem-count,supp:lem-dense}). \cref{supp:lem-separation} combines the two cases, and \cref{supp:lem-noise} bounds the probability that noise cancels the loss of one correspondence. \cref{supp:thm-recovery,supp:thm-gaussian} treat bounded and Gaussian model points. Each averages these probabilities over the model points and sums them over all correspondences, using the second bound of \cref{supp:lem-path} for correspondences that move few points. \cref{supp:thm-lower} proves the converse from a close pair of projections that noise of order $1/N$ confuses. The proof of \cref{thm:threshold} combines the three theorems.

\para{Setting.}
The proof allows model points in $\R^q$ for any dimension $q\ge2$, with $q=3$ in \cref{thm:threshold}. For $N\ge2$, the model points $\vect y_1,\dots,\vect y_N\in\R^q$ are independent and identically distributed. The camera is an affine map $(A,\vect b)$ with $A\in\R^{2\times q}$ of rank two and $\vect b\in\R^2$, fixed independently of the sample. Scaled orthography is the case in which $A$ is a positive multiple of a matrix with orthonormal rows. The dimension, the distribution and the camera stay fixed as $N$ grows. The true projections are $\vect p_i=A\vect y_i+\vect b$. We relabel the observations so that the true correspondence is the identity $\mathrm{id}$, which gives
\begin{equation}\label{eq:supp-observation}
 \vect u_i=\vect p_i+\vect w_i,\quad\vect w_i\sim\mathcal N(0,\noise^2I_2),
\end{equation}
with noise vectors that are independent of each other and of the model. From here on, $\sigma$ denotes any candidate correspondence, that is, any permutation of $\{1,\dots,N\}$.

\para{Least-squares losses.}
We score a candidate correspondence $\sigma$ by its least-squares loss over affine maps,
\begin{equation}\label{eq:supp-losses}
 L(\sigma)=\min_{A',\vect b'}\sum_{i=1}^N\norm{\vect u_i-A'\vect y_{\sigma(i)}-\vect b'}^2,
\end{equation}
and its noiseless loss $L_0(\sigma)$ is the same minimum with $\vect p_i$ in place of $\vect u_i$. The joint least-squares correspondence minimizes $L$ over all permutations. The proof also covers restricted camera families. Let $\mathcal F_{\mathrm{cam}}$ be a family of affine maps that contains the true map $(A,\vect b)$, such as scaled orthography when the true camera has this form. The restricted loss $L^{\mathcal F_{\mathrm{cam}}}(\sigma)$ is the infimum of the sum in \cref{eq:supp-losses} over the family. \cref{thm:threshold} takes all affine maps as the family. To measure how far $\sigma$ moves the points, we write $\supp(\sigma)=\{i:\sigma(i)\ne i\}$ for the indices that $\sigma$ moves and $k(\sigma)=\abs{\supp(\sigma)}$ for their number. The total squared displacement of $\sigma$ and the smallest distance between two true projections are
\begin{equation}\label{eq:supp-counts}
\begin{aligned}
 S_\sigma&=\sum_{i=1}^N\norm{\vect p_i-\vect p_{\sigma(i)}}^2,\\
 d_{\min}&=\min_{i\ne j}\norm{\vect p_i-\vect p_j}.
\end{aligned}
\end{equation}
None of these quantities depends on $\vect b$.

\begin{definition}[Bounded model]\label{def:supp-bounded}
The model is bounded if \textup{(i)} $\norm{\vect y_i}\le Y_{\max}$ almost surely, \textup{(ii)} the covariance of $\vect y_i$ is positive definite and \textup{(iii)} $\vect p_i$ has a density on $\R^2$ bounded by $\fmax$.
\end{definition}

The bounded distributions of \cref{thm:threshold} satisfy these conditions. If $\vect y_i$ has a density bounded by a constant and supported in the ball of radius $Y_{\max}$, then (i) and (ii) hold. Condition (iii) then holds with $\fmax$ the product of the bound on the density of $\vect y_i$, the volume of the unit ball in $\R^{q-2}$ and $Y_{\max}^{q-2}/\det(AA^\top)^{1/2}$. A sample from a surface need not satisfy (iii), because its projection can have an unbounded density at the contour. In the bounds below, all constants depend on the distribution and on $A$.

\para{Least squares in matrix form.}
Writing the data as matrices turns each loss into a squared distance to a column space. We put $\vect y_i^+=(\vect y_i^\top,1)^\top\in\R^{q+1}$ and let $Z\in\R^{N\times(q+1)}$ have rows $\vect y_i^{+\top}$. The $N\times2$ matrices $P^\star$, $U$ and $W$ have rows $\vect p_i^\top$, $\vect u_i^\top$ and $\vect w_i^\top$, so that $U=P^\star+W$. A subscript $\sigma$ reorders matrix rows, so that row $i$ of $Z_\sigma$ is $\vect y_{\sigma(i)}^{+\top}$ and row $i$ of $P^\star_\sigma$ is $\vect p_{\sigma(i)}^\top$. An affine map is the matrix $\Theta'=[A'\ \vect b']\in\R^{2\times(q+1)}$, and the points it predicts under $\sigma$ are the rows of $Z_\sigma\Theta'^\top$. The true map $\Theta=[A\ \vect b]$ gives $P^\star=Z\Theta^\top$. Let $\Pi$ and $\Pi_\sigma$ be the orthogonal projections onto the column spaces of $Z$ and $Z_\sigma$. For each image coordinate, least squares projects the observations onto the column space of $Z_\sigma$, and the loss is the squared Frobenius norm of the residual,
\begin{equation}\label{eq:supp-projector}
\begin{aligned}
 L(\sigma)&=\norm{(I-\Pi_\sigma)U}^2,\\
 L_0(\sigma)&=\norm{(I-\Pi_\sigma)P^\star}^2 .
\end{aligned}
\end{equation}
The diagonal entries of $\Pi$, called leverages, measure the influence of each point on the fit. We write $\Pi_{\max}=\max_i\Pi_{ii}$ for the largest, and we have $(\Pi_\sigma)_{ii}=\Pi_{\sigma(i)\sigma(i)}$ because $Z_\sigma$ reorders the rows of $Z$.

\para{Constants.}
The bounds use the augmented second-moment matrix $\Sigma_+=\E[\vect y_i^+\vect y_i^{+\top}]$. It is positive definite under (ii), because for $(\vect a_{\rm th},a_{{\rm th},0})\ne0$ its quadratic form $\operatorname{Var}(\vect a_{\rm th}\cdot\vect y_i)+(\vect a_{\rm th}\cdot\E\vect y_i+a_{{\rm th},0})^2$ is positive. We use the quantities
\begin{equation*}
\begin{aligned}
 \lambda_\Sigma&=\tfrac12\lambda_{\min}(\Sigma_+), & \Lambda&=\sqrt{Y_{\max}^2+1},\\
 \alpha_{\rm th}&=\lambda_\Sigma/(2\Lambda^2), & \projrad&=Y_{\max}\norm{A}_{\mathrm{op}},\\
 \beta_{\rm th}&=256\pi\fmax \noise^2, & d_0&=\min\Bigl\{1,\frac{\alpha_{\rm th}^2}{1024e^2\pi\fmax}\Bigr\}.
\end{aligned}
\end{equation*}
All of them except $\beta_{\rm th}$, which is proportional to $\noise^2$, are constants. With high probability, $Z^\top Z\succeq N\lambda_\Sigma I_{q+1}$, and the leverage argument then applies to permutations that move at most $\alpha_{\rm th}N$ points (\cref{supp:lem-design}). Also with high probability, permutations that move more points have noiseless loss above $d_0$ (\cref{supp:lem-dense}). The quantity $\beta_{\rm th}$ enters the final sum over permutations. By construction, $\norm{\vect y_i^+}\le\Lambda$ and $\norm{\vect p_i-\vect b}\le\projrad$. The last diagonal entry of $\Sigma_+$ is one and $\Lambda\ge1$, so $\lambda_\Sigma\le\frac12$ and $\alpha_{\rm th}\le\frac14$.

\begin{lemma}[Noiseless loss and permutation displacement]\label{supp:lem-leverage}
Let $x_+=\max\{x,0\}$ denote the positive part of a real number $x$. For every sample and every $\sigma$, we have
\begin{equation}\label{eq:supp-leverage}
 L_0(\sigma)\ge\bigl(1-k(\sigma)\Pi_{\max}\bigr)_+S_\sigma .
\end{equation}
In particular, if $k(\sigma)\Pi_{\max}\le\frac12$, then $L_0(\sigma)\ge S_\sigma/2\ge k(\sigma)d_{\min}^2/2$.
\end{lemma}
\begin{proof}
The true camera maps the reordered model to the reordered projections $P^\star_\sigma$, so $L_0(\sigma)$ is the squared norm of the part of $P^\star-P^\star_\sigma$ that remains after the affine map is refitted. The columns of $P^\star_\sigma=Z_\sigma\Theta^\top$ lie in the column space of $Z_\sigma$, so $(I-\Pi_\sigma)P^\star_\sigma=0$, and \cref{eq:supp-projector} gives
\begin{equation}
\begin{aligned}
 L_0(\sigma)&=\norm{(I-\Pi_\sigma)(P^\star-P^\star_\sigma)}^2\\
 &=S_\sigma-\tr\bigl((P^\star-P^\star_\sigma)^\top\Pi_\sigma(P^\star-P^\star_\sigma)\bigr).
\end{aligned}
\end{equation}
The rows of $P^\star-P^\star_\sigma$ are $(\vect p_i-\vect p_{\sigma(i)})^\top$. They vanish outside $\supp(\sigma)$, and $\norm{P^\star-P^\star_\sigma}^2=S_\sigma$. Only the block of $\Pi_\sigma$ indexed by $\supp(\sigma)$ therefore contributes to the trace, which is at most the largest eigenvalue of that block times $S_\sigma$. The block is positive semidefinite, so its largest eigenvalue is at most its trace,
\begin{equation}
\begin{aligned}
 \sum_{i\in\supp(\sigma)}(\Pi_\sigma)_{ii}&=\sum_{i\in\supp(\sigma)}\Pi_{\sigma(i)\sigma(i)}\\
 &\le k(\sigma)\Pi_{\max} ,
\end{aligned}
\end{equation}
where the equality holds because $Z_\sigma$ reorders the rows of $Z$. This bound and $L_0(\sigma)\ge0$ give \cref{eq:supp-leverage}. Finally, each of the $k(\sigma)$ rows of $P^\star-P^\star_\sigma$ indexed by $\supp(\sigma)$ has norm at least $d_{\min}$, because $\sigma(i)\ne i$ for these rows, so $S_\sigma\ge k(\sigma)d_{\min}^2$.
\end{proof}

\needspace{7\baselineskip}
\begin{lemma}[Design matrix concentration]\label{supp:lem-design}
For a bounded model (\cref{def:supp-bounded}), the event $\Omega_{\mathrm{cond}}=\{Z^\top Z\succeq N\lambda_\Sigma I_{q+1}\}$ satisfies
\begin{equation}\label{eq:supp-design}
 \Pr[\Omega_{\mathrm{cond}}^c]\le2(q+1)^2\exp\Bigl(-\frac{N\lambda_\Sigma^2}{2(q+1)^2\Lambda^4}\Bigr).
\end{equation}
On $\Omega_{\mathrm{cond}}$, we have $\Pi_{\max}\le\Lambda^2/(N\lambda_\Sigma)$, and every $\sigma$ with $k(\sigma)\le\alpha_{\rm th} N$ has $L_0(\sigma)\ge S_\sigma/2$.
\end{lemma}
\begin{proof}
We compare $\frac1NZ^\top Z$ with $\Sigma_+$ entry by entry and then bound the leverages. Each entry of $\vect y_i^+\vect y_i^{+\top}$ lies in $[-\Lambda^2,\Lambda^2]$. By Hoeffding's inequality~\cite{vershynin2018high}, an entry of $\frac1NZ^\top Z-\Sigma_+$ exceeds $\lambda_\Sigma/(q+1)$ in absolute value with probability at most $2\exp(-N\lambda_\Sigma^2/(2(q+1)^2\Lambda^4))$, and there are $(q+1)^2$ entries. If no entry exceeds this threshold, the Frobenius norm of the difference is at most $\lambda_\Sigma$, so $\lambda_{\min}(\frac1NZ^\top Z)\ge2\lambda_\Sigma-\lambda_\Sigma$. On $\Omega_{\mathrm{cond}}$, every leverage satisfies $\Pi_{ii}=\vect y_i^{+\top}(Z^\top Z)^{-1}\vect y_i^+\le\Lambda^2/(N\lambda_\Sigma)$. If $k(\sigma)\le\alpha_{\rm th} N$, then $k(\sigma)\Pi_{\max}\le\frac12$, and \cref{supp:lem-leverage} applies.
\end{proof}

\cref{supp:lem-design} gives no bound when $\sigma$ moves more than $\alpha_{\rm th} N$ points, because the factor $(1-k(\sigma)\Pi_{\max})_+$ in \cref{eq:supp-leverage} can then vanish. For these permutations, we show that a fixed affine map rarely fits an incorrect correspondence (\cref{supp:lem-path}), count permutations by their moved indices and cycles (\cref{supp:lem-count}), and cover all relevant affine maps by finitely many fixed maps (\cref{supp:lem-dense}). For $\sigma\ne\mathrm{id}$, we let $\cyc(\sigma)$ be the number of cycles of $\sigma$ of length at least two and put $r(\sigma)=k(\sigma)-\cyc(\sigma)$. Every such cycle has at least two elements, so $r(\sigma)\ge k(\sigma)/2$.

\begin{lemma}[Residual bound for a fixed incorrect permutation]\label{supp:lem-path}
Suppose that the $\vect y_i$ are independent and that each $\vect p_i$ has a density bounded by $\fmax$. Fix an affine map $(A',\vect b')$ and a permutation $\sigma\ne\mathrm{id}$, and let $r=r(\sigma)$. Then, for every $\tau_{\rm th}>0$, the residuals of $(A',\vect b')$ satisfy the first bound below, and the residuals $\vect p_i-\vect p_{\sigma(i)}$ of the true map, whose squared norms sum to $S_\sigma$, satisfy the second:
\begin{align}
 \Pr\Bigl[\sum_{i=1}^N\norm{\vect p_i-A'\vect y_{\sigma(i)}-\vect b'}^2\le\tau_{\rm th}\Bigr]&\le\frac{(\pi\fmax\tau_{\rm th})^r}{r!},\label{eq:supp-smallball}\\
 \E\exp(-S_\sigma/\tau_{\rm th})&\le(\pi\fmax\tau_{\rm th})^r.\label{eq:supp-laplace}
\end{align}
\end{lemma}
\begin{proof}
We write $\vect\varepsilon_i=\vect p_i-A'\vect y_{\sigma(i)}-\vect b'$ for the residual of observation $i$. A moved model point enters two residuals, so the residuals are not independent. We therefore drop one residual from each cycle of length at least two and integrate the remaining $r$ residuals one at a time, each against a model point that no later residual involves.

\emph{Step 1. Choice of residuals.} We write a cycle of $\sigma$ of length at least two as $i_0\to i_1\to\dots\to i_0$, with $\sigma(i_a)=i_{a+1}$. We discard the residual at $i_0$ and order the remaining residuals of the cycle as $i_1,i_2,\ldots$. Repeating this for every cycle retains $r$ residuals. For $a\ge1$, the point $\vect y_{i_a}$ enters the residual $\vect\varepsilon_{i_a}$ through $\vect p_{i_a}$, the residual $\vect\varepsilon_{i_{a-1}}$ through $A'\vect y_{i_a}$, and no other residual. The residual at $i_{a-1}$ is either discarded or earlier in the order. The retained residuals after $\vect\varepsilon_{i_a}$ therefore do not depend on $\vect y_{i_a}$.

\emph{Step 2. Integration.} We denote the retained residuals, in this order, by $\vect\varepsilon_{(1)},\dots,\vect\varepsilon_{(r)}$. We claim that every measurable $\chi\ge0$ on $(\R^2)^r$ satisfies
\begin{equation}\label{eq:supp-integrate}
 \E\chi\bigl(\vect\varepsilon_{(1)},\dots,\vect\varepsilon_{(r)}\bigr)\le\fmax^r\int_{(\R^2)^r}\chi.
\end{equation}
We condition on all model points except the one with the first retained index. By Step~1, the residuals $\vect\varepsilon_{(2)},\dots,\vect\varepsilon_{(r)}$ are then fixed, and $\vect\varepsilon_{(1)}$ is a translate of the corresponding $\vect p_i$. The model points are independent, so the conditional density of this $\vect p_i$ is bounded by $\fmax$. The conditional expectation of $\chi$ is therefore at most $\fmax\int_{\R^2}\chi(\vect\varepsilon_{(1)},\vect\varepsilon_{(2)},\dots,\vect\varepsilon_{(r)})\,d\vect\varepsilon_{(1)}$, which by Tonelli's theorem~\cite{rana2002measure} is a nonnegative measurable function of the remaining residuals. Repeating the argument for the remaining retained indices in order proves \cref{eq:supp-integrate}.

\emph{Step 3. The two bounds.} The sum of the squared norms of all residuals is at least the sum over the retained residuals. Taking for $\chi$ the indicator of the ball of radius $\sqrt{\tau_{\rm th}}$ in $\R^{2r}$, whose volume is $\pi^r\tau_{\rm th}^r/r!$, gives \cref{eq:supp-smallball}. For \cref{eq:supp-laplace}, we take $(A',\vect b')=(A,\vect b)$, so that $\vect\varepsilon_i=\vect p_i-\vect p_{\sigma(i)}$. We also take $\chi(\vect\varepsilon)=\exp(-\norm{\vect\varepsilon}^2/\tau_{\rm th})$, whose integral over $\R^{2r}$ is $(\pi\tau_{\rm th})^r$.
\end{proof}

\begin{lemma}[Number of permutations by moved points and cycles]\label{supp:lem-count}
For integers $2\le k\le N$ and $1\le j\le\lfloor k/2\rfloor$, the number $N_{N,k,j}$ of permutations with $k(\sigma)=k$ and $\cyc(\sigma)=j$ satisfies $N_{N,k,j}\le N^k2^k/j!$. Consequently, for $0<\varpi\le1$ we have
\begin{equation}\label{eq:supp-count-perm}
 N_{N,k,j}\frac{\varpi^{k-j}}{(k-j)!}\le\Bigl(\frac{4eN\sqrt\varpi}{k}\Bigr)^{k}.
\end{equation}
\end{lemma}
\begin{proof}
We can choose an ordered list of $k$ distinct indices in at most $N^k$ ways. We can cut the list into $j$ consecutive blocks of length at least two in fewer than $2^k$ ways and read each block as a cycle. Every permutation with $k(\sigma)=k$ and $\cyc(\sigma)=j$ arises in this way at least $j!$ times, once for each ordering of its cycles, and dividing by this multiplicity gives the first bound. For the second, we use $j!\,(k-j)!=k!/\binom kj\ge k!/2^k$, then $\varpi^{k-j}\le\varpi^{k/2}$, which holds because $k-j\ge k/2$ and $\varpi\le1$, and finally $k!\ge(k/e)^k$.
\end{proof}

\begin{lemma}[Uniform loss bound for permutations moving many points]\label{supp:lem-dense}
For a bounded model, let $\Omega_{\mathrm{dense}}$ be the event that every $\sigma$ with $k(\sigma)>\alpha_{\rm th} N$ has $L_0(\sigma)>d_0$. Then
\begin{equation}\label{eq:supp-dense}
\begin{aligned}
 \Pr[\Omega_{\mathrm{cond}}\setminus\Omega_{\mathrm{dense}}]&\le4^{-\alpha_{\rm th} N}N^2N_{\mathrm{net}},\\
 N_{\mathrm{net}}&=\Bigl(1+\frac{2(\projrad+1)\Lambda\sqrt N}{\sqrt{\lambda_\Sigma d_0}}\Bigr)^{2(q+1)} .
\end{aligned}
\end{equation}
\end{lemma}
\begin{proof}
The noiseless loss is a minimum over all affine maps, whereas \cref{supp:lem-path} applies to one fixed map. We therefore first show that every map with a small residual is bounded, and then approximate each bounded map by one of finitely many fixed maps. Neither $L_0$ nor $\Omega_{\mathrm{cond}}$ depends on $\vect b$, so we take $\vect b=0$, which gives $\norm{P^\star}\le\projrad\sqrt N$.

\emph{Step 1. The minimizer is bounded.} On $\Omega_{\mathrm{cond}}$, suppose that some $\sigma$ with $k(\sigma)>\alpha_{\rm th} N$ has $L_0(\sigma)\le d_0\le1$, and let $\Theta'$ attain this minimum. Since $Z_\sigma^\top Z_\sigma=Z^\top Z\succeq N\lambda_\Sigma I_{q+1}$, we obtain
\begin{equation}
 \sqrt{N\lambda_\Sigma}\,\norm{\Theta'}\le\norm{Z_\sigma\Theta'^\top}\le\norm{P^\star}+\sqrt{d_0}\le\projrad\sqrt N+1 ,
\end{equation}
so $\norm{\Theta'}\le(\projrad+1)/\sqrt{\lambda_\Sigma}$, uniformly in $\sigma$.

\emph{Step 2. Finite approximation of the affine maps.} We choose a finite subset $\mathcal G$ of the ball of radius $(\projrad+1)/\sqrt{\lambda_\Sigma}$ in $\R^{2\times(q+1)}$ such that every point of the ball lies within $\sqrt{d_0}/(\Lambda\sqrt N)$ of $\mathcal G$. Such a covering set, called a net, needs at most $N_{\mathrm{net}}$ points~\cite{vershynin2018high} and can be fixed before sampling. We replace $\Theta'$ by its nearest point $\Theta''\in\mathcal G$. This replacement changes the matrix of predicted points by at most $\norm Z_{\mathrm{op}}\norm{\Theta'-\Theta''}\le\Lambda\sqrt N\cdot\sqrt{d_0}/(\Lambda\sqrt N)=\sqrt{d_0}$, so the residual of $\Theta''$ has norm at most $2\sqrt{d_0}$. Hence, if $\Omega_{\mathrm{cond}}$ holds and $\Omega_{\mathrm{dense}}$ fails, there are a point $\Theta''\in\mathcal G$ and a permutation $\sigma$ with $k(\sigma)>\alpha_{\rm th} N$ such that
\begin{equation}\label{eq:supp-net}
 \sum_{i=1}^N\norm{\vect p_i-\Theta''\vect y_{\sigma(i)}^+}^2\le4d_0.
\end{equation}

\emph{Step 3. Union bound.} The points of $\mathcal G$ do not depend on the sample, so \cref{supp:lem-path} applies to each of them. Fix $\Theta''$ and $\sigma$. By \cref{eq:supp-smallball} with $\tau_{\rm th}=4d_0$, the event of \cref{eq:supp-net} has probability at most $\varpi^{r(\sigma)}/r(\sigma)!$, where $\varpi=4\pi\fmax d_0$. The choice of $d_0$ gives $\sqrt\varpi\le\alpha_{\rm th}/(16e)\le1$. By \cref{eq:supp-count-perm}, the permutations with $k(\sigma)=k$ and $\cyc(\sigma)=j$ contribute at most $(4eN\sqrt\varpi/k)^k$ in total. For $k>\alpha_{\rm th} N$, this is at most $(4e\sqrt\varpi/\alpha_{\rm th})^k\le4^{-k}\le4^{-\alpha_{\rm th} N}$. Summing over the at most $N^2$ pairs $(k,j)$ and the at most $N_{\mathrm{net}}$ points of $\mathcal G$ gives \cref{eq:supp-dense}.
\end{proof}

\begin{lemma}[Lower bounds on noiseless loss]\label{supp:lem-separation}
For a bounded model, let $\Omega_{\mathrm{sep}}=\Omega_{\mathrm{cond}}\cap\Omega_{\mathrm{dense}}$. Then
\begin{equation}\label{eq:supp-pides}
\begin{aligned}
 \Pr[\Omega_{\mathrm{sep}}^c]&\le p_{\mathrm{des}},\\
 p_{\mathrm{des}}&=2(q+1)^2\exp\Bigl(-\frac{N\lambda_\Sigma^2}{2(q+1)^2\Lambda^4}\Bigr)\\
 &\quad+4^{-\alpha_{\rm th} N}N^2N_{\mathrm{net}},
\end{aligned}
\end{equation}
and $p_{\mathrm{des}}\le C_{\rm des}e^{-c_{\rm th}N}$ for constants $C_{\rm des},c_{\rm th}>0$. On $\Omega_{\mathrm{sep}}$, every $\sigma\ne\mathrm{id}$ satisfies $L_0(\sigma)\ge S_\sigma/2\ge k(\sigma)d_{\min}^2/2$ if $k(\sigma)\le\alpha_{\rm th} N$, and $L_0(\sigma)>d_0$ otherwise.
\end{lemma}
\begin{proof}
The complement of $\Omega_{\mathrm{sep}}$ is contained in the union of $\Omega_{\mathrm{cond}}^c$ and $\Omega_{\mathrm{cond}}\setminus\Omega_{\mathrm{dense}}$, so its probability is at most the sum of the bounds in \cref{eq:supp-design,eq:supp-dense}. This sum is exponentially small because $N^2N_{\mathrm{net}}$ grows polynomially in $N$. On $\Omega_{\mathrm{sep}}$, the first claim follows from \cref{supp:lem-design,supp:lem-leverage} and the second from the definition of $\Omega_{\mathrm{dense}}$.
\end{proof}

\begin{lemma}[Noise perturbation of the loss]\leavevmode\label{supp:lem-noise}
Let $\noise>0$. For every fixed model sample and every $\sigma$, the probability over the noise satisfies
\begin{equation}\label{eq:supp-noise-one}
\begin{aligned}
 &\Pr\bigl[L(\sigma)-\norm W^2<L_0(\sigma)/2\bigr]\\
 &\quad\le\bigl(1+2^{q+1}\bigr)\exp\Bigl(-\frac{L_0(\sigma)}{128\noise^2}\Bigr).
\end{aligned}
\end{equation}
\end{lemma}
\begin{proof}
By \cref{eq:supp-projector} and $U=P^\star+W$, we have
\begin{equation}
\begin{aligned}
 L(\sigma)-\norm W^2&=L_0(\sigma)+2\bigl\langle(I-\Pi_\sigma)P^\star,W\bigr\rangle\\
 &\quad-\norm{\Pi_\sigma W}^2.
\end{aligned}
\end{equation}
The middle term is linear in the noise, and the last term is the reduction in loss obtained by fitting the noise. The middle term is a centered Gaussian variable with variance $4\noise^2L_0(\sigma)$, so it is below $-L_0(\sigma)/4$ with probability at most $\exp(-L_0(\sigma)/(128\noise^2))$. The variable $\norm{\Pi_\sigma W}^2/\noise^2$ is chi-square with at most $2(q+1)$ degrees of freedom, and its moment generating function at $\frac14$ is at most $2^{q+1}$. By Markov's inequality, $\norm{\Pi_\sigma W}^2$ exceeds $L_0(\sigma)/4$ with probability at most $2^{q+1}\exp(-L_0(\sigma)/(16\noise^2))$. When neither deviation occurs, the loss difference is at least $L_0(\sigma)/2$. Since $e^{-L_0(\sigma)/(16\noise^2)}\le e^{-L_0(\sigma)/(128\noise^2)}$, the sum of the two probabilities is at most the bound in \cref{eq:supp-noise-one}.
\end{proof}

\begin{theorem}[Exact recovery for bounded model points]\label{supp:thm-recovery}
For a bounded model with $\noise>0$ and $N\beta_{\rm th}<1$, let
\begin{equation}\label{eq:supp-event}
\begin{aligned}
 \Omega_{\mathrm{rec}}=\bigl\{&L^{\mathcal F_{\mathrm{cam}}}(\sigma)-L^{\mathcal F_{\mathrm{cam}}}(\mathrm{id})\ge L_0(\sigma)/2>0\\
 &\text{for all }\sigma\ne\mathrm{id}\text{ and all }\mathcal F_{\mathrm{cam}}\ni(A,\vect b)\bigr\}.
\end{aligned}
\end{equation}
Then $\Omega_{\mathrm{rec}}\cap\Omega_{\mathrm{sep}}$ has probability at least
\begin{equation}\label{eq:supp-bound}
\begin{aligned}
 1-p_{\mathrm{des}}-\bigl(1+2^{q+1}\bigr)\Bigl[&\exp\Bigl(\frac{N^2\beta_{\rm th}}{2(1-N\beta_{\rm th})}\Bigr)-1\\
 &+N!\,e^{-d_0/(128\noise^2)}\Bigr].
\end{aligned}
\end{equation}
Consequently, there are constants $c_1,C_{\rm th},c_{\rm th}>0$ such that, for all $N$ and all $0<\noise^2\le c_1/N^2$, we have
\begin{equation}\label{eq:supp-rate}
 \Pr[\Omega_{\mathrm{rec}}\cap\Omega_{\mathrm{sep}}]\ge1-C_{\rm th}N^2\noise^2-C_{\rm th}e^{-c_{\rm th}N}.
\end{equation}
On $\Omega_{\mathrm{rec}}$, the identity is the unique minimizer of $L^{\mathcal F_{\mathrm{cam}}}$, so the least-squares correspondence is the true one.
\end{theorem}
\begin{proof}
We bound the probability that some incorrect correspondence has a loss too close to or below the true loss. Bounding each such probability through $d_{\min}$ alone would add a logarithmic factor to the required signal-to-noise ratio, giving order $N^2\log N$. We avoid this factor by averaging each error probability over the model before summing over the $N!-1$ incorrect correspondences, and by counting permutations by their cycles.

\emph{Step 1. Reduction to the unrestricted loss.} Restricting the family of affine maps cannot decrease the loss, so $L^{\mathcal F_{\mathrm{cam}}}(\sigma)\ge L(\sigma)$. Evaluating at the true map gives $L^{\mathcal F_{\mathrm{cam}}}(\mathrm{id})\le\norm W^2$. It therefore suffices to bound the probability that $L(\sigma)-\norm W^2<L_0(\sigma)/2$ for some $\sigma\ne\mathrm{id}$. Almost surely $d_{\min}>0$, so on $\Omega_{\mathrm{sep}}$ every $L_0(\sigma)$ with $\sigma\ne\mathrm{id}$ is positive by \cref{supp:lem-separation}.

\emph{Step 2. Conditioning on the model.} The noise is independent of the model, so we apply \cref{supp:lem-noise} with the model fixed and then average the conditional bound over the model. A union bound shows that the probability that $\Omega_{\mathrm{sep}}$ fails, or that some $\sigma\ne\mathrm{id}$ has $L(\sigma)-\norm W^2<L_0(\sigma)/2$, is at most
\begin{equation}\label{eq:supp-sum}
 p_{\mathrm{des}}+\bigl(1+2^{q+1}\bigr)\sum_{\sigma\ne\mathrm{id}}\E\Bigl[\mathbf 1_{\Omega_{\mathrm{sep}}}e^{-L_0(\sigma)/(128\noise^2)}\Bigr].
\end{equation}

\emph{Step 3. Correspondences that move many points.} For $k(\sigma)>\alpha_{\rm th} N$, we have $L_0(\sigma)>d_0$ on $\Omega_{\mathrm{sep}}$, so the summand is at most $e^{-d_0/(128\noise^2)}$. There are fewer than $N!$ such $\sigma$.

\emph{Step 4. Correspondences that move few points.} For $k(\sigma)\le\alpha_{\rm th} N$, we have $L_0(\sigma)\ge S_\sigma/2$ on $\Omega_{\mathrm{sep}}$. We use this inequality, drop the indicator and apply \cref{eq:supp-laplace} with $\tau_{\rm th}=256\noise^2$, which gives
\begin{equation}
\begin{aligned}
 \E\Bigl[\mathbf 1_{\Omega_{\mathrm{sep}}}e^{-L_0(\sigma)/(128\noise^2)}\Bigr]&\le\E e^{-S_\sigma/(256\noise^2)}\\
 &\le\beta_{\rm th}^{r(\sigma)}.
\end{aligned}
\end{equation}

\emph{Step 5. Sum over cycle types.} To sum $\beta_{\rm th}^{r(\sigma)}$ over all $\sigma\ne\mathrm{id}$, which include those of Step~4, we group the permutations by their cycle lengths. A permutation with $\mu_a$ cycles of length $a$ for each $a\ge2$ has $k=\sum_aa\mu_a$ and $r=\sum_a(a-1)\mu_a$. By Cauchy's formula for cycle types, the number of such permutations is $N!/((N-k)!\prod_aa^{\mu_a}\mu_a!)\le\prod_a(N^a/a)^{\mu_a}/\mu_a!$. A cycle type is a choice of $\mu_2,\dots,\mu_N\ge0$, not all zero. We allow every such choice, even when the cycles would use more than $N$ indices. The added terms are nonnegative, and the sum then factors over $a$. Subtracting the term of the identity, for which every $\mu_a$ is zero, gives
\begin{equation}
\begin{aligned}
 \sum_{\sigma\ne\mathrm{id}}\beta_{\rm th}^{r(\sigma)}&\le\prod_{a=2}^N\sum_{\mu\ge0}\frac1{\mu!}\Bigl(\frac{N^a\beta_{\rm th}^{a-1}}{a}\Bigr)^{\mu}-1\\
 &=\exp\Bigl(\sum_{a=2}^N\frac{N^a\beta_{\rm th}^{a-1}}{a}\Bigr)-1\\
 &\le\exp\Bigl(\frac{N^2\beta_{\rm th}}{2(1-N\beta_{\rm th})}\Bigr)-1 .
\end{aligned}
\end{equation}
Together with $p_{\mathrm{des}}$ from Step~2 and the bound of Step~3, this bound proves \cref{eq:supp-bound}.

\emph{Step 6. The rate.} Let $c_1=\min\{1/(512\pi\fmax),d_0/256\}$ and suppose that $\noise^2\le c_1/N^2$. Then $N^2\beta_{\rm th}\le\frac12$ and $N\beta_{\rm th}\le\frac14$, so the exponent in \cref{eq:supp-bound} is at most $N^2\beta_{\rm th}$. Since $e^x-1\le2x$ on $[0,1]$, the first term in the bracket is at most $512\pi\fmax N^2\noise^2$. For the second term, the function $\noise^2\mapsto e^{-d_0/(128\noise^2)}/\noise^2$ is increasing for $\noise^2\le d_0/128$. Its value is therefore at most its value at $\noise^2=c_1/N^2$, where $d_0/(128\noise^2)\ge2N^2$. Hence $N!\,e^{-d_0/(128\noise^2)}\le(N^2\noise^2/c_1)\exp(N\log N-2N^2)\le N^2\noise^2/c_1$. \cref{eq:supp-bound} and the bound $p_{\mathrm{des}}\le C_{\rm des}e^{-c_{\rm th}N}$ of \cref{supp:lem-separation} give \cref{eq:supp-rate} with $C_{\rm th}=\max\{C_{\rm des},(1+2^{q+1})(512\pi\fmax+1/c_1)\}$.
\end{proof}

\begin{theorem}[Exact recovery for Gaussian model points]\label{supp:thm-gaussian}
Suppose that the $\vect y_i$ are independent and identically distributed Gaussian vectors with a positive definite covariance $\Sigma$, and that $A$ has rank two. There are constants $c_1,C_{\rm th}>0$ such that, for all $N\ge2$ and all $0<\noise^2\le c_1/N^2$, the event $\Omega_{\mathrm{rec}}$ of \cref{eq:supp-event} has probability at least $1-C_{\rm th}\,N^2\noise^2-C_{\rm th}/N$.
\end{theorem}
\begin{proof}
A Gaussian distribution has unbounded support, so \cref{supp:thm-recovery} does not apply directly. We repeat its proof on an event on which every sampled point lies within a deterministic radius that grows slowly with $N$. The losses and $d_{\min}$ do not change when the model is translated, so we suppose that the $\vect y_i$ have zero mean. Then $\vect p_i$ is Gaussian with covariance $A\Sigma A^\top$, its density is at most $\fmax=(2\pi)^{-1}\det(A\Sigma A^\top)^{-1/2}$, and $\Sigma_+=\operatorname{diag}(\Sigma,1)$.

\emph{Step 1. Truncation.} We put $Y_N^2=(2q+6\log N)\norm\Sigma_{\mathrm{op}}$, $\Lambda_N^2=Y_N^2+1$, $\projrad_N=Y_N\norm{A}_{\mathrm{op}}$, $\alpha_{{\rm th},N}=\lambda_\Sigma/(2\Lambda_N^2)$ and $d_{0,N}=\min\{1,\alpha_{{\rm th},N}^2/(1024e^2\pi\fmax)\}$. Let $\Omega_{\mathrm{cond},N}$ be the event that $\max_i\norm{\vect y_i}\le Y_N$ and $Z^\top Z\succeq N\lambda_\Sigma I_{q+1}$. By the chi-square tail bound of Laurent and Massart~\cite{laurent2000adaptive}, each $\norm{\vect y_i}^2$ exceeds $Y_N^2$ with probability at most $N^{-2}$, so the first condition fails with probability at most $1/N$. The entries of $\vect y_i^+\vect y_i^{+\top}$ have finite moment generating functions near zero. By the Chernoff bound, each entry of $\frac1NZ^\top Z-\Sigma_+$ therefore exceeds $\lambda_\Sigma/(q+1)$ in absolute value with probability at most $2e^{-c_{\rm G}N}$ for some $c_{\rm G}>0$. Hence $\Pr[\Omega_{\mathrm{cond},N}^c]\le1/N+2(q+1)^2e^{-c_{\rm G}N}$.

\emph{Step 2. Separation.} Apart from the concentration of $Z^\top Z$, which Step~1 replaces, the proofs of \cref{supp:lem-design,supp:lem-dense} use $Y_{\max}$ only through three bounds, namely $\norm{\vect y_i^+}\le\Lambda$ for the leverages, $\norm{P^\star}\le\projrad\sqrt N$ for the minimizer and $\norm Z_{\mathrm{op}}\le\Lambda\sqrt N$ for the net. On $\Omega_{\mathrm{cond},N}$, all three hold with $(Y_{\max},\Lambda,\alpha_{\rm th},d_0,\projrad)$ replaced by the deterministic quantities $(Y_N,\Lambda_N,\alpha_{{\rm th},N},d_{0,N},\projrad_N)$. \cref{supp:lem-path} needs only the independence of the model points and the density bound, which hold without truncation. Thus, on $\Omega_{\mathrm{cond},N}$, every $\sigma$ with $k(\sigma)\le\alpha_{{\rm th},N}N$ satisfies $L_0(\sigma)\ge S_\sigma/2$. The event that $\Omega_{\mathrm{cond},N}$ holds and some $\sigma$ with $k(\sigma)>\alpha_{{\rm th},N}N$ has $L_0(\sigma)\le d_{0,N}$ has probability at most $4^{-\alpha_{{\rm th},N}N}N^2N_{\mathrm{net},N}$, where $N_{\mathrm{net},N}$ is $N_{\mathrm{net}}$ with the same replacements. We define $\Omega_{\mathrm{sep},N}$ by removing this event from $\Omega_{\mathrm{cond},N}$. Since $\log N\ge\log2$ for $N\ge2$, there is a constant $a_{\rm G}>0$ such that $\alpha_{{\rm th},N}\ge a_{\rm G}/\log N$ and $d_{0,N}\ge a_{\rm G}/\log^2N$ for every $N\ge2$. Hence $N_{\mathrm{net},N}$, in which $\projrad_N$ and $\Lambda_N$ grow like $\sqrt{\log N}$, is at most a polynomial in $N$, and $4^{-\alpha_{{\rm th},N}N}\le\exp(-a_{\rm G}N\log4/\log N)$. The probability of the removed event, multiplied by $N$, therefore tends to zero and is bounded over $N\ge2$. With Step~1, this gives a constant $C_{\rm G}$ such that $\Pr[\Omega_{\mathrm{sep},N}^c]\le C_{\rm G}/N$ for every $N\ge2$.

\emph{Step 3. Sum over the correspondences.} Steps~1 to~5 of the proof of \cref{supp:thm-recovery} apply with $\Omega_{\mathrm{sep},N}$ in place of $\Omega_{\mathrm{sep}}$ and $\Pr[\Omega_{\mathrm{sep},N}^c]$ in place of $p_{\mathrm{des}}$. The correspondences with $k(\sigma)\le\alpha_{{\rm th},N}N$ contribute at most $(1+2^{q+1})[\exp(N^2\beta_{\rm th}/(2(1-N\beta_{\rm th})))-1]$, which is at most a multiple of $N^2\noise^2$ as in Step~6 of that proof. The others contribute at most $(1+2^{q+1})\,N!\,e^{-d_{0,N}/(128\noise^2)}$. We choose $c_1\le\min\{1/(512\pi\fmax),\ \inf_{N\ge2}N^2d_{0,N}/256\}$, where the infimum is positive because every term is positive and $N^2d_{0,N}\to\infty$. Then $\noise^2\le c_1/N^2\le d_{0,N}/128$, and the monotonicity argument of Step~6 gives
\begin{equation}
 N!\,e^{-d_{0,N}/(128\noise^2)}\le\frac{N^2\noise^2}{c_1}\exp\Bigl(N\log N-\frac{N^2d_{0,N}}{128c_1}\Bigr).
\end{equation}
The exponent tends to $-\infty$ because $N^2d_{0,N}\ge a_{\rm G}N^2/\log^2N$. The exponential therefore tends to zero and is bounded over $N\ge2$, so this contribution is also at most a multiple of $N^2\noise^2$. Adding $\Pr[\Omega_{\mathrm{sep},N}^c]\le C_{\rm G}/N$ gives the bound of the theorem.
\end{proof}

\begin{theorem}[No exact recovery above the noise threshold]\label{supp:thm-lower}
Suppose that the true projections $\vect p_1,\dots,\vect p_N$ are independent with a common density $f_{\rm p}$ on $\R^2$ bounded by $\fmax$, and let $I_{\rm p}=\int f_{\rm p}^2$ be the squared $L^2$ norm of this density. The observations are $\vect u_i=\vect p_{\sigma(i)}+\vect w_i$, where $\sigma$ is the true correspondence and the noise vectors $\vect w_i\sim\mathcal N(0,\noise^2I_2)$ are independent of each other and of the model. Suppose that $\sigma$ is uniformly distributed over all permutations and independent of the model and the noise, and let the estimator know the model and the camera $(A,\vect b)$. For every $c_0>0$ there is an $N_0(f_{\rm p},c_0)$ such that, whenever $N\ge N_0$ and $N\noise\ge c_0$, every measurable estimator errs with probability at least
\begin{equation}\label{eq:supp-lower}
 \Phi\bigl(-1/\sqrt2\bigr)\cdot\frac{(\pi c_0^2I_{\rm p}/8)^2}{\pi\fmax c_0^2/2+\tfrac14\pi^2\fmax^2c_0^4}>0,
\end{equation}
where $\Phi$ is the standard normal distribution function. The same bound holds for the worst case over the true correspondence.
\end{theorem}
\begin{proof}
With probability bounded away from zero, two true projections lie within $c_0/N$ of each other. At noise level $\noise\ge c_0/N$, an estimator confuses this pair with positive probability even when it knows all other labels.

\emph{Step 1. A close pair occurs with probability bounded away from zero.} Let $N_{\mathrm{close}}$ be the number of unordered pairs with $\norm{\vect p_i-\vect p_j}\le c_0/N$. For $d>0$, let $f_{{\rm p},d}(\vect u_{\rm d})$ be the average of $f_{\rm p}$ over the disk of radius $d$ about $\vect u_{\rm d}$, so that $f_{{\rm p},d}\le\fmax$ and $\Pr[\norm{\vect p_1-\vect p_2}\le d]=\pi d^2\int f_{\rm p}f_{{\rm p},d}$. By the Lebesgue differentiation theorem~\cite{rana2002measure}, $f_{{\rm p},d}\to f_{\rm p}$ almost everywhere as $d\to0$. Since $f_{\rm p}f_{{\rm p},d}\le\fmax f_{\rm p}$, dominated convergence gives $\int f_{\rm p}f_{{\rm p},d}\to I_{\rm p}>0$. For $N\ge N_0$, therefore,
\begin{equation}
 \frac{\pi c_0^2I_{\rm p}}{8}\le\E N_{\mathrm{close}}\le\binom N2\pi\fmax\frac{c_0^2}{N^2}\le\frac{\pi\fmax c_0^2}{2}.
\end{equation}
Two distinct pairs are both close with probability at most $(\pi\fmax c_0^2/N^2)^2$. For disjoint pairs this follows from independence, and for pairs that share a point it follows by conditioning on that point. There are fewer than $N^4/4$ ordered choices of two distinct pairs, so $\E N_{\mathrm{close}}^2\le\E N_{\mathrm{close}}+\frac14\pi^2\fmax^2c_0^4$. By the second-moment method, which is the Cauchy--Schwarz inequality, $\Pr[N_{\mathrm{close}}>0]\ge(\E N_{\mathrm{close}})^2/\E N_{\mathrm{close}}^2$. This is at least the fraction that multiplies $\Phi(-1/\sqrt2)$ in \cref{eq:supp-lower}.

\emph{Step 2. A close pair forces a positive error probability.} We choose $\{i,j\}$ as a closest pair of true projections, with the choice determined by the model. We reveal to the estimator the true correspondence of every other model point, together with the unordered pair of observations generated by $\vect p_i$ and $\vect p_j$. Given the revealed labels, the two remaining assignments have equal prior probabilities. Under each assignment, the ordered pair of observations is Gaussian in $\R^4$ with covariance $\noise^2I_4$. The two means are distinct almost surely and lie $\sqrt2\,d_{\min}$ apart. With equal priors and a common isotropic covariance, the nearest-mean test is optimal. Projection onto the line through the means gives its error probability $\Phi(-d_{\min}/(\sqrt2\noise))$, which is at least $\Phi(-1/\sqrt2)$ when $d_{\min}\le c_0/N\le \noise$, that is, on $\{N_{\mathrm{close}}>0\}$. An estimator without the revealed information cannot do better, so it errs with at least this probability on $\{N_{\mathrm{close}}>0\}$. Integrating over the model gives \cref{eq:supp-lower}. The worst-case error over the true correspondence is at least the average.
\end{proof}

\begin{proof}[Proof of \cref{thm:threshold}]
We combine the three preceding theorems for a fixed distribution and camera, using the constants of each. For $0<\noise^2\le c_1/N^2$, the probability that the least-squares correspondence is wrong is at most $C_{\rm th}(N^2\noise^2+e^{-c_{\rm th}N})$ for a bounded model by \cref{supp:thm-recovery}, and at most $C_{\rm th}(N^2\noise^2+1/N)$ for Gaussian model points by \cref{supp:thm-gaussian}. For $\noise=0$, the loss $L$ equals $L_0$, which vanishes at the identity. When $d_{\min}>0$, which holds almost surely, every $\sigma\ne\mathrm{id}$ has positive noiseless loss on $\Omega_{\mathrm{sep}}$ for a bounded model (\cref{supp:lem-separation}) and on $\Omega_{\mathrm{sep},N}$ for Gaussian model points (proof of \cref{supp:thm-gaussian}). The error probability is then at most $C_{\rm des}e^{-c_{\rm th}N}$ or $C_{\rm G}/N$, respectively. Since the proof of \cref{supp:thm-recovery} takes $C_{\rm th}\ge C_{\rm des}$, the bounded case gives the stated bound whenever $N\noise\le\sqrt{c_1}$. If $N\noise\to0$, then eventually $\noise^2\le c_1/N^2$, and all these bounds tend to zero. If $\limsup N\noise>0$, \cref{supp:thm-lower} applies along a subsequence in both models, so no estimator recovers a uniformly distributed correspondence with probability tending to one, even when the camera is known. If ties among minimizing correspondences are broken uniformly, relabeling the input relabels the output distribution in the same way. The error probability of least squares is then the same for every true correspondence, so it does not recover any fixed true correspondence with probability tending to one either.
\end{proof}

\subsection{Outliers on the structure of the inliers}
\label{sec:supp_proof_colocated}

Moment matching models outliers as uniform. Real outliers often lie on the same structure as the inliers (\cref{sec:experiments}), and we treat this case for the model points. Suppose that the outlier model points project with the same image distribution as the inliers. Including them then changes the feature average of the projections by at most $1/\sqrt{m}$ in root mean square, which is small when the number $m$ of inliers is large. \cref{sec:supp_ablation} tests the method when outliers among the 2D points lie on the scene structure.

The assumption concerns the distribution of the projections in the image. Model points on the same physical surface need not have a common projected distribution, for example when visibility or sampling density differs between them.

\needspace{7\baselineskip}
\begin{lemma}[Inliers \& outliers from a common distribution]
\label{lem:colocated}
We fix a measurable feature map $\vect\phi:\R^2\to\R^{d_\phi}$ with $\norm{\vect\phi(\vect p)}\le1$ for every $\vect p\in\R^2$. We fix integers $1\le m\le N$ and draw the projections $\vect p_1,\dots,\vect p_N$ independently from one image distribution. We designate the first $m$ projections as inliers, independently of their sampled positions. Then, in expectation over these samples,
\begin{equation}
\E\,\Bigl\lVert\frac1N\sum_{i\le N}\vect\phi(\vect p_i)-\frac1{m}\sum_{i\le m}\vect\phi(\vect p_i)\Bigr\rVert^2
\le\frac1{m}.
\end{equation}
\end{lemma}

\begin{proof}
We write $N_{\mathrm{out}}=N-m$ for the number of outliers. When $N_{\mathrm{out}}\ge1$, we define the inlier and outlier feature averages
\begin{equation}
\bar{\vect\phi}_{\mathrm{in}}=\frac1{m}\sum_{i\le m}\vect\phi(\vect p_i),\qquad
\bar{\vect\phi}_{\mathrm{out}}=\frac1{N_{\mathrm{out}}}\sum_{i>m}\vect\phi(\vect p_i).
\end{equation}
We write the difference in the lemma as a multiple of the difference between these two averages, which are independent and have the same expectation.

\emph{Step 1. Decomposition.} If $N_{\mathrm{out}}=0$, the two averages in the lemma coincide. Otherwise, the average over all $N$ points is $(m\bar{\vect\phi}_{\mathrm{in}}+N_{\mathrm{out}}\bar{\vect\phi}_{\mathrm{out}})/N$, so the difference in the lemma equals
\begin{equation}
\frac1N\sum_{i\le N}\vect\phi(\vect p_i)-\bar{\vect\phi}_{\mathrm{in}}
=\frac{N_{\mathrm{out}}}{N}\bigl(\bar{\vect\phi}_{\mathrm{out}}-\bar{\vect\phi}_{\mathrm{in}}\bigr).
\label{eq:supp-colocated-split}
\end{equation}

\emph{Step 2. Two independent averages.} We put $\vect\mu=\E\,\vect\phi(\vect p_1)$ and $v=\E\,\norm{\vect\phi(\vect p_1)-\vect\mu}^2\le\E\,\norm{\vect\phi(\vect p_1)}^2\le1$. The two averages use disjoint sets of independent points, so they are independent and both have expectation~$\vect\mu$. The mean squares of their deviations from $\vect\mu$ are $v/m$ and $v/N_{\mathrm{out}}$. The cross term vanishes by independence, so
\begin{equation}
\E\,\norm{\bar{\vect\phi}_{\mathrm{out}}-\bar{\vect\phi}_{\mathrm{in}}}^2
=v\Bigl(\frac1{m}+\frac1{N_{\mathrm{out}}}\Bigr)\le\frac{N}{m\,N_{\mathrm{out}}}.
\end{equation}

\emph{Step 3. Conclusion.} Multiplying by $(N_{\mathrm{out}}/N)^2$ gives the exact mean square $v\,N_{\mathrm{out}}/(N\,m)\le N_{\mathrm{out}}/(N\,m)\le1/m$.
\end{proof}

\para{Outliers from another distribution.}
Without the common distribution, \cref{eq:supp-colocated-split} and $\norm{\vect\phi}\le1$ give the deterministic bound $2N_{\mathrm{out}}/N$, as in \cref{supp:lem-contamination}. For a fixed fraction $N_{\mathrm{out}}/N$ of outliers, this bound does not decrease as the number of points grows. Suppose that the projections are independent, with the inliers drawn from one distribution with feature mean $\vect\mu$ and the outliers from another with feature mean $\vect\mu_{\mathrm{out}}$. The computation of Steps~2 and~3 then shows that the difference in the lemma lies within $1/\sqrt m$ of $(N_{\mathrm{out}}/N)(\vect\mu_{\mathrm{out}}-\vect\mu)$ in root mean square. For a fixed outlier fraction and $\vect\mu_{\mathrm{out}}\ne\vect\mu$, this vector is nonzero and does not change as the point sets grow. Differences in visibility or sampling density between the inlier and the outlier model points can produce this case, even when the outliers lie on the same structure as the inliers.

\subsection{Camera initialization}
\label{sec:supp_camera_localization}
Camera pose can be estimated by aligning point distributions~\cite{campbell2019gosma}. We initialize the camera position and orientation from observations $\dets$, a 3D model $\pts$ and known intrinsics $\Kmat$. Model points outside the observed view may have no image counterparts. We search for poses whose projections concentrate near the observations and select the model points that project into the observed image region. These points form the model for refining the pose by expectation--maximization (EM) and for the alignment check.

\para{Camera centers and orientations.}
We parameterize a pose by its rotation $\Rot$ and camera center $\vect c$, with $\trans=-\Rot\vect c$. We form a box from the 1st and 99th percentiles of the model coordinates along the principal axes of the model. The diagonal of this box defines the scene extent, and we enlarge the box by half of each side length on both sides. Three full grids cover the enlarged box with maximum cell side lengths of 20\%, 10\% and 5\% of the scene extent. Along a box axis of length $L_k$, a level with cell-size limit $\Delta_{\rm c}$ has $\lceil L_k/\Delta_{\rm c}\rceil$ intervals of equal length. We test the center of every resulting 3D cell. At each center, we test 11{,}520 rotations, using 640 viewing directions distributed over the sphere and 18 roll angles separated by $20\degree$. The box bounds the initial grid, but local refinement can move the center beyond it.

\para{Projection densities.}
The calibrated observation $\vect u_{{\rm cal},i}$ consists of the first two coordinates of $\Kmat^{-1}(\vect u_i^\top,1)^\top$. At a pose $\xi=(\Rot,\trans)$, let $\vect p_{{\rm cal},j}(\xi)$ be the first two coordinates of $\Rot(\vect y_j-\vect c)$ divided by its depth. We use a narrow and a broad Gaussian bandwidth $0<\tau_{\rm n}<\tau_{\rm b}$. Let $\mathcal J_{\rm cam}(\xi)$ contain the indices of model points at positive depth whose projections lie in the bounding rectangle of the calibrated observations, extended by $4.25\tau_{\rm b}$ on each side. For a bandwidth $\tau>0$, the smoothed projection density is
\begin{equation}
\label{eq:camera-density}
f_\tau(\vect z;\xi)=\sum_{j\in\mathcal J_{\rm cam}(\xi)}
\frac{\exp\!\big(-\norm{\vect z-\vect p_{{\rm cal},j}(\xi)}^2/(2\tau^2)\big)}{2\pi\tau^2}.
\end{equation}

The density ratio at observation $i$ is
\begin{equation}
q_{{\rm s},i}(\xi)=\frac{f_{\tau_{\rm n}}(\vect u_{{\rm cal},i};\xi)}{f_{\tau_{\rm b}}(\vect u_{{\rm cal},i};\xi)+\varepsilon_{\rm cam}},
\label{eq:camera-density-ratio}
\end{equation}
where $\varepsilon_{\rm cam}>0$ prevents division by zero. Each Gaussian kernel has unit integral, so the ratio compares concentration at an observation with concentration in its neighborhood. We jointly vary the rotation and camera center to minimize
\begin{equation}
\label{eq:camera-search}
\mathcal L_{\rm cam}(\xi)=-\frac1n\sum_{i=1}^n\log\big(1-\alpha+\alpha q_{{\rm s},i}(\xi)\big),
\end{equation}
with fixed $\alpha=0.1$.

\para{Numerical evaluation.}
We approximate the densities by bilinear rasterization, normalized Gaussian convolution and bilinear evaluation at the observations. The raster stores projected counts without division by their total, and we add $\varepsilon_{\rm cam}=10^{-12}$ to the broad density in raster-count units. At each smoothing stage, the raster covers the extended rectangle that defines $\mathcal J_{\rm cam}$. We discard poses with fewer than eight projected points contributing to the raster. The coarse stage evaluates a fixed random sample of at most 65{,}536 model points, and the medium and fine stages evaluate one of at most 262{,}144.

\para{Local refinement.}
At each center, we keep up to eight of the 64 rotations of lowest coarse objective, at pairwise angles above $20\degree$. We refine the kept candidates of a level in order of increasing objective, at most 2{,}048, 4{,}096 and 8{,}192 of them at the three full grids and 2{,}048 at each further level, in batches of 256. Refinement uses the quasi-Newton method of Broyden, Fletcher, Goldfarb and Shanno (BFGS)~\cite{nocedal2006numerical}, jointly updating the rotation and camera center through three smoothing stages. The coarse, medium and fine stages use narrow widths $\tau_{\rm n}=0.04$, $0.015$ and $0.00375$ and broad widths $\tau_{\rm b}=0.16$, $0.08$ and $0.08$ on rasters of $128\times128$, $256\times256$ and $1{,}024\times1{,}024$, respectively. They run 15, 20 and 20 iterations, and the fine stage refines the 32 candidates of a batch with the lowest medium objective. These widths are in calibrated image coordinates, while the center grid uses model-coordinate units scaled by the scene extent.

\para{Three-point correction.}
For the four candidates of a batch with the lowest fine objective, we pair every observation with its nearest projected model point in the observed image region. We take the 24 closest pairs with distinct model points and solve the three-point problem~\cite{ke2017ap3p} for every triple of them. We rank the solutions by the medium objective and evaluate the best eight by the fine objective. The best of these replaces the candidate when it lowers the fine objective.

\para{Refinement by EM.}
The observed image region is the rectangle spanned by the 0.5th and 99.5th percentiles of the calibrated observations, enlarged by 5\% of each side length on both sides. Each candidate selects the model points at positive depth whose projections lie inside this region. At most 262{,}144 of these points, drawn at random, form the model for EM and the alignment check, and at least eight are required. EM in its form with outliers then refines the pose twice, from noise levels of $0.03$ and then $0.01$ in calibrated image coordinates, each time with the CPU refinement of \cref{tab:supp_search_settings}, before the alignment check of \cref{sec:supp_acceptance}.

\para{Acceptance and finer grids.}
After EM, we accept a pose when it passes the alignment check and at least $\max\{8,0.3\min(n,N)\}$ observations are matched, where $N$ counts all model points of the scene. A pose with this many matches is also accepted when the fraction of observations within $10^{-3}$ of a projected model point in the observed image region, in calibrated image coordinates, exceeds by at least $0.2$ its largest value at 32 random poses. These random poses have centers drawn uniformly from the enlarged box and rotations drawn from the rotation grid. The search stops at the first accepted pose and otherwise moves to the next candidate. If no pose is accepted after the three full grids, up to three further levels halve the cell size but subdivide only the 512 cells with the best scan scores. If all levels are used, the search returns the pose with the most matched 2D points and reports that it was not accepted.

\subsection{Coverage of the camera-center grid}
\label{sec:supp_proof_coverage}

Camera search supplies the initial poses that EM refines (\cref{sec:supp_camera_localization}). It places candidate camera centers at the centers of a grid of cells in a box around the 3D model. We show how fine the grid must be for some candidate to lie within a given distance of the true camera center, and how many candidates this requires.

\begin{lemma}[Coverage of the camera-center grid]
\label{lem:coverage}
Let $\Delta_{\mathrm c}>0$, and for $k=1,2,3$ let $a_k\in\R$ and $L_k>0$. We divide side~$k$ of the box $\prod_{k=1}^{3}[a_k,a_k+L_k]$ into $N_k=\lceil L_k/\Delta_{\mathrm c}\rceil$ intervals of equal length $L_k/N_k\le\Delta_{\mathrm c}$, and the products of these intervals form the cells. Then every point of the box is within $\sqrt3\,\Delta_{\mathrm c}/2$ of a cell center, and the number of centers is
\begin{equation}
N_1N_2N_3\le\prod_{k=1}^{3}\Bigl(\frac{L_k}{\Delta_{\mathrm c}}+1\Bigr).
\end{equation}
\end{lemma}

\begin{proof}
Since $L_k>0$ and $\Delta_{\mathrm c}>0$, every $N_k$ is positive. The $N_k$ intervals along side $k$ have length $L_k/N_k$ and midpoints $a_k+(i+\tfrac12)L_k/N_k$ for $0\le i<N_k$. A cell center consists of one midpoint per coordinate, so we can bound the distance to the nearest center one coordinate at a time.

\emph{Step 1. One coordinate.} Let $\vect c$ be a point of the box. Its coordinate $c_k$ lies in one of the intervals along side $k$, within half the interval length of the midpoint of that interval. Since $N_k\ge L_k/\Delta_{\mathrm c}$, the interval length is at most~$\Delta_{\mathrm c}$.

\emph{Step 2. Three coordinates.} Choosing such a midpoint in each coordinate gives a grid center $\vect c_{\mathrm{grid}}$ with $\norm{\vect c-\vect c_{\mathrm{grid}}}^2=\sum_k(c_k-c_{\mathrm{grid},k})^2\le3\Delta_{\mathrm c}^2/4$.

\emph{Step 3. Count.} The centers are the $N_1N_2N_3$ triples of midpoints, and $N_k<L_k/\Delta_{\mathrm c}+1$ by the definition of the ceiling.
\end{proof}

Let $r_{\rm c}>0$, and let the viewing distance $d_{\rm v}$ be the median depth of the model points observed in the image. Suppose that the true camera center $\vect c^\star$ lies in the box. Suppose also that the rotation grid contains a rotation $\Rot_0$ such that refinement converges to the true pose from the rotation $\Rot_0$ together with every camera center within distance $r_{\rm c}d_{\rm v}$ of $\vect c^\star$. By \cref{lem:coverage}, the choice $\Delta_{\mathrm c}=2r_{\rm c}d_{\rm v}/\sqrt3$ places a grid center within this distance, so refinement from this center and $\Rot_0$ converges to the true pose. Whether the search refines this candidate also depends on its ranking. \cref{sec:supp_whole_scene} uses this choice of $\Delta_{\mathrm c}$ to bound the cost of the search.

\section{Implementation details}
\label{sec:supp_config}
\subsection{Search, refinement \& acceptance}
\label{sec:supp_search}
\para{Hardware.} The implementation requires one Intel Xeon (Cascade Lake) central processing unit (CPU), 128~GB of random-access memory (RAM) and one RTX 4090 graphics processing unit (GPU).

This appendix details the four steps of \cref{sec:estimation} and then the complete search of \cref{alg:momentpnp}, with its settings and complexity. Steps 1 and 2 work in calibrated image coordinates, and Steps 3 and 4 measure residuals, distances and noise in pixels. The solver also takes as input whether outliers may be present. This choice selects one of two searches, each organized in stages. In a GPU stage, the GPU generates initial camera poses and fits them by moment matching. It then runs EM on a batch of candidates, and the CPU reruns EM on at most three of them. Without outliers, a CPU search follows the GPU stages if none of their fits is accepted. \cref{sec:supp_independent_edges} adapts the steps to edge observations.

\para{Camera initialization.}
When the observations cover an unknown part of a scene reconstruction, Step 1 searches over camera centers and orientations for the initial poses. Each candidate pose selects the model points that project into the observed image region, and these points form the model for Steps 3 and 4; further details in \cref{sec:supp_camera_localization}.

\para{Monomial features.}
\label{sec:supp_moment_vectors}
On outlier-free inputs, the monomial feature map is $\vect\phi_4\circ h$, where $h$ is the fixed image normalization of \cref{sec:prelim}. For a point $\vect z=(z_1,z_2)$, the monomial vector lists the fourteen monomials of degrees one to four,
\begin{equation}
\label{eq:phi4}
\begin{aligned}
\vect\phi_4(\vect z)=\big(&z_1,\,z_2,\,z_1^2,\,z_1z_2,\,z_2^2,\,z_1^3,\,z_1^2z_2,\,z_1z_2^2,\,z_2^3,\\
&z_1^4,\,z_1^3z_2,\,z_1^2z_2^2,\,z_1z_2^3,\,z_2^4\big),
\end{aligned}
\end{equation}
so its average over a point set contains the fourteen distinct entries of the moments $\mom_1$ to $\mom_4$ of \cref{eq:moment}. The averages over the observations and over the projections at a pose $\xi$ are
\[
\vect\mu_{\dets}=\frac1n\sum_{i=1}^n\vect\phi_4(h(\vect u_i)),\enspace
\vect\mu_{\pts}(\xi)=\frac1N\sum_{j=1}^N\vect\phi_4(h(\vect p_j(\xi))),
\]
and their squared distance is $\mathcal L_4$ of \cref{def:moments}. If all observations coincide, their root-mean-square radius is zero, and we set $r_{\rm cal}=1$ in $h$.

\para{Fourier features with outliers.}
\label{sec:supp_outlier_features}
With outliers, a few distant points can dominate the fourth-order moments, so moment matching uses bounded Fourier features. Let the bounding rectangle $D_{\rm F}\subset\R^2$ of the calibrated observations have positive area, midpoint $\vect b_F$ and larger half-width $q_F$. We evaluate the features at the box coordinates $\vect z=(\vect u_{\mathrm{cal}}-\vect b_F)/q_F$, in which $D_{\rm F}$ becomes a rectangle with half-widths $\vect h_F=(h_{F,1},h_{F,2})$, those of $D_{\rm F}$ divided by $q_F$. Each feature is multiplied by the spatial taper $W_F(\vect z)=W_{F,1}(z_1)W_{F,2}(z_2)$. With $\operatorname{Clip}(t)=\min\{1,\max\{0,t\}\}$, padding $\delta_F$ and taper width $\tau_F>0$, each factor is
\[
W_{F,r}(t)=
\operatorname{Clip}\!\left(\frac{t+h_{F,r}+\delta_F}{\tau_F}\right)\!
\operatorname{Clip}\!\left(\frac{h_{F,r}+\delta_F-t}{\tau_F}\right).
\]
Along each axis, $W_{F,r}$ vanishes outside the padded interval $[-h_{F,r}-\delta_F,\,h_{F,r}+\delta_F]$ and rises linearly over a distance $\tau_F$ from each of its ends. Projections far outside $D_{\rm F}$ therefore contribute nothing, and the linear rise keeps each feature continuous as a projection crosses the ends of the interval.

For a cutoff $K_F$, the frequency set $\mathrm F$ contains the vectors $\vect k\in\frac\pi2\mathbb Z^2$ with $\norm{\vect k}\le\frac\pi2K_F$ whose first coordinate is positive, or zero with a positive second coordinate. The set thus excludes $\vect k=0$, and no frequency appears together with its negative. Each frequency contributes the cosine and sine pair $a_{\vect k}W_F(\vect z)\big(\cos(\vect k^\top\vect z),\sin(\vect k^\top\vect z)\big)$. The amplitude $a_{\vect k}=\exp(-\zeta_F^2\norm{\vect k}^2/2)$ with spectral width $\zeta_F>0$ damps high frequencies. The concatenated pairs form a vector of norm at most $\featbound=(\sum_{\vect k}a_{\vect k}^2)^{1/2}$. Evaluating this vector at the box coordinates of an image point $\vect u$ and dividing it by $\featbound$ gives the feature map $\vect\phi_F(\vect u)$, with $\norm{\vect\phi_F}\le1$ as \cref{lem:colocated,supp:lem-contamination} require. The division scales $\mathcal L_F$ by the positive constant $\featbound^{-2}$, so it changes neither the minimizing poses at each frequency cutoff nor the estimate $\hat\unmatched$ below.

The Fourier feature averages of \cref{def:moments} are $\vect\mu_{\dets,F}=\frac1n\sum_i\vect\phi_F(\vect u_i)$ and $\vect\mu_{\pts,F}(\xi)=\frac1N\sum_j\vect\phi_F(\vect p_j(\xi))$. In the second average, a model point at nonpositive depth contributes a zero vector but still counts in the denominator $N$. The outlier model places uniform image outliers on $D_{\rm F}$ with density $1/|D_{\rm F}|$ and model outliers on the bounding box $\mathcal B\subset\R^3$ of the model points with density $1/\operatorname{vol}(\mathcal B)$. The box must have positive volume. We compute the image average $\vect\mu_{\dets,0}$ once by midpoint quadrature on $D_{\rm F}$. The image of a uniform distribution on $\mathcal B$ depends on the pose and is generally not uniform, so we recompute $\vect\mu_{\pts,0}(\xi)$ at each pose from the projections of a midpoint lattice of $\mathcal B$. Lattice points at nonpositive depth also contribute zero. \cref{supp:lem-contamination} gives the identity behind \cref{eq:moment-matching}. When the inlier observations equal their projections and both point sets contain the same fraction $\unmatched$ of uniform outliers, the expected feature averages at the true pose $\xi$ differ by $\unmatched(\vect\mu_{\dets,0}-\vect\mu_{\pts,0}(\xi))$. Noise and unequal outlier fractions add a further residual at the true pose. EM then refines the pose from the individual observations and re-estimates their outlier fraction.

\para{Camera initialization and moment matching.}
On the point-set benchmarks, Step 1 places initial cameras at the orientations of a rotation grid, with translations from the centroids and spreads as described below, and scores them by $\mathcal L_F$ at the lowest cutoff. Whole-scene queries use the camera search of \cref{sec:supp_camera_localization}. Step 2 proceeds as in \cref{sec:estimation}. Moment matching minimizes $\mathcal L_F$ of \cref{def:moments} with outliers and its special case $\mathcal L_4$ on complete inputs. To initialize the translation at a grid rotation, we match the projected centroid and root-mean-square radius to those of the observations, in closed form under scaled orthography~\cite{hartley2004multiple} and then by Gauss--Newton~\cite{nocedal2006numerical} under perspective, keeping every model point in front of the camera. Scaled orthography provides these starting values, and all objectives and refinements use the pinhole projection of \cref{sec:problem}. \cref{tab:supp_search_settings} gives the grid sizes, the numbers of retained poses and their minimum separations. With outliers, we set the outlier fraction at each pose to the least-squares value
\begin{equation}
\begin{aligned}
\hat\unmatched(\xi)&=\frac{\langle\vect d,\vect d_0\rangle}{\norm{\vect d_0}^2},\\
\vect d&=\vect\mu_{\dets,F}-\vect\mu_{\pts,F}(\xi),\\
\vect d_0&=\vect\mu_{\dets,0}-\vect\mu_{\pts,0}(\xi),
\end{aligned}
\end{equation}
clipped to $[0,0.75]$. We score each pose by the residual $\norm{\vect d-\hat\unmatched\vect d_0}^2$ that remains after subtracting the estimated outlier contribution. Uniform outliers shift the centroids and root-mean-square radii from which we solve for the translation at a grid rotation. Before solving, we therefore subtract the expected contribution of a fraction $\unmatched_0$ of uniform outliers on $D_{\rm F}$ from the moments of the observations. We make the same correction to the moments of the model points using the midpoint lattice of $\mathcal B$. Uniform outliers fill every cell of a $5\times5$ grid on $D_{\rm F}$, so the emptiest cells contain mainly outliers. We set $\unmatched_0$ to $25/n$ times the mean count of the four emptiest cells, clipped to $[0,0.75]$. If the fitted fraction at the best pose differs from $\unmatched_0$ by more than $0.05$, we repeat the search from that fraction. The fits then proceed through increasing frequency cutoffs $K_F$. If no fit is accepted, the search with outliers repeats this fitting from the further starting fractions of \cref{tab:supp_search_settings}. It finally obtains further camera poses from the rotation grid, optimizes their features and refines them by EM.

\para{Affine initial poses.}
\label{sec:supp_moment_initialization}
On outlier-free inputs, the CPU search begins from the affine initial poses of \cref{thm:identifiability}. We write $\dets_{\mathrm{cal}}=\{\vect u_{\mathrm{cal},i}\}$ for the calibrated observations, with mean $\bar{\vect u}_{\mathrm{cal}}$ and covariance $\Sigma_{\dets_{\mathrm{cal}}}$. The model points have mean $\bar{\vect y}$ and covariance $\Sigma_{\pts}$. When both covariances are positive definite, the normalized points are $\tilde{\vect u}_i=\Sigma_{\dets_{\mathrm{cal}}}^{-1/2}(\vect u_{\mathrm{cal},i}-\bar{\vect u}_{\mathrm{cal}})$ and $\tilde{\vect y}_j=\Sigma_{\pts}^{-1/2}(\vect y_j-\bar{\vect y})$. For complete noiseless affine projections, they are related by an orientation $O\in\Vset$ (\cref{supp:lem-normalization}). We identify a normalized point $\tilde{\vect u}=(\tilde u_1,\tilde u_2)$ with the complex number $w=\tilde u_1+\mathrm i\tilde u_2$. For $N_c>0$ complex numbers $w_1,\ldots,w_{N_c}$ and integers $a\ge b\ge0$, the complex moments are $c_{a,b}=N_c^{-1}\sum_{i=1}^{N_c}|w_i|^{2b}w_i^{a-b}$.

We fit nine real values, the real and imaginary parts of $c_{3,0}$ and $c_{2,1}$ and of $c_{4,0}$ and $c_{3,1}$, together with the real value $c_{2,2}$, the average of $|w|^4$. These values are invertible linear combinations of the distinct entries $(\vect m_3,\vect m_4)$ of $\mom_3$ and $\mom_4$, so matching them matches the cubic and quartic moments of \cref{thm:identifiability}. Starting from a grid of orientations, we fit the values of the projected normalized model $O\tilde\pts$ to those of the normalized observations by batched Levenberg--Marquardt refinement~\cite{nocedal2006numerical}. We rank the resulting orientations by their residuals. Undoing the normalization converts an orientation $O$ into the affine camera
\begin{equation}
A=\Sigma_{\dets_{\mathrm{cal}}}^{1/2}O\Sigma_{\pts}^{-1/2},\qquad
\vect b=\bar{\vect u}_{\mathrm{cal}}-A\bar{\vect y}.
\end{equation}
The thin singular-value decomposition of $A$ is $A=U_A\operatorname{diag}(\kappa_1,\kappa_2)V_A^\top$, where $U_A\in\R^{2\times2}$ and $V_A\in\R^{3\times2}$ have orthonormal columns and $\kappa_1\ge\kappa_2>0$. We set $O_0=U_AV_A^\top$ and $\bar\kappa=(\kappa_1+\kappa_2)/2$. The initial rotation has $O_0$ as its first two rows and their cross product as its third, and the initial translation is $\trans_0=(b_1,b_2,1)^\top/\bar\kappa$. From this initial pose, we fit the six pose parameters to $\mathcal L_4$, with the same normalization $h$ for the observations and the projections.

The fit weights the nine real values equally, whereas \cref{thm:identifiability} measures moment differences in the norm of the full symmetric tensors. For two normalized 2D point sets, the differences of their tensors and the differences $\Delta c_{a,b}$ of their complex moments are related by
\begin{align*}
\norm{\Delta\mom_3}^2
 &=\tfrac14|\Delta c_{3,0}|^2+\tfrac34|\Delta c_{2,1}|^2,\\
\norm{\Delta\mom_4}^2
 &=\tfrac18|\Delta c_{4,0}|^2+\tfrac12|\Delta c_{3,1}|^2
   +\tfrac38(\Delta c_{2,2})^2.
\end{align*}
The residual for the nine fitted values and the tensor residual therefore vanish at the same poses but weight nonzero differences differently. The residual $\fitres$, the error $\obserr$ and the modulus $\Gmod$ of \cref{thm:identifiability} use the tensor norm.

With four model points and positive definite covariances, the normalized model is a regular tetrahedron. Each of its 24 orthogonal symmetries $Q_{\rm s}$ maps it to itself, so the orientations $O$ and $OQ_{\rm s}$ give the same moments and the affine fit cannot distinguish them. For each symmetry, we therefore undo the normalization and fit a rigid pose under the pinhole projection model. The CPU search compares all of these poses within its limits, even if an earlier pose is accepted. For any number of model points, if none of the first affine initial poses is accepted, the CPU search continues on successively finer rotation grids. At each grid rotation, the translation is fitted to the centroid and spread of the observations as in Step 1, which keeps every model point in front of the camera. It then tries the remaining affine initial poses.

\para{Planar point models.}
\label{sec:supp_planar}
We treat a 3D point set as numerically planar when the smallest eigenvalue of its covariance is at most $10^{-8}$ times the largest. Normalization would then divide the coordinate along the plane normal by a near-zero standard deviation, so the affine fit uses plane coordinates. Let $Q_{\rm p}\in\R^{3\times2}$ have orthonormal columns along the two leading covariance directions, and let $\vect\nu$ be their unit normal. We fit $A_2\in\R^{2\times2}$ from centered plane coordinates to the calibrated observations. Let $\kappa_1$ be the largest singular value of $A_2$, and let $\vect a_{\rm p}\in\R^2$ satisfy $\vect a_{\rm p}\vect a_{\rm p}^\top=\kappa_1^2I_2-A_2A_2^\top$. We complete $A_2$ to the two matrices $A=A_2Q_{\rm p}^\top\pm\vect a_{\rm p}\vect\nu^\top$. Then $AA^\top=\kappa_1^2I_2$, so both completions are scaled-orthographic cameras. They project points that lie exactly in the plane identically, so we compare them by perspective fitting and refinement on the original 3D coordinates.

\para{EM likelihood.}
At a pose $\xi$, let $\forwardcount(\xi)$ be the number of model points in front of the camera with finite image projections, $\noisefit>0$ the current noise standard deviation per image coordinate in pixels, $p_d$ the probability that each of these model points is observed, and $n_{\rm cl}$ the expected number of outlier observations. Step 3 models each observation $\vect u$ as an outlier, uniform on a rectangle $D_{\rm im}$, with probability $\unmatched$, or as a projection of one of these model points, chosen uniformly, with isotropic Gaussian noise:
\begin{equation}
\label{eq:em-mixture}
p(\vect u)=\frac{\unmatched}{|D_{\rm im}|}+\frac{1-\unmatched}{\forwardcount}\sum_{j:\,\text{front}}\mathcal N\big(\vect u;\vect p_j(\xi),\noisefit^2I_2\big).
\end{equation}
The uniform component has support $D_{\rm im}$, the bounding rectangle of the observations with each side extended at both ends by its length divided by $\sqrt n$. A side of zero length is given the length $10^{-300}$ pixels, so the fixed area $|D_{\rm im}|$ in square pixels is positive. EM evaluates only the projections near each observation. On the GPU, the neighborhood $\mathcal J_i$ of observation $i$ contains the projections within $4.5\noisefit$. Beyond this distance, the Gaussian weight $k_{ij}$ defined below is less than $e^{-10}$. On the CPU, it contains the projections within $3\noisefit$ and also the nearest projection, even when that projection lies outside this radius. For each pair in a neighborhood, let $k_{ij}=\exp[-\norm{\vect u_i-\vect p_j(\xi)}^2/(2\noisefit^2)]$. We score poses by the log-likelihood of a Poisson point process, with each inner sum restricted to $\mathcal J_i$,
\begin{equation}
\label{eq:em-likelihood}
\begin{aligned}
\lik_{\mathrm{EM}}&=\sum_{i=1}^n\log\!\left(\frac{n_{\rm cl}}{|D_{\rm im}|}
 +\frac{p_d}{2\pi \noisefit^2}\sum_{j\in\mathcal J_i}k_{ij}\right)\\
&\quad-n_{\rm cl}-p_d\forwardcount(\xi).
\end{aligned}
\end{equation}
The expected number of observations under this process is $\Lambda_{\rm EM}=n_{\rm cl}+p_d\forwardcount(\xi)$. Dividing the argument of the logarithm by $\Lambda_{\rm EM}$ gives the mixture density of \cref{eq:em-mixture} with $\unmatched=n_{\rm cl}/(n_{\rm cl}+p_d\forwardcount)$, evaluated with the same neighborhoods $\mathcal J_i$. This division leaves the responsibilities unchanged. The Poisson score is therefore
\begin{equation}
\label{eq:pose-ranking}
\lik_{\mathrm{EM}}=\sum_{i=1}^n\log p(\vect u_i)+n\log\Lambda_{\rm EM}-\Lambda_{\rm EM}.
\end{equation}
The first term measures agreement with observation locations, and the remaining terms compare expected and observed counts. The term $n\log\Lambda_{\rm EM}-\Lambda_{\rm EM}$ is constant for fixed $\Lambda_{\rm EM}$, but with outliers $\Lambda_{\rm EM}$ can vary across poses.

\para{Step 3: EM updates.}
The expectation step assigns to each pair in a neighborhood the posterior probability that observation $i$ comes from projection $j$. This responsibility is
\[
r_{ij}=\frac{k_{ij}}{2\pi\noisefit^2n_{\rm cl}/(|D_{\rm im}|p_d)+\sum_{j'\in\mathcal J_i}k_{ij'}}.
\]
Pairs outside the neighborhoods have zero responsibility, and the remainder $1-\sum_jr_{ij}$ is the responsibility of the uniform component. The sum $n_{\rm m}=\sum_{ij}r_{ij}$ is the expected number of observations that come from projections. The maximization step updates the pose by a damped Gauss--Newton step on the weighted squared reprojection errors. With the cross-product matrix $[\cdot]_\times$, the projection Jacobians in the local pose coordinates $\posecoord=(\vect\omega,\trans)$ are $J_j=\partial\proj(\Rot\vect y_j+\trans)[-[\Rot\vect y_j]_\times\ I_3]$, and they give
\[
H_{\rm GN}=\sum_{ij}r_{ij}J_j^\top J_j,\qquad
\vect g_{\rm GN}=\sum_{ij}r_{ij}J_j^\top(\vect p_j-\vect u_i).
\]
With damping $\lambda_{\rm GN}>0$ and $D_{\rm GN}=\operatorname{diag}(\operatorname{diag}H_{\rm GN})$, a trial step $\Delta\posecoord=(\Delta\vect\omega^\top,\Delta\trans^\top)^\top$ solves the damped system $(H_{\rm GN}+\lambda_{\rm GN}D_{\rm GN})\Delta\posecoord=-\vect g_{\rm GN}$. It gives the rotation $\Rot'=\exp([\Delta\vect\omega]_\times)\Rot$ and, separately, the translation $\trans'=\trans+\Delta\trans$. On both the GPU and the CPU, we accept a trial step only if it lowers this weighted cost at fixed responsibilities. Then, at the updated pose $\xi'$ and for $n_{\rm m}>0$, the noise estimate and, with outliers, $p_d$ and $n_{\rm cl}$ are
\begin{align*}
 \noisefit_{\rm EM}^2&=\frac{\sum_{ij}r_{ij}\norm{\vect u_i-\vect p_j(\xi')}^2}{2n_{\rm m}},\\
 p_d'&=\min\{1,n_{\rm m}/\forwardcount(\xi')\},\\
 n_{\rm cl}'&=n-n_{\rm m}.
\end{align*}
Without outliers, $p_d=1$ and $n_{\rm cl}=10^{-3}n$ stay fixed, and this small expected outlier count keeps a single gross error from dominating the likelihood. EM starts from the noise level of \cref{tab:supp_search_settings} and anneals it. While $\noisefit_{\rm EM}<\noisefit$, each iteration sets $\noisefit'=\min\{\noisefit_{\rm EM},0.9\noisefit\}$. At the first iteration with $\noisefit_{\rm EM}\ge\noisefit$, annealing ends, and from then on $\noisefit'=\noisefit_{\rm EM}$, bounded on the GPU by the noise ceiling of \cref{tab:supp_search_settings}. When a pose is given without a noise level, such as an initial pose of the CPU search or an alternative of Step 4, EM starts from the root-mean-square distance of the observations to their nearest projections divided by $\sqrt2$, without annealing. EM has converged when annealing has ended, when $|\noisefit'-\noisefit|\le10^{-4}\noisefit$, and when no projection coordinate moves by more than $10^{-4}\noisefit$ between iterations.

\para{Valid fits \& ranking.}
\label{sec:supp_ranking}
Step 4 considers only valid fits. A valid fit has finite parameters and, without outliers, positive depths for all model points. With outliers, every model point with a positive responsibility must stay in front of the camera, and at least three model points must lie in front of the camera with finite image projections. Before CPU refinement, we rank a stage's valid fits, without outliers first by convergence and then by likelihood. With outliers, we rank them by likelihood alone and, if some of them have $n_{\rm m}\ge\max\{8,0.3\min(n,N)\}$, rank only those, because a fit that explains few observations provides little evidence about the pose. If no fit is accepted, the solver returns the valid fit of highest likelihood and reports that it was not accepted. It reports failure when no valid fit remains.

\para{Association level.}
\label{sec:supp_association_implementation}
For the association level of \cref{def:association}, we use $\nnb=\min\{8,\forwardcount-1\}$ neighbors and measure the distances $d_{i,k}$ in pixels. For observation $i$, the transformed distance ratio is $\assoc_i=-\nnb\log(1-d_{i,1}^2/d_{i,\nnb+1}^2)$. \cref{supp:lem-association} proves that each $\assoc_i$ has the unit exponential distribution under the Poisson model of \cref{def:association}. Without outliers, the association level is the lower tail
\[
\begin{aligned}
p_{\rm assoc}&=\Pr\Big[G_\Gamma\le\sum_{i=1}^n\assoc_i\Big],\\
G_\Gamma&\sim\operatorname{Gamma}(n-3,1).
\end{aligned}
\]
The reduced shape $n-3$ accounts for the six fitted pose parameters. With outliers, we count the observations with $\assoc_i\le\pi/16$. When the nearest projection is much closer than the $(\nnb+1)$th, the ratio $\assoc_i$ is close to $\pi d_{i,1}^2$ times the density of projections, so the condition places the nearest projection within about a quarter of the average spacing of the projections, the inverse square root of that density. Under the Poisson model, each observation meets it with probability $1-e^{-\pi/16}$, and the association level is the upper tail
\[
\begin{aligned}
p_{\rm assoc}&=\Pr\big[N_{\rm B}\ge|\{i:\assoc_i\le\pi/16\}|\big],\\
N_{\rm B}&\sim\operatorname{Binomial}(n,1-e^{-\pi/16}).
\end{aligned}
\]
These reference distributions treat the $\assoc_i$ as independent. Neighboring observations share projections, and the pose is fitted to the same observations, so neither distribution is exact (\cref{sec:supp_association}). We therefore use $p_{\rm assoc}$ as a score and compare it with fixed thresholds far in the tail. A pose that EM has not yet refined has no parameters fitted to the distances. The CPU search and the comparison of alternative poses therefore test such a pose by $p^0_{\rm assoc}$, which uses the Gamma shape $n$ without outliers and equals $p_{\rm assoc}$ with outliers.

\para{Step 4: alignment check.}
\label{sec:supp_acceptance}
The empirical acceptance criteria of \cref{alg:momentpnp} test whether the observations have projections of their own. When the noise is small compared with the spacing of the projections, an inlier observation at the true pose is typically much closer to its generating projection than to any other. The ratio $d_{i,1}/d_{i,2}$ of the distances from observation $i$ to its two nearest projections is then typically small, as is the association level. On outlier-free inputs, a further test asks whether the moment residual $\mathcal L_4$ is no larger than noise of standard deviation $\noisefit$ would make it at the true pose. Let $\vect z_j=h(\vect p_j)$ be the normalized projections, $\partial\vect\phi_4$ the $14\times2$ Jacobian of $\vect\phi_4$, and $f_K=\sqrt{K_{11}K_{22}}$ the geometric-mean focal length. With the radius $r_{\rm cal}$ of \cref{sec:prelim}, we approximate the first-order expected value of $\mathcal L_4$ at the true pose by
\begin{equation}
    V_4=\frac{\noisefit^2}{f_K^2r_{\rm cal}^2N^2}\sum_{j=1}^N\norm{\partial\vect\phi_4(\vect z_j)}^2.
\end{equation}
When the focal lengths differ, this approximation uses one noise scale in place of the two calibrated-coordinate noise scales.

Without outliers, we accept a fit in either of two ways. The first requires convergence, at least 90\% as many distinct nearest projections as distinct observation locations, and a median ratio $d_{i,1}/d_{i,2}$ over the distinct observation locations of at most $0.25$. Repeated observations at one location share their nearest projection, so both quantities count each location once. With fewer than thirty distinct observation locations, the first way requires all nearest projections to be distinct and replaces the median ratio by its 75th percentile with the same threshold. The second way accepts a pose whose correspondences are not resolved because the noise is comparable to the spacing of the projections, provided the observations lie far closer to the projections than chance would place them. It applies to converged fits with $n\ge48$. It requires $p_{\rm assoc}\le10^{-14}$, relaxed to $10^{-12}$ for $n\ge200$, together with $\mathcal L_4\le9V_4$ unless $p_{\rm assoc}\le10^{-30}$.

With outliers, we accept a converged fit with $n_{\rm m}\ge\max\{8,0.3\min(n,N)\}$ and $p_{\rm assoc}\le10^{-9}$ if its median ratio $d_{i,1}/d_{i,2}$ over the matched observations is at most $0.12$. Matched observations are those to which the correspondences of \cref{alg:momentpnp} assign a model point. For $n\ge500$, where dense projections can prevent resolved correspondences, an association level of at most $10^{-100}$ allows a median ratio of at most $0.25$. With or without outliers, when the standard deviation of the image noise is given as an optional input, acceptance also requires $\noisefit$ to be at most three times that value.

\para{Step 4: pose ambiguity resolution.}
\label{sec:supp_alternative_poses}
Step 4 compares the selected pose, the accepted fit of highest likelihood, with the alternative poses described in \cref{par:symmetric} and characterized by \cref{supp:lem-companions}. We build the alternative poses from selected model points, which are all model points without outliers and, with outliers, those associated with matched observations. Each alternative keeps the centroid of the selected points fixed in camera coordinates.

The seven half-turn alternatives of \cref{supp:lem-companions} are the poses $\Rot\halfturn_k$ for $k=1,2,3$ and $\opthalf\Rot\halfturn_k$ for $k=0,\dots,3$, with $\halfturn_k$ taken about the principal axes of the selected points. We form them when at least three points are selected and their principal axes can be computed. Each must have valid projections and $p^0_{\rm assoc}\le10^{-3}$. Without outliers, we omit alternatives whose rotations differ by less than $3\degree$ from the accepted pose, fit the others to the image moments, and keep those whose residual is at most the larger of the accepted pose's residual and $9V_4$, with $V_4$ evaluated at the alternative.

The planar alternative uses the plane through the centroid of the selected points whose normal is a covariance eigenvector of smallest eigenvalue. It reflects the selected points through this plane, and it reflects the camera coordinates through the plane that contains the centroid and is parallel to the image plane. The two reflections compose to the rotation $\opthalf\Rot\halfturn_k$, with $\halfturn_k$ the half-turn about that eigenvector, so this alternative starts from the half-turn alternative that \cref{supp:lem-companions}(c) identifies for planar models. It differs from that alternative in its refinement. For a planar model, the accepted pose and its alternative give nearly the same image with the same correspondences, so we first fit this alternative under perspective projection to the correspondences of the accepted pose. We then refine it by EM only if its projections are valid, its rotation differs by more than $3\degree$ from the accepted pose and $p^0_{\rm assoc}\le10^{-3}$.

After EM, we discard alternatives of either kind whose rotation differs by less than $3\degree$ from any accepted pose. A remaining alternative replaces the selected pose only if it satisfies the acceptance criteria and has higher likelihood. The selected pose then gives the correspondences as in \cref{alg:momentpnp}.

\para{Correspondences.}
At the returned pose, each matched observation corresponds to its nearest projection, and every other observation is an outlier. When several observations share a nearest projection, only the closest keeps it. With outliers, an accepted pose with at least six matches is refitted to these correspondences by least squares, and the refit replaces it if the matched points stay in front of the camera.

\para{Search stages.}
\cref{alg:momentpnp} expands \cref{alg:main} with the search stages of the point-set benchmarks, the division of work between the GPU and the CPU, the early exits and the acceptance criteria, and \cref{tab:supp_search_settings} gives the settings of each stage. Inputs must have finite coordinates and meet the requirements of \cref{sec:problem} and the point-count limits in the table. The table states noise levels in units of $\rmsrad$, the root-mean-square distance of the observations from their centroid, in pixels. The search stops after the first stage that yields an accepted fit. A fit that is not accepted but has an association level at most $10^{-100}$ ends the GPU stages early, because its pose already places the observations far closer to the projections than chance would. Without outliers, the CPU search then follows. It defers initial poses with $p^0_{\rm assoc}>10^{-3}$ and, if none of its other initial poses reaches EM, refines up to three deferred poses from a larger noise level.

\begin{algorithm*}[tp]
\fontsize{8}{9.4}\selectfont
\caption{\method, expanding \cref{alg:main}; GPU lines handle all poses at once, CPU lines one at a time}
\label{alg:momentpnp}
\KwIn{Observations $\dets=\{\vect u_i\}_{i=1}^n$ in pixels, model $\pts=\{\vect y_j\}_{j=1}^N$, intrinsics $\Kmat$, whether outliers may be present, optional noise level $\noise$}
\KwOut{Pose $\hat\xi$, noise estimate $\noisefit$, correspondences $\hat\sigma$ and the flag $a$, true if $\hat\xi$ is accepted; no pose if no valid fit exists}
\tcp{$\mathtt{ObservedMoments}$ (CPU); on complete inputs, the moments $\vect\mu_{\dets}=\frac1n\sum_i\vect\phi_4(h(\vect u_i))$ replace both averages}
$\vect\mu_{\dets,F}\gets\frac1n\sum_i\vect\phi_F(h_F(\vect u_i))$,\quad
$\vect\mu_{\dets,0}\gets$ average of $\vect\phi_F$ over the rectangle $D_{\rm F}$ that bounds the calibrated observations\;
\tcp{$\mathtt{Accepted}(\xi)$ (CPU), with $d_{i,1}\le d_{i,2}\le\cdots$ the distances from $\vect u_i$ to the projections, $\assoc_i=-\nnb\log(1-d_{i,1}^2/d_{i,\nnb+1}^2)$, $n_{\rm m}$ from Step 3 and $V_4$ the expected $\mathcal L_4$ under noise $\noisefit$ at the true pose (\cref{sec:supp_acceptance}); a fit is valid when its parameters are finite and its matched points have positive depth}
Without outliers, $p_{\rm assoc}\gets\Pr[\mathrm{Gamma}(n-3,1)\le\sum_i\assoc_i]$, and a valid converged fit is accepted if 90\% of the observations have distinct nearest projections and $\operatorname{median}_i\,d_{i,1}/d_{i,2}\le0.25$, or if $n\ge48$, $p_{\rm assoc}\le10^{-14}$ ($10^{-12}$ for $n\ge200$) and $\mathcal L_4\le9V_4$ or $p_{\rm assoc}\le10^{-30}$\;
With outliers, $p_{\rm assoc}\gets\Pr[\mathrm{Binomial}(n,1-e^{-\pi/16})\ge|\{i:\assoc_i\le\pi/16\}|]$, and a valid converged fit is accepted if $n_{\rm m}\ge\max\{8,0.3\min(n,N)\}$, $p_{\rm assoc}\le10^{-9}$ and $\operatorname{median}_{i\,\rm matched}\,d_{i,1}/d_{i,2}\le0.12$ ($0.25$ if $n\ge500$ and $p_{\rm assoc}\le10^{-100}$)\;
Any other fit is not accepted, in particular an invalid or unconverged fit, or one with $\noisefit>3\noise$ when $\noise$ is given\;
\lIf{the inputs are outlier-free}{$S\gets$ GPU stages on the $800\times48$ and then the $3{,}200\times96$ rotation grid with $\unmatched_0=0$, then the CPU search}
\lElse{$S\gets$ GPU stages from the estimated outlier fraction $\unmatched_0$, from $\unmatched_0=0,0.15,0.30,0.45,0.60$, and on the $800\times48$ grid with outlier-corrected and cross-correlated translations}
$\Pi\gets\varnothing$,\quad $a\gets\mathrm{false}$\tcp*{the stages of $S$ are the levels of $\mathtt{WidenCameraSearch}$ in \cref{alg:main}}
\ForEach{stage in $S$, while $a$ is false}{
  \lIf{this is a later GPU stage and $p_{\rm assoc}(\hat\xi)\le10^{-100}$}{skip it: $\hat\xi$ already places the observations far closer to the projections than chance}
  \eIf{this is a GPU stage}{
    \tcp{Step 1, $\mathtt{CameraInitialization}$ (GPU)}
    $\Pi'\gets$ for each grid rotation $\Rot$, the pose whose translation matches the projected centroid and spread of the model to those of the observations, both corrected for a fraction $\unmatched_0$ of uniform outliers; keep those of lowest $\mathcal L_F(\xi,\hat\unmatched(\xi))$ at $K_F=4$ with distinct orientations\;
    \tcp{Step 2, $\mathtt{MomentMatching}$ (GPU); on complete inputs, $\unmatched=0$ and $\mathcal L_4(\xi)=\norm{\vect\mu_{\dets}-\vect\mu_{\pts}(\xi)}^2$ replaces $\mathcal L_F$}
    \For{$K_F=4,8,\dots,20$ and each $\xi\in\Pi'$}{
      $\vect d\gets\vect\mu_{\dets,F}-\vect\mu_{\pts,F}(\xi)$,\quad $\vect d_0\gets\vect\mu_{\dets,0}-\vect\mu_{\pts,0}(\xi)$,\quad $\hat\unmatched\gets\min\bigl\{0.75,\max\{0,\langle\vect d,\vect d_0\rangle/\norm{\vect d_0}^2\}\bigr\}$\;
      $\xi\gets$ Levenberg--Marquardt on $\mathcal L_F(\xi,\hat\unmatched)=\norm{\vect d-\hat\unmatched\,\vect d_0}^2$; after each cutoff, $\Pi'\gets$ the fits of lowest $\mathcal L_F$ with distinct orientations\;
    }
    \tcp{Step 3, $\mathtt{EMRefinement}$ (GPU, all of $\Pi'$ in one batch), under the mixture $p(\vect u)=\unmatched/|D_{\rm im}|+(1-\unmatched)\forwardcount^{-1}\sum_{j\,\text{in front}}\mathcal N(\vect u;\vect p_j(\xi),\noisefit^2I_2)$}
    \Repeat{the most likely fit of the batch has converged or 120 iterations have run, which also stops the other fits}{
      $r_{ij}\gets$ posterior probability that $\vect u_i$ comes from $\vect p_j(\xi)$,\quad $n_{\rm m}\gets\sum_{ij}r_{ij}$, for each $\xi\in\Pi'$\tcp*{E-step}
      $\xi\gets$ Gauss--Newton step on $\sum_{ij}r_{ij}\norm{\vect p_j(\xi)-\vect u_i}^2$,\quad $\noisefit^2\gets\sum_{ij}r_{ij}\norm{\vect u_i-\vect p_j(\xi)}^2/(2n_{\rm m})$, update $\unmatched$\tcp*{M-step}
    }
    Drop invalid fits and fits whose $\noisefit$ reached $0.999$ times its ceiling (they have not found the observations); for the rest, $\lik_{\rm EM}(\xi)\gets\sum_i\log p(\vect u_i)+n\log\Lambda-\Lambda$, with $\Lambda$ the expected number of observations\;
    Rank the fits by $\lik_{\rm EM}$, converged fits first without outliers, and among those with $n_{\rm m}\ge\max\{8,0.3\min(n,N)\}$, if any, with outliers\;
    \tcp{Step 3 continued, and Step 4, $\mathtt{AlignmentCheck}$ (CPU)}
    \ForEach{$\xi$ among the three highest-ranked fits}{
      $\xi\gets$ EM from $\xi$ on the CPU until convergence or the iteration limit,\quad $\Pi\gets\Pi\cup\{\xi\}$;\quad \lIf{$\mathtt{Accepted}(\xi)$}{leave this loop}
    }
  }{
    \tcp{CPU search, on outlier-free inputs only; its fits replace the earlier ones}
    $\Pi\gets$ CPU EM refinements (Step 3) of the affine initial poses of \cref{thm:identifiability}, then of three rotation grids with translations as in Step 1, then of the other affine poses, deferring those with $p^0_{\rm assoc}>10^{-3}$, until one is accepted (all 24 symmetric poses of four nonplanar points)\;
    \lIf{no pose reached EM}{refine up to three deferred poses, starting EM from a larger noise level}
  }
  $\hat\xi\gets$ the most likely accepted pose in $\Pi$, or the most likely pose in $\Pi$ if none is accepted,\quad $a\gets\mathtt{Accepted}(\hat\xi)$\;
}
\lIf{$\Pi=\varnothing$}{\Return no pose, because no valid fit was found}
\If{$a$}{
  \tcp{Step 4, $\mathtt{Alternatives}$ (CPU); $\halfturn_k$ and $\opthalf$ are the model and camera half-turns of \cref{par:symmetric}}
  $\Pi'\gets$ EM refinements (Step 3, CPU) of the poses with rotations $\Rot\halfturn_1,\Rot\halfturn_2,\Rot\halfturn_3,\opthalf\Rot,\opthalf\Rot\halfturn_1,\opthalf\Rot\halfturn_2,\opthalf\Rot\halfturn_3$ and the same centroid, for the rotation $\Rot$ of $\hat\xi$, when they have valid projections, $p^0_{\rm assoc}\le10^{-3}$ and rotations more than $3\degree$ from $\hat\xi$\;
  $\hat\xi\gets$ the most likely accepted pose in $\{\hat\xi\}\cup\Pi'$\;
}
\tcp{$\mathtt{Correspondences}$ (CPU)}
For $i=1,\dots,n$, $\hat\sigma(i)\gets\arg\min_j\norm{\vect u_i-\vect p_j(\hat\xi)}$ if that projection explains $\vect u_i$ better than the outlier term, else $\hat\sigma(i)\gets\varnothing$; keep only the closest of several observations sharing a model point\;
\lIf{there are outliers, $a$ holds and at least six matches remain}{keep the least-squares refit of $\hat\xi$ to $\hat\sigma$ if its matched points stay in front of the camera}
\Return $(\hat\xi,\noisefit,\hat\sigma,a)$\tcp*{if $a$ is false, $\hat\xi$ is the most likely fit and is reported as not accepted}
\end{algorithm*}

\begin{table*}[tp]
\centering\footnotesize
\setlength{\tabcolsep}{4pt}
\caption{\textbf{Search settings on the point-set benchmarks (ours).} Each rotation-grid size gives the number of viewing directions times the number of roll angles. Moment matching and EM use every input point. Scoring the rotation grid uses at most 4{,}096 model points, and cross-correlation at most 1{,}024. The CPU search has no time limit.}
\label{tab:supp_search_settings}
\begin{tabular*}{\textwidth}{@{\extracolsep{\fill}}lp{0.77\textwidth}@{}}
\toprule
Stage & Settings \\
\midrule
Inputs & Outlier-free: $n=N\ge4$. With outliers: $\min(n,N)\ge6$, and EM needs at least three model points in front of the camera with finite image projections. \\
Outlier-free search & We use $800\times48$ rotations, refined to $3{,}200\times96$ in the second stage, and keep at most $1{,}024$ or $2{,}048$ initial poses separated by $6\degree$. Moment matching uses 30 iterations, keeping at most 96 or 192 fits separated by $1\degree$. \\
Fourier moment matching & $200\times24$ rotations scored with $K_F=4$ and $\zeta_F=0.15$, keeping 48 initial poses separated by $10\degree$. Five fitting stages use $(K_F,\zeta_F)=(4,0.15)$, $(8,0.08)$, $(12,0.05)$, $(16,0.04)$ and $(20,0.03)$, and fit the best 48, 20, 20, 8 and 8 poses for 20, 12, 12, 10 and 10 iterations. Padding $\delta_F=0.02$ and taper width $\tau_F=0.06$. The image average $\vect\mu_{\dets,0}$ uses $48\times48$ midpoints of $D_{\rm F}$, and $\vect\mu_{\pts,0}$ uses an $8\times8\times8$ midpoint lattice of $\mathcal B$. The second stage starts from the outlier fractions 0, 0.15, 0.30, 0.45 and 0.60. \\
Other poses with outliers & $800\times48$ rotations. Translations come from centroids and spreads corrected for observed outlier fractions 0, 0.05, 0.1, 0.15, 0.2, 0.3, 0.4 and 0.5 and model outlier fractions 0, 0.2 and 0.4, keeping the best score per rotation, or from cross-correlation at seven scales from 0.5 to 2 times the ratio of the spreads. Each translation estimate keeps 2{,}048 initial poses separated by $6\degree$, and after 16 GPU EM iterations the 256 of highest likelihood continue. \\
GPU EM & At most 120 iterations. Without outliers, initial noise $0.02\rmsrad$ and ceiling $0.05\rmsrad$. With outliers, initial noise and ceiling $0.1\rmsrad$. The batch stops when its fit of highest likelihood has converged, counting with outliers only fits with $n_{\rm m}\ge\max\{8,0.3\min(n,N)\}$, and the other fits stop early. Fits whose estimated noise reaches $0.999$ times the ceiling have not found the observations and do not go to CPU refinement. \\
CPU refinement & We refine at most three candidates per stage, selected by the ranking rules of Step 3. Without outliers, fits stopped early receive at most 60 CPU EM iterations, and converged fits or fits at the iteration limit go to the acceptance criteria without further EM. With outliers, we first run 30 CPU EM iterations. If the fit has not converged and its association level is at most $10^{-3}$, we run at most 60 further iterations. \\
CPU affine search & Planar point sets (covariance eigenvalue ratio at most $10^{-8}$) use both completions of each distinct planar affine fit from the closed-form initialization and 720 initial rolls. Other point sets use $3{,}000$ viewing directions and 72 rolls, and the nine real values of the complex moments. We initially try six fitted orientations and increase this to at most 48 if no estimate satisfies the acceptance criteria. \\
CPU rigid search & $200\times24$, $800\times48$ and $3{,}200\times96$ rotations, keeping 24, 64 and 128 initial poses separated by $10\degree$. A finer level requires its rotation count times $N\le4\times10^7$. Other initial poses on the CPU use at most 600 EM iterations, and the final attempts initialized with a larger $\noisefit$ use 150 with initial noise $0.05\rmsrad$. \\
\bottomrule
\end{tabular*}
\end{table*}

\para{Complexity.}
For fixed numbers of features, candidates and iterations, computing the features costs $O(n+N)$. A GPU EM iteration then costs $O(n+N+N_{\rm pair})$, where $N_{\rm pair}$ counts the observation--projection pairs that the GPU examines in a grid of square image cells and can reach $O(nN)$. Sorting the observations into these cells adds $O(n\log n)$, and GPU storage is $O(n+N)$ for fixed batch size and grid limits. CPU refinement takes $O(N\log N+nN)$ time per iteration and can store $O(nN)$ neighbor pairs. With fixed search and iteration limits, the solver therefore has worst-case time $O(nN+(n+N)\log(n+N))$ and space $O(nN)$.

\begin{figure}[tp]
\centering
\includegraphics[width=217.07426pt,height=133.28622pt]{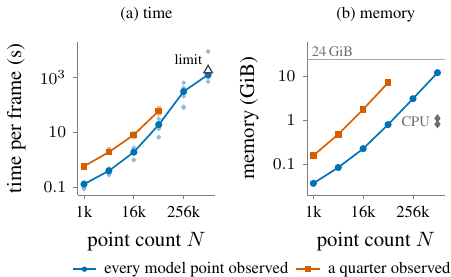}
\caption{\textbf{Measured time and memory on rendered ETH3D scans.} (a)~Time per frame. Light markers are single views and lines are medians. The triangle marks the Courtyard view that returned no pose within its 1{,}800-second limit. (b)~GPU allocator memory, which is equal across the views of each size, and the peak CPU memory of the four one-million-point runs that recorded it (gray). The horizontal line is the memory of the specified GPU.}
\label{fig:resources}
\end{figure}

\para{Measured time and memory.}
The complexity bounds above are worst cases. In \cref{fig:resources}, we measure time and memory on the rendered ETH3D scans of \cref{sec:supp_scaling}, from 1{,}000 to one million model points, with every model point or a quarter of them observed. GPU memory grows linearly with the number of model points. From 64{,}000 points on, the GPU allocator memory grows by about 12.4~KiB per added point when every model point is observed and reaches 11.87~GiB at one million points. With a quarter observed, it holds about 115~KiB per model point, or 7.0~GiB at 64{,}000 points. CPU memory stays small, and the peak resident memory of the four one-million-point runs that recorded it is 0.81 to 1.11~GiB. Time grows faster than memory. Up to 16{,}000 points, the median time per frame grows about linearly with the number of points. From 64{,}000 to 256{,}000 points, it grows about as the square of that number. This growth is consistent with the term $N_{\rm pair}$, because denser projections in a fixed image place more of them in each grid cell around an observation. At one million points, the seven completed views take 689 to 8{,}735~s, with a median of 1{,}244~s.

\para{Where the solver stops.}
No run of \method\ exhausted memory. Up to one million points, the only failure is one Courtyard view that returned no pose within its 1{,}800-second limit (\cref{sec:supp_scaling}). The input size is therefore limited by time. Linear extrapolation of the measured memory places the limit of a 24~GiB GPU near two million points with every model point observed and near 200{,}000 model points with a quarter observed. The learned baselines, by contrast, run out of memory from 16{,}000 or 64{,}000 points (\cref{tab:render_baselines}).

\subsection{Edge observations}
\label{sec:supp_independent_edges}

In the study of \cref{sec:supp_observations}, we reduce a red--green--blue (RGB) image and a computer-aided design (CAD) mesh to straight edges. The image edges are line segments found by a detector. In the mesh, an edge is a crease when the normals of its two adjacent faces differ by at least $30\degree$ or when it has a single adjacent face. An edge is a contour at a candidate pose when one adjacent face is turned toward the camera and the other away from it. At a candidate pose, the model edges are the visible creases and contours. We extract the image edges and the model edges independently. The inputs thus contain no pairing between image and model edges, and the two edge sets may sample the same visible boundary at different densities.

The method uses these edges in two forms. For moment matching, it averages the features over image arclength, the length along the edges in the image, rather than over points. Perspective projection shortens the parts of a 3D edge that are farther from the camera, so a uniform average along the 3D edge would not match an average along the detected image edge. \cref{lem:edge-agreement} shows that the two image-arclength feature averages agree at the true pose when the detected edges coincide with the visible projected edges. For Step 3, EM needs point observations. We place the 2D observations at equal image-arclength spacing along the detected image edges. At each initial pose, we place points at equal image-arclength spacing along the visible projected model edges and map each one back to its 3D point on the edge by inverting the change of parameter in the proof of \cref{lem:edge-agreement}. These 3D points form the model points and stay fixed during refinement. Because the two sets of samples are placed independently, an observation and a model point need not correspond to the same physical point, and EM treats them as two unordered point sets.

\begin{lemma}[Feature agreement along projected edges]
\label{lem:edge-agreement}
For a pose $\xi=(\Rot,\trans)$, let $f_\xi(\vect y)=\proj(\Rot\vect y+\trans)$ be the projection of a 3D point $\vect y$. At a candidate pose $\xi$, let $\{(\vect y^0_j,\vect y^1_j)\}_{j=1}^{N_E}$ be a finite family of visible straight 3D edge pieces whose endpoints lie in front of the camera. Their image endpoints are $\vect p^0_j=f_\xi(\vect y^0_j)$ and $\vect p^1_j=f_\xi(\vect y^1_j)$, and their projected lengths are $\edgelen_j=\norm{\vect p^1_j-\vect p^0_j}$. For a fixed measurable image feature map $\vect\phi:\R^2\to\R^{d_\phi}$ that is bounded on both edge sets, and for a positive total projected length, we define
\begin{equation}
\label{eq:edge-mean}
 \vect\mu_E(\xi)=\frac{\sum_j \edgelen_j\int_0^1\vect\phi((1-r_E)\vect p^0_j+r_E\vect p^1_j)\,dr_E}{\sum_j \edgelen_j}.
\end{equation}
Let $\mathcal C_E$ be the union of the independently detected image edges, with positive finite arclength. With the arclength element $d\edgelen$ in the image, let $\vect\mu_C=\int_{\mathcal C_E}\vect\phi\,d\edgelen\big/\int_{\mathcal C_E}d\edgelen$ be the average of $\vect\phi$ over $\mathcal C_E$. If, at the true pose $\xi^\star$, the union of the visible projected edge pieces equals $\mathcal C_E$ up to a set of zero arclength, and distinct projected pieces overlap only on sets of zero arclength, then $\vect\mu_E(\xi^\star)=\vect\mu_C$. The equality holds for the fourteen raw monomials through degree four and for bounded Fourier features, without point assignments or shared edge identities.
\end{lemma}
\begin{proof}
We show that each term of the numerator in \cref{eq:edge-mean} is the arclength integral of $\vect\phi$ over one projected piece. For one 3D segment $\vect y(r_E)=(1-r_E)\vect y^0_j+r_E\vect y^1_j$, we write $x^0_{j3}=(\Rot\vect y^0_j+\trans)_3>0$ and $x^1_{j3}=(\Rot\vect y^1_j+\trans)_3>0$ for the depths of its endpoints in camera coordinates. Its image is
\[
 \begin{aligned}
 f_\xi(\vect y(r_E))&=(1-\alpha_E(r_E))\vect p^0_j+\alpha_E(r_E)\vect p^1_j,\\
 \alpha_E(r_E)&=\frac{r_Ex^1_{j3}}{(1-r_E)x^0_{j3}+r_Ex^1_{j3}},
 \end{aligned}
\]
whose derivative is $\alpha_E'(r_E)=x^0_{j3}x^1_{j3}/((1-r_E)x^0_{j3}+r_Ex^1_{j3})^2>0$. The map $\alpha_E$ is therefore an increasing bijection of $[0,1]$, so the projected piece is the image segment from $\vect p^0_j$ to $\vect p^1_j$, traversed once. Along this segment, the arclength element is $\edgelen_j\,d\alpha_E$, and the arclength integral of $\vect\phi$ over the piece is $\edgelen_j\int_0^1\vect\phi((1-\alpha_E)\vect p^0_j+\alpha_E\vect p^1_j)\,d\alpha_E$, the $j$th term of the numerator. A piece of zero projected length contributes zero. Because distinct projected pieces overlap only on sets of zero arclength, the sum of these integrals counts each point of the projected union once, except on a set of zero arclength. At the true pose, the projected union is $\mathcal C_E$, so the numerator equals $\int_{\mathcal C_E}\vect\phi\,d\edgelen$ and the denominator equals the length of $\mathcal C_E$, which gives the equality.
\end{proof}

Detected edges in general differ from the projected ones. \cref{supp:lem-edge-stability} bounds the effect of three departures. These are displacement by the detector, edges present in only one of the two sets, and quadrature error. The lemma then turns the resulting feature discrepancy into a bound on pose error near the true pose. A coupling makes the comparison precise without pairing edges. It is a joint distribution of a point on the true projected edges and a point on the observed image edges, each distributed uniformly with respect to image arclength.

\begin{lemma}[Edge mismatch \& local pose stability]
\label{supp:lem-edge-stability}
We couple the arclength distributions on the true projected edges and the observed image edges, using a joint distribution with these two marginals. We assume that the coupling places at least a fraction $0\le\edgemass\le1$ of its mass on pairs at image distance at most $\edgedisp\ge0$. We let the fixed measurable feature map have norm bound $\featbound\ge0$ and Lipschitz constant $\featlip\ge0$ on a bounded image region containing the observed edges and the projected edges at all poses in a fixed neighborhood of $\xi^\star$. We let $\hat{\vect\mu}_E$ and $\hat{\vect\mu}_C$ be quadrature approximations of $\vect\mu_E$ and $\vect\mu_C$, and we let $\quaderr$ bound their combined image and model quadrature error throughout that neighborhood. The empirical feature discrepancy at the true pose then satisfies
\begin{equation}
\label{eq:edge-discrepancy}
\norm{\hat{\vect\mu}_E(\xi^\star)-\hat{\vect\mu}_C}\le\epsilon_E=\edgemass \featlip\edgedisp+2\featbound(1-\edgemass)+\quaderr.
\end{equation}
For a pose chart $\xi(\posecoord)$ in fixed six-dimensional local coordinates, we define $\vect g_E(\posecoord)=\vect\mu_E(\xi(\posecoord))$. We assume that $\vect g_E$ is continuously differentiable and that its Jacobian $J^\star\in\R^{d_\phi\times6}$ at the true pose coordinates $\posecoord^\star$ has full column rank, with least singular value $\jacsv>0$. An estimate $\hat{\posecoord}$ sufficiently close to the true pose, with fitted residual $\hat\epsilon_E=\norm{\hat{\vect\mu}_E(\xi(\hat{\posecoord}))-\hat{\vect\mu}_C}$, satisfies
\begin{equation}
\label{eq:edge-pose-bound}
\norm{\hat{\posecoord}-\posecoord^\star}\le\frac{2(\hat\epsilon_E+\epsilon_E)}{\jacsv}.
\end{equation}
\end{lemma}
\begin{proof}
We first bound the difference of the exact averages through the coupling. On at least a fraction $\edgemass$ of the coupled mass, the two points lie within $\edgedisp$ of each other, so their features differ by at most $\min(\featlip\edgedisp,2\featbound)$. On the remaining mass, the features differ by at most $2\featbound$. Since $\min(\featlip\edgedisp,2\featbound)\le2\featbound$, a larger $\edgemass$ can only lower this bound. Integration over the coupling therefore bounds the difference of the exact averages by $\epsilon_{\mathrm{law}}=\edgemass \featlip\edgedisp+2\featbound(1-\edgemass)$. Adding the quadrature errors by the triangle inequality gives \cref{eq:edge-discrepancy}. At the estimate, the triangle inequality through $\hat{\vect\mu}_E(\xi(\hat{\posecoord}))$, then $\hat{\vect\mu}_C$ and then $\vect\mu_C$ gives $\norm{\vect g_E(\hat{\posecoord})-\vect g_E(\posecoord^\star)}\le\hat\epsilon_E+\quaderr+\epsilon_{\mathrm{law}}=\hat\epsilon_E+\epsilon_E$.

We next show that $\vect g_E$ separates poses near $\posecoord^\star$. We write $J(\posecoord)$ for the Jacobian of $\vect g_E$, so $J^\star=J(\posecoord^\star)$. By continuity, $J(\posecoord)$ stays within $\jacsv/2$ of $J^\star$ in operator norm on some convex coordinate neighborhood of $\posecoord^\star$. For $\Delta\posecoord=\posecoord-\posecoord^\star$, the fundamental theorem of calculus gives
\[
 \vect g_E(\posecoord)-\vect g_E(\posecoord^\star)
 =J^\star\Delta\posecoord+\int_0^1\bigl(J(\posecoord^\star+\nu\Delta\posecoord)-J^\star\bigr)\Delta\posecoord\,d\nu.
\]
The first term has norm at least $\jacsv\norm{\Delta\posecoord}$, and the integral has norm at most $(\jacsv/2)\norm{\Delta\posecoord}$. Thus $\norm{\vect g_E(\posecoord)-\vect g_E(\posecoord^\star)}\ge(\jacsv/2)\norm{\Delta\posecoord}$ in this neighborhood. For an estimate in the neighborhood, combining this lower bound with the upper bound of the first part gives \cref{eq:edge-pose-bound}.
\end{proof}

The bound $\epsilon_E$ separates the three departures. The fraction $\edgemass$ of arclength that the coupling pairs within $\edgedisp$ contributes at most $\edgemass\featlip\edgedisp$. The remaining fraction $1-\edgemass$, which includes arclength present in only one of the two edge sets, contributes up to $2\featbound(1-\edgemass)$. Since $\epsilon_E\ge2\featbound(1-\edgemass)$, the bound is small only when nearly all arclength is shared. \cref{eq:edge-pose-bound} converts the feature discrepancy into a pose error, and the factor $2/\jacsv$ is large when the features respond weakly to some pose change. The bound is local and applies only to an estimate that the search has already brought near the true pose.

The following example shows that the full-column-rank condition of \cref{supp:lem-edge-stability} holds for a simple set of edges. Consider the normalized calibration $\Kmat=I_3$, the true pose with $\Rot^\star=I_3$ and $\trans^\star=0$, and the seven segments $(x,a_E,1)$ with $-1\le x\le1$ and $a_E=-3,\ldots,3$. For nonnegative integers $p$ and $q$, let $m_{p,q}$ denote the image-arclength average of $x^py^q$. The derivatives of $(m_{0,1},m_{0,2},m_{0,3},m_{1,0},m_{1,1},m_{1,2})$ in the local pose coordinates $\posecoord=(\omega_x,\omega_y,\omega_z,t_x,t_y,t_z)$ of the EM updates are
\[
 J_0=\begin{pmatrix}
 -9&0&0&0&1&0\\
 0&0&0&0&0&-8\\
 -124&0&0&0&12&0\\
 0&2&0&1&0&0\\
 0&0&-11/3&0&0&0\\
 0&32/3&0&4&0&0
 \end{pmatrix}.
\]
We obtain these entries by differentiating the projected endpoints and lengths and integrating, using $\mathbb E x^2=1/3$ along a segment and $\mathbb E a_E^2=4$ and $\mathbb E a_E^4=28$ over the segments. The determinant is $\det J_0=11264/9\ne0$, and it stays nonzero under small endpoint perturbations because it is continuous in the endpoints. \cref{sec:supp_observations} evaluates this edge variant on images and CAD models.

\FloatBarrier

\section{Experiments}
\label{sec:supp_experiments}
This appendix follows the order of \cref{sec:experiments}. \cref{sec:supp_baselines} gives the benchmark inputs and the learned baselines, and each later subsection gives the protocol and full results of one study in the main experiments section.
\subsection{Datasets \& baselines}
\label{sec:supp_baselines}
This subsection describes the inputs of the three benchmarks in \cref{tab:main} and how we run the learned baselines on them.

\para{Benchmark inputs.}
For MegaDepth~\cite{li2018megadepth}, ModelNet40~\cite{wu2015shapenets} and Data61~\cite{namin2015data61}, we read the released per-frame 2D and 3D coordinates of the blind PnP benchmarks~\cite{campbell2020bpnpnet} and keep their original point selection. MegaDepth pairs 2D points with model points that COLMAP reconstructs by structure from motion (SfM). Its test frames contain 10 to 12{,}125 points. ModelNet40 samples 1{,}000 points uniformly from each CAD model, projects them with a virtual pinhole camera and perturbs the projections with Gaussian image noise. Data61 pairs panoramas with light detection and ranging (LiDAR) point sets, and we use its released frames of 1{,}000 points. We remove the repeated padding entries indicated by the released counts, leave all other coordinates unchanged and shuffle both point sets. Each test set supplies 1{,}000 random outlier-free frames and 300 frames with at most 2{,}000 inliers. To the latter, we add outliers drawn uniformly in the bounding box of each point set. We call these frames the comparison frames. The outlier ratio is the number of outliers added to each point set divided by its inlier count.

\para{Learned baselines.}
We run Deep Blind PnP~\cite{liu2020learning}, BPnPNet~\cite{campbell2020bpnpnet} and GoMatch~\cite{zhou2022gomatch} with their released code and weights. BPnPNet returns three poses, from RANSAC on its predicted correspondences, from its declarative layer and after Levenberg--Marquardt (LM) refinement. On each dataset, we run BPnPNet checkpoints trained with the correspondence loss alone ($L_c$) and with the pose loss. The Deep Blind PnP release passes the 2{,}000 most probable matches to the P3P-RANSAC of OpenCV. We also evaluate posest, the pose-estimation library recommended in that release, which reproduces its reported MegaDepth median rotation error. The released variant with a fixed match count fails on frames with fewer than about 45 points, because these frames have fewer than 2{,}000 candidate pairs. Our top-$n$ variants therefore pass the $1.5\min(n,N)$ most probable matches, up to 2{,}000. The variants marked +LM refine the RANSAC pose by LM on its inliers. For GoMatch, we apply its coordinate-only MegaDepth checkpoint to the input point sets without retrieved reference images. \cref{tab:supp_variants} lists every pose output that we evaluate, and \cref{tab:main} reports the most accurate one.

\subsection{Benchmarks}
\label{sec:supp_tables}
This subsection completes the benchmark comparison of \cref{tab:main}. It reports error quartiles and distributions, accuracy at every outlier ratio and for every pose output, runtimes, results on held-out frames, and examples.

\begin{figure*}[tp]
\centering
\pgfplotsset{curveaxis/.style={mpPlot, title style={font=\footnotesize, yshift=-1.2ex},
  xlabel near ticks, ylabel near ticks, ylabel shift=-3pt, xlabel shift=-2pt,
  ymin=-4, ymax=104, ytick={0,25,50,75,100}, unbounded coords=jump, axis background/.style={fill=white}},
  cOurs/.style={colOurs, line width=0.9pt, mark=star, mark size=1.9pt}, cLiu/.style={colLiu, line width=0.65pt, mark=square*, mark size=1.1pt}, cBPnP/.style={colBPnP, line width=0.65pt, mark=triangle*, mark size=1.3pt},
  cCFPE/.style={oiOrange, line width=0.65pt, mark=*, mark size=1.3pt},
  cEM/.style={black!80, line width=0.65pt, mark=pentagon*, mark size=1.3pt},
  cGoM/.style={oiPurple, line width=0.65pt, mark=diamond*, mark size=1.3pt}}
{\footnotesize\begin{tabular}{@{}llllll@{}}
\tikz[baseline=-0.6ex]{\draw[colOurs, line width=0.9pt] (0,0) -- (0.32,0); \draw[colOurs] plot[mark=star, mark size=1.9pt] coordinates {(0.16,0)};}~\method & \tikz[baseline=-0.6ex]{\draw[colLiu, line width=0.65pt] (0,0) -- (0.32,0); \draw[colLiu] plot[mark=square*, mark size=1.1pt] coordinates {(0.16,0)};}~Deep Blind PnP & \tikz[baseline=-0.6ex]{\draw[colBPnP, line width=0.65pt] (0,0) -- (0.32,0); \draw[colBPnP] plot[mark=triangle*, mark size=1.3pt] coordinates {(0.16,0)};}~BPnPNet &
\tikz[baseline=-0.6ex]{\draw[oiOrange, line width=0.65pt] (0,0) -- (0.32,0); \draw[oiOrange] plot[mark=*, mark size=1.3pt] coordinates {(0.16,0)};}~CFPE (reproduced) & \tikz[baseline=-0.6ex]{\draw[oiPurple, line width=0.65pt] (0,0) -- (0.32,0); \draw[oiPurple] plot[mark=diamond*, mark size=1.3pt] coordinates {(0.16,0)};}~GoMatch & \tikz[baseline=-0.6ex]{\draw[black!80, line width=0.65pt] (0,0) -- (0.32,0); \draw[black!80] plot[mark=pentagon*, mark size=1.3pt] coordinates {(0.16,0)};}~EM-ICP
\end{tabular}}\\[2pt]
\includegraphics[width=494.15222pt,height=114.24873pt]{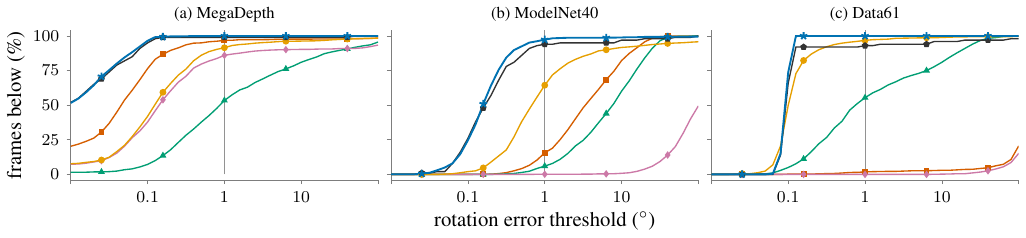}
\caption{\textbf{Rotation-error distributions without outliers} on the test sets of MegaDepth~\cite{li2018megadepth}, ModelNet40~\cite{wu2015shapenets} and Data61~\cite{namin2015data61}. Each curve gives the percentage of frames with rotation error below each value, and the vertical line marks $1\degree$. EM-ICP uses 100 of the comparison frames per dataset, and the other methods the 1{,}000 outlier-free frames.}
\label{fig:supp_curves_outlier_free}
\end{figure*}

\begin{figure*}[tp]
\centering
\pgfplotsset{curveaxis/.style={mpPlot, title style={font=\footnotesize, yshift=-1.3ex},
  xlabel near ticks, ylabel near ticks, ylabel shift=-3pt, xlabel shift=-2pt,
  ymin=-4, ymax=104, ytick={0,25,50,75,100}, unbounded coords=jump,
  legend style={font=\footnotesize, draw=none, fill=none, inner sep=1pt, /tikz/every even column/.append style={column sep=8pt}}, legend cell align=left, legend columns=-1},
  cOurs/.style={colOurs, line width=0.9pt, mark=star, mark size=1.9pt}, cLiu/.style={colLiu, line width=0.65pt, mark=square*, mark size=1.2pt}, cBPnP/.style={colBPnP, line width=0.65pt, mark=triangle*, mark size=1.5pt},
  cCFPE/.style={oiOrange, line width=0.65pt, mark=*, mark size=1.3pt},
  cEM/.style={black!80, line width=0.65pt, mark=pentagon*, mark size=1.3pt},
  cGoM/.style={oiPurple, line width=0.65pt, mark=diamond*, mark size=1.5pt}}
{\footnotesize\begin{tabular}{@{}lll@{}}
\tikz[baseline=-0.6ex]{\draw[colOurs, line width=0.9pt] (0,0) -- (0.32,0); \draw[colOurs] plot[mark=star, mark size=1.9pt] coordinates {(0.16,0)};}~\method & \tikz[baseline=-0.6ex]{\draw[colLiu, line width=0.65pt] (0,0) -- (0.32,0); \draw[colLiu] plot[mark=square*, mark size=1.1pt] coordinates {(0.16,0)};}~Deep Blind PnP & \tikz[baseline=-0.6ex]{\draw[colBPnP, line width=0.65pt] (0,0) -- (0.32,0); \draw[colBPnP] plot[mark=triangle*, mark size=1.1pt] coordinates {(0.16,0)};}~BPnPNet \\
\tikz[baseline=-0.6ex]{\draw[oiOrange, line width=0.65pt] (0,0) -- (0.32,0); \draw[oiOrange] plot[mark=*, mark size=1.3pt] coordinates {(0.16,0)};}~CFPE (reproduced) & \tikz[baseline=-0.6ex]{\draw[oiPurple, line width=0.65pt] (0,0) -- (0.32,0); \draw[oiPurple] plot[mark=diamond*, mark size=1.1pt] coordinates {(0.16,0)};}~GoMatch & \tikz[baseline=-0.6ex]{\draw[black!80, line width=0.65pt] (0,0) -- (0.32,0); \draw[black!80] plot[mark=pentagon*, mark size=1.1pt] coordinates {(0.16,0)};}~EM-ICP
\end{tabular}}\\[2pt]
\includegraphics[width=459.78197pt,height=137.48947pt]{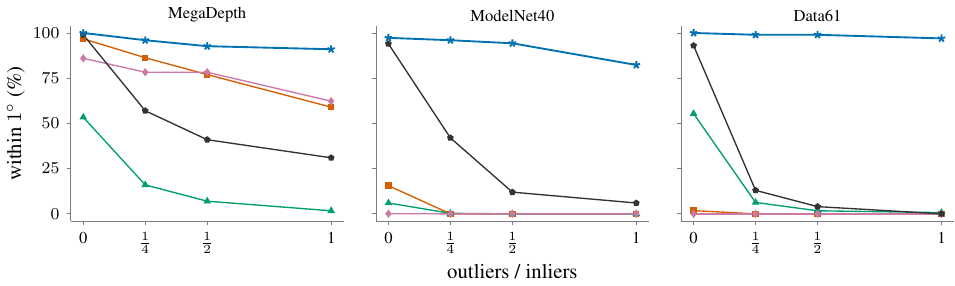}\\[1pt]
{\footnotesize (a) Accuracy against the outlier ratio}\\[5pt]
\includegraphics[width=462.83652pt,height=134.07138pt]{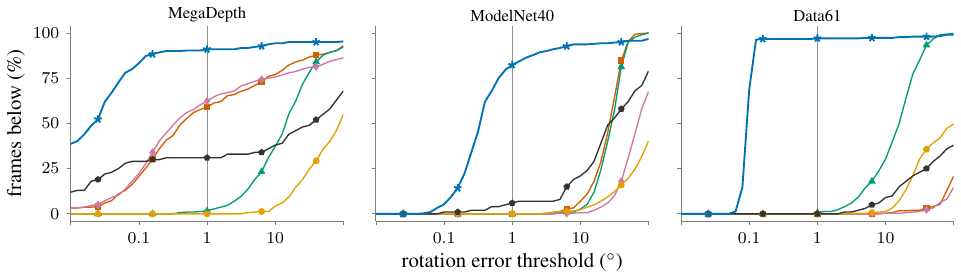}\\[1pt]
{\footnotesize (b) Rotation-error distribution with as many outliers as inliers}\\[5pt]
\includegraphics[width=462.85345pt,height=134.0054pt]{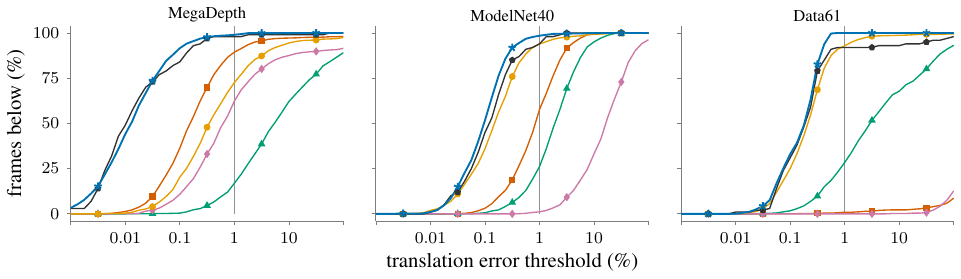}\\[1pt]
{\footnotesize (c) Translation-error distribution without outliers}
\caption{\textbf{Accuracy with outliers and further error distributions} on MegaDepth~\cite{li2018megadepth}, ModelNet40~\cite{wu2015shapenets} and Data61~\cite{namin2015data61}. (a)~Percentage of frames with rotation error below $1\degree$ against the outlier ratio. (b)~Percentage of frames below each rotation error, with as many outliers as inliers. (c)~Percentage of outlier-free frames below each relative translation error. Vertical lines mark $1\degree$ and 1\%. EM-ICP uses 100 of the comparison frames per condition, and the other methods 300 frames with outliers and 1{,}000 without. CFPE appears in (b) and (c).}
\label{fig:supp_curves}
\end{figure*}

\para{Errors without outliers.}
\cref{tab:supp_errors} gives the quartiles of rotation error and of relative translation error on the outlier-free test sets, together with the percentage of frames within both $1\degree$ and 1\%. \cref{fig:supp_curves_outlier_free} plots the full rotation-error distribution on each test set, and \cref{fig:supp_curves}(c) the distribution of relative translation error. A symmetric object can have a large rotation error at a correct alignment, so on ModelNet40 we also measure ADD-S~\cite{hinterstoisser2012model}. It averages the distance from each model point under the estimated pose to the nearest model point under the reference pose. ADD-S is below 10\% of the model diameter on \MNaddsOK\ of all 2{,}466 outlier-free ModelNet40 test frames.

\para{Outliers.}
\cref{tab:supp_variants} gives the accuracy of every evaluated pose output at each outlier ratio, at thresholds of $1\degree$ and $5\degree$. \cref{fig:supp_curves}(a) plots accuracy against the outlier ratio, and \cref{fig:supp_curves}(b) the rotation-error distribution with as many outliers as inliers.

\para{Runtime.}
\cref{tab:supp_timing} gives the median time per frame at each outlier ratio. Without outliers, GoMatch alone is faster than \method. With outliers, \method\ is the slowest of the four methods, at 0.2 to 7.1~s per frame. \cref{fig:supp_cumulative} plots accuracy against time, comparing methods that return a single pose with searches that improve their best pose until the time limit. \cref{tab:supp_bysize} groups the outlier-free MegaDepth test frames by their number of points. In every group, 99.8 to 100\% of frames are within $1\degree$. The median time grows from 17~ms for 30 to 99 points to 227~ms for 3{,}000 points or more, and it is 800~ms for the six frames with fewer than 30 points.

\para{Held-out frames.}
We tuned \method\ on the benchmark test sets, so we also evaluate 1{,}000 frames per dataset from the training split not used for tuning. \cref{tab:supp_details} gives the accuracy, acceptance outcomes and runtimes on these held-out frames and on the full test sets. On ModelNet40 with outliers, no held-out pose meets the acceptance criteria, so these rows score the most likely pose, which the solver returns without acceptance (\cref{sec:supp_search}).

\begin{figure*}[tp]
\centering
\includegraphics[width=494.77615pt,height=252.49762pt]{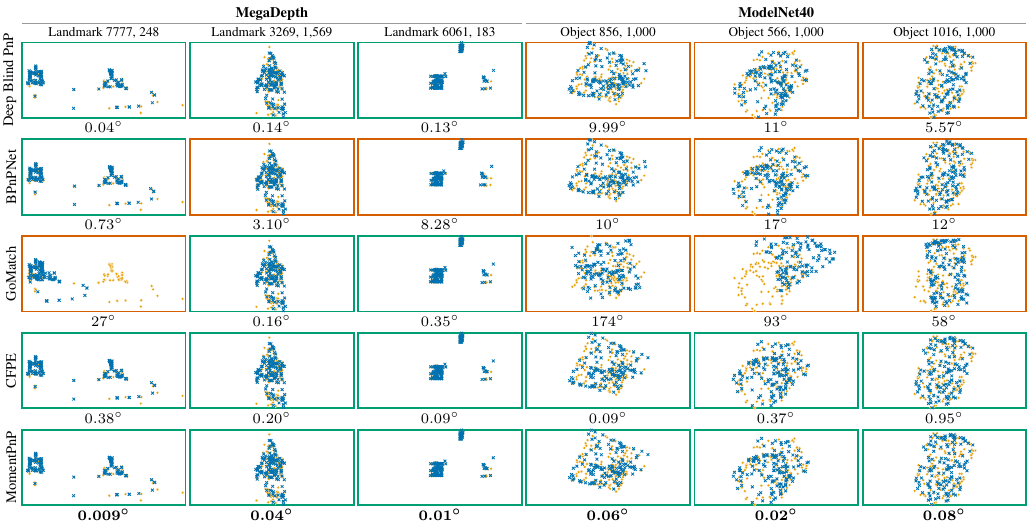}
\caption{\textbf{Qualitative results on point-only benchmarks.} MegaDepth and ModelNet40 provide coordinates without photographs. Orange dots are the 2D points, and blue crosses are the model points projected under the returned pose. Titles give the landmark or object and the point count. Labels give rotation error, with the lowest in bold, and green borders mark errors below $1\degree$.}
\label{fig:gallery_points}
\end{figure*}

\begin{figure*}[tp]
\centering
\includegraphics[width=475.32477pt,height=200.42232pt]{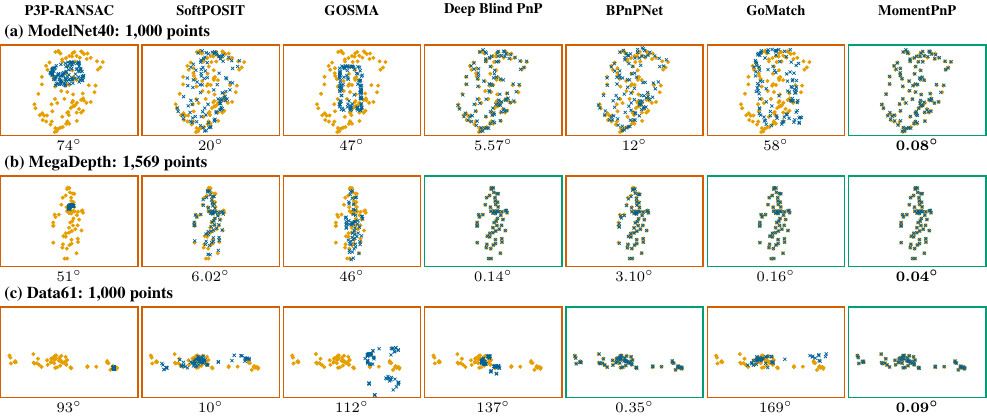}
\caption{\textbf{One benchmark frame per dataset for each method.} Rows (a)--(c) show ModelNet40~\cite{wu2015shapenets}, MegaDepth~\cite{li2018megadepth} and Data61~\cite{namin2015data61}. Orange dots are the 2D points, and blue crosses are the model points projected under the returned pose. Labels give rotation error, with the lowest in bold, and green borders mark errors below $1\degree$.}
\label{fig:supp_methods}
\end{figure*}

\para{Examples.}
\cref{fig:gallery_points} shows MegaDepth and ModelNet40 frames, which have no photographs, for the learned methods, CFPE and \method. \cref{fig:supp_methods} shows one frame of each benchmark for P3P-RANSAC, SoftPOSIT, GOSMA, the learned methods and \method.
\begin{table}[tp]
\centering
\footnotesize
\setlength{\tabcolsep}{4pt}
\renewcommand{\arraystretch}{0.86}
\captionsetup{hypcap=false,type=table,skip=4pt}
\caption{\textbf{Median time per frame (ms, $\downarrow$)} on the comparison frames of \cref{tab:main}, at each outlier ratio. Each learned method is timed with its most accurate pose output. Bold marks the lowest time in each column of each dataset.}
\label{tab:supp_timing}
\begin{tabular}{llrrrr}
\toprule
Set & Method & 0 & $1/4$ & $1/2$ & $1$ \\
\midrule
MegaDepth~\cite{li2018megadepth} & Deep Blind PnP & 240 & 119 & 149 & 133 \\
 & BPnPNet & 223 & 180 & 180 & 174 \\
 & GoMatch & \textbf{21} & \textbf{28} & \textbf{29} & \textbf{39} \\
 & \method\ (ours) & 31 & 206 & 259 & 394 \\
\addlinespace[3pt]
ModelNet40~\cite{wu2015shapenets} & Deep Blind PnP & 330 & \textbf{27} & \textbf{34} & \textbf{46} \\
 & BPnPNet & 159 & 95 & 97 & 100 \\
 & GoMatch & \textbf{36} & 44 & 64 & 97 \\
 & \method\ (ours) & 56 & 3{,}035 & 4{,}350 & 7{,}110 \\
\addlinespace[3pt]
Data61~\cite{namin2015data61} & Deep Blind PnP & 289 & 128 & 128 & 137 \\
 & BPnPNet & 159 & 114 & 108 & 117 \\
 & GoMatch & \textbf{35} & \textbf{45} & \textbf{61} & \textbf{94} \\
 & \method\ (ours) & 48 & 268 & 360 & 526 \\
\bottomrule
\end{tabular}
\par\vspace{6pt}
\setlength{\tabcolsep}{5pt}
\caption{\textbf{\method\ on MegaDepth~\cite{li2018megadepth} by point count (ours).} Outlier-free test frames grouped by their number of points. Frames counts the frames in each group, $<1\degree$ gives the percentage with rotation error below $1\degree$, and the last two columns give the median and 95th-percentile time per frame in milliseconds.}
\label{tab:supp_bysize}
\begin{tabular}{lrrrr}
\toprule
Points & Frames & $<1\degree$ (\%)\,$\uparrow$ & Median\,$\downarrow$ & 95th pct.\,$\downarrow$ \\
\midrule
fewer than 30 & 6 & 100.0 & 800 & 2{,}741 \\
30--99 & 726 & 99.9 & 17 & 23 \\
100--299 & 2{,}345 & 99.8 & 19 & 26 \\
300--999 & 4{,}596 & 100.0 & 31 & 54 \\
1{,}000--2{,}999 & 2{,}501 & 100.0 & 68 & 192 \\
3{,}000 or more & 621 & 100.0 & 227 & 715 \\
\bottomrule
\end{tabular}

\end{table}

\begin{table*}[tp]
\centering\footnotesize
\setlength{\tabcolsep}{3pt}
\caption{\textbf{Accuracy at every outlier ratio ($\uparrow$).} Entries give the percentage of all evaluated frames with rotation error below $1\degree$ (top) or $5\degree$ (bottom), for the MegaDepth~\cite{li2018megadepth}, ModelNet40~\cite{wu2015shapenets} and Data61~\cite{namin2015data61} comparison frames of \cref{tab:main}. The outlier ratio is the number of outliers added to each point set divided by its inlier count. We list every evaluated pose output of the learned methods (\cref{sec:supp_baselines}). The label $L_c$ marks BPnPNet weights trained with the correspondence loss alone. Bold marks the highest accuracy in each column of each panel.}
\label{tab:supp_variants}
\begin{tabular*}{\textwidth}{@{\extracolsep{\fill}}l cccc cccc cccc@{}}
\toprule
 & \multicolumn{4}{c}{MegaDepth~\cite{li2018megadepth}} & \multicolumn{4}{c}{ModelNet40~\cite{wu2015shapenets}} & \multicolumn{4}{c}{Data61~\cite{namin2015data61}} \\
\cmidrule(lr){2-5}\cmidrule(lr){6-9}\cmidrule(l){10-13}
Method & 0 & $1/4$ & $1/2$ & $1$ & 0 & $1/4$ & $1/2$ & $1$ & 0 & $1/4$ & $1/2$ & $1$ \\
\midrule
\multicolumn{13}{@{}l}{\emph{Rotation error below $1\degree$ (\%)}} \\
\addlinespace[2pt]
BPnPNet~\cite{campbell2020bpnpnet} \\
\quad RANSAC & 50.1 & 13.0 & 6.3 & 1.0 & 5.4 & 0.0 & 0.0 & 0.0 & 51.1 & 5.7 & 1.0 & 0.3 \\
\quad declarative & 19.2 & 0.0 & 0.0 & 0.3 & 1.9 & 0.3 & 0.0 & 0.0 & 9.6 & 1.3 & 1.7 & 0.3 \\
\quad RANSAC+LM & 53.4 & 16.0 & 7.0 & 1.7 & 6.0 & 0.0 & 0.0 & 0.0 & 55.4 & 6.0 & 1.0 & 0.7 \\
\quad $L_c$, RANSAC & 41.8 & 9.7 & 3.3 & 0.7 & 2.6 & 0.0 & 0.0 & 0.0 & 47.9 & 5.7 & 1.7 & 0.3 \\
\quad $L_c$, declarative & 11.4 & 0.0 & 1.0 & 0.3 & 1.6 & 0.0 & 0.0 & 0.0 & 8.7 & 1.0 & 0.7 & 0.3 \\
\quad $L_c$, RANSAC+LM & 44.8 & 12.0 & 5.3 & 1.3 & 3.1 & 0.0 & 0.0 & 0.0 & 52.4 & 6.3 & 1.7 & 0.3 \\
Deep Blind PnP~\cite{liu2020learning} \\
\quad released & 79.3 & 64.7 & 53.3 & 34.0 & 10.3 & 0.0 & 0.0 & 0.0 & 0.1 & 0.0 & 0.0 & 0.0 \\
\quad released+LM (ours) & 79.7 & 65.3 & 54.3 & 33.0 & 9.3 & 0.0 & 0.0 & 0.0 & 0.1 & 0.0 & 0.0 & 0.0 \\
\quad top-$n$+LM (ours) & 92.7 & 72.7 & 63.3 & 39.0 & 12.9 & 0.0 & 0.0 & 0.0 & 0.1 & 0.0 & 0.0 & 0.0 \\
\quad posest & 91.7 & 85.0 & 77.0 & 57.3 & 15.6 & 0.0 & 0.0 & 0.0 & 1.0 & 0.0 & 0.0 & 0.0 \\
\quad posest top-$n$ (ours) & 96.8 & 86.3 & 76.0 & 59.0 & 15.6 & 0.0 & 0.0 & 0.0 & 1.8 & 0.0 & 0.0 & 0.0 \\
GoMatch~\cite{zhou2022gomatch} & 86.0 & 78.3 & 78.3 & 62.3 & 0.1 & 0.0 & 0.0 & 0.0 & 0.0 & 0.0 & 0.0 & 0.0 \\
\method (ours) & \textbf{100.0} & \textbf{96.0} & \textbf{92.7} & \textbf{91.0} & \textbf{97.3} & \textbf{96.0} & \textbf{94.3} & \textbf{82.3} & \textbf{100.0} & \textbf{99.0} & \textbf{99.0} & \textbf{97.0} \\
\midrule
\multicolumn{13}{@{}l}{\emph{Rotation error below $5\degree$ (\%)}} \\
\addlinespace[2pt]
BPnPNet~\cite{campbell2020bpnpnet} \\
\quad RANSAC & 75.5 & 40.3 & 27.7 & 18.0 & 35.3 & 2.0 & 0.3 & 0.7 & 72.7 & 35.3 & 24.7 & 13.7 \\
\quad declarative & 60.7 & 21.3 & 14.0 & 12.7 & 26.0 & 0.7 & 0.3 & 0.3 & 55.4 & 22.3 & 22.0 & 15.3 \\
\quad RANSAC+LM & 74.3 & 42.0 & 30.3 & 19.0 & 37.3 & 2.0 & 0.7 & 1.0 & 73.0 & 36.0 & 24.7 & 14.0 \\
\quad $L_c$, RANSAC & 69.2 & 40.0 & 27.0 & 15.3 & 22.9 & 1.7 & 0.3 & 0.3 & 70.2 & 37.0 & 26.0 & 15.0 \\
\quad $L_c$, declarative & 52.5 & 15.0 & 11.7 & 9.3 & 20.4 & 1.0 & 0.0 & 0.0 & 52.7 & 25.3 & 20.7 & 16.3 \\
\quad $L_c$, RANSAC+LM & 69.2 & 43.3 & 30.3 & 16.3 & 25.8 & 1.3 & 0.3 & 0.3 & 70.4 & 37.0 & 26.0 & 15.0 \\
Deep Blind PnP~\cite{liu2020learning} \\
\quad released & 90.3 & 84.7 & 72.0 & 58.3 & 58.4 & 8.0 & 1.3 & 1.3 & 1.2 & 1.0 & 0.3 & 0.0 \\
\quad released+LM (ours) & 90.0 & 84.7 & 72.0 & 58.7 & 59.2 & 7.3 & 1.3 & 1.3 & 1.2 & 1.0 & 0.3 & 0.0 \\
\quad top-$n$+LM (ours) & 96.8 & 88.3 & 80.3 & 61.3 & 59.8 & 6.7 & 1.3 & 1.3 & 1.4 & 0.7 & 0.3 & 0.0 \\
\quad posest & 94.9 & 91.3 & 85.3 & 71.0 & 61.4 & 6.7 & 1.7 & 1.0 & 2.7 & 1.0 & 0.0 & 0.3 \\
\quad posest top-$n$ (ours) & 97.6 & 92.0 & 85.7 & 71.0 & 64.1 & 7.7 & 1.7 & 1.0 & 2.6 & 1.7 & 0.0 & 0.3 \\
GoMatch~\cite{zhou2022gomatch} & 90.0 & 83.7 & 84.7 & 73.3 & 0.4 & 0.0 & 0.3 & 0.3 & 0.0 & 0.0 & 0.0 & 0.0 \\
\method (ours) & \textbf{100.0} & \textbf{96.3} & \textbf{93.0} & \textbf{92.3} & \textbf{98.7} & \textbf{99.3} & \textbf{99.0} & \textbf{91.7} & \textbf{100.0} & \textbf{99.0} & \textbf{99.0} & \textbf{97.3} \\
\bottomrule
\end{tabular*}
\end{table*}

\begin{table*}[tp]
\centering\footnotesize
\setlength{\tabcolsep}{6pt}
\caption{\textbf{Rotation \& translation errors without outliers} on the test sets of MegaDepth~\cite{li2018megadepth}, ModelNet40~\cite{wu2015shapenets} and Data61~\cite{namin2015data61}. Error columns give the first quartile (Q1), median and third quartile (Q3) over frames with a returned pose. Translation-vector error is divided by the norm of the reference translation. The final column gives the percentage of all evaluated frames with both rotation error below $1\degree$ and relative translation error below 1\%. Each learned method uses its most accurate pose output under the rotation criterion. Bold marks the lowest quartiles and the highest percentage in each dataset.}
\label{tab:supp_errors}
\begin{tabular}{llccc}
\toprule
 & & Rotation error ($\degree$)\,$\downarrow$ & Translation error (\%)\,$\downarrow$ & Both below\,$\uparrow$ \\
Set & Method & Q1 / median / Q3 & Q1 / median / Q3 & $1\degree$ and 1\% (\%) \\
\midrule
MegaDepth~\cite{li2018megadepth} & \method (ours) & \textbf{0.000} / \textbf{0.009} / \textbf{0.031} & \textbf{0.005} / \textbf{0.013} / \textbf{0.035} & \textbf{99.0} \\
 & Deep Blind PnP (posest top-$n$) & 0.018 / 0.045 / 0.096 & 0.065 / 0.155 / 0.398 & 89.4 \\
 & BPnPNet (RANSAC+LM) & 0.277 / 0.853 / 5.530 & 1.571 / 5.402 / 26.087 & 16.9 \\
 & GoMatch (released) & 0.059 / 0.136 / 0.348 & 0.228 / 0.590 / 1.880 & 62.0 \\
\addlinespace[3pt]
ModelNet40~\cite{wu2015shapenets} & \method (ours) & \textbf{0.101} / \textbf{0.155} / \textbf{0.250} & \textbf{0.046} / \textbf{0.096} / \textbf{0.171} & \textbf{96.5} \\
 & Deep Blind PnP (posest) & 1.582 / 3.350 / 8.058 & 0.392 / 0.833 / 1.678 & 13.0 \\
 & BPnPNet (RANSAC+LM) & 3.209 / 7.497 / 15.387 & 0.963 / 1.909 / 3.626 & 2.6 \\
 & GoMatch (released) & 56.857 / 100.473 / 151.686 & 7.593 / 16.255 / 33.558 & 0.1 \\
\addlinespace[3pt]
Data61~\cite{namin2015data61} & \method (ours) & \textbf{0.084} / \textbf{0.092} / \textbf{0.101} & \textbf{0.075} / \textbf{0.183} / \textbf{0.282} & \textbf{100.0} \\
 & Deep Blind PnP (posest top-$n$) & 104.450 / 129.940 / 157.006 & 212.138 / 347.411 / 617.287 & 0.9 \\
 & BPnPNet (RANSAC+LM) & 0.288 / 0.761 / 6.275 & 0.861 / 2.903 / 20.331 & 27.8 \\
 & GoMatch (released) & 117.717 / 143.316 / 163.898 & 161.679 / 268.018 / 468.110 & 0.0 \\
\bottomrule
\end{tabular}
\end{table*}

\begin{table*}[tp]
\centering\footnotesize
\setlength{\tabcolsep}{5pt}
\caption{\textbf{\method (ours) on every evaluated split} of MegaDepth~\cite{li2018megadepth}, ModelNet40~\cite{wu2015shapenets} and Data61~\cite{namin2015data61}. The outlier ratio is the number of added outliers per point set divided by its inlier count. Rotation-accuracy percentages use every evaluated frame, independently of the acceptance test. Acceptance columns count frames, and timing columns give milliseconds per frame. Test rows use the full released test sets. Held-out rows use 1{,}000 randomly selected training-split frames per dataset that were not used for tuning.}
\label{tab:supp_details}
\begin{tabular}{llrrrrrrrrr}
\toprule
 & & & \multicolumn{2}{c}{Frames below (\%)\,$\uparrow$} & \multicolumn{4}{c}{Acceptance (frames)} & \multicolumn{2}{c}{Time (ms)\,$\downarrow$} \\
\cmidrule(lr){4-5}\cmidrule(lr){6-9}\cmidrule(l){10-11}
Split & Outliers & Frames & $<1\degree$ & $<5\degree$ & accepted & accepted, $\ge 5\degree$ & declined, $<1\degree$ & no pose & median & mean \\
\midrule
MegaDepth~\cite{li2018megadepth}, test & 0 & 10{,}795 & 99.9 & 100.0 & 10{,}788 & 0 & 7 & 0 & 31 & 58 \\
 & $1/4$ & 10{,}795 & 96.8 & 97.4 & 9{,}559 & 1 & 921 & 7 & 219 & 597 \\
\addlinespace[3pt]
ModelNet40~\cite{wu2015shapenets}, test & 0 & 2{,}466 & 97.0 & 98.7 & 2{,}366 & 13 & 60 & 0 & 55 & 256 \\
\addlinespace[3pt]
Data61~\cite{namin2015data61}, test & 0 & 2{,}640 & 100.0 & 100.0 & 2{,}640 & 0 & 0 & 0 & 48 & 52 \\
 & $1/4$ & 2{,}640 & 98.9 & 99.1 & 2{,}601 & 0 & 11 & 1 & 261 & 391 \\
\addlinespace[3pt]
MegaDepth, held-out & 0 & 1{,}000 & 100.0 & 100.0 & 1{,}000 & 0 & 0 & 0 & 34 & 35 \\
 & $1/4$ & 1{,}000 & 97.0 & 97.3 & 916 & 1 & 55 & 0 & 205 & 327 \\
 & $1/2$ & 1{,}000 & 94.7 & 95.2 & 900 & 0 & 48 & 0 & 263 & 452 \\
 & $1$ & 1{,}000 & 91.3 & 92.4 & 866 & 1 & 50 & 0 & 369 & 728 \\
\addlinespace[3pt]
ModelNet40, held-out & 0 & 1{,}000 & 98.6 & 99.1 & 962 & 5 & 32 & 0 & 53 & 227 \\
 & $1/4$ & 1{,}000 & 98.1 & 100.0 & 0 & 0 & 981 & 0 & 3{,}184 & 3{,}288 \\
 & $1/2$ & 1{,}000 & 95.4 & 99.4 & 0 & 0 & 954 & 0 & 4{,}531 & 4{,}602 \\
 & $1$ & 1{,}000 & 72.4 & 85.7 & 0 & 0 & 724 & 0 & 7{,}224 & 7{,}318 \\
\addlinespace[3pt]
Data61, held-out & 0 & 1{,}000 & 100.0 & 100.0 & 1{,}000 & 0 & 0 & 0 & 48 & 55 \\
 & $1/4$ & 1{,}000 & 96.6 & 97.7 & 891 & 0 & 75 & 2 & 294 & 619 \\
 & $1/2$ & 1{,}000 & 94.7 & 96.3 & 866 & 0 & 81 & 1 & 388 & 1{,}006 \\
 & $1$ & 1{,}000 & 91.4 & 94.1 & 830 & 0 & 86 & 0 & 536 & 1{,}685 \\
\bottomrule
\end{tabular}
\end{table*}

\input{fig_supp_cumulative}

\subsection{Role of moment matching}
\label{sec:supp_moment_role}
This subsection gives the protocols and full results of the three comparisons in \cref{sec:experiments} that test what moment matching contributes to \method. They match moments of different orders before the same EM refinement, restart EM from random poses, and search globally over poses. The subsection ends with point sets smaller than those covered by \cref{thm:pinhole}.

\begin{table}[tp]
\centering\small
\setlength{\tabcolsep}{3pt}
\caption{\textbf{Moment order with shared refinement (ours)} on outlier-free inputs. Entries count accurate poses before / after the same EM refinement, out of the $n_r$ inputs in each row. Direct EM refines the initial poses without moment matching, and Degree $\order$ matches the average of every monomial through degree $\order$. A pose is accurate when its rotation error is below $1\degree$ and its translation error is below 1\% of the reference translation norm.}
\label{tab:supp_feature_control}
\begin{tabular*}{\columnwidth}{@{\extracolsep{\fill}}lrcccc@{}}
\toprule
Dataset & $n_r$ & Direct EM & Degree 2 & Degree 3 & Degree 4 \\
\midrule
MegaDepth & 100 & 0 / 68 & 0 / 57 & 53 / 70 & 59 / 70 \\
ModelNet40 & 100 & 0 / 39 & 2 / 26 & 56 / 78 & 54 / 66 \\
Data61 & 100 & 0 / 64 & 1 / 48 & 69 / 75 & 78 / 83 \\
Synthetic & 30 & 0 / 12 & 0 / 15 & 27 / 27 & 24 / 25 \\
\bottomrule
\end{tabular*}
\end{table}

\para{Moment order.}
To isolate the effect of moment order, every variant starts from the same initial orientations for each input, uses the same search effort and fits translations from the centroid and spread of the 2D points. We use 100 outlier-free test frames from each benchmark and a separate sample of thirty independent synthetic models of 64 points. Their projections receive Gaussian image noise of standard deviation $0.000625$ per calibrated coordinate. The moment variants match the averages of all five, nine or fourteen nonconstant monomials through degree two, three or four. All orders share one image normalization. Each moment variant ranks and refines poses by its own moment residual. Direct EM takes the initial poses in a fixed random order. All variants retain four poses with the same minimum angular separation, refine them with the same EM settings and select the final pose by likelihood. For the counts before EM, we estimate the noise and mixture parameters and select a pose by likelihood without updating it. Before EM, the third- and fourth-order variants bring 53 to 78 of the 100 poses on each benchmark within $1\degree$ rotation error and 1\% relative translation error, and the second-order variant at most two (\cref{tab:supp_feature_control}). After EM, both still recover more poses than direct EM and the second-order variant on every dataset. Which of the two is more accurate varies across datasets.

\para{EM from random poses.}
EM-ICP in \cref{tab:main} restarts the EM refinement of \method\ from random poses until the 30-second limit and keeps the most likely pose. It restarts a median of \EmRestarts\ times per MegaDepth frame, and its 30 seconds per frame are over five hundred times the median time of \method\ without outliers (\cref{tab:supp_timing}). \cref{fig:supp_cumulative} shows how its accuracy grows with time.

\para{Global search.}
\label{sec:supp_global_search}
GOPAC~\cite{campbell2020gopac} and GOSMA~\cite{campbell2019gosma} search over poses by branch and bound within a given domain of camera centers. We run them and \method\ on thirty complete, outlier-free pairs of point sets, ten fixed frames from each of the MegaDepth, Data61 and ModelNet40 test sets, without the released time limits. We remove the external clock and the internal duration limits of the released programs, 6{,}000~s for GOPAC and 600~s for GOSMA, and otherwise keep their code and settings. A search then ends on its own only when GOPAC proves its optimum or GOSMA reaches its released queue-size limit. On a frame where both end on their own within the earlier limits, the modified and released programs return the same poses. We stopped the GOPAC searches still running after 5.2 to 6.0 hours and scored their best pose so far. In a first attempt, four GOPAC searches exceeded a per-process memory cap of 3~GB, and we restarted them without the cap.
Both global methods receive unit bearings and the same normalized model points. We center each model at its 3D centroid $\bar{\vect y}$ and scale it to a fixed root-mean-square distance from that centroid. For GOPAC, we use the released angular tolerance of 0.017 radians ($0.974\degree$) and compare two initial lower bounds on the number of inliers, zero and an annealed bound that starts from one fewer than the number of 2D points. GOSMA minimizes a normalized squared $L^2$ distance between densities, and we keep its released stopping tolerance of 0.1 on the gap between the best objective value and the global lower bound.

\para{Search domain.}
Both global methods need a bounded domain of camera centers that contains the true center, which we compute from the inputs. Let $a_{\max}$ be the largest distance from a model point to the centroid $\bar{\vect y}$, let $\vect c$ be the camera center, and let $\beta>0$ be the angle between two observed unit bearings. The bound below holds for every such pair, and we take a widely separated pair because a larger angle tightens it. For noiseless complete observations, the camera distance $d_{\rm c}=\lVert\vect c-\bar{\vect y}\rVert$ satisfies
\[
 d_{\rm c}\leq a_{\max}\bigl(1+\csc(\beta/2)\bigr).
\]
The bound is immediate when $d_{\rm c}\leq a_{\max}$. When $d_{\rm c}>a_{\max}$, the two model points observed along these bearings are at least $d_{\rm c}-a_{\max}>0$ from the camera, so by the cosine rule they are at least $2(d_{\rm c}-a_{\max})\sin(\beta/2)$ apart. They are also at most $2a_{\max}$ apart, which gives the bound. Both global methods search the same cuboid centered at $\bar{\vect y}$, with half-width twice the right-hand side of the bound to allow for measurement noise. This cuboid contains all thirty reference camera centers.

\begin{table*}[tp]
\centering
\footnotesize
\setlength{\tabcolsep}{4pt}
\caption{\textbf{Global search on complete inputs without the released time limits, all outcomes.} Each method runs on the thirty outlier-free frames of \cref{tab:global_comparison}. Columns count the frames with rotation error below $1\degree$ and $5\degree$, the searches that ended with GOPAC's optimality flag or at GOSMA's queue-size limit, and the stagnated GOPAC searches that we stopped after five to six hours. We score them at their best pose so far. Times are the median and the largest wall time of the finished searches.}
\label{tab:supp_gopac_comparison_all}
\begin{tabular}{llrrrrrrr}
\toprule
Method & Initial bound & Rot. $<1\degree$ $\uparrow$ & Rot. $<5\degree$ $\uparrow$ & Optimality flag & Queue limit & Stopped & Median time (s) $\downarrow$ & Largest time (s) $\downarrow$ \\
\midrule
GOPAC & Zero & 0/30 & 0/30 & 12/30 & 0/30 & 18/30 & 2{,}562 & 4{,}088 \\
GOPAC & Annealed & 0/30 & 1/30 & 12/30 & 0/30 & 18/30 & 3{,}015 & 4{,}770 \\
GOSMA & \textemdash & 0/30 & 1/30 & 0/30 & 30/30 & 0/30 & 68.8 & 293 \\
\method (ours) & \textemdash & \textbf{30/30} & \textbf{30/30} & \textemdash & \textemdash & 0/30 & \textbf{1.92} & \textbf{2.58} \\
\bottomrule
\end{tabular}
\end{table*}

\para{Global search results.}
We evaluate the final pose of each finished search and the best pose so far of each stopped one, in the original coordinates. Timings include program startup, input processing and model preparation. \cref{tab:supp_gopac_comparison_all} reports every outcome. \method\ recovers all thirty frames within $1\degree$, and its largest errors are $0.753\degree$ in rotation and $0.303\%$ in relative translation. Neither global method brings a frame within $1\degree$. GOSMA ends every search at its queue-size limit, after a median of 69~s. In each setting, GOPAC finishes twelve of the thirty searches, on ten ModelNet40 and two MegaDepth frames, and none of the Data61 searches. In the two settings, the finished searches take a median of 43 and 50 minutes. Every finished GOPAC search certifies a pose at which every observation is an angular inlier, with rotation errors from $5.4\degree$ to $179\degree$. The number of angular inliers that GOPAC maximizes therefore does not identify the pose. On a 113-point MegaDepth frame, for example, both settings certify 113 angular inliers at rotation errors of $19.689\degree$ and $8.552\degree$. For these poses, all 113 model points lie in front of the camera, and the maximum nearest-neighbor angular errors are $0.812\degree$ and $0.882\degree$, below the tolerance of $0.974\degree$. An inaccurate pose can therefore make every point an angular inlier. Since every finished search is at least $5.4\degree$ off, the one annealed GOPAC pose within $5\degree$ comes from a stopped search.

\para{Small point sets.}
\label{sec:supp_small_sets}
\cref{thm:pinhole} assumes at least twenty points, so we also evaluate \method\ on smaller complete inputs. For each of $N=8$ and $N=16$, we draw one model whose 3D coordinates are independent and Gaussian with mean zero and standard deviation 0.2. We then apply a fixed pose, project with identity intrinsics and independently shuffle the exact projections. The global methods keep the parameters and normalization above and search all rotations, with the camera center in the fixed box $[-5,5]^3$ in model units. Here, unlike above, GOSMA stops at a 30-second search limit. On both models, the unchanged solver recovers the generating pose within $1\degree$ on every run, with lower median rotation error and runtime than either global method (\cref{tab:small_sets}). Repeated GOPAC runs can return different poses with the same maximal number of angular inliers. GOSMA reaches its search-time limit on both models.

\begin{table}[tp]
\centering\footnotesize
\setlength{\tabcolsep}{2pt}
\caption{\textbf{Small complete point sets.} One noiseless model at each point count $N$. Rotation error and solver time are medians of three runs per method, and $<1\degree$ counts runs with error below $1\degree$. Times cover GOSMA's mixture construction and the complete \method\ solve, but not file input and output, process startup, GPU initialization or compilation. Stop counts runs reaching GOSMA's 30-second search limit, which is checked between iterations. Bold marks the lowest median rotation error and runtime.}
\label{tab:small_sets}
\begin{tabular*}{\columnwidth}{@{\extracolsep{\fill}}rlrrrr@{}}
\toprule
$N$ & Method & Rot. ($\degree$) & $<1\degree$ & Time (s) & Stop \\
\midrule
8 & GOPAC zero & 1.838 & 1/3 & 18.749 & 0/3 \\
8 & GOPAC annealed & 2.469 & 1/3 & 16.263 & 0/3 \\
8 & GOSMA & 165.053 & 0/3 & 30.471 & 3/3 \\
8 & \method (ours) & $\mathbf{<10^{-10}}$ & 3/3 & \textbf{0.100} & 0/3 \\
\midrule
16 & GOPAC zero & 1.866 & 0/3 & 2.589 & 0/3 \\
16 & GOPAC annealed & 1.866 & 0/3 & 1.234 & 0/3 \\
16 & GOSMA & 38.121 & 0/3 & 32.912 & 3/3 \\
16 & \method (ours) & $\mathbf{<10^{-10}}$ & 3/3 & \textbf{0.069} & 0/3 \\
\bottomrule
\end{tabular*}
\end{table}

\subsection{Further datasets}
\label{sec:supp_qualitative}
\cref{tab:real} tests whether \method, with its benchmark settings, recovers poses from structure-from-motion inputs of four further datasets. This subsection describes the reconstructions, image selection and preprocessing for these inputs. It then gives the errors and runtimes per dataset, the comparison with outliers on all images, and examples.

\begin{table*}[tp]
\centering\footnotesize
\setlength{\tabcolsep}{1.5pt}
\caption{\textbf{SfM inputs per dataset}, extending \cref{tab:real}. Images/scenes counts the evaluated images and scenes, and $m$ gives the range of inlier counts per image. For each method, the columns give the percentage of all images with rotation error below $1\degree$ ($\uparrow$), the median rotation error in degrees over returned poses ($\downarrow$), and the median time per image in milliseconds ($\downarrow$). The lower panel adds $\lfloor m/2\rfloor$ uniform outliers to each point set and keeps the images with $m\leq2{,}000$. Times of Deep Blind PnP and BPnPNet include all evaluated pose outputs, and their errors use the selected output. Without outliers, every method returns all 169 poses. With outliers, \method\ returns 125 of 126 poses and the other methods all 126. Bold marks the highest accuracy and the lowest error and time within each dataset and panel.}
\label{tab:real_full}
\begin{tabular*}{\textwidth}{@{\extracolsep{\fill}}lcr*{5}{rrr}@{}}
\toprule
 & & & \multicolumn{3}{c}{Deep Blind PnP} & \multicolumn{3}{c}{BPnPNet} & \multicolumn{3}{c}{GoMatch} & \multicolumn{3}{c}{\shortstack{CFPE\\(reproduced)}} & \multicolumn{3}{c}{\shortstack{\method\\(ours)}} \\
\cmidrule(lr){4-6}\cmidrule(lr){7-9}\cmidrule(lr){10-12}\cmidrule(lr){13-15}\cmidrule(l){16-18}
Dataset & Images/scenes & $m$ & $<1\degree$ & Error & Time & $<1\degree$ & Error & Time & $<1\degree$ & Error & Time & $<1\degree$ & Error & Time & $<1\degree$ & Error & Time \\
\midrule
\multicolumn{18}{l}{\textit{Without outliers}} \\
Cambridge Landmarks~\cite{kendall2015posenet} & 27/4 & 326--2{,}120 & 33 & 88.80 & 351 & 0 & 122.44 & 147 & 15 & 103.63 & \textbf{37} & 81 & 0.539 & 585 & \textbf{100} & \textbf{0.125} & 85 \\
Phototourism~\cite{snavely2006phototourism,jin2021imc} & 36/3 & 469--2{,}607 & 92 & 0.07 & 377 & 19 & 6.80 & 135 & 78 & 0.23 & \textbf{58} & 92 & 0.200 & 941 & \textbf{100} & \textbf{0.006} & 67 \\
7-Scenes~\cite{shotton2013scene} & 36/3 & 128--582 & 69 & 0.53 & 303 & 36 & 21.41 & 103 & 17 & 7.36 & 29 & \textbf{100} & 0.360 & 1{,}029 & \textbf{100} & \textbf{0.134} & \textbf{25} \\
ETH3D~\cite{schops2017eth3d} & 70/6 & 202--5{,}010 & 20 & 94.48 & 355 & 0 & 108.32 & 111 & 1 & 113.42 & 72 & \textbf{100} & 0.015 & 956 & \textbf{100} & \textbf{0.004} & \textbf{66} \\
\cmidrule(lr){1-18}
All & 169/16 & 128--5{,}010 & 48 & 1.23 & 344 & 12 & 82.73 & 122 & 23 & 76.39 & \textbf{42} & 95 & 0.110 & 916 & \textbf{100} & \textbf{0.008} & 60 \\
\midrule
\multicolumn{18}{l}{\textit{With $m/2$ outliers per point set, $m\leq2{,}000$}} \\
Cambridge Landmarks & 26/4 & 326--1{,}928 & 4 & 98.65 & 382 & 0 & 116.95 & 139 & 8 & 116.85 & \textbf{50} & 0 & 34.46 & 1{,}211 & \textbf{100} & \textbf{0.148} & 724 \\
Phototourism & 26/3 & 469--1{,}975 & 69 & 0.31 & 374 & 0 & 15.69 & 121 & 62 & 0.41 & \textbf{84} & 0 & 126.20 & 1{,}877 & \textbf{96} & \textbf{0.016} & 452 \\
7-Scenes & 36/3 & 128--582 & 50 & 1.05 & 326 & 3 & 28.21 & 98 & 6 & 19.65 & \textbf{30} & 0 & 61.63 & 1{,}779 & \textbf{100} & \textbf{0.137} & 213 \\
ETH3D & 38/6 & 202--1{,}875 & 13 & 99.79 & 350 & 0 & 110.73 & 115 & 3 & 118.57 & \textbf{53} & 0 & 50.27 & 1{,}290 & \textbf{87} & \textbf{0.005} & 303 \\
\cmidrule(lr){1-18}
All & 126/16 & 128--1{,}975 & 33 & 45.98 & 352 & 1 & 65.14 & 114 & 17 & 82.32 & \textbf{42} & 0 & 52.25 & 1{,}594 & \textbf{95} & \textbf{0.072} & 380 \\
\bottomrule
\end{tabular*}
\end{table*}

\begin{table*}[tp]
\centering\footnotesize
\setlength{\tabcolsep}{4pt}
\caption{\textbf{SfM inputs with outliers, without the inlier limit.} All 169 images of \cref{tab:real}, with $\lfloor m/2\rfloor$ outliers added to each point set for $m$ inliers. Inlier counts range from 128 to 5{,}010 per image, which gives 192 to 7{,}515 points in each point set. Within $1\degree$ gives the number of images with rotation error below $1\degree$ and their percentage of all 169. OOM means out of GPU memory, and Other exceptions counts the remaining runs without a pose. Times include failed runs and, for Deep Blind PnP and BPnPNet, all evaluated pose outputs. Bold marks the highest recovery and the lowest runtimes.}
\label{tab:real_uncapped}
\begin{tabular*}{\textwidth}{@{\extracolsep{\fill}}lrrrrrrr@{}}
\toprule
 & \multicolumn{2}{c}{Within $1\degree$ ($\uparrow$)} & & \multicolumn{2}{c}{Recorded failures ($\downarrow$)} & \multicolumn{2}{c}{Time per image ($\downarrow$)} \\
\cmidrule(lr){2-3}\cmidrule(lr){5-6}\cmidrule(l){7-8}
Method & Images & \% & Returned/attempted & OOM & Other exceptions & Median (ms) & Maximum (s) \\
\midrule
Deep Blind PnP & 49 & 29.0 & 169/169 & 0 & 0 & 366 & 0.60 \\
BPnPNet & 1 & 0.6 & 169/169 & 0 & 0 & 108 & \textbf{0.22} \\
GoMatch & 26 & 15.4 & 160/169 & 9 & 0 & \textbf{55} & 0.60 \\
CFPE (reproduced) & 0 & 0.0 & 169/169 & 0 & 0 & 1{,}976 & 11.27 \\
\method (ours) & \textbf{147} & \textbf{87.0} & 167/169 & 0 & 2 & 547 & 57.15 \\
\bottomrule
\end{tabular*}
\end{table*}

\para{Reconstructions.}
For Cambridge Landmarks~\cite{kendall2015posenet}, we use the VisualSFM reconstructions~\cite{wu2013visualsfm} of King's College, Old Hospital, Shop Facade and St.\ Mary's Church. For ETH3D~\cite{schops2017eth3d}, we use the released reconstructions of Courtyard, Delivery Area, Electro, Facade, Office and Terrace. For Phototourism, we use the COLMAP reconstructions~\cite{schonberger2016sfm} of the Image Matching Challenge~\cite{jin2021imc} for Trevi Fountain, Brandenburg Gate and Sacr\'e-C\oe ur. For 7-Scenes~\cite{shotton2013scene}, we use the depth-assisted SfM models of Sarlin \etal~\cite{sarlin2021pixloc} for Heads, Fire and Chess.

\para{Queries.}
Each query contains the tracked 2D points of one image and the model points observed in that image. We undistort the 2D points, keep the tracks with positive depth and reprojection error below three pixels, and shuffle both point sets. On Cambridge Landmarks, we take the four test images with the most points in each scene and eleven further test images spread over the rest. On the other datasets, we take up to twelve landscape images per scene with 100 to 3{,}000 points, or up to 4{,}000 points on 7-Scenes and 6{,}000 on ETH3D. With outliers, we keep the images with at most 2{,}000 inliers and add half as many outliers as inliers to each point set, as in the benchmark protocol. The learned methods keep their MegaDepth weights, and we report the released output with the most poses within $1\degree$.

\para{Results.}
\cref{tab:real_full} extends \cref{tab:real} with the median rotation error and runtime of each method on each dataset. \cref{tab:real_uncapped} removes the limit of 2{,}000 inliers and evaluates all 169 images with outliers. On these inputs, \method\ returns 167 poses, 147 of them within $1\degree$. GoMatch runs out of memory on nine images, and the other baselines return a pose for every image.

\begin{figure*}[tp]
\centering
\resizebox{0.96\textwidth}{!}{%
\includegraphics[width=495.43903pt,height=604.00307pt]{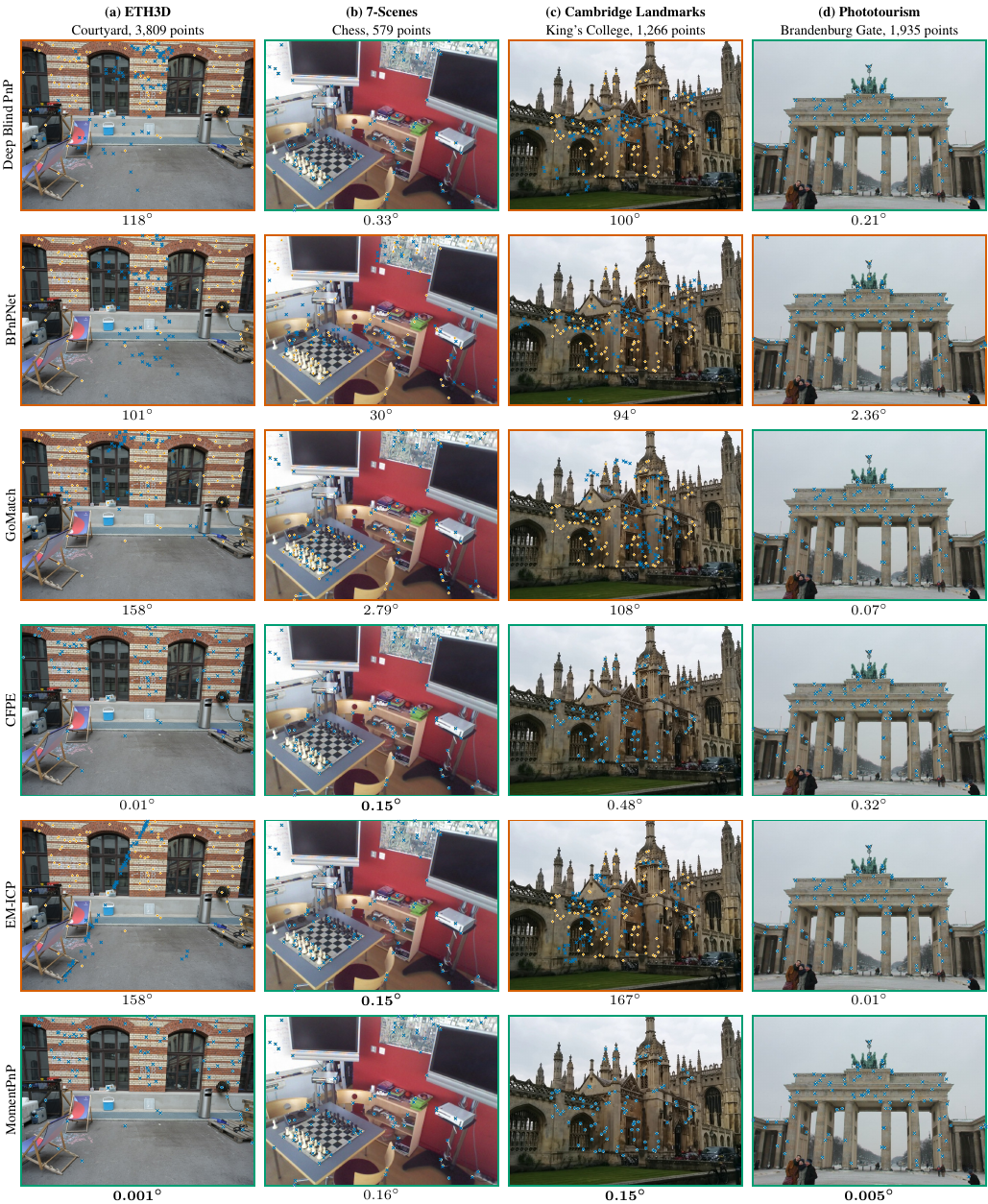}%
}
\caption{\textbf{Qualitative results on SfM inputs.} Each dataset is shown in its first selected outlier-free image, with one row per method. Orange dots are the tracked 2D points, and blue crosses are the model points projected under the returned pose. Titles give the scene and the point count. Labels give rotation error, with the lowest in bold, and green borders mark errors below $1\degree$.}
\label{fig:gallery}
\end{figure*}

\begin{figure*}[tp]
\centering
\includegraphics[width=494.77283pt,height=252.49762pt]{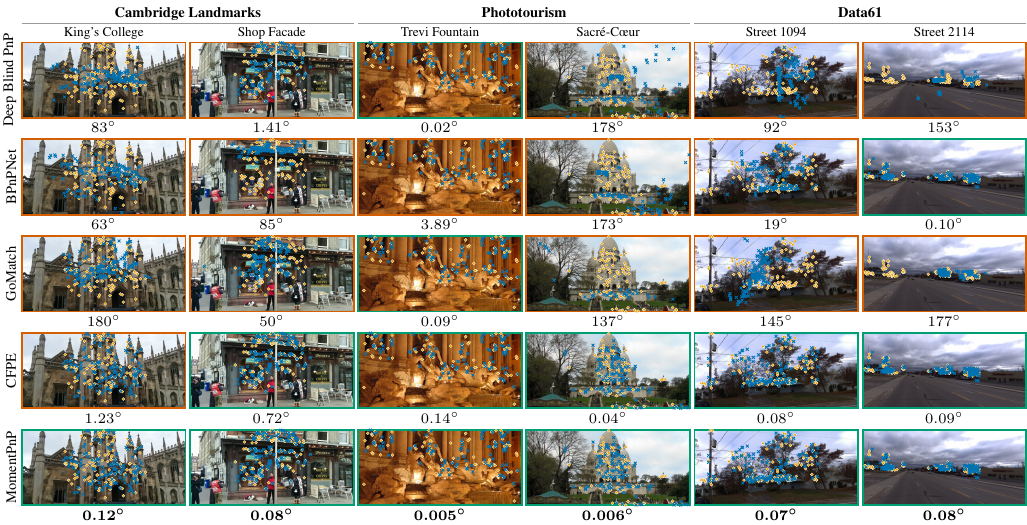}
\caption{\textbf{Qualitative results on outdoor scenes} from Cambridge Landmarks, Phototourism and Data61. Orange dots are the 2D points, and blue crosses are the model points projected under the returned pose. Labels give rotation error, with the lowest in bold, and green borders mark errors below $1\degree$. Data61 model points come from LiDAR.}
\label{fig:gallery_outdoor}
\end{figure*}

\begin{figure*}[tp]
\centering
\includegraphics[width=494.7734pt,height=252.49762pt]{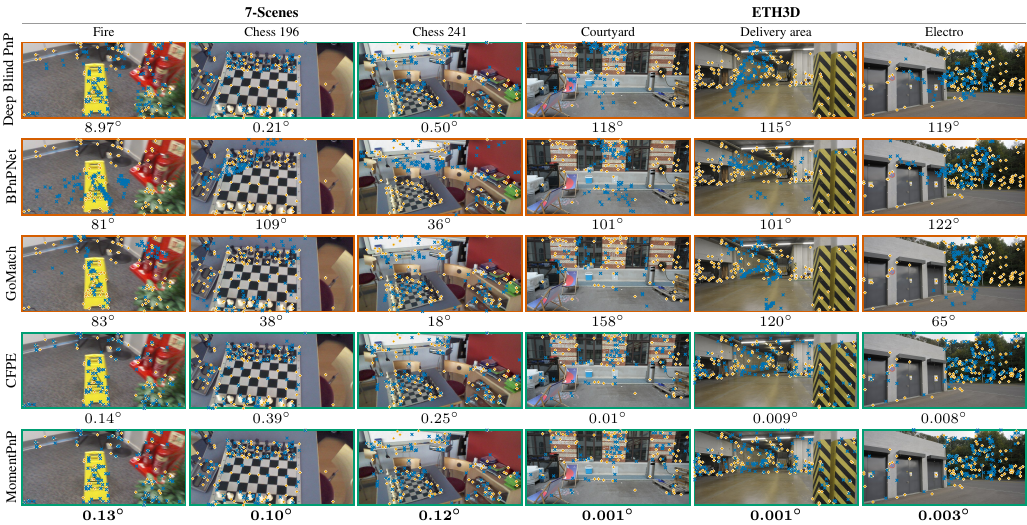}
\caption{\textbf{Qualitative results on 7-Scenes and ETH3D.} Orange dots are the 2D points, and blue crosses are the model points projected under the returned pose. Labels give rotation error, with the lowest in bold, and green borders mark errors below $1\degree$.}
\label{fig:gallery_scenes}
\end{figure*}

\para{Examples.}
\cref{fig:gallery} shows the first selected outlier-free image of each dataset for every method. \cref{fig:gallery_outdoor} adds images from Cambridge Landmarks, Phototourism and Data61, and \cref{fig:gallery_scenes} from 7-Scenes and ETH3D. \method\ is within $1\degree$ on every image and has the lowest rotation error on all but one. On the Chess image of \cref{fig:gallery}, CFPE and EM-ICP are $0.01\degree$ lower. In each figure, exactly two images have a pose within $1\degree$ from at least one learned method.

\subsection{Scaling to large point sets}
\label{sec:supp_scaling}
This subsection gives the protocol behind \cref{fig:scaling,fig:scale_gallery} and the results per scan and per view up to one million points. It also reports the percentage of correct labels, the runtimes and the results of the learned baselines.

\para{Rendered 2D points.}
The experiment varies the number of points, and the fraction of model points that generate 2D points, on real scene geometry. We use the ETH3D laser scans~\cite{schops2017eth3d} of Facade, Courtyard and Office, which hold 45.0, 37.8 and 48.6 million scan points. For each scan, three registered cameras give three reference poses, and 2.6 to 10.6 million scan points fall within the view frustum of a reference pose. Within this frustum, we sample nested subsets of the scan points with positive depth as model points and project them to form the 2D points, ignoring occlusion. We add independent isotropic Gaussian noise with standard deviation $\noise=1$ pixel per coordinate in a $6{,}200\times4{,}100$ image and shuffle both point sets. In a second setting, with partial detection, a uniformly sampled quarter of the model points generates 2D points. Every method receives the same point sets, and the learned baselines use their MegaDepth weights.

Times cover complete inference runs and exclude input and model loading. \method\ runs a small GPU warm-up before timing. The times of the learned methods include compilation on first use where it is required. The Courtyard and Office runs have a limit of 1{,}800 seconds per view and point count.

\begin{table}[tp]
\centering\footnotesize
\setlength{\tabcolsep}{1.9pt}
\caption{\textbf{Rendered 2D points on ETH3D scans (ours).} Medians, over the three reference poses of each scan, of time per frame in seconds ($\downarrow$), rotation error in thousandths of a degree ($\downarrow$), and Labels, the percentage of 2D points whose nearest projected model point generated them ($\uparrow$). Points counts the model points. In the upper block, every model point generates a 2D point, and in the lower block a uniformly sampled quarter does.}
\label{tab:render}
\begin{tabular*}{\columnwidth}{@{\extracolsep{\fill}}rrrrrrrrrr@{}}
\toprule
 & \multicolumn{3}{c}{Facade} & \multicolumn{3}{c}{Courtyard} & \multicolumn{3}{c}{Office} \\
\cmidrule(lr){2-4}\cmidrule(lr){5-7}\cmidrule(l){8-10}
Points & Time & Error & Labels & Time & Error & Labels & Time & Error & Labels \\
\midrule
\multicolumn{10}{@{}l}{\emph{Every model point has a 2D point}}\\
1{,}000 & 0.09 & 1.5 & 100 & 0.14 & 1.9 & 100 & 0.14 & 2.1 & 100 \\
4{,}000 & 0.28 & 0.62 & 100 & 0.46 & 1.2 & 100 & 0.39 & 1.2 & 100 \\
16{,}000 & 0.97 & 0.30 & 99 & 2.4 & 0.43 & 100 & 1.9 & 0.81 & 99 \\
64{,}000 & 6.6 & 0.21 & 98 & 22.7 & 0.22 & 98 & 18.6 & 0.21 & 98 \\
256{,}000 & 83.8 & 0.044 & 93 & 355 & 0.12 & 93 & 302 & 0.091 & 92 \\
1{,}000{,}000 & 1{,}606 & 0.066 & 79 & -- & -- & -- & 990 & 0.11 & 76 \\
\addlinespace
\multicolumn{10}{@{}l}{\emph{A quarter of the model points have a 2D point}}\\
1{,}000 & 0.52 & 2.7 & 100 & 0.59 & 4.2 & 100 & 0.58 & 3.4 & 100 \\
4{,}000 & 1.7 & 1.9 & 100 & 2.0 & 2.4 & 100 & 1.9 & 3.7 & 100 \\
16{,}000 & 7.4 & 0.96 & 100 & 8.8 & 0.80 & 100 & 7.8 & 1.2 & 99 \\
64{,}000 & 59.1 & 0.59 & 98 & 69.6 & 0.40 & 98 & 50.5 & 0.58 & 98 \\
\bottomrule
\end{tabular*}
\end{table}

\begin{figure*}[tp]
\centering
\pgfplotsset{mpRecoveryAxis/.style={mpPlot, scale only axis, width=0.36\textwidth, height=0.205\textwidth,
  xlabel near ticks, ylabel near ticks, ylabel shift=-3pt, xlabel shift=-2pt,
  title style={at={(axis description cs:0,1.04)}, anchor=south west,
    font=\footnotesize\bfseries, inner sep=0pt, yshift=-2pt},
  legend columns=3,
  legend style={at={(0.5,-0.36)}, anchor=north, inner sep=1pt,
    row sep=0pt, /tikz/column sep=0.9em}},
  mpFacade/.style={colOurs, line width=0.9pt, mark=*, mark size=1.1pt},
  mpCourtyard/.style={colOurs, line width=0.9pt, mark=square*, mark size=1.1pt, mark options={fill=white}},
  mpOffice/.style={colOurs, line width=0.9pt, mark=triangle*, mark size=1.3pt, mark options={fill=colOurs!35}},
  mpLiu/.style={colLiu, line width=0.65pt, mark=square*, mark size=1.1pt},
  mpBPnP/.style={colBPnP, line width=0.65pt, mark=triangle*, mark size=1.3pt},
  mpGoMatch/.style={oiPurple, line width=0.65pt, mark=diamond*, mark size=1.3pt},
  mpMoment/.style={colOurs, line width=0.9pt, mark=*, mark size=1.2pt},
  mpCFPE/.style={oiOrange, line width=0.65pt, mark=star, mark size=1.5pt},
  mpEM/.style={black!80, line width=0.65pt, mark=pentagon*, mark size=1.3pt}}
\includegraphics[width=477.35905pt,height=174.96912pt]{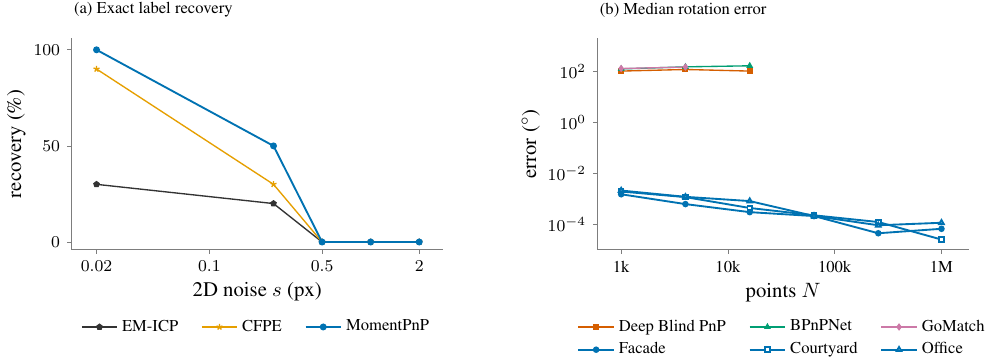}
\caption{\textbf{Exact label recovery and pose error per scan.} (a)~Percentage of synthetic frames with $N=1024$ in which every 2D point is labeled with its generating model point, against the noise standard deviation $\noise$ (\cref{sec:supp_ablation}). (b)~Median rotation error on rendered 2D points (\cref{sec:supp_scaling}). Blue curves give \method\ on each scan, and the learned baselines use their returned poses from all nine views.}
\label{fig:recovery_supp}
\end{figure*}

\para{Pose accuracy.}
\cref{tab:render} gives the median rotation error, percentage of correct labels and time over the three reference poses of each scan. \cref{fig:recovery_supp}(b) plots the median rotation error of \method\ on each scan, together with the learned baselines over all nine views. At one million points, all three Facade runs return poses, and one of them takes 8{,}735 seconds. For Courtyard, one view contains repeated 3D coordinates at this size and is excluded, and the solver returns no pose for a second view within the 1{,}800-second limit. The Courtyard markers at one million points thus show a single pose. \cref{tab:render} lists one million points for Facade and Office, and \cref{tab:courtyard_office_scaling} gives every Courtyard and Office view through one million points. \cref{tab:million_examples} gives one view of each scan at one million points, and \cref{fig:scale_gallery} shows these three views. For Facade~1, the estimated projections differ from the reference projections by at most \ScaleMillionShift\ pixels. With partial detection, every pose up to 64{,}000 points has rotation error below $0.006\degree$.

\begin{table}[tp]
\centering\footnotesize
\caption{\textbf{One view of each scan at one million points (ours).} Each view has 1{,}000{,}000 distinct model points and as many 2D points. \cref{fig:scale_gallery} shows these views. Relative translation is the translation error divided by the reference translation norm, and Labels is the percentage of 2D points whose nearest projected model point generated them.}
\label{tab:million_examples}
\begin{tabular*}{\columnwidth}{@{\extracolsep{\fill}}lrrr@{}}
\toprule
View & Rotation ($\degree$)\,$\downarrow$ & Rel. translation\,$\downarrow$ & Labels (\%)\,$\uparrow$ \\
\midrule
Facade 1 & $5.3705\times10^{-5}$ & $1.1990\times10^{-6}$ & 78.0767 \\
Courtyard 3 & $2.4968\times10^{-5}$ & $4.5102\times10^{-7}$ & 77.4822 \\
Office 1 & $4.8244\times10^{-5}$ & $3.1206\times10^{-7}$ & 77.6801 \\
\bottomrule
\end{tabular*}
\end{table}

\begin{table*}[tp]
\centering\footnotesize
\setlength{\tabcolsep}{3pt}
\caption{\textbf{Every Courtyard and Office view (ours).} Rotation error in $10^{-4}$ degrees and relative translation error in $10^{-6}$ for each view and point count, and at one million points the solve time and Labels as in \cref{tab:million_examples}. Invalid inputs contain repeated 3D coordinates and are excluded. Timeout marks a run that returned no pose within the 1{,}800-second limit.}
\label{tab:courtyard_office_scaling}
\begin{tabular*}{\textwidth}{@{\extracolsep{\fill}}lrrrrrrrr@{}}
\toprule
 & \multicolumn{3}{c}{Rotation ($10^{-4}$ deg)\,$\downarrow$} & \multicolumn{3}{c}{Relative translation ($10^{-6}$)\,$\downarrow$} & \multicolumn{2}{c}{1M points} \\
\cmidrule(lr){2-4}\cmidrule(lr){5-7}\cmidrule(l){8-9}
View & 64k & 256k & 1M & 64k & 256k & 1M & Solve (s)\,$\downarrow$ & Labels (\%)\,$\uparrow$ \\
\midrule
Courtyard 1 & 2.2351 & 2.1025 & Invalid & 2.3367 & 2.9767 & -- & -- & -- \\
Courtyard 2 & 1.2694 & 1.0114 & Timeout & 1.3939 & 1.1487 & -- & -- & -- \\
Courtyard 3 & 2.8570 & 1.2198 & 0.2497 & 2.4586 & 1.7281 & 0.4510 & 688.7 & 77.48 \\
Office 1 & 2.0979 & 2.4092 & 0.4824 & 8.0690 & 5.6997 & 0.3121 & 985.1 & 77.68 \\
Office 2 & 2.6994 & 0.9101 & 1.1428 & 7.5577 & 3.8684 & 3.7290 & 1243.7 & 76.32 \\
Office 3 & 1.8078 & 0.8665 & 1.4484 & 7.0205 & 5.3789 & 8.0800 & 990.0 & 75.35 \\
\bottomrule
\end{tabular*}
\end{table*}

\begin{figure}[tp]
\centering
\includegraphics[width=213.63068pt,height=142.85606pt]{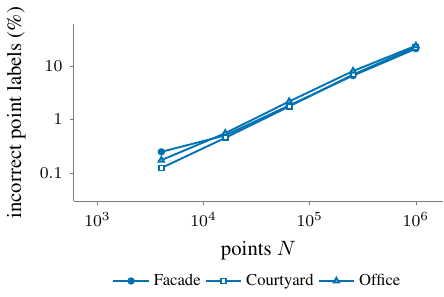}
\caption{\textbf{Incorrect labels on rendered 2D points.} Median percentage, over the reference poses of each scan, of 2D points whose nearest projected model point under the recovered pose did not generate them.}
\label{fig:scaling_label}
\end{figure}

\para{Correspondences.}
Each 2D point is labeled with its nearest projected model point. When the noise is small compared with the spacing between nearby projections, nearly every label is correct. This spacing shrinks as the point sets grow, so more 2D points lie nearer to another projection than to their own. \cref{fig:scaling_label} plots the median percentage of 2D points whose nearest projected model point under the recovered pose did not generate them. The Labels columns of \cref{tab:render,tab:million_examples,tab:courtyard_office_scaling} give the percentage that did. The percentage of correct labels falls from 100\% at 1{,}000 points to 75 to 79\% at one million. Over the same range, the rotation error decreases up to 256{,}000 points and stays near that level at one million (\cref{tab:render}). \cref{sec:supp_proof_threshold} relates these percentages to the probability that every label is correct.

\begin{figure}[!tp]
\centering
\includegraphics[width=236.91228pt,height=155.06216pt]{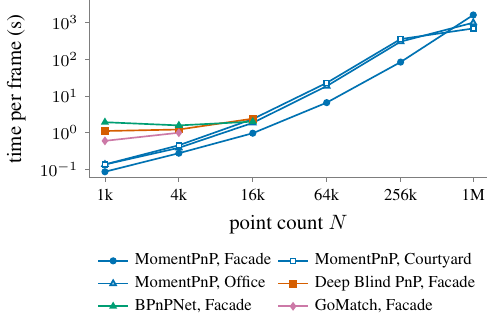}
\caption{\textbf{Inference time on rendered 2D points}~\cite{schops2017eth3d}. Median time per frame over the reference poses of each scan for \method, and over the three Facade poses for the learned baselines. Their curves end where they run out of memory (\cref{tab:render_baselines}).}
\label{fig:scaling-time}
\end{figure}

\para{Runtime.}
\cref{fig:scaling-time} gives the median time per frame of \method\ on each scan and compares it with the learned baselines. Across the scans, the median time of \method\ is 0.09 to 0.14~s at 1{,}000 points and 84 to 355~s at 256{,}000 points (\cref{tab:render}).

\para{Learned baselines.}
\cref{tab:render_baselines} counts the poses of the learned baselines within $1\degree$ over the nine views. GoMatch runs out of memory from 16{,}000 points and the other learned baselines from 64{,}000 points. Below these sizes, they bring at most two of the nine poses within $1\degree$. Their curves in \cref{fig:scaling-time} use the three Facade poses.

\begin{table}[tp]
\centering\footnotesize
\caption{\textbf{Learned baselines on rendered 2D points.} Poses with rotation error below $1\degree$ over the nine views of \cref{tab:render}. OOM marks point counts at which a method runs out of memory and returns no pose. Bold marks the highest count at each point count.}
\label{tab:render_baselines}
\begin{tabular*}{\columnwidth}{@{\extracolsep{\fill}}rcccc@{}}
\toprule
 & \multicolumn{4}{c}{Recovered poses $\uparrow$} \\
\cmidrule(l){2-5}
Points & Deep Blind PnP & BPnPNet & GoMatch & \shortstack{\method\\(ours)} \\
\midrule
1{,}000 & 2/9 & 0/9 & 0/9 & \textbf{9/9} \\
4{,}000 & 0/9 & 0/9 & 0/9 & \textbf{9/9} \\
16{,}000 & 0/9 & 0/9 & OOM & \textbf{9/9} \\
64{,}000 & OOM & OOM & OOM & \textbf{9/9} \\
256{,}000 & OOM & OOM & OOM & \textbf{9/9} \\
\bottomrule
\end{tabular*}
\end{table}

\subsection{Complete scene reconstructions}
\label{sec:supp_whole_scene}
We evaluate \method\ on observations in one image and the complete 3D reconstruction. Neither the visible model points nor a pose prior is given. \cref{sec:supp_camera_localization} specifies the camera search and the refinement by EM. Here we give the query protocol, the results of \cref{tab:scene} and the search cost.

\para{Queries.}
We use the thirteen ETH3D high-resolution training scenes of \cref{tab:scene}. Each query contains the tracked 2D points of one image and every finite model point of the released sparse reconstruction~\cite{schops2017eth3d}, which starts from COLMAP~\cite{schonberger2016sfm}, with both point sets shuffled. We exclude image detections without a triangulated track. In each scene, the tuning queries are the first three images in file-name order with at least 200 tracked 2D points, and we fix the settings on them. We test the settings on the test queries, three further images spaced evenly over the remaining eligible images. A query is recovered when the returned pose has rotation error below $1\degree$. \cref{sec:supp_camera_localization} specifies the camera search and its settings.

\begin{table}[tp]
\centering\footnotesize
\setlength{\tabcolsep}{2.5pt}
\caption{\textbf{Recovery from full scene reconstructions on ETH3D.} Each query contains the tracked 2D points of one image and every finite SfM model point of the scene, without the visible subset or a pose. Points is the number of model points, and Ratio is the scene extent divided by the viewing distance. Recovery columns count queries with rotation error below $1\degree$. The learned baselines and Tuning use the three queries on which we chose the settings, and Test uses three held-out queries with these settings fixed. DBP, BPnP and GoM denote Deep Blind PnP, BPnPNet and GoMatch, and OOM means out of memory. Bold marks the highest recovery on the tuning queries in each scene.}
\label{tab:scene}
\begin{tabular*}{\columnwidth}{@{\extracolsep{\fill}}lrrccccc@{}}
\toprule
 & & & \multicolumn{3}{c}{Recovery $\uparrow$} & \multicolumn{2}{c}{\shortstack{\method\\(ours) $\uparrow$}} \\
\cmidrule(lr){4-6}\cmidrule(l){7-8}
Scene & Points & Ratio & DBP & BPnP & GoM & Tuning & Test \\
\midrule
Kicker & 15{,}897 & 2.7 & 0/3 & 0/3 & OOM & \textbf{3/3} & 3/3 \\
Office & 3{,}461 & 3.4 & 2/3 & 0/3 & 0/3 & \textbf{3/3} & 3/3 \\
Facade & 85{,}096 & 4.8 & OOM & 0/3 & OOM & \textbf{3/3} & 3/3 \\
Terrace & 10{,}676 & 5.3 & 0/3 & 0/3 & OOM & \textbf{3/3} & 3/3 \\
Pipes & 2{,}473 & 5.4 & 0/3 & 0/3 & 0/3 & \textbf{3/3} & 3/3 \\
Relief & 26{,}516 & 5.6 & 0/3 & 0/3 & OOM & \textbf{3/3} & 3/3 \\
Courtyard & 33{,}487 & 7.1 & 0/3 & 0/3 & OOM & \textbf{3/3} & 3/3 \\
Relief 2 & 20{,}514 & 9.3 & 0/3 & 0/3 & OOM & \textbf{3/3} & 3/3 \\
Meadow & 2{,}194 & 10.8 & 0/3 & 0/3 & 0/3 & \textbf{3/3} & 2/3 \\
Delivery Area & 31{,}978 & 13.2 & 0/3 & 0/3 & OOM & \textbf{3/3} & 3/3 \\
Electro & 20{,}268 & 14.6 & 0/3 & 0/3 & OOM & \textbf{3/3} & 1/3 \\
Terrains & 18{,}553 & 17.9 & 0/3 & 0/3 & OOM & \textbf{3/3} & 3/3 \\
Playground & 15{,}663 & 40.3 & 0/3 & 0/3 & OOM & \textbf{3/3} & 2/3 \\
\midrule
All & & & 2/39 & 0/39 & 0/39 & \textbf{39/39} & 35/39 \\
\bottomrule
\end{tabular*}
\end{table}

\begin{figure}[tp]
\centering
\includegraphics[width=209.25372pt,height=129.53297pt]{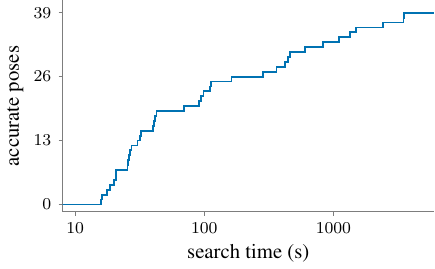}
\caption{\textbf{Search time on complete scene reconstructions.} Number of the 39 tuning queries of \cref{tab:scene} whose search has returned a pose within $1\degree$ by a given search time.}
\label{fig:scaling_supp}
\end{figure}

\para{Results.}
\cref{tab:scene} gives the recovery in each scene for the tuning and test queries, and for the learned baselines on the tuning queries. \method\ recovers all 39 tuning queries within $1\degree$, with rotation errors of at most \SceneMaxErr, and 35 of the 39 test queries. \cref{fig:scaling_supp} plots the number of tuning queries recovered by each search time. The search recovers 19 of the 39 within 43~s and the last after 3{,}531~s.

\para{Cost of the search.}
The number of grid centers that the search must visit sets its cost. Let $d_{\rm v}>0$ be the viewing distance, the depth of the observed part of the scene. Suppose that the true camera center lies in the enlarged box of \cref{sec:supp_camera_localization} and that, for some fraction $r_{\rm c}>0$, refinement succeeds from any grid center within $r_{\rm c}d_{\rm v}$ of it. Suppose also that the rotation grid covers the basin of convergence. By \cref{lem:coverage}, cells of side $\Delta_{\mathrm c}=2r_{\rm c}d_{\rm v}/\sqrt3$ place some grid center within $r_{\rm c}d_{\rm v}$ of the true camera center, and the grid has at most $\prod_k(\sqrt3L_k/(2r_{\rm c}d_{\rm v})+1)$ centers, where $L_k$ are the side lengths of the enlarged box. The bound grows roughly as the cube of the ratio of scene extent to viewing distance. Recovery also requires that the search keep this grid center after ranking. At each center, scoring the 11{,}520 rotations takes work linear in the number of model points in the coarse sample, which is at most 65{,}536. Beyond the three full grids, the search subdivides only the best cells. \SceneCellRankSentence\ For evaluation, the viewing distance is the median depth of the model points observed in the query image, and the Ratio column of \cref{tab:scene} divides the scene extent by it.

\subsection{Ablation study}
\label{sec:supp_ablation}
This subsection gives the protocols and results of the controls in the ablation study of \cref{sec:experiments}, which vary the depths of the model points, the image noise and the kind of outliers. It then reports three further controls, on the search implementation and the pose ambiguity resolution, on the selection of visible points and on the acceptance test on held-out frames.

\begin{table*}[tp]
\centering\small
\caption{\textbf{Depth variation with complete point sets.} Ten paired trials per condition, each with 128 points and Gaussian noise of standard deviation $\noise=2$ pixels per coordinate at a focal length of 800 pixels. Depths vary about $\refdepth=4.5$. The second column gives the median over trials of the ratio $x_{3,\max}/x_{3,\min}$ of the largest to the smallest camera-frame depth $x_{j3}=(\Rot\vect y_j+\trans)_3$. Error is the median rotation error and Time the median total runtime. Affine initialization returns the candidate with the lowest moment residual, before perspective fitting and EM. The full \method\ completes its search and refinement. Bold marks the lower error and runtime in each row.}
\label{tab:depth}
\begin{tabular*}{\textwidth}{@{\extracolsep{\fill}}lrrrrrr@{}}
\toprule
& & \multicolumn{2}{c}{Affine initialization (ours)} & \multicolumn{3}{c}{Full \method (ours)} \\
\cmidrule(lr){3-4}\cmidrule(l){5-7}
Depth condition & $x_{3,\max}/x_{3,\min}$ & Error ($\degree$)\,$\downarrow$ & Time (ms)\,$\downarrow$ & $<1\degree$\,$\uparrow$ & Error ($\degree$)\,$\downarrow$ & Time (ms)\,$\downarrow$ \\
\midrule
Spread $f_{\rm sp}=0.02$ & 1.04 & 20.88 & \textbf{51} & 8/10 & \textbf{0.374} & 108 \\
Spread $f_{\rm sp}=0.1$ & 1.22 & 15.43 & \textbf{51} & 10/10 & \textbf{0.227} & 109 \\
Spread $f_{\rm sp}=0.3$ & 1.85 & 12.88 & \textbf{51} & 10/10 & \textbf{0.089} & 108 \\
Spread $f_{\rm sp}=0.6$ & 3.95 & 156.55 & \textbf{50} & 10/10 & \textbf{0.062} & 110 \\
Spread $f_{\rm sp}=0.9$ & 18.04 & 161.26 & \textbf{51} & 10/10 & \textbf{0.051} & 109 \\
13 close points at $0.8\refdepth$ & 1.25 & 22.30 & \textbf{50} & 10/10 & \textbf{0.182} & 109 \\
13 close points at $0.5\refdepth$ & 2.00 & 14.99 & \textbf{51} & 10/10 & \textbf{0.072} & 110 \\
13 close points at $0.2\refdepth$ & 5.00 & 9.49 & \textbf{50} & 10/10 & \textbf{0.047} & 112 \\
13 close points at $0.1\refdepth$ & 10.00 & 8.76 & \textbf{50} & 10/10 & \textbf{0.046} & 112 \\
\bottomrule
\end{tabular*}
\end{table*}

\para{Depth variation.}
The affine initial poses of Step 1 assume an affine camera (\cref{thm:identifiability}), and a perspective camera departs further from an affine camera as the depths of the model points spread. In each paired trial of \cref{tab:depth}, we vary the depths about a reference depth $\refdepth=4.5$ while keeping the noise-free image bearings and the noise realization fixed. In the first set of conditions, the depths are $\refdepth(1+f_{\rm sp}q_{{\rm sp},j})$ for fixed $q_{{\rm sp},j}\in[-1,1]$ and a spread factor $f_{\rm sp}$. In the second, 13 of the 128 points move closer to the camera and the other depths stay at $\refdepth$. We draw calibrated image coordinates independently from scaled and shifted Beta distributions. The first set draws $q_{{\rm sp},j}$ uniformly from $[-1,1]$ and uses $f_{\rm sp}\in\{0.02,0.1,0.3,0.6,0.9\}$. The second places the same 13 points at $0.8$, $0.5$, $0.2$ and $0.1$ times $\refdepth$ in separate trials. The affine initialization in this study tests 3{,}000 viewing directions with 72 rolls each, keeps the affine map with the lowest residual and converts it to an initial pose under scaled orthography, without perspective fitting or EM.
The affine initial pose has rotation error above $1\degree$ in every trial. After perspective fitting and EM, the full solver is within $1\degree$ in 88 of the 90 trials. At the largest spread, with a median depth ratio of 18, the affine initial pose has a median error of $161\degree$, and the full solver is within $1\degree$ in all ten. As the spread shrinks, the median error of the full solver rises from $0.05\degree$ to $0.37\degree$. Its two inaccurate poses, at $1.0\degree$ and $1.3\degree$, are accepted and occur at the smallest spread.

\begin{figure*}[tp]
\centering
\begin{minipage}[b]{\textwidth}\centering
{\footnotesize\begin{tikzpicture}[baseline=-0.6ex]
  \draw[oiBlue, line width=0.9pt] (0.00,0) -- (0.34,0);
  \draw[oiBlue] plot[mark=*, mark size=1.2pt] coordinates {(0.17,0)};
  \node[anchor=west, inner sep=0.5pt] at (0.40,0) {$N{=}16$};
  \draw[oiVermilion, line width=0.9pt] (1.32,0) -- (1.66,0);
  \draw[oiVermilion] plot[mark=square*, mark size=1.1pt] coordinates {(1.49,0)};
  \node[anchor=west, inner sep=0.5pt] at (1.72,0) {$N{=}32$};
  \draw[oiGreen, line width=0.9pt] (2.64,0) -- (2.98,0);
  \draw[oiGreen] plot[mark=triangle*, mark size=1.4pt] coordinates {(2.81,0)};
  \node[anchor=west, inner sep=0.5pt] at (3.04,0) {$N{=}64$};
  \draw[oiPurple, line width=0.9pt] (3.96,0) -- (4.30,0);
  \draw[oiPurple] plot[mark=diamond*, mark size=1.4pt] coordinates {(4.13,0)};
  \node[anchor=west, inner sep=0.5pt] at (4.36,0) {$N{=}128$};
  \draw[oiOrange, line width=0.9pt] (5.44,0) -- (5.78,0);
  \draw[oiOrange] plot[mark=pentagon*, mark size=1.3pt] coordinates {(5.61,0)};
  \node[anchor=west, inner sep=0.5pt] at (5.84,0) {$N{=}256$};
  \draw[oiSky, line width=0.9pt] (6.91,0) -- (7.25,0);
  \draw[oiSky] plot[mark=otimes*, mark size=1.2pt] coordinates {(7.08,0)};
  \node[anchor=west, inner sep=0.5pt] at (7.31,0) {$N{=}1024$};
  \draw[black] plot[mark=x, mark size=2pt] coordinates {(8.71,0)};
  \node[anchor=west, inner sep=0.5pt] at (8.94,0) {reference};
\end{tikzpicture}}\\[2pt]
\includegraphics[width=433.79184pt,height=134.99141pt]{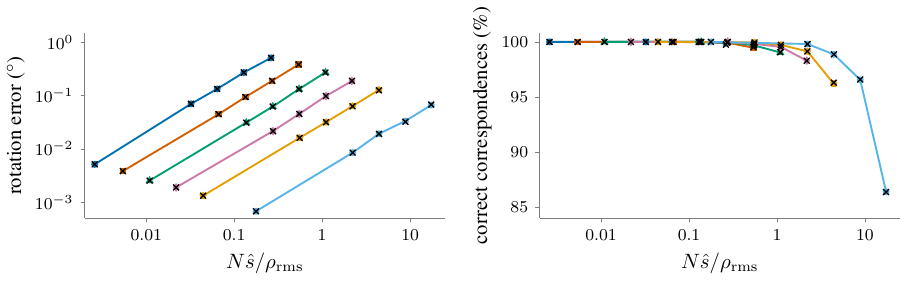}\\[1pt]
{\footnotesize (a) Increasing noise: rotation error (left) and correct correspondences (right)}
\end{minipage}\\[8pt]
\begin{minipage}[b]{0.48\textwidth}\centering
{\footnotesize\begin{tikzpicture}[baseline=-0.6ex]
  \draw[colOurs, line width=0.9pt] (0.00,0) -- node[midway, circle, fill=colOurs, draw=colOurs, minimum size=2.4pt, inner sep=0pt] {} (0.34,0);
  \node[anchor=west, inner sep=0.5pt] at (0.40,0) {\method};
  \draw[colBPnP, line width=0.65pt] (2.09,0) -- node[midway, circle, fill=white, draw=colBPnP, minimum size=2.4pt, inner sep=0pt] {} (2.43,0);
  \node[anchor=west, inner sep=0.5pt] at (2.49,0) {BPnPNet};
\end{tikzpicture}}\\[0pt]
{\footnotesize\begin{tikzpicture}[baseline=-0.6ex]
  \draw[oiGrey] plot[mark=*, mark size=1.2pt] coordinates {(0.17,0)};
  \node[anchor=west, inner sep=0.5pt] at (0.40,0) {$N{=}48$};
  \draw[oiGrey] plot[mark=square*, mark size=1.1pt] coordinates {(1.49,0)};
  \node[anchor=west, inner sep=0.5pt] at (1.72,0) {$N{=}100$};
  \draw[oiGrey] plot[mark=triangle*, mark size=1.4pt] coordinates {(2.96,0)};
  \node[anchor=west, inner sep=0.5pt] at (3.19,0) {$N{=}200$};
  \draw[oiGrey] plot[mark=diamond*, mark size=1.4pt] coordinates {(4.44,0)};
  \node[anchor=west, inner sep=0.5pt] at (4.67,0) {$N{=}500$};
\end{tikzpicture}}\\[2pt]
\includegraphics[width=209.74715pt,height=119.12138pt]{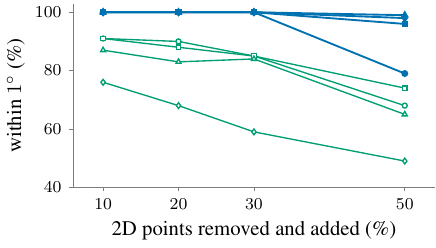}\\[1pt]
{\footnotesize (b) Removed \& added 2D points}
\end{minipage}\hfill
\begin{minipage}[b]{0.48\textwidth}\centering
{\footnotesize
\includegraphics[width=215.16028pt,height=149.03143pt]{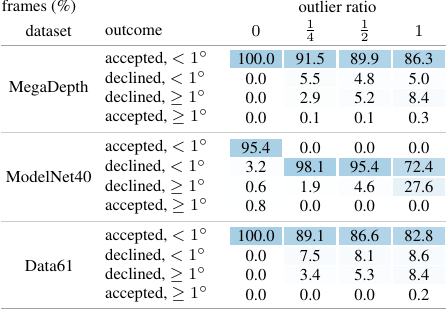}}\\[1pt]
{\footnotesize (c) Acceptance outcomes by outlier ratio}
\end{minipage}
\caption{\textbf{Noise, structured outliers and acceptance.} (a)~Rotation error (left) and percentage of correct labels (right) of \method\ on synthetic inputs against $N\noisefit/\rmsrad$, one curve per point count $N$. Crosses mark the pose fitted to the true correspondences (left) and minimum-cost bijective matching at the true pose (right). (b)~Percentage of Data61~\cite{namin2015data61} frames with rotation error below $1\degree$ for \method\ and BPnPNet, against the percentage of 2D points replaced by other 2D points on the scene, one curve per $N$. (c)~Percentage of the 1{,}000 held-out frames of MegaDepth~\cite{li2018megadepth}, ModelNet40~\cite{wu2015shapenets} and Data61 at each outlier ratio, split jointly by acceptance and by rotation error below $1\degree$.}
\label{fig:supp_degradation}
\end{figure*}

\para{Image noise.}
These synthetic controls measure how image noise affects the pose and the labels. We draw $N$ model points independently and uniformly from $[-0.1,0.1]^3$. The rotation axis is uniform on the unit sphere, and the rotation angle is uniform in $[0,\pi]$. The translation is $(t_x,t_y,1)^\top$ with $t_x$ and $t_y$ independent and uniform in $[-0.15,0.15]$. The camera has focal lengths of 1{,}400 pixels and principal point $(640,512)$ in a $1280\times1024$ image. We add independent isotropic Gaussian noise with standard deviation $\noise\in\{0.02,0.25,0.5,1,2\}$ pixels per coordinate. Each noise level uses 200 trials for $N\in\{16,32,64\}$, 100 for $N\in\{128,256\}$ and 40 for $N=1024$. In \cref{thm:threshold}, exact matching depends on the noise through $N\noise$ relative to the spread of the points. Let $\noisefit$ be the noise standard deviation fitted by \method, and let $\rmsrad=(n^{-1}\sum_i\norm{\vect u_i-\bar{\vect u}}^2)^{1/2}$ be the root-mean-square radius of the 2D points about their mean $\bar{\vect u}$, both in pixels. To compare point counts on one scale, we plot results against $N\noisefit/\rmsrad$. Over this noise range, the rotation error of \method\ follows that of the pose fitted to the true correspondences. Its percentage of correct labels follows that of minimum-cost bijective matching at the true pose (\cref{fig:supp_degradation}(a)). The second agreement is consistent with \cref{prop:plugin}, which bounds the labeling error of nearest-projection matching at an accurate pose. It is also consistent with \cref{supp:thm-lower}, under which, for large $N$, even an estimator that knows the camera errs with probability bounded away from zero once $N\noise$ is at least a fixed positive constant.

\begin{figure*}[tp]
\centering
\includegraphics[width=489.95314pt,height=118.15868pt]{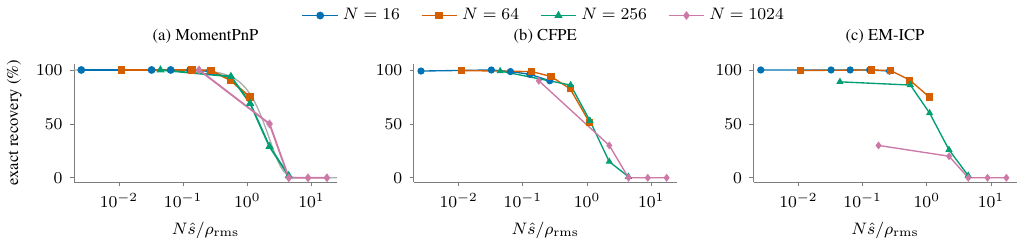}
\caption{\textbf{Exact label recovery across point counts.} Percentage of synthetic frames in which every 2D point is labeled with its generating model point, for (a)~\method, (b)~CFPE and (c)~EM-ICP. All panels place each noise level at $N\noisefit/\rmsrad$ computed from the noise $\noisefit$ fitted by \method. The gray curve in (a) is the fit $e^{-\kappa_{\mathrm{fit}}(N\noisefit/\rmsrad)^2}$ with $\kappa_{\mathrm{fit}}=\ThresholdKappa$.}
\label{fig:threshold_all_sizes}
\end{figure*}

\para{Exact label recovery.}
A stricter measure is the percentage of frames in which every 2D point is labeled with its generating model point. \cref{fig:threshold_all_sizes} plots it for \method, CFPE and EM-ICP against $N\noisefit/\rmsrad$, and \cref{fig:recovery_supp}(a) against the noise at $N=1024$. EM-ICP restarts on the CPU for up to 30~s. At $N=1024$ and $\noise=0.02$ pixels, \method\ recovers every label in all 40 frames, compared with 90\% of the frames for CFPE and 30\% for EM-ICP. At $N=1024$ and $\noise=0.5$ pixels, none of these methods recovers every label in any frame.

\para{Structured outliers.}
Real outliers often lie on scene structure, which the uniform outlier model of \cref{def:moments} does not describe. To test such outliers, we take Data61 frames with $N$ model points and replace a fraction of the 2D points of each frame with as many other 2D points on the same scene structures. \method\ is within $1\degree$ on every frame with up to 30\% replaced and on 79\% to 99\% of the frames at 50\% (\cref{fig:supp_degradation}(b)). At every fraction, BPnPNet is within $1\degree$ on fewer frames.

\begin{table}[tp]
\centering\footnotesize
\setlength{\tabcolsep}{3pt}
\caption{\textbf{Search implementation and pose ambiguity resolution (ours)} on all 10{,}795 MegaDepth, 2{,}466 ModelNet40 and 2{,}640 Data61 outlier-free test frames. The top rows give the percentage of frames with rotation error below $1\degree$ for the GPU search and for the CPU search alone, and the ratio of their median times per frame. Bold marks the more accurate search. The lower rows count the frames in which an alternative pose of Step 4 is examined, the frames in which it replaces the selected pose, and the frames with the stated accuracy and acceptance outcomes.}
\label{tab:ablation}
\begin{tabular*}{\columnwidth}{@{\extracolsep{\fill}}lrrr@{}}
\toprule
 & MegaDepth & ModelNet40 & Data61 \\
\midrule
GPU search (ours) (\%)\,$\uparrow$ & \textbf{99.9} & \textbf{97.0} & \textbf{100.0} \\
CPU search only (ours) (\%)\,$\uparrow$ & 99.8 & 96.8 & 99.9 \\
CPU/GPU runtime ratio & 4.7 & 6.4 & 5.0 \\
\midrule
Ambiguous pose examined (frames) & 154 & 715 & 0 \\
\quad replaces the pose, GPU search & 0 & 2 & 0 \\
\quad replaces the pose, CPU search & 15 & 47 & 0 \\
Accepted with error $\ge5\degree$ (frames) & 0 & 13 & 0 \\
Within $1\degree$ but declined (frames) & 7 & 60 & 0 \\
\bottomrule
\end{tabular*}
\end{table}

\para{Search implementation \& pose ambiguity resolution.}
\cref{tab:ablation} measures the effect of the batched GPU search, the pose ambiguity resolution and the acceptance test on all 15{,}901 outlier-free test frames. The CPU search keeps moment matching, EM and acceptance, but replaces the batched GPU search with a coarser search. The GPU search places 97.0 to 100\% of the frames of each test set within $1\degree$. The coarser CPU search loses at most 0.2 percentage points and takes 4.7 to 6.4 times the median GPU runtime. The rows on pose ambiguity resolution count the frames in which an alternative pose of Step 4 is examined and the frames in which it replaces the selected pose (\cref{supp:lem-companions}). These replacements occur mainly in the CPU search, on flat MegaDepth scenes and ModelNet40 objects. Both replacements in the GPU search and 61 of the 62 in the CPU search are within $1\degree$. Alternatives within $3\degree$ of an accepted pose are discarded, so each of these replacements corrects a pose at least $2\degree$ from the true one. Model symmetry has a related effect. Under exact symmetry, different poses give identical projections, and near-symmetry can make the moment residual less informative about orientation error (\cref{sec:supp_auxiliary,sec:supp_proof_companions}). On ModelNet40, 27 of the 31 frames with rotation error of at least $5\degree$ meet the ADD-S criterion of 10\% of the model diameter (\cref{sec:supp_tables}), as do 11 of the 13 accepted with such errors. The last two rows count the errors of the acceptance test in both directions. The test accepts 13 ModelNet40 poses with rotation error of at least $5\degree$ and none on the other test sets. It declines \MDdeclinedRight\ MegaDepth and \MNdeclinedRight\ ModelNet40 poses within $1\degree$.

\begin{table}[!tp]
\centering\small
\setlength{\tabcolsep}{4pt}
\caption{\textbf{Selection of visible points (ours).} Accurate poses in 30 paired runs with $n$ 2D points, 96 model points and 24 inliers, with model points chosen at random or by directional deletion. Both counts require rotation error below $1\degree$. Column $n_t$ also requires translation error below 1\% of the reference translation norm, and $n_C$ camera-center error below 1\% of the root-mean-square radius of the model.}
\label{tab:supp_overlap_control}
\begin{tabular*}{\columnwidth}{@{\extracolsep{\fill}}rcccc@{}}
\toprule
 & \multicolumn{2}{c}{Random selection} & \multicolumn{2}{c}{Directional deletion} \\
\cmidrule(lr){2-3}\cmidrule(lr){4-5}
$n$ & $n_t$ & $n_C$ & $n_t$ & $n_C$ \\
\midrule
24 & 29 & 5 & 1 & 0 \\
32 & 27 & 4 & 0 & 0 \\
48 & 26 & 5 & 0 & 0 \\
96 & 0 & 0 & 0 & 0 \\
\bottomrule
\end{tabular*}
\end{table}

\para{Selection of visible points.}
Moment matching needs both point sets to sample the same structure, so recovery can depend on which model points are visible as well as on how many. We compare two selections in paired trials on thirty independent 3D models of 96 model points (\cref{tab:supp_overlap_control}). Each pair uses the same 3D point set, intrinsics, pose, measurement noise and unmatched 2D points. The noise is Gaussian with standard deviation $0.001$ per calibrated coordinate. In both selections, 24 model points generate 2D points. Random selection chooses them uniformly at random, and directional deletion chooses them from one end of the model along a random image direction. Uniform unmatched 2D points then bring the 2D point set to $n$ points. With $n\le48$, random selection gives 26 to 29 of the 30 poses within $1\degree$ and 1\% translation error, and directional deletion at most one. Recovery thus depends on which points are visible.

\para{Acceptance on held-out frames.}
\cref{fig:supp_degradation}(c) splits the 1{,}000 held-out frames of each dataset and outlier ratio by acceptance and by rotation error below $1\degree$. In every setting, at most 8 frames yield an accepted pose with rotation error of at least $1\degree$.

\subsection{Independently selected inputs}
\label{sec:supp_observations}

\begin{figure}[tp]
\centering
\includegraphics[width=232.45886pt,height=46.66246pt]{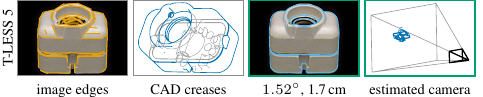}
\caption{\textbf{Image edges and a CAD model.} For T-LESS object 5, we show detected image edges, CAD creases with hidden creases in gray, the projected creases at the recovered pose, and the estimated camera. Rotation and camera-center errors appear below the projection.}
\label{fig:supp_independent_rows}
\end{figure}

In every study above, each inlier 2D point comes from a model point, through a track, a released pair or rendering. In the studies of this subsection, the 2D points and the model points are selected independently, so no 2D--3D correspondences are given. The studies pair scale-invariant feature transform (SIFT) detections with a 3D model built from other photographs, image edges with a CAD model, and detected motion-capture markers or infrared light-emitting diodes (LEDs) with their 3D positions. The SIFT study uses the search over camera centers and rotations of \cref{sec:supp_whole_scene}. The study with CAD models uses the edge features of \cref{sec:supp_independent_edges}. The marker studies apply \method\ to the two point sets. \cref{tab:independent_summary} summarizes recovery in these studies, with a SIFT example in \cref{fig:failure}(a) and a CAD example in \cref{fig:supp_independent_rows}.
\begin{table}[tp]
\centering\footnotesize
\setlength{\tabcolsep}{3pt}
\caption{\textbf{Recovery from independently selected inputs (ours).} A case is recovered when its rotation and position errors are below the listed thresholds. Position error is the camera-center error or, for the infrared LEDs, the translation error of the motion relative to frame 61. Raw evaluates the returned pose, and Symmetry applies the official T-LESS model symmetry. We count every selected case, including extraction failures and poses returned at the search limit.}
\label{tab:independent_summary}
\begin{tabular*}{\columnwidth}{@{\extracolsep{\fill}}lccc@{}}
\toprule
Observations & Thresholds & Raw\,$\uparrow$ & Symmetry\,$\uparrow$ \\
\midrule
All SIFT detections & $1\degree$, 1\% extent & 2/6 & --- \\
RGB/CAD edges & $5\degree$, 5~cm & 6/20 & 8/20 \\
\quad New objects & $5\degree$, 5~cm & 3/16 & 5/16 \\
Motion-capture centroids & $5\degree$, 5~cm & 6/8 & --- \\
Infrared LEDs (relative) & $5\degree$, 5~cm & 19/31 & --- \\
\bottomrule
\end{tabular*}
\end{table}

\para{SIFT detections \& reconstructed maps.}
\label{sec:supp_all_keypoints}
We use the ETH3D scenes Office, Courtyard and Facade. In each scene, the first photograph is a tuning query, and two photographs evenly spaced through the scene's list of photographs are test queries. We compute new SIFT detections~\cite{lowe2004sift} and triangulate the model points~\cite{schonberger2016sfm} from source photographs that exclude all nine queries, so no query photograph takes part in feature extraction, matching or triangulation. The Office, Courtyard and Facade maps contain 3{,}272, 42{,}714 and 96{,}909 model points from 23, 35 and 73 source photographs. The search uses the grids of \cref{sec:supp_whole_scene}, with three subdivision levels.

\para{Acceptance.}
We tune the acceptance thresholds for these inputs on the three tuning queries and fix them before evaluating the six test queries. A pose is accepted only if at least 30\% of the points in the smaller point set are matched. This applies both to the expected number $n_{\rm m}$ of 2D points matched under EM (\cref{sec:supp_config}) and to the final correspondences. We require the median, over the matched 2D points, of the ratio of the distances to the nearest and second-nearest projections to be at most 0.12. For queries with at least 500 detections and an association level of at most $10^{-100}$, we relax this limit to 0.50. The benchmark acceptance test with outliers permits 0.25 in this case (\cref{sec:supp_acceptance}). We run the search on each test query with a time limit of 3{,}600 seconds, checked between completed operations. At the limit, it returns its best valid pose. A query is recovered when its rotation error is below $1\degree$ and its camera-center error below 1\% of the scene extent.

\begin{table*}[tp]
\centering\footnotesize
\setlength{\tabcolsep}{3pt}
\caption{\textbf{SIFT queries against independently built maps (ours).} Column 2D points counts every SIFT detection in the query image, including several orientations at one location, and Locations counts distinct image locations. Center gives the camera-center error as a percentage of the scene extent (\cref{sec:supp_whole_scene}). Recovered requires rotation error below $1\degree$ and camera-center error below 1\%, and Accepted states whether the returned pose passes the acceptance tests. Times above 3{,}600~s mark searches stopped at the limit.}
\label{tab:supp_all_keypoints}
\begin{tabular*}{\textwidth}{@{\extracolsep{\fill}}lrrrrrrcc@{}}
\toprule
Query & 2D points & Locations & Model points & Rot. ($\degree$) $\downarrow$ & Center (\%) $\downarrow$ & Time (s) $\downarrow$ & Recovered $\uparrow$ & Accepted \\
\midrule
Office 9 & 335 & 287 & 3{,}272 & 70.58 & 54.546 & 3600.1 & no & no \\
Office 18 & 1{,}327 & 1{,}134 & 3{,}272 & 0.11 & 0.054 & 3604.2 & yes & no \\
Courtyard 13 & 13{,}123 & 11{,}525 & 42{,}714 & 103.08 & 15.811 & 3618.6 & no & no \\
Courtyard 26 & 11{,}079 & 9{,}399 & 42{,}714 & 93.82 & 7.460 & 3600.2 & no & no \\
Facade 26 & 9{,}432 & 8{,}205 & 96{,}909 & 0.0027 & 0.001 & 870.6 & yes & yes \\
Facade 51 & 12{,}525 & 10{,}517 & 96{,}909 & 177.34 & 21.350 & 3600.2 & no & no \\
\bottomrule
\end{tabular*}
\end{table*}

\para{SIFT results.}
\cref{tab:supp_all_keypoints} gives every returned pose. Two of the six test queries are recovered. The accepted Facade 26 pose of \cref{fig:failure}(a) gives 5{,}666 distinct pairs, all in front of the camera. Their median and root-mean-square reprojection distances are 0.763 and 1.168 pixels. The accurate Office 18 pose is returned at the time limit without being accepted.

\para{Why such inputs are harder.}
\cref{fig:failure}(b,c) shows two measurements that explain this difficulty. First, the two point sets can sample different structure. Of the 12{,}069 SIFT points detected in a Meadow image, 255 belong to reconstruction tracks, and many of the other detections lie on vegetation and sky that the reconstruction does not contain (\cref{fig:failure}(b)). Second, a dense 3D model places projections near most 2D points at more than one pose. In a Facade image, 2.6 million laser-scan points fall within the view. At the reference pose, 80.6\% of the 9{,}754 2D points lie within 3.4 pixels of a projected scan point. At a pose rotated by $2\degree$, 76.8\% still do (\cref{fig:failure}(c)). Distances to the nearest projections therefore barely distinguish the two poses.

\para{Image edges \& CAD models.}
We fix all filter, search and refinement settings on four tuning views of T-LESS~\cite{hodan2017tless} object 25. Before evaluating any pose error, we select views 72 and 360 of objects 1, 5, 9, 13, 17, 21, 26 and 29, and views 72, 216, 360 and 504 of object 25. The twenty test views thus contain sixteen views of new objects and four held-out views of the tuning object. Each run receives one RGB image, the intrinsics and a CAD mesh in its own coordinate frame, and no reference pose, visibility, model symmetries or edge identities.

\para{Edge features \& search.}
We use eight-pixel segments from the line segment detector (LSD)~\cite{grompone2012lsd} in the image, and creases and candidate contours (\cref{sec:supp_independent_edges}) on the CAD model. We normalize image coordinates by the arclength-weighted centroid and root-mean-square radius of the detected image edges. We integrate Fourier features analytically along each projected segment using sinc functions. We score an initial grid of poses with these features and keep the best poses for EM. Scoring on this grid ignores occlusion. For the retained poses, exact visibility then clips occluded edges, and all CAD vertices must lie in front of the camera and project inside the image. For point-based EM, we sample each projected edge once by image arclength and keep these samples during refinement. For the final ranking, we recompute visibility and sampling and score each pose by an edge-curve likelihood. An improvement of the point-based EM objective thus need not improve this likelihood.

\begin{table}[tp]
\centering\footnotesize
\setlength{\tabcolsep}{3pt}
\caption{\textbf{Held-out T-LESS views (ours).} Rotation error in degrees and camera-center error in millimeters for each of the twenty test views. Raw pose evaluates the returned pose, and Model symmetry applies the same official model symmetry to both errors, sampling continuous symmetries in $1\degree$ steps. Bold marks views with rotation error below $5\degree$ and camera-center error below 50~mm. The views of object 25 are held-out views of the tuning object.}
\label{tab:supp_tless_holdout}
\begin{tabular*}{\columnwidth}{@{\extracolsep{\fill}}rrrrrr@{}}
\toprule
 & & \multicolumn{2}{c}{Raw pose} & \multicolumn{2}{c}{Model symmetry} \\
\cmidrule(lr){3-4}\cmidrule(l){5-6}
Object & View & Rot. $\downarrow$ & Center $\downarrow$ & Rot. $\downarrow$ & Center $\downarrow$ \\
\midrule
1 & 72 & 159.94 & 1169.4 & 159.88 & 1169.7 \\
1 & 360 & 17.81 & 188.4 & 14.70 & 167.0 \\
5 & 72 & 128.37 & 1141.4 & 128.37 & 1141.4 \\
5 & 360 & \textbf{1.52} & \textbf{17.4} & \textbf{1.52} & \textbf{17.4} \\
9 & 72 & 179.65 & 353.4 & \textbf{3.33} & \textbf{35.8} \\
9 & 360 & \textbf{1.93} & \textbf{22.7} & \textbf{1.93} & \textbf{22.7} \\
13 & 72 & 131.84 & 72.6 & 33.66 & 337.4 \\
13 & 360 & 176.58 & 911.4 & 142.36 & 1140.8 \\
17 & 72 & 75.74 & 176.2 & 4.92 & 52.4 \\
17 & 360 & 26.32 & 256.6 & \textbf{1.95} & \textbf{17.4} \\
21 & 72 & 46.42 & 505.3 & 46.42 & 505.3 \\
21 & 360 & 133.74 & 755.7 & 133.74 & 755.7 \\
26 & 72 & 6.45 & 68.2 & 6.45 & 68.2 \\
26 & 360 & \textbf{3.14} & \textbf{28.8} & \textbf{3.14} & \textbf{28.8} \\
29 & 72 & 179.81 & 437.4 & 7.42 & 84.2 \\
29 & 360 & 75.03 & 781.3 & 75.03 & 781.3 \\
25 & 72 & \textbf{4.52} & \textbf{47.0} & \textbf{4.52} & \textbf{47.0} \\
25 & 216 & \textbf{3.28} & \textbf{32.7} & \textbf{3.28} & \textbf{32.7} \\
25 & 360 & \textbf{1.88} & \textbf{20.2} & \textbf{1.88} & \textbf{20.2} \\
25 & 504 & 114.00 & 1071.1 & 66.53 & 700.1 \\
\midrule
\multicolumn{2}{l}{Within both thresholds $\uparrow$} & \multicolumn{2}{c}{6/20} & \multicolumn{2}{c}{8/20} \\
\bottomrule
\end{tabular*}
\end{table}

\begin{table}[tp]
\centering\footnotesize
\setlength{\tabcolsep}{3pt}
\caption{\textbf{Motion-capture markers and infrared LEDs (ours).} A comparison is accurate when its rotation error is below $5\degree$ and its position error below 5~cm. For motion capture, these are the errors of the camera pose. For the LEDs, they are the errors of the motion of each of the 31 other selected frames relative to frame 61. Accepted outputs and extraction failures count frames, and inaccurate accepted comparisons count motions. Error medians use all eight motion-capture poses and the 24 LED motions for which both poses are available.}
\label{tab:supp_independent_sensors}
\begin{tabular*}{\columnwidth}{@{\extracolsep{\fill}}lrr@{}}
\toprule
 & Motion capture & Infrared LEDs \\
\midrule
Accurate comparisons\,$\uparrow$ & 6/8 & 19/31 \\
Accepted pose outputs & 5/8 & 25/32 \\
Inaccurate accepted comparisons\,$\downarrow$ & 1/8 & 5/31 \\
Extraction failures\,$\downarrow$ & 0/8 & 7/32 \\
Median rotation error ($\degree$)\,$\downarrow$ & 0.324 & 1.063 \\
Median position error (mm)\,$\downarrow$ & 20.9 & 25.9 \\
\bottomrule
\end{tabular*}
\end{table}

\para{Edge results.}
\cref{tab:supp_tless_holdout} gives the rotation and camera-center errors of every test view, for the returned pose and after applying the official model symmetry. A view is recovered when its rotation error is below $5\degree$ and its camera-center error below 50~mm. To find whether a failure comes from the search or from the objective, we score the reference and estimated poses with the same final likelihood, using a Gaussian standard deviation of two pixels and a uniform outlier fraction of 0.05. Every reference pose satisfies the constraints of the search, with every CAD vertex in front of the camera and projecting inside the image. At these poses, we sample the detected image edges and the visible projected CAD edges by image arclength, at most two pixels apart. A median of 72.9\% of the image-edge samples and 83.1\% of the CAD-edge samples lie within three pixels of a sample of the other set. The detected and projected edges thus overlap only in part. Of the twelve views that fail under model symmetry, four return poses with lower final likelihood than the reference pose, so the search misses a more likely pose. In the other eight, the estimated pose has higher likelihood than the reference, so the objective itself favors these inaccurate estimates.

\para{Motion-capture markers.}
We use the static trial of a public running motion-capture dataset.\footnote{\url{https://github.com/MaxwellM34/Running_Motion_Capture_Dataset}} For the query camera, we keep all unlabeled marker centroids, corrected for radial distortion. We tune on one camera at the first frame and evaluate all eight other cameras at frame 501, using \method\ with one-pixel noise and a 30-second time limit. Six of the eight returned poses are within $5\degree$ and 5~cm, and none is within $1\degree$ and 1~cm. One of the five accepted poses fails the $5\degree$ and 5~cm thresholds. As a control, efficient perspective-n-point (EPnP)~\cite{lepetit2009epnp} with the reference assignments is within $5\degree$ and 5~cm on all eight cameras and within $1\degree$ and 1~cm on one.

\para{Infrared LEDs.}
We use the raw infrared images and the surveyed coordinates of four noncoplanar LEDs from a public demonstration~\cite{faessler2014leds}. A fixed filter on intensity, area and contour shape extracts the LED centroids, and the point solver runs in its outlier-free setting with one-pixel noise and a ten-second time limit. The transform between the LED model and the attached AprilTag~\cite{olson2011apriltag} is unknown, so we compare motions relative to frame 61 with independent tag estimates. With $\Xi^{\mathrm{tag}}_i$ and $\Xi^{\mathrm{led}}_i$ the rigid transforms from the tag frame and from the LED frame to the camera frame at frame $i$, $\Xi^{\mathrm{tag}}_i=\Xi^{\mathrm{led}}_i\Xi^{\mathrm{obj}}$ for a fixed transform $\Xi^{\mathrm{obj}}$ between the object frames, so the relative transforms $\Xi^{\mathrm{tag}}_j(\Xi^{\mathrm{tag}}_i)^{-1}=\Xi^{\mathrm{led}}_j(\Xi^{\mathrm{led}}_i)^{-1}$ do not depend on $\Xi^{\mathrm{obj}}$ (\cref{tab:supp_independent_sensors}). A fixed error in the camera frame does not cancel in this comparison.

\end{document}